\documentclass[10pt,twocolumn,letterpaper]{article}

\usepackage{iccv}
\usepackage{times}
\usepackage{epsfig}
\usepackage{graphicx}
\usepackage{amsmath}
\usepackage{amssymb}
\usepackage{booktabs}
\usepackage{multirow}
\usepackage{makecell}
\usepackage{bbding}
\usepackage{upgreek}
\usepackage{enumitem}
\usepackage{paralist}
\usepackage{float}
\usepackage{subcaption}
\usepackage{threeparttable}
\usepackage{fontawesome}

\usepackage[pagebackref=true,breaklinks=true,letterpaper=true,colorlinks,bookmarks=false]{hyperref}

\iccvfinalcopy

\usepackage{soul}
\newcommand{\JHdel}[1]{\textcolor{red}{\st{#1}}}
\newcommand{\JHNOTE}[1]{\textcolor{red}{[JH: #1]}}
\newcommand{\WP}[1]{\textcolor{blue}{[WP: #1]}}

\usepackage[capitalize]{cleveref}

\def\iccvPaperID{9440} 

\ificcvfinal\fi
\usepackage{tikz}

\begin{document}
	
	\title{GeoUDF: Surface Reconstruction from 3D Point Clouds via Geometry-guided Distance Representation}
	
	\author{Siyu Ren$^{1,2}$ \quad Junhui Hou$^{1} \thanks{Corresponding author. This work was supported by the Hong Kong Research Grants Council under Grants 11202320 and 11219422.}$  \quad Xiaodong Chen$^2$ \quad Ying He$^3$ \quad Wenping Wang$^4$  \\
		$^1$City University of Hong Kong \quad $^2$Tianjin University \\ $^3$ Nanyang Technological University \quad $^4$Texas A\&M University  \\
		{\tt\small siyuren2-c@my.cityu.edu.hk, jh.hou@cityu.edu.hk, xdchen@tju.edu.cn,} \\ {\tt\small yhe@ntu.edu.sg, wenping@cs.hku.hk}
	}

	
	\maketitle
	
	\ificcvfinal\thispagestyle{empty}\fi
	
	\begin{abstract}

		We present a learning-based method, namely GeoUDF, to tackle the long-standing and challenging problem of reconstructing a discrete surface from a sparse point cloud.
		To be specific, we propose a geometry-guided learning method for UDF and its gradient estimation that explicitly formulates 
		the unsigned distance of a query point as the learnable affine 
		averaging of its distances to the tangent planes of neighboring points on the surface.
		Besides, we model the local geometric structure of the input point clouds by explicitly learning a quadratic polynomial for each point. This not only facilitates upsampling the input sparse point cloud 
		but also naturally induces unoriented normal, which further augments UDF estimation. 
		Finally, to extract triangle meshes from the predicted UDF 
		we propose a customized 
		edge-based marching cube module. We conduct extensive experiments and ablation studies
		to demonstrate the significant advantages of our method over state-of-the-art methods in terms of reconstruction accuracy, efficiency, and generality.
		The source code is publicly available at \href{https://github.com/rsy6318/GeoUDF}{https://github.com/rsy6318/GeoUDF}.
	\end{abstract}
	
	\section{Introduction}
	\label{sec:intro}
	Surface reconstruction from point clouds is a fundamental and challenging problem in 3D vision, graphics and robotics \cite{APP0,APP1,APP2}. 
	Traditional approaches, such as Poisson surface reconstruction~\cite{SPSR}, compute an occupancy or signed distance field (SDF) by solving Poisson's equation, resulting in a sparse line system. Then they utilize the Marching Cubes (MC) \cite{MARCHINGCUBE} algorithm to extract iso-surfaces as the reconstructed meshes. These methods are computationally efficient,  scalable, resilient to noise and tolerant to registration artifacts. However, they work only for oriented point samples. Recent developments in deep learning have demonstrated that implicit fields, such as binary occupancy fields (BOFs) and signed distance fields (SDFs), can be learned directly from raw point clouds, making them a promising tool for reconstructing surfaces from unoriented point samples. 
	
	Although BOFs and SDFS are effective at reconstructing watertight and manifold surfaces, they have limitations when it comes to representing more general types of surfaces, such as those with open boundaries or interior structures. In contrast, the unsigned distance field (UDF) is more powerful in terms of representation ability, and can overcome the above-mentioned limitations of BOFs and SDFs.
	The existing deep UDF-based methods, such as \cite{NDF,GIFS,MESHUDF}, employ neural networks to regress UDFs from input point clouds. Due to their reliance on the training data, these methods have limited generalizability
	and cannot deal with unseen objects well. Another technical challenge diminishing UDFs' practical usage is that one cannot use the conventional MC to extract iso-surfaces directly because UDFs do not have zero-crossings and thus do not provide the required sign-change information of inside or outside. Some methods \cite{MESHUDF, CAP-UDF} locally convert a UDF to an SDF in a cubic cell and then apply MC to extract the surface. However, they often fail at corners or edges. Recently, GIFS \cite{GIFS} modifies MC by predicting whether the surface interacts with each cube edge via a neural network. 
	
	In this paper, we propose \textit{GeoUDF}, a new learning-based framework reconstructing the underlying surface of a sparse point cloud. 
	We particularly aim at addressing two fundamental issues of UDFs, namely, (\textbf{1}) how to predict an accurate UDF
	and its gradients for arbitrary 3D inputs? and (\textbf{2}) how to extract triangle meshes, given a UDF?
	More importantly, the constructed methods should be well generalized to unseen data. To solve the first issue, we propose a geometry-guided learning module,
	adaptively approximating the UDF and its gradient of a  point cloud 
	by leveraging its local geometry 
	in a decoupled manner.
	As for the second issue, we propose an edge-based MC algorithm, extracting triangle mesh from any UDF according to the intersection relationship between the surface and the connections of any two vertices of the cubes. In addition, we propose to enhance input point cloud quality by explicitly and concisely modeling its local geometry structure,
	which is beneficial for surface reconstruction.
	The elegant formulations distinguish our GeoUDF from existing methods, making it \textit{more lightweight, efficient}, and \textit{accurate} and \textit{better generalizability}.  
	Besides, the proposed modules contained in GeoUDF can be independently
	used as a general method in its own right.
	Extensive experiments on various datasets show the significant superiority of our method over state-of-the-art methods.
	
	In summary, the main contributions of this paper are three-fold: 
	
	\begin{compactitem}
		\item novel accurate, compact, efficient, and explainable learning-based UDF and its gradient estimation methods that are \textit{decoupled};
		\item a concise yet effective learning-based point cloud upsampling and representation method; 
		and
		\item a general method for extracting triangle meshes from any unsigned distance field.
		
	\end{compactitem}
	
	\section{Related Work}
	\label{SEC:RELATED:WORKS}
	
	\begin{figure*}[tbp]
		\centering
		\includegraphics[width=0.95\textwidth]{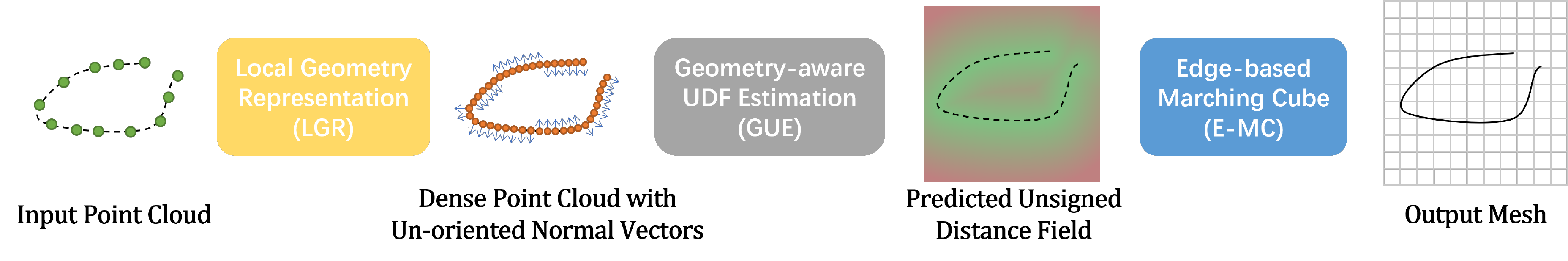}\vspace{-0.3cm}
		\caption{
			The pipeline of GeoUDF, a lightweight, explainable, and efficient  learning-based surface reconstruction framework that produces 3D meshes with high accuracy and strong generalizability. 
		}
		\label{PIPELINE} \vspace{-0.5cm}
	\end{figure*}
	
	\noindent\textbf{Learning Implicit Surface Representation}.
	With the development of deep learning, many advances have been made in the implicit representation of 3D shapes in recent years. For example, 
	Binary Occupancy Field (BOF) casts the problem into binary classification,  i.e.,  
	any point in the 3D space can be classified as inside or outside of a closed shape. ONet \cite{OCCNET} uses a latent vector to represent the whole input, then uses a decoder to get the occupancy of a query point. Subsequently, IF-NET \cite{IFNET} and CONet \cite{CONVOCCNET} adopt convolutions on  voxelized features, which increase the reconstruction accuracy. Referring to SPSR \cite{SPSR}, SAP \cite{SAP} introduces a differentiable formulation of SPSR and then incorporates it into a neural network to reconstruct the surfaces from point clouds. Recently, POCO \cite{POCO} designs an attention-based convolution and further improves the reconstruction quality. SDFs, distinguishing interior and exterior by assigning a sign to the distance,  are a more accurate representation. DeepSDF \cite{DEEPSDF} uses an auto-decoder to optimize the latent vector in order to refine the SDF. Based on DeepSDF, DeepLS \cite{DEEPLS} uses a grid to store the independent latent codes, each only responsible for a local area. Neural-Pull \cite{NeuralPull} and OSP \cite{OSP} concentrate on the differential property and use the gradient of the SDF to pull the point onto the surface, where they estimate and refine the SDF. Some traditional methods can also be adapted through neural networks, such as DeepIMLS \cite{DEEPIMLS} and DOG \cite{DOG}, which incorporate implicit moving least-squares (IMLS) \cite{IMLS} into neural networks to estimate the SDF of a point cloud.
	
	BOF and SDF can only deal with watertight shapes because they assume a surface that partitions the whole space inside and outside, making it impossible to represent general shapes, such as those with open surfaces with boundaries or non-manifold surfaces.
	UDF can represent general shapes since it represents the (unsigned) distance from a query point to the surface. NDF \cite{NDF} and GIFS \cite{GIFS} use 3D convolution to regress the UDF from the voxelized point cloud. CAP-UDF \cite{CAP-UDF} employs a field consistency constraint to get consistency-aware UDF. To our best knowledge, there are no methods of learning 
	a UDF from geometric information explicitly.
	
	\noindent\textbf{Surface Extraction from Implicit Fields}.
	After learning implicit  fields, such as BOFs or SDFs, the MC algorithm \cite{MARCHINGCUBE} is commonly used to obtain a triangle mesh. However, MC cannot be applied to UDFs because there is no inside/outside information.
	NDF \cite{NDF} uses the gradient of the UDF to generate a much denser point cloud and then uses a ball-pivoting algorithm  \cite{BALLPIVOTING} to extract the triangle mesh, which is inefficient with poor surface quality. GIFS \cite{GIFS} not only predicts the UDF, but also determines whether any two vertices of the cubes are separated by the surface, which may be regarded as a binary classification. Then it modifies the MC to extract a triangle mesh from the UDF. Note that the Marching Cubes method modified by GIFS cannot be applied to any UDF because it needs an extra network to achieve the binary classification on each edge of the cube. 
	MeshUDF \cite{MESHUDF} 
	uses one vertex of the cube as a reference and computes the inner product of the gradients of UDF with the other seven points, thus converting UDF to SDF in this cube according to the sign of the inner product. But such a criterion is not robust enough and often predicts wrong faces at the edges or corners.
	
	\noindent\textbf{Learning-based Point Cloud Upsampling}. 
	Point cloud upsampling (PU) aims to densify sparse point clouds to recover more geometry details, which will be beneficial to downstream surface reconstruction.
	As the first learning-based method for point cloud upsampling, PU-Net \cite{PUNET}  employs PointNet++ \cite{POINTNET2} to extract features and then expands the features through multi-branch MLPs to achieve the upsampling.
	Furthermore, the generative adversarial network-based PU-GAN \cite{PUGAN}
	achieves impressive results on non-uniform data, but the network is difficult to train. PU-GCN \cite{PUGCN}, a graph convolutional network-based approach, introduces the NodeShuffle module into the existing upsampling pipeline. The first geometry-centric method PUGeo-Net \cite{PUGEO} 
	links
	deep learning with differentiable geometry to learn the local geometry structure of point clouds.
	MPU \cite{PATCHPU} appends a 1D code $\{-1,~1\}$ to separate replicated the point-wise features.
	Built upon the linear approximation theorem, MAFU \cite{DMFU} adaptively learns the interpolation weights in a flexible manner, as well as high-order approximation errors. Alternatively, PU-Flow \cite{mao2022pu} conducts learnable interpolation in feature space regularized by normalizing flows.
	Recently, NP \cite{NP} adopts a neural network to represent the local geometric shape in a continuous manner, 
	embedding more shape information.
	
	\section{Proposed Method}
	\label{PRO:METHOD}
	\noindent\textbf{Overview}. 
	Our GeoUDF consists of three modules: local geometry representation (LGR), geometry-guided UDF estimation (GUE), and edge-based marching cube (E-MC), as shown in Fig. \ref{PIPELINE}. Specifically, given a sparse 3D point cloud, we first model its local geometry through LGR, producing a dense point cloud associated with un-oriented normal vectors. Then, we predict the  unsigned distance field of the resulting dense point cloud via GUE 
	from which we customize an E-MC module to extract the triangle mesh for the zero-level set. \textbf{Note} that
	our GeoUDF works for both watertight and open surfaces with interior structures. Besides, 
	each of the three modules can be independently used as a general method in its own right.  In what follows, we will detail each module. 
	
	\subsection{Local Geometry Representation} \label{SEC:LGR}
	\noindent \textbf{Problem formulation}. Let $\mathcal{P}=\{\mathbf{p}_i|\mathbf{p}_i\in\mathbb{R}^3\}_{i=1}^N$ be an input sparse point cloud of $N$ points sampled from a surface $\mathcal{S}$ to be reconstructed.  
	The local theory of surfaces in differential geometry \cite{FUNDAMENTAL} shows that 
	the local geometry of any point of a regular surface is uniquely determined by the first and second fundamental forms up to rigid motion, and can be expressed as a quadratic function. 
	Therefore we use a polynomial to approximate the local patch centered at each point: \vspace{-0.2cm}
	\begin{equation}
		\label{2ND:POLY}
		f_i(\mathbf{u})=\mathbf{p}_i+\mathbf{A}_i\texttt{E}(\mathbf{u}), \vspace{-0.15cm}
	\end{equation}
	where $\mathbf{u}=[u_1,u_2]^\mathsf{T}\in\mathbb{R}^2$ is the coordinate in the 2D local parameter domain,  $\texttt{E}(\mathbf{u}):=[1\ u_1\ u_2\ u_1^2\ u_1u_2\ u_2^2]^\mathsf{T}\in\mathbb{R}^{6}$,
	and $\mathbf{A}_i\in\mathbb{R}^{3\times 6}$ is the coefficient matrix.
	
	With this representation, we can perform the following two operations, which will bring  benefits to the subsequent module.  \textbf{(1)} \textit{Densifying $\mathcal{P}$}.
	For each point $\mathbf{p}_i\in\mathcal{P}$, we can uniformly sample $M$ 2D coordinates  
	from a pre-defined local parameterization 
	$\mathcal{D}=[-\delta, \delta]^2\subset\mathbb{R}^2$, 
	which are then substituted into  Eq. (\ref{2ND:POLY}) 
	to generate additional $M$
	3D points  
	around $\mathbf{p}_i$, 
	producing a dense point cloud  $\mathcal{P}_M=\{\mathbf{p}_j|\mathbf{p}_j\in\mathbb{R}^3\}_{j=1}^{NM}$. \textbf{(2)} \textit{Inducing un-oriented normal vectors of $\mathcal{P}_M$}. From the explicit parameterization function in
	Eq. \eqref{2ND:POLY}, we can calculate the un-oriented normal vector for each point of $\mathcal{P}_M$,  denoted as $\mathcal{N}_M=\{\mathbf{n}_j|\mathbf{n}_j\in\mathbb{R}^3,\|\mathbf{n}_j\|=1\}$,  based on the differential geometry property \cite{DIFF_PRO}. 
	
\begin{figure}
		\centering
		\includegraphics[width=0.47\textwidth]{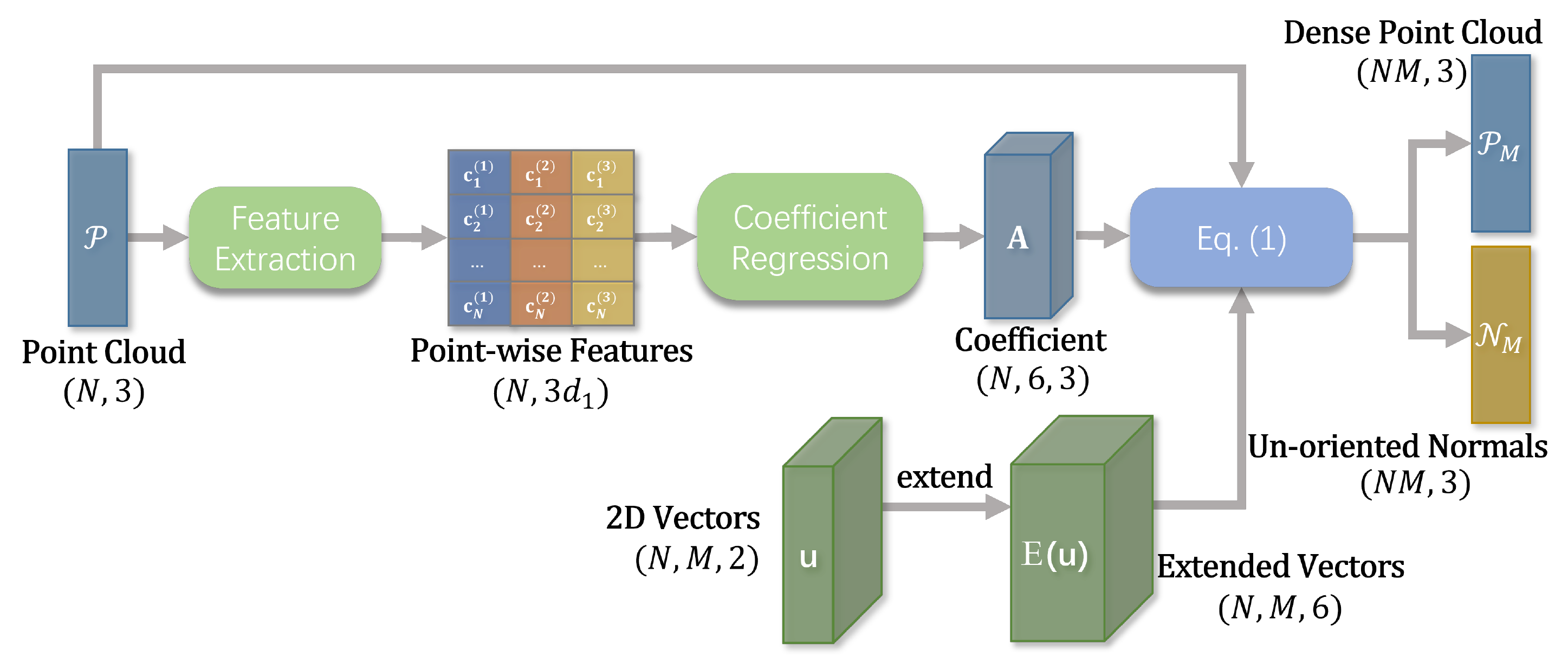}\vspace{-0.3cm}
		\caption{Flowchart of the LGR module. Note that the 2D coordinates are randomly sampled from $\mathcal{D}$ via PDS \cite{POISSONSAMPLING} at each iteration. We refer readers to the \textit{supplementary material} for the detailed network architecture.
		}
		\label{PU:ARC}
		\vspace{-0.4cm}
	\end{figure}

	\noindent\textbf{Learning-based implementation}. As depicted in Fig. \ref{PU:ARC}, we design a sub-network to realize this representation process in a data-driven manner, which is boiled down to predicting the coefficient matrix $\{\mathbf{A}_i\}_{i=1}^N$. Specifically, we adopt 3-layer EdgeConvs \cite{DGCNN}, which can capture  
	structural information of point cloud data,
	to embed $\mathcal{P}$ 
	into a high-dimensional feature space, generating  point-wise features $\{\mathbf{c}^{(l)}_{i}\in\mathbb{R}^{d_1}\}_{i=1}^N$
	at the $l$-th layer. Then, we concatenate the features of three layers and feed them into an MLP  
	to predict 
	18-dimension vectors, which are further reshaped into $\{\mathbf{A}_{i}\}_{i=1}^N$.  
	Besides, in Sec. \ref{SEC:ABLATION}, we experimentally demonstrate the advantages of this concise operation over state-of-the-art point cloud upsampling methods.

	\subsection{Geometry-guided  UDF Estimation}
	This module aims to estimate the unsigned distance field of $\mathcal{P}_M$, where the iso-surface with the value of zero indicates the surface. As aforementioned,
	existing learning-based UDF estimation methods, such as NDF \cite{NDF} and GIFS \cite{GIFS},  utilize a neural network to implicitly regress the UDFs from point clouds, thus limiting their accuracy. Besides, the trained models often yield poor results on unseen data.
	In sharp contrast to the regression-based methods, GUE  leverages the inherent geometric property of input point clouds, leading to a more accurate UDF estimation method with better generalizability.
	
	\begin{figure}[tbp]
		\centering
		\subfloat[]{\includegraphics[width=0.12\textwidth]{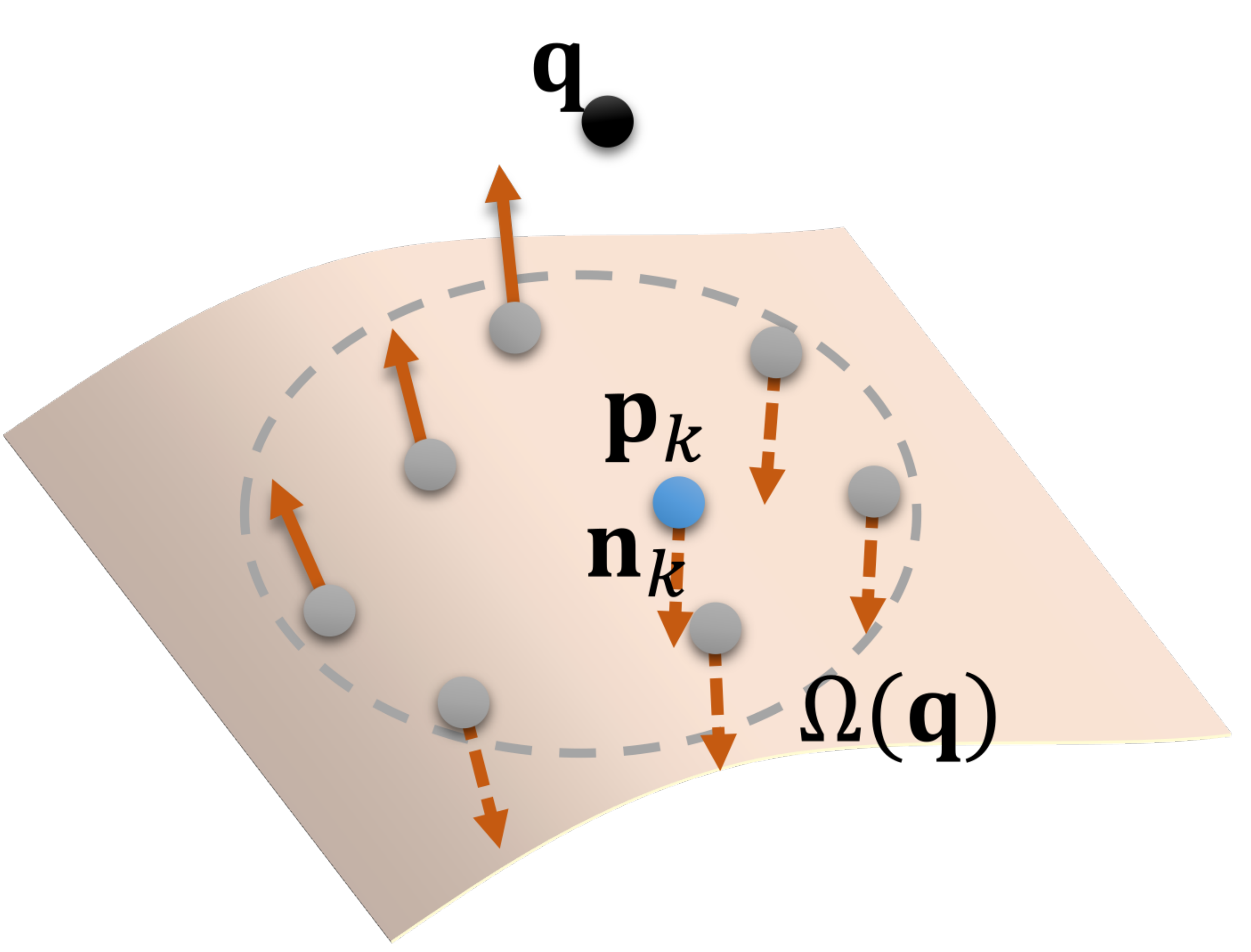}\label{NORMAL:UNORIENT}} 
		\subfloat[]{\includegraphics[width=0.12\textwidth]{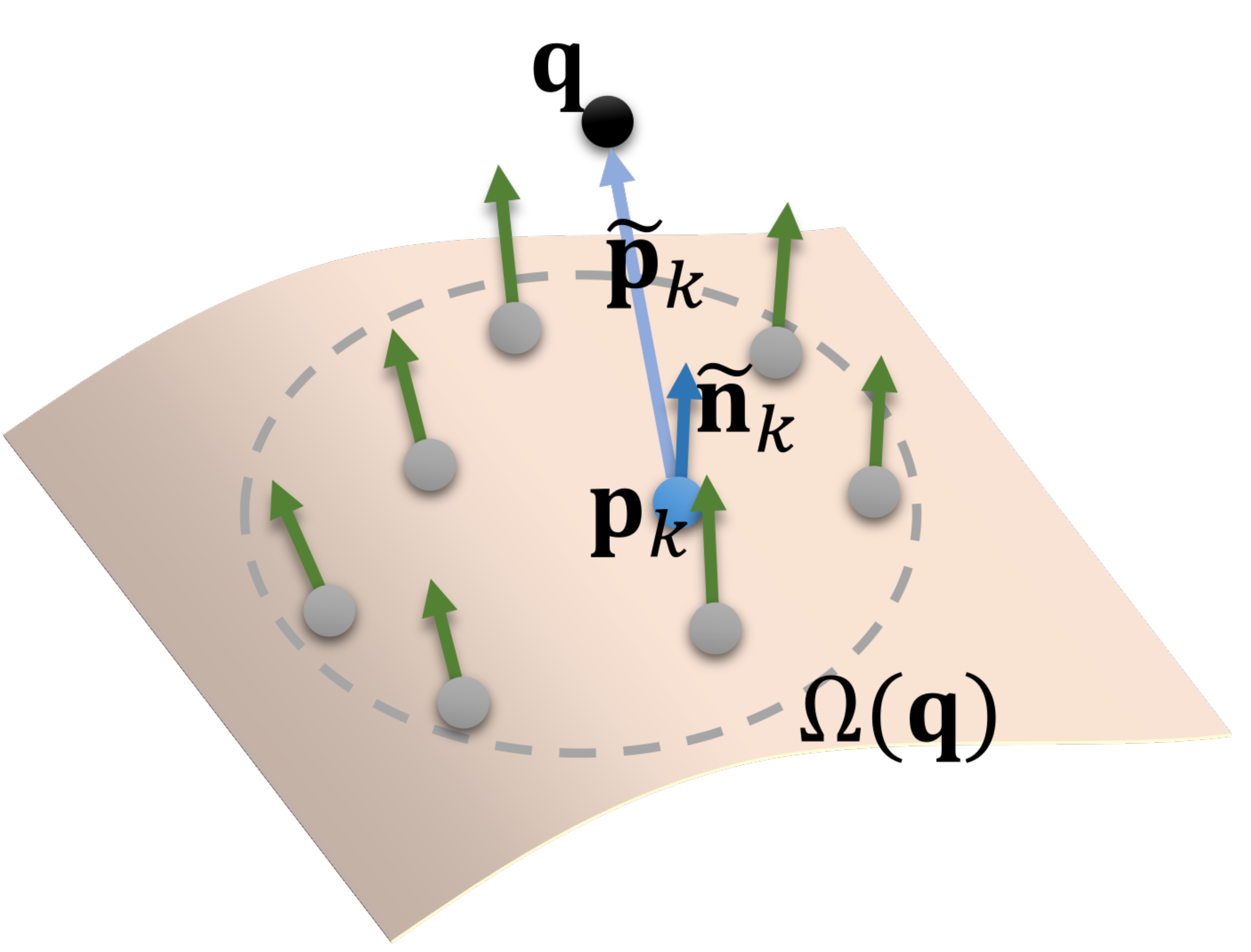}\label{NORMAL:ORIENT}} 
		\subfloat[]{\includegraphics[width=0.12\textwidth]{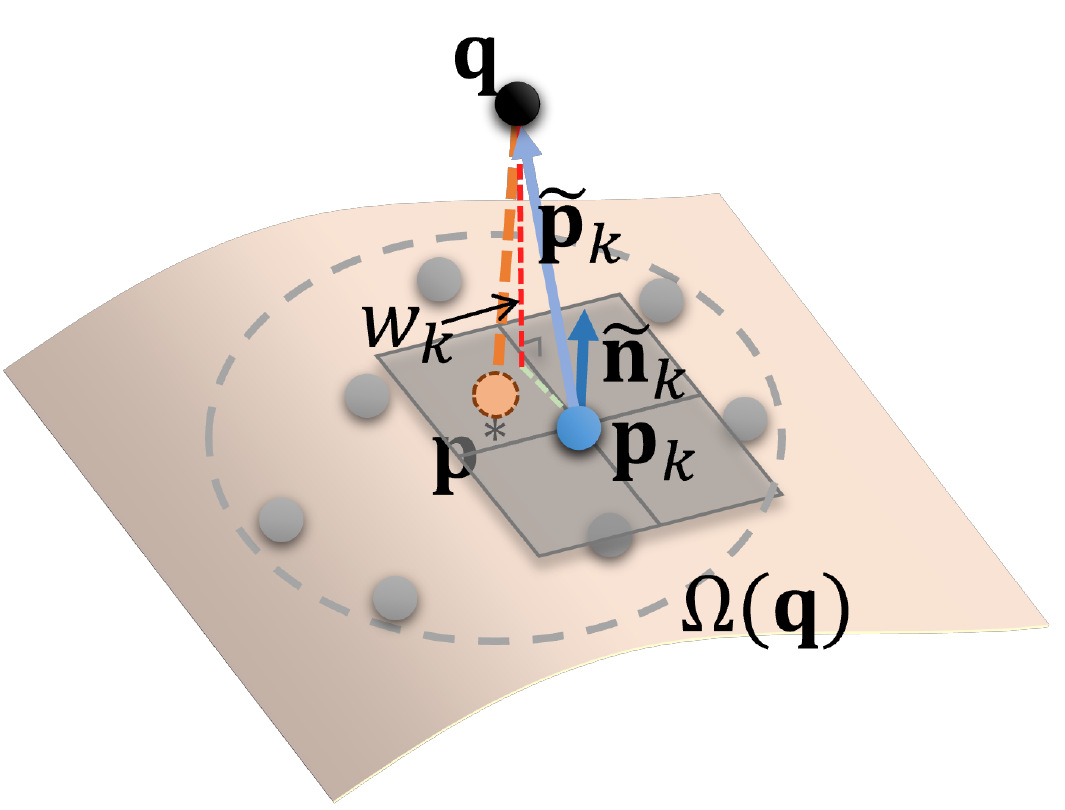}\label{UDF}} 
		\subfloat[]{\includegraphics[width=0.12\textwidth]{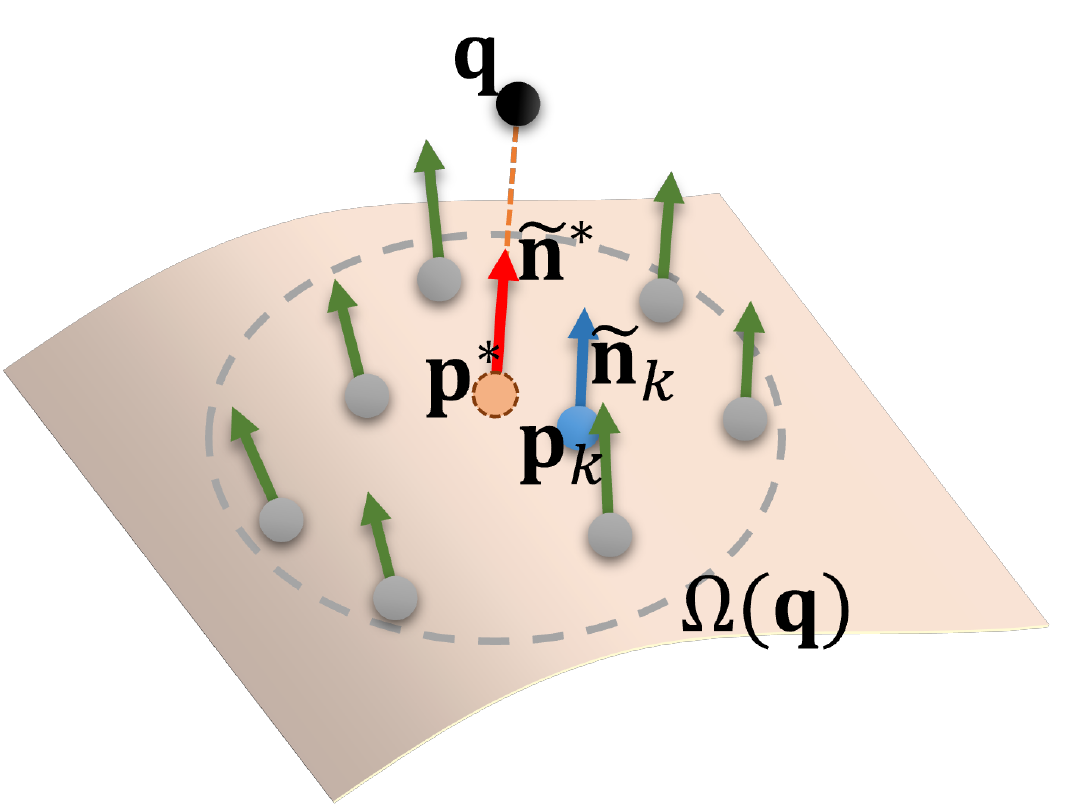}\label{UDF_GRAD}} 
		\vspace{-0.15cm}
		\caption{Illustration of the UDF and its gradient estimation processes. (a) Un-oriented normal vectors. (b) Aligned normal vectors. (c) UDF estimation. (d) UDF gradient estimation.
		}
		\label{UDF:ESTIMATION}  \vspace{-0.2cm}
	\end{figure}

	\noindent \textbf{Formulation of UDF and its gradient estimation}. Given a query point $\mathbf{q}\in\mathbb{R}^3$, we can find its $K$ nearest points from $\mathcal{P}_M$ in a Euclidean distance sense, denoted as $\Omega(\mathbf{q})=\{\mathbf{p}_k|\mathbf{p}_k\in\mathcal{P}_R\}_{k=1}^K$. Denote by  $\mathbf{n}_k$ the un-oriented normal of $\mathbf{p}_k$.
	Before calculating the UDF, as shown in Fig. \ref{NORMAL:ORIENT}, we first align 
	$\{\mathbf{n}_k\}_{k=1}^K$ with reference to  $\mathbf{q}$ via \vspace{-0.15cm}
	\begin{equation}\label{EQ:NORMAL:ORIENTATION}
		\mathbf{\tilde{n}}_k(\mathbf{q})=\texttt{sgn}\left(\langle\mathbf{n}_k, \mathbf{\tilde{p}}_k\rangle\right)\mathbf{n}_k, \vspace{-0.15cm}
	\end{equation}
	where $\mathbf{\tilde{p}}_k:=\mathbf{q}-\mathbf{p}_k$, $\langle\cdot, \cdot\rangle$ computes the inner product of two vectors, and $\texttt{sgn}(\cdot)$ extracts the sign of the input.
	Eq. \eqref{EQ:NORMAL:ORIENTATION} makes $\mathbf{\tilde{n}}_k$ align with $\mathbf{\tilde{p}}_k$, i.e., $\langle\mathbf{\tilde{n}}_k,\mathbf{\tilde{p}}_k \rangle>0$. 
	
\begin{figure}[t]
		\centering
		\subfloat[]{\includegraphics[width=0.2\textwidth]{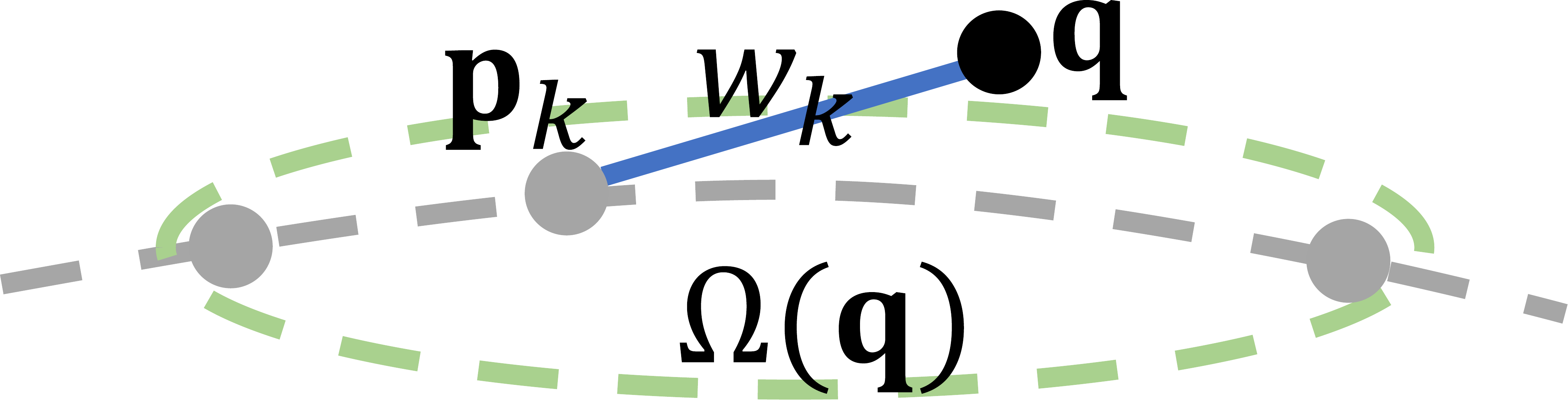}\label{UDF:DIST:ERROR:P2P}}  \quad
		\subfloat[]{\includegraphics[width=0.2\textwidth]{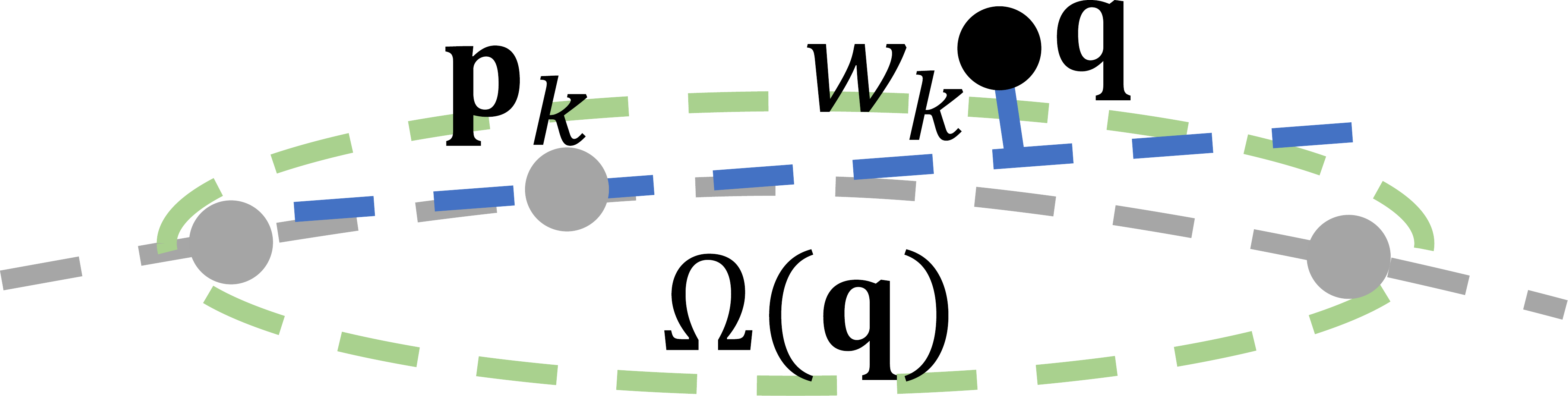}} \vspace{-0.3cm}
		\caption{
			Visual illustration of the difference between (\textbf{a}) P2P and (\textbf{b}) P2T distances when $\Omega (\mathbf{q})$ is sparse. We also refer readers to Table \ref{ABLATION:UDF} for the quantitative comparisons.}
		\label{UDF:DIST:ERROR}\vspace{-0.3cm}
	\end{figure}

	
	Denote by $\mathbf{p}^{*}$ the point on surface $\mathcal{S}$ closest to $\mathbf{q}$, 
	and $\mathbf{\tilde{n}}^{*}$ its 
	aligned normal vector.
	As $\Omega(\mathbf{q})$ generally covers a small region/area shown in Fig. \ref{UDF}, 
	we could simply adopt the weighted average of the Euclidean distances between $\mathbf{q}$ and each of $\{\mathbf{p}_k\}$, namely point-to-point (P2P) distance,
	to approximate $\mathbf{q}$'s UDF:\vspace{-0.2cm}
	\begin{equation}\label{EQ:UDF:DIST}
		\texttt{U}(\mathbf{q})\approx\sum_{\mathbf{p}_k\in\Omega (\mathbf{q})}w_1(\mathbf{q},\mathbf{p}_k)\cdot\|\mathbf{\tilde{p}}_k\|_2, 
		\vspace{-0.2cm}
	\end{equation}
	where the weights $\{w_1(\mathbf{q},\mathbf{p}_k)\}_{k=1}^K$ are non-negative and satisfying $\sum_{k=1}^Kw_1(\mathbf{q},\mathbf{p}_k)=1$. 
	However, despite upsampling, $\Omega(\mathbf{q})$ may still  be sparse, and the distance of $\mathbf{q}$ to each $\mathbf{p}_k$ is much larger than its unsigned distance  value, as shown in Fig. \ref{UDF:DIST:ERROR:P2P}, resulting in a large approximation error. 
	Instead, we propose to approximate $\texttt{U}(\mathbf{q})$ through the weighted average of the Euclidean distances from $\mathbf{q}$ to the tangent planes of each of $\{\mathbf{p}_k\}$, namely 
	point-to-tangent plane (P2T) distance
	:\vspace{-0.2cm}
	\begin{equation}\label{EQ:UDF}
		\texttt{U}(\mathbf{q})\approx\Phi(\mathbf{q}):=\sum_{\mathbf{p}_k\in\Omega(\mathbf{q})}w_1(\mathbf{q},\mathbf{p}_k)\cdot \langle \mathbf{\tilde{n}}_k, \mathbf{\tilde{p}}_k \rangle.
		\vspace{-0.2cm}
	\end{equation}
	
	\begin{figure}[t]
		\centering
		\includegraphics[width=0.47\textwidth]{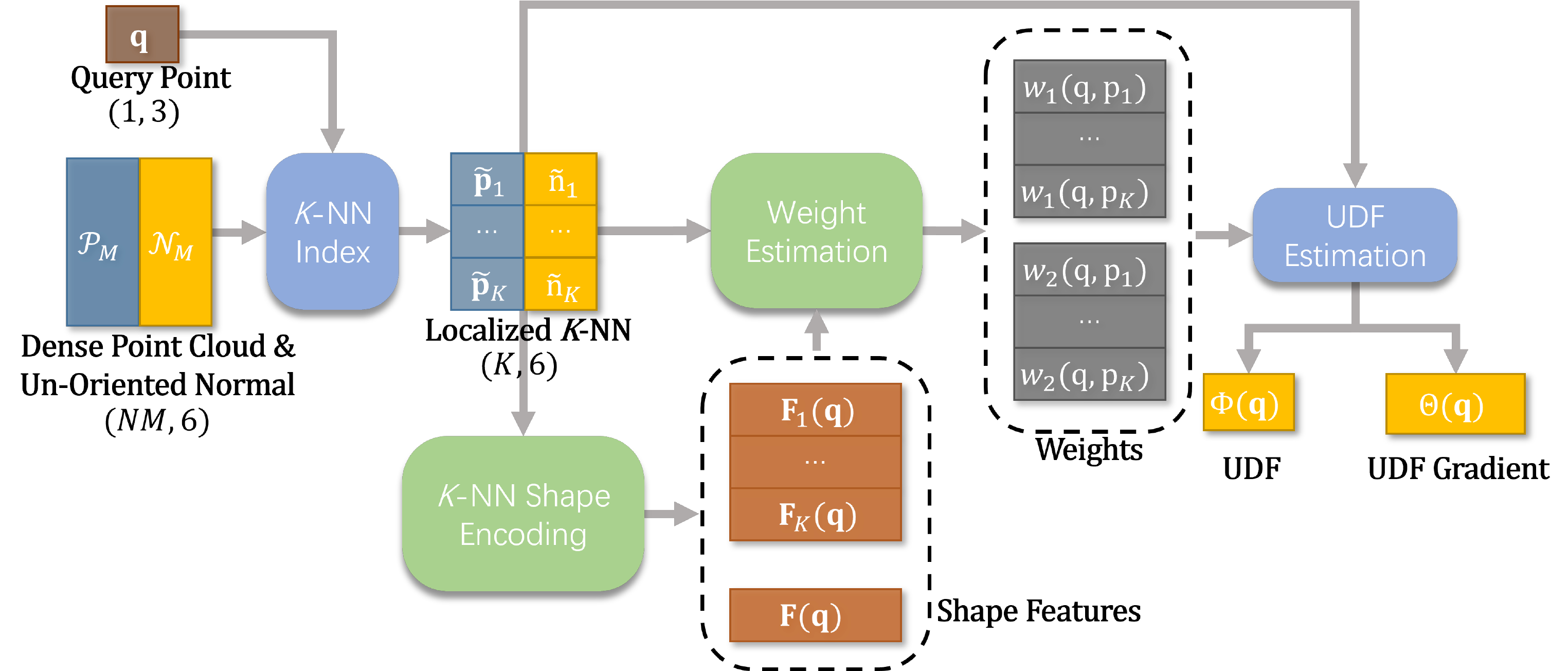}\vspace{-0.3cm}
		\caption{Flowchart of the GUE module. We refer readers to the \textit{supplementary material} for the detailed network architecture.}
		\label{REC:ARC}
		\vspace{-0.4cm}
	\end{figure}
	
	
	Similarly, we propose to approximate the gradient of $\texttt{U}(\mathbf{q})$,
	denoted as $\nabla\texttt{U}(\mathbf{q})$,  
	using the continuity property of the normal vectors of a surface and the fact $\nabla\texttt{U}(\mathbf{q})=\mathbf{\tilde{n}}^{*}$.  See Fig. \ref{UDF_GRAD}.  We use $\mathbf{\tilde{n}}_k$,  each of which is around $\mathbf{\tilde{n}}^{*}$, to approximate
	$\nabla\texttt{U}(\mathbf{q})$ as \vspace{-0.15cm}
	\begin{equation}  \label{EQ:UDF:GRAD}
		\nabla\texttt{U}(\mathbf{q})\approx\Theta(\mathbf{q}):=\frac{\sum_{\mathbf{p}_k\in\Omega(\mathbf{q})}w_2(\mathbf{q},\mathbf{p}_k)\cdot\tilde{\mathbf{n}}_k}{\|\sum_{\mathbf{p}_k\in\Omega(\mathbf{q})}w_2(\mathbf{q},\mathbf{p}_k)\cdot\tilde{\mathbf{n}}_k\|},\vspace{-0.15cm}
	\end{equation}
	where 
	the weights $\{w_2(\mathbf{q}, \mathbf{p}_k)\}_{k=1}^K$ are non-negative and satisfy  $\sum_{k=1}^{K}w_2(\mathbf{q},\mathbf{p}_k)=1$. 
	
	\noindent\textbf{Learning the weights}. Based on the above formulation, the problem of UDF estimation is boiled down to obtaining  $\{w_1(\mathbf{q}, \mathbf{p}_k),~ w_2(\mathbf{q}, \mathbf{p}_k)\}_{k=1}^K$. As illustrated in Fig. \ref{REC:ARC}, 
	we construct a sub-network to learn them adaptively.
	Intuitively, the values of the weights should be relevant to 
	the relative position between $\mathbf{q}$ and $\mathbf{p}_k$  and the overall shape of $\Omega(\mathbf{q})$. Thus, we embed $\mathbf{\tilde{p}}_k$ and $ \mathbf{\tilde{n}}_k$ via an MLP to obtain point-wise features for all neighbors, {$\mathbf{F}_k(\mathbf{q})$}, which are then maxpooled, leading to a global feature, {$\mathbf{F}(\mathbf{q})$}, to encode the overall shape of $\Omega(\mathbf{q})$.  Finally, we concatenate the above-mentioned features and feed them into two separated MLPs followed with the Softmax layer to regress the two sets of weights. 

	\noindent \textbf{\textit{Remark}}. 
	Although $\Theta(\mathbf{q})$ could be calculated through the back-propagation (BP) of the neural network (the NDF method \cite{NDF} adopts this manner), such a process is time-consuming. 
	Compared with BP-based UDF gradient estimation, our method is more efficient. Besides, owing to the \textit{separate}
	estimation processes, 
	the UDF error would not be transferred to its gradient, 
	making it more accurate. To the best of our knowledge, this is the \textit{first} to decouple the estimation processes of UDF and its gradient. 
	See the experimental comparison in Sec. \ref{SEC:ABLATION}.

	\subsection{Edge-based Marching Cube}
	Inspired by the classic marching cube algorithm \cite{MARCHINGCUBE} that extracts triangles in a cube according to the occupancy of its eight vertices, we propose \textit{edge}-based marching cube (E-MC) to extract triangle meshes from the predicted unsigned distance field in the preceding module. 
	Generally, we first detect whether the connection between any pair of vertices of a cube intersects with the surface, then find the most matched condition with the detection results from the lookup table of Marching Cube.

	\begin{figure}[tbp]
		\centering
		\subfloat[]{\includegraphics[width=0.12\textwidth]{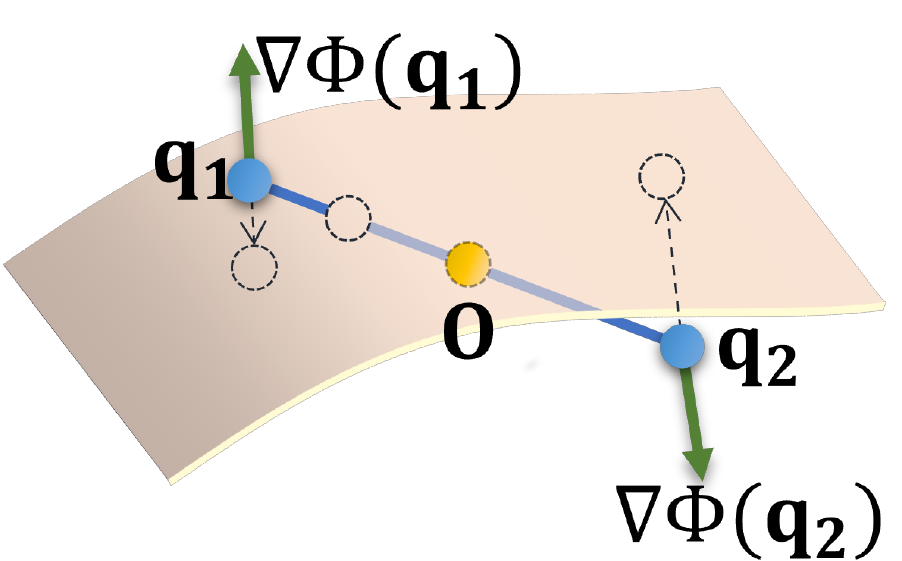}} 
		\subfloat[]{\includegraphics[width=0.12\textwidth]{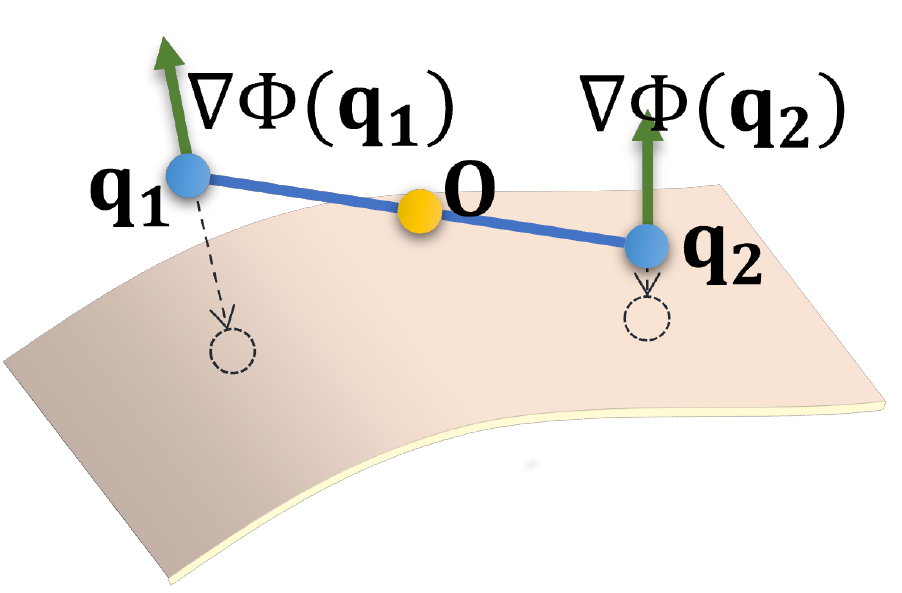}} 
		\subfloat[]{\label{EDGE:DETECT3}\includegraphics[width=0.12\textwidth]{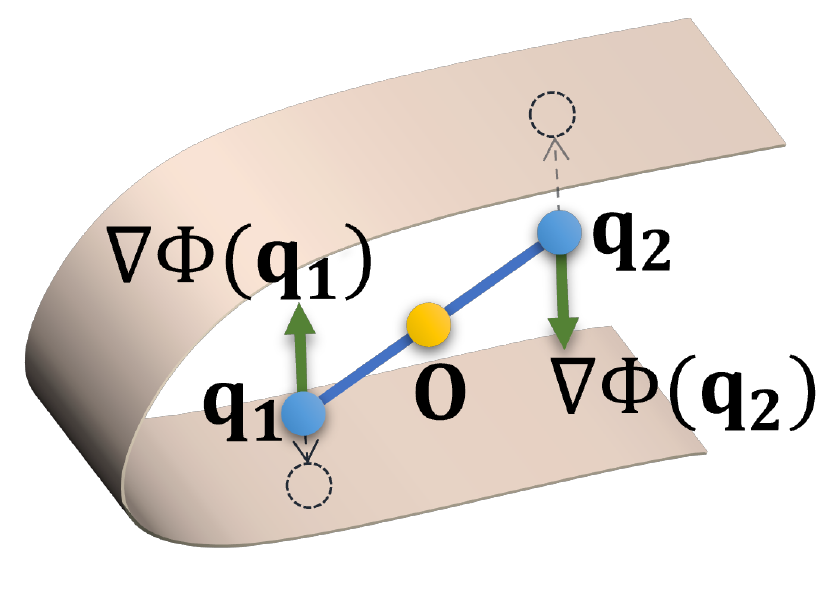}} 
		\subfloat[]{\label{EDGE:DETECT4}\includegraphics[width=0.12\textwidth]{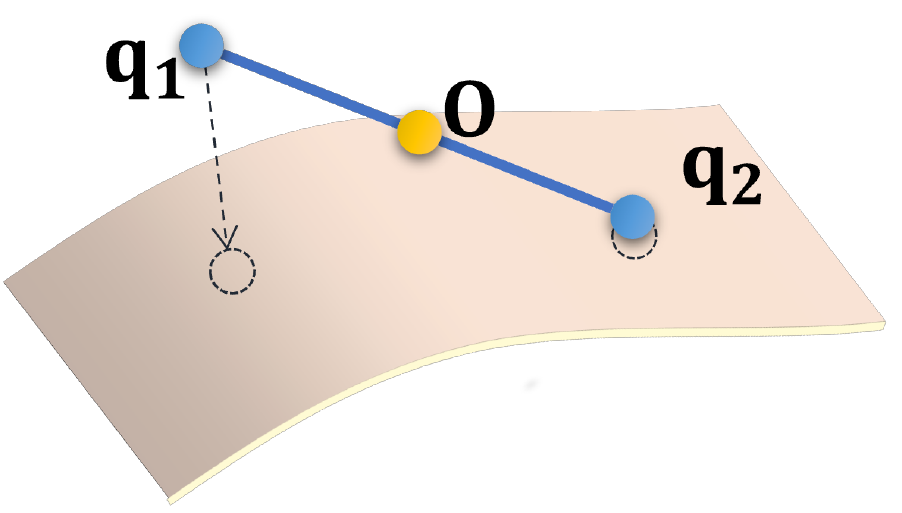}} 
		\vspace{-0.2cm}
		\caption{The position relationship between the surface and edge. (a) The surface interacts with the edge. (b)(c) The surface does not interact with the edge. (d) One vertex of the edge is extremely close to the surface. See the \textit{supplementary material} for comprehensive examination of different cases.}
		\label{EDGE:DETEC}
		\vspace{-0.5cm}
	\end{figure}

	\noindent\textbf{Edge intersection detection}. As illustrated in Fig. \ref{EDGE:DETEC}, if the surface interacts with the connection between
	vertices $\mathbf{q}_1$ and $\mathbf{q}_2$, denoted as $\mathbf{q}_1\mathbf{q}_2$, 
	the following constraints must be satisfied: \vspace{-0.3cm}
	\begin{align}\label{EDGE:DETECTION:CONDITION}
		\centering
		&\langle\Theta(\mathbf{q}_1), \Theta(\mathbf{q}_2)\rangle<0, \nonumber\\
		&\langle\Theta(\mathbf{q}_1), \overrightarrow{\mathbf{o}\mathbf{q}_2}\rangle>0, ~~\langle\Theta(\mathbf{q}_2), \overrightarrow{\mathbf{o}\mathbf{q}_1}\rangle>0,\vspace{-0.3cm}
	\end{align}
	where $\mathbf{o}$ is the midpoint between $\mathbf{q}_1$ and $\mathbf{q}_2$.
	Specifically, the first constraint indicates the directions of the UDF gradients of $\mathbf{q}_1$ and $\mathbf{q}_2$ must be opposite, i.e., the angle between them is larger than $90^\circ$. 
	The last two constraints ensure that $\mathbf{q}_1$ and $\mathbf{q}_2$ are located at different sides of an identical part of a surface to eliminate the case shown in Fig. \ref{EDGE:DETECT3}. Once the above three constraints are satisfied, the intersection of the surface with $\mathbf{q}_1\mathbf{q}_2$,
	denoted as $\mathbf{v}(\mathbf{q}_1,\mathbf{q}_2)$, can be calculated via \vspace{-0.2cm}
	\begin{equation} \label{INTER:VERT}
		\mathbf{v}(\mathbf{q}_1,\mathbf{q}_2)=\frac{\mathbf{q}_2\Phi(\mathbf{q}_1)+\mathbf{q}_1\Phi(\mathbf{q}_2)}{\Phi(\mathbf{q}_1)+\Phi(\mathbf{q}_2)}.\vspace{-0.15cm}
	\end{equation}
	
	In addition, if at least one of $\Phi(\mathbf{q}_1)$ and $\Phi(\mathbf{q}_2)$ is less than a small threshold $\tau$, their connection is determined to interact with the surface, and the vertex with the smaller UDF is the intersection.
	
	\noindent \textbf{Triangle extraction}. After utilizing the edge intersection detection on the connections between any two vertices of a cube, we find the most similar condition to the detection results in the lookup table of Marching Cube to extract the triangles. We refer the readers to the \textit{supplementary material} for more details.

	\noindent \textbf{\textit{Remark}}. With reference to a typical vertex of the 3D cube, MeshUDF \cite{MESHUDF} converts UDFs to SDFs in a small cube by using only the first constraint of Eq. \eqref{EDGE:DETECTION:CONDITION}. GIFS \cite{GIFS} utilizes a sub-network to determine whether 
	$\mathbf{q}_1\mathbf{q}_2$ intersects the surface,
	which is not as general as ours. In Sec. \ref{SEC:ABLATION}, we experimentally demonstrate the advantage of our E-MC over MeshUDF.  

	\subsection{Loss Function}
	We design a joint loss function to train GeoUDF in an end-to-end manner. 
	\textbf{(1)} \textit{Loss for upsampling} $\mathcal{L}_\text{PU}$. We compute the Chamfer Distance (CD) between the upsampled point cloud $\mathcal{P}_M$ and its ground-truth dense point cloud to drive the learning of the LGR module. 
	\textit{Note that we do not supervise the normal vectors}.
	\textbf{(2)} \textit{Losses for UDF and its gradient learning} $\mathcal{L}_{\rm UDF}$ and $\mathcal{L}_{\rm Grad}$. 
	We use the mean absolute error between predicted
	and ground-truth UDFs
	and the cosine distance between predicted
	and ground-truth gradients
	to supervise the learning of the GUE module.
	The overall loss function is finally written as \vspace{-0.2cm}
	\begin{equation}
		\mathcal{L}=\lambda_1\mathcal{L}_{\rm PU}+\lambda_2\mathcal{L}_{\rm UDF}+\lambda_3\mathcal{L}_{\rm Grad}, \vspace{-0.15cm}
	\end{equation}
	where $\lambda_1$, $\lambda_2$, and $\lambda_3$ are hyper-parameters for balancing the three terms.

	\section{Experiments}
	
	\noindent\textbf{Implementation details}.
	During training, 
	we set the upsampling factor $M=16$.
	For each shape, we sampled 2048 query points near the surface in one training iteration. We set the size of neighbourhood 
	$K=10$.
	We trained our framework in two stages: 
	first, we only trained LGR, i.e., $\lambda_1=100$ and $\lambda_2=\lambda_3=0$, with the learning rate $10^{-3}$ for 100 epochs; second, we trained the whole network with $\lambda_1=100$, $\lambda_2=1$, and $\lambda_3=0.1$ for 300 epochs with the learning rate $10^{-4}$. We conducted all experiments 
	on an NVIDIA RTX 3090 GPU with Intel(R) Xeon(R) CPU.

	\begin{table*}[tbp] 
		\centering
		\caption{{\small Quantitative comparison of different methods on the 13 classes of the ShapeNet dataset. Note that the MC of resolution 128 was applied to all methods for surface reconstruction, except POCO$^{\dag}$ and GIFS$^{\dag}$, where the MC resolutions were set to 256 and 160, respectively, to keep the same as their original papers. 
			}
		}\vspace{-0.3cm}
		\resizebox{.7\textwidth}{!}{
			\setlength{\tabcolsep}{3.5mm}{
				\begin{tabular}{l|c c|c c|c c|c c}
					\toprule[1.2pt]
					\multirow{3}{0.11\textwidth}{Method}  & \multicolumn{4}{c|}{Clean}&  \multicolumn{4}{c}{Noisy (0.005)} \\
					\cline{2-9}
					& \multicolumn{2}{c|}{CD $(\times 10^{-2})$ $\downarrow$ } &  \multicolumn{2}{c|}{F-Score $\uparrow$} & \multicolumn{2}{c|}{CD $(\times 10^{-2})$ $\downarrow$ } &  \multicolumn{2}{c}{F-Score $\uparrow$}\\
					\cline{2-9}
					& Mean & Median & $\text{F1}^{0.5\%}$ & $\text{F1}^{1\%}$ & Mean & Median & $\text{F1}^{0.5\%}$ & $\text{F1}^{1\%}$\\
					\hline
					ONet \cite{OCCNET}  & {$0.894$} & {$0.716$} & {$0.535$} & {$0.756$} & $0.875$ & $0.764$& $0.642$& $0.785$\\
					CONet \cite{CONVOCCNET}  & {$0.549$} & {$0.489$} & {$0.624$} & {$0.910$} & $0.454$ & $0.444$& $0.766$& $0.942$\\
					SAP \cite{SAP} &  $0.411$ & $0.370$ & $0.841$ & $0.953$ & $0.380$ & $0.358$& $0.821$& $0.959$\\
					POCO \cite{POCO}            & 0.321 & 0.283 & 0.913 & 0.971 & 0.369 & 0.337 & 0.863 & 0.965 \\
					POCO$^{\dag}$ \cite{POCO}   & 0.290 & 0.221 & 0.921 & 0.985 & 0.304 & 0.280 & 0.875 & 0.984 \\
					{DOG \cite{DOG}} &$0.429$ & $0.353$ &$0.792$ &$0.956$ & $0.381$ & $0.337$ & $0.814$ & $0.967$  \\
					\hline
					NDF \cite{NDF}   & $0.341$   & $0.320$    & $0.840$    & $0.976$    & $0.431$   & $0.419$   & $0.685$   & $0.961$  \\
					GIFS \cite{GIFS}  & $0.328$ & $0.276$ & $0.860$ & $0.974$ & $0.418$ & $0.358$ & $0.731$ & $0.958$ \\ 
					GIFS$^{\dag}$ \cite{GIFS} & $0.281$ & $0.243$ & $0.914$ & $0.985$ & $0.376$ & $0.348$ & $0.780$ & $0.968$ \\ 
					Ours  &$\pmb{0.234}$ & $\pmb{0.226}$ & $\pmb{0.938}$ & $\pmb{0.992}$ & $\pmb{0.289}$ & $\pmb{0.278}$& $\pmb{0.893}$& $\pmb{0.987}$\\
					\bottomrule[1.2pt]
				\end{tabular}
		}}\vspace{-0.3cm}
		\label{SHAPENET:TAB}
	\end{table*} \vspace{-0.3cm}
	
	\begin{figure*}[t] \small
		\centering
		{
			\begin{tikzpicture}[]
				\node[] (a) at (0,0.6) {\includegraphics[width=0.09\textwidth]{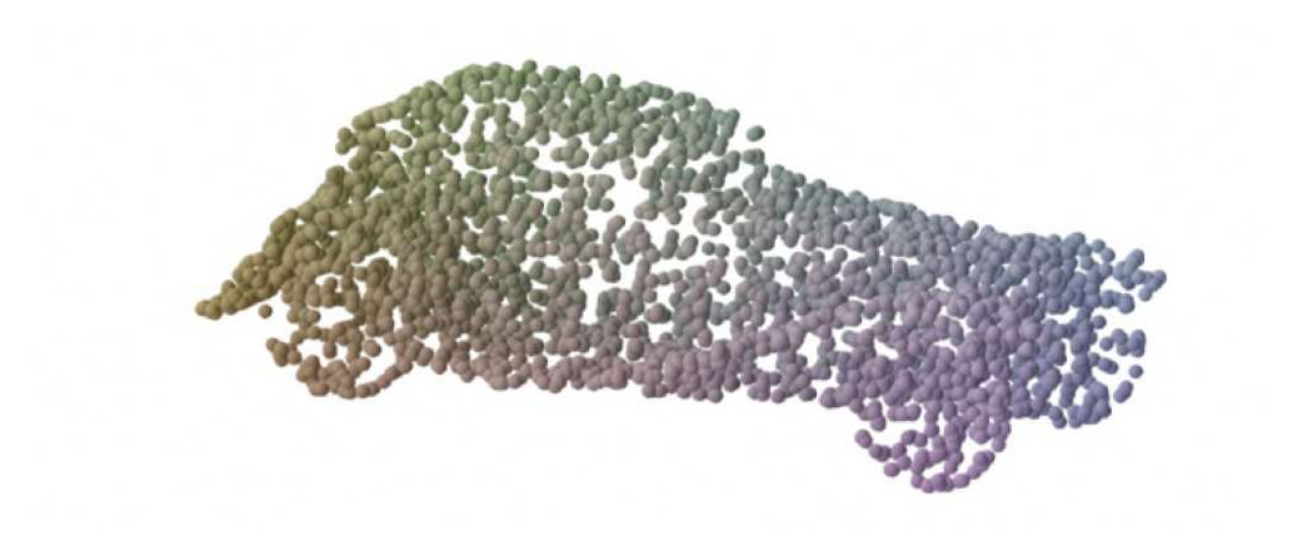} };
				\node[] (a) at (0,1.4) {\includegraphics[width=0.09\textwidth]{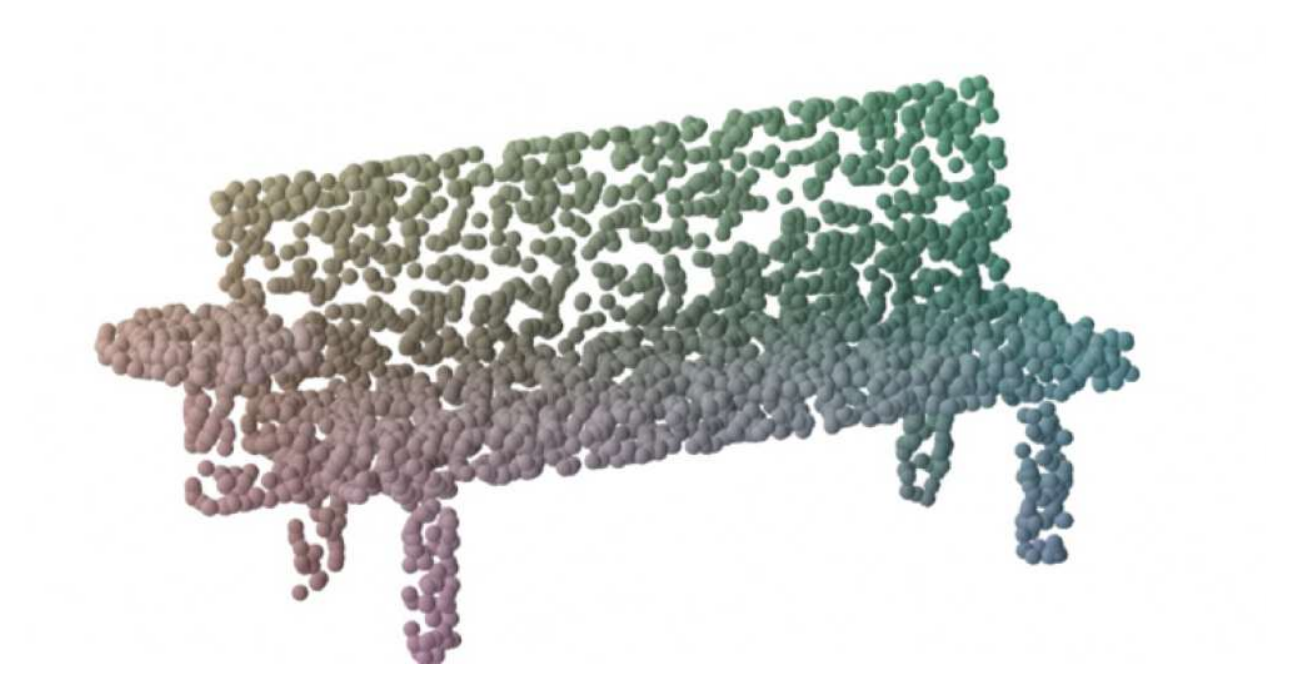} };
				\node[] (a) at (0,2.6) {\includegraphics[width=0.09\textwidth]{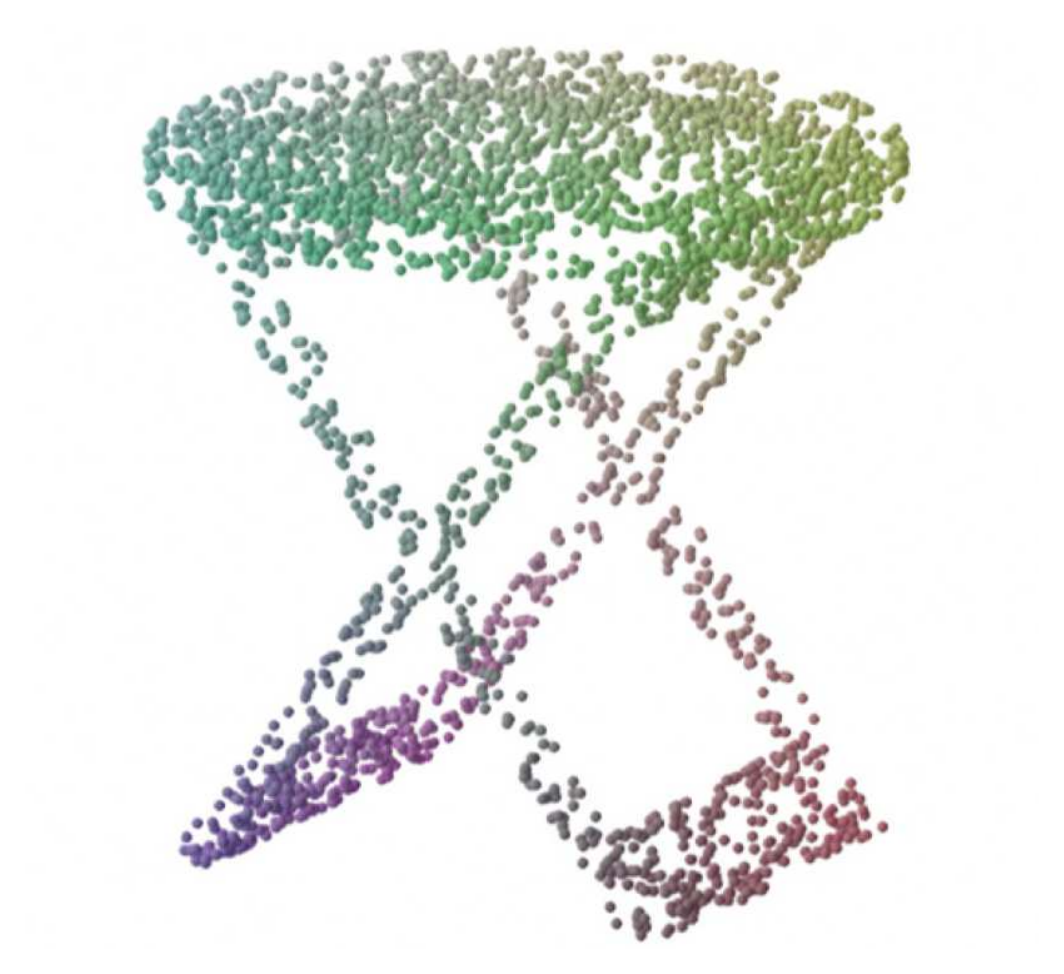} };
				\node[] (a) at (0,4) {\includegraphics[width=0.09\textwidth]{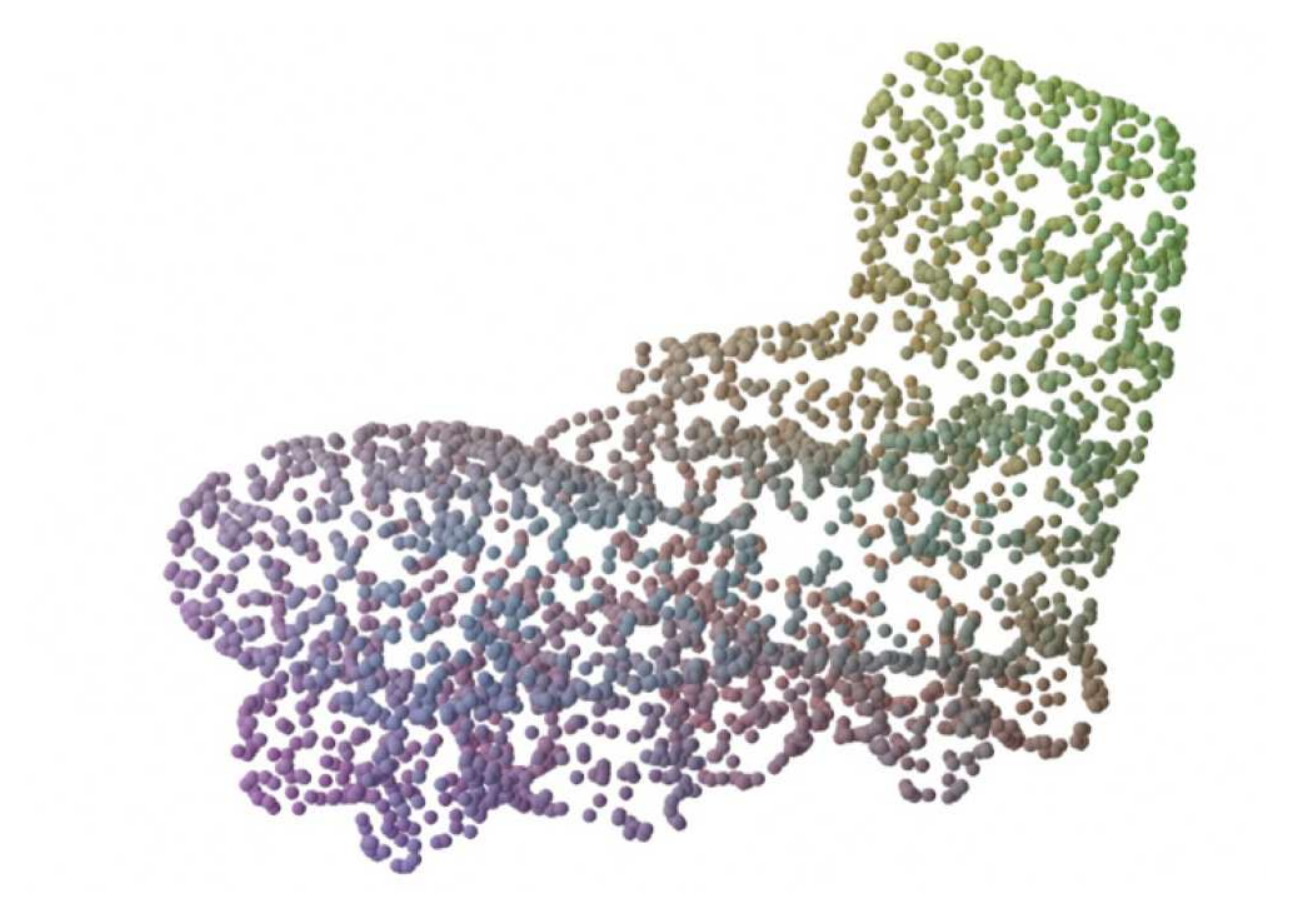} };
				\node[] (a) at (0,5.4) {\includegraphics[width=0.09\textwidth]{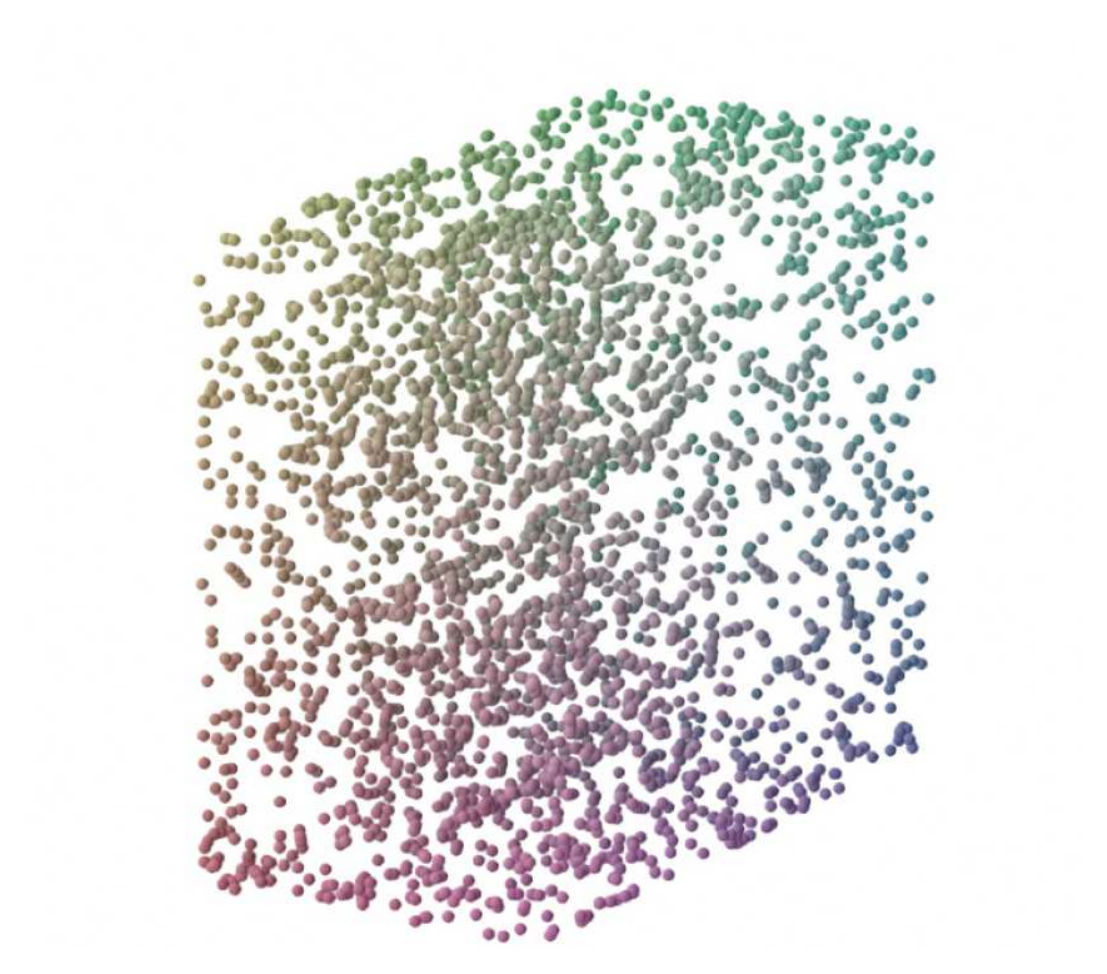} };
				\node[] (a) at (0,0) {\small (a) Input};

				\node[] (b) at (15/8,0.6) {\includegraphics[width=0.09\textwidth]{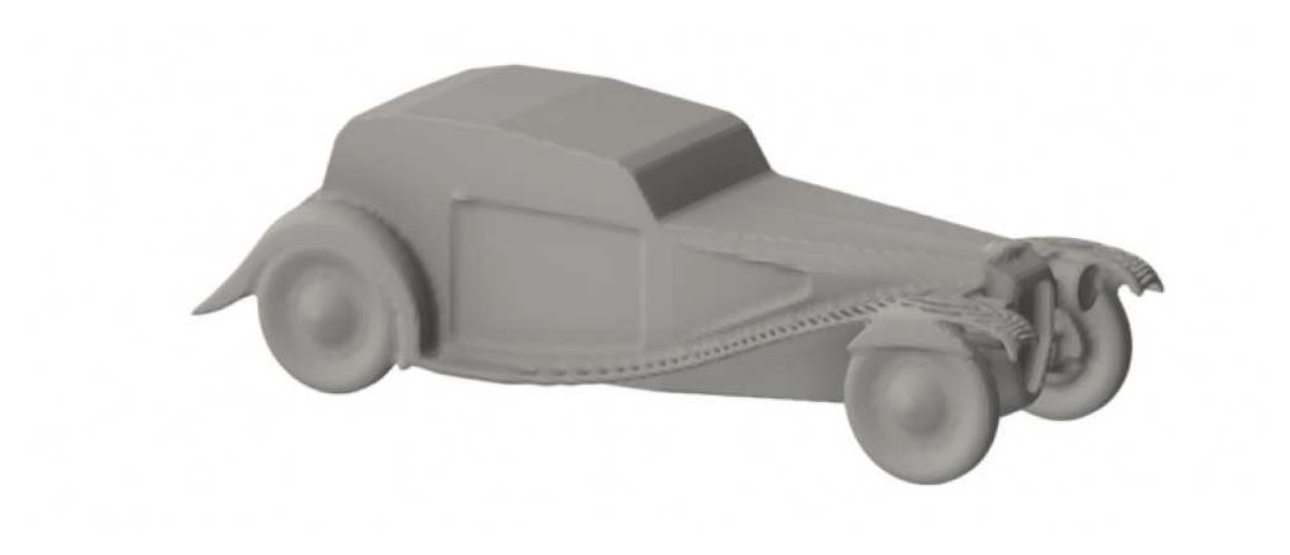}};
				\node[] (b) at (15/8,1.4) {\includegraphics[width=0.09\textwidth]{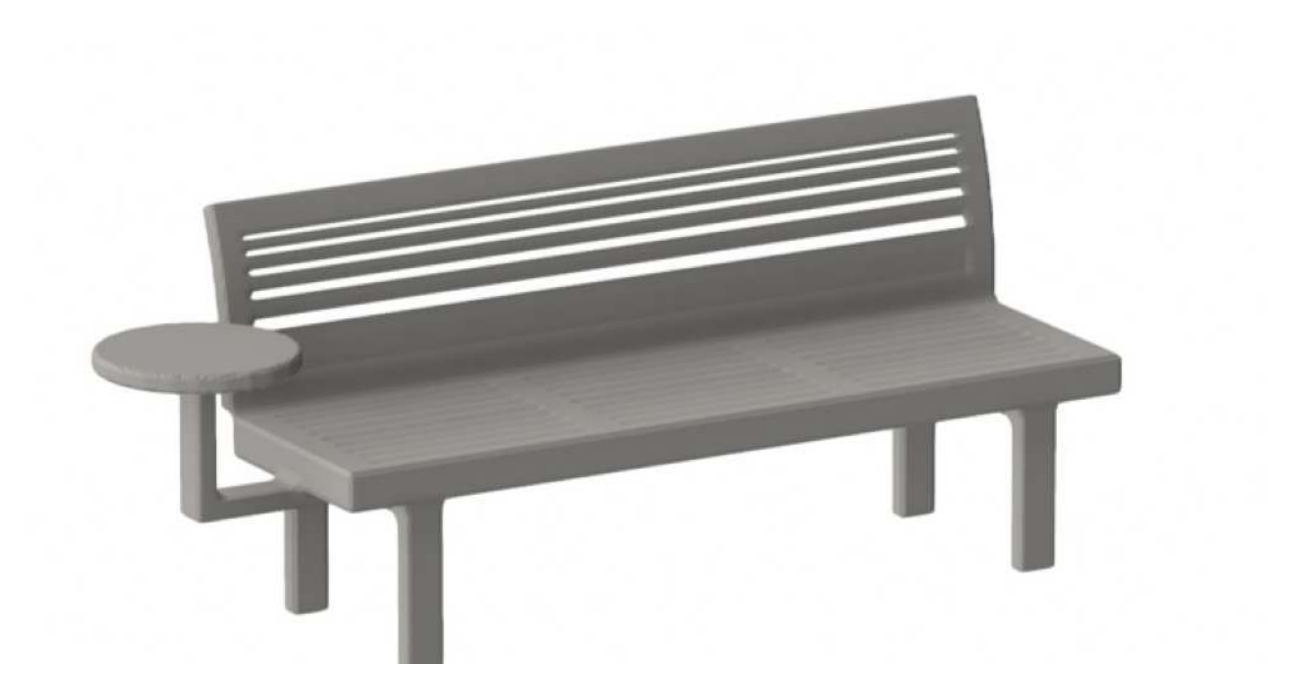}};
				\node[] (b) at (15/8,2.6) {\includegraphics[width=0.09\textwidth]{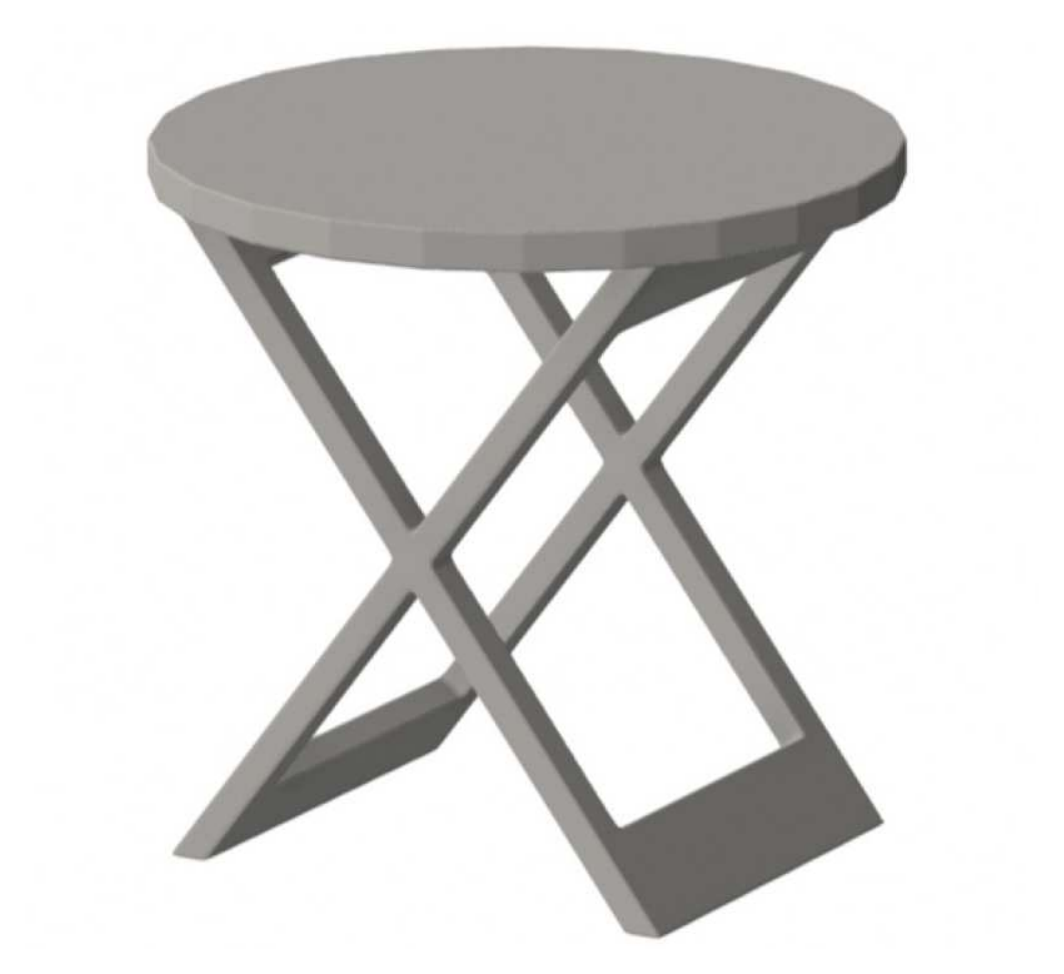}};
				\node[] (b) at (15/8,4) {\includegraphics[width=0.09\textwidth]{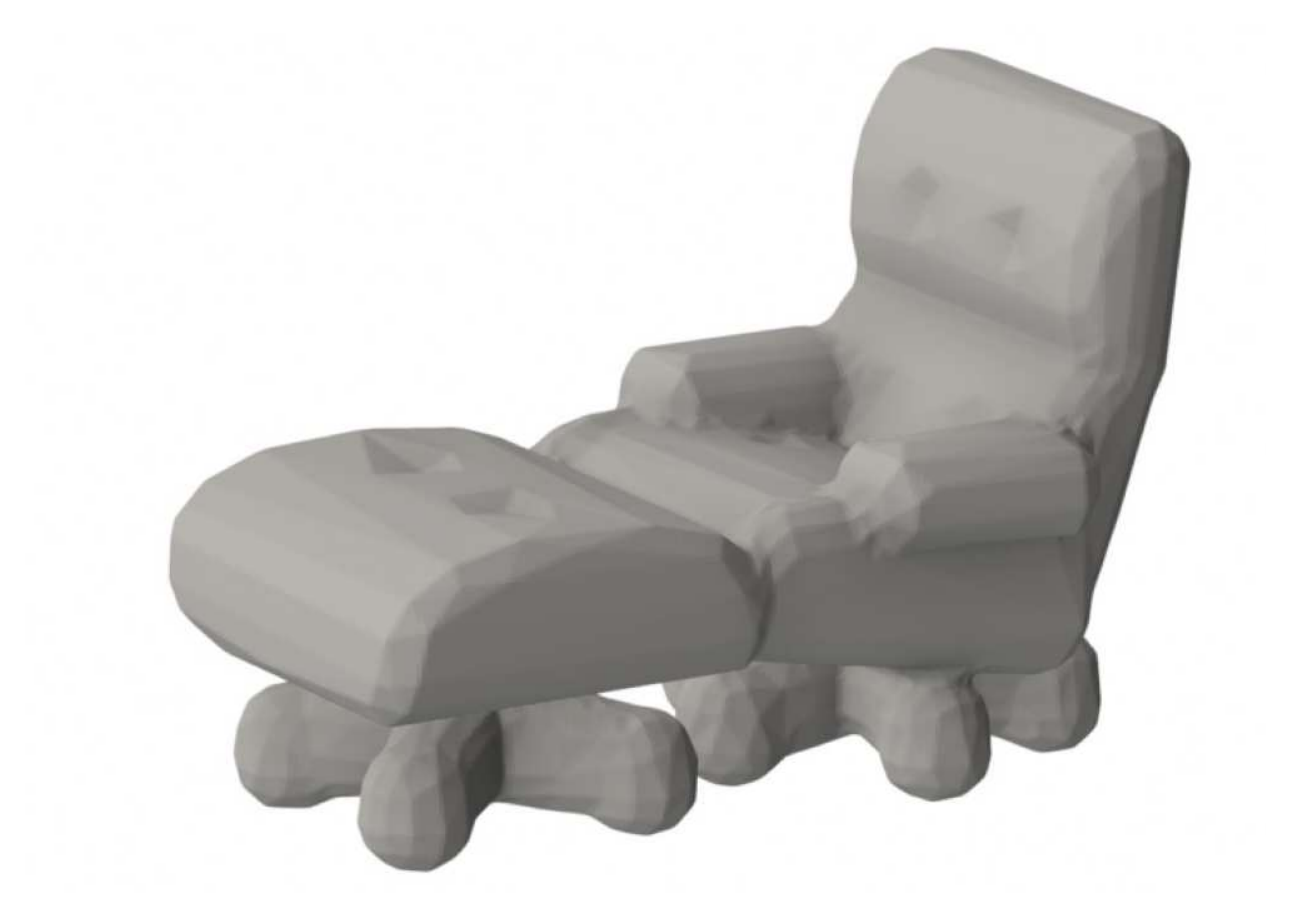}};
				\node[] (b) at (15/8,5.4) {\includegraphics[width=0.09\textwidth]{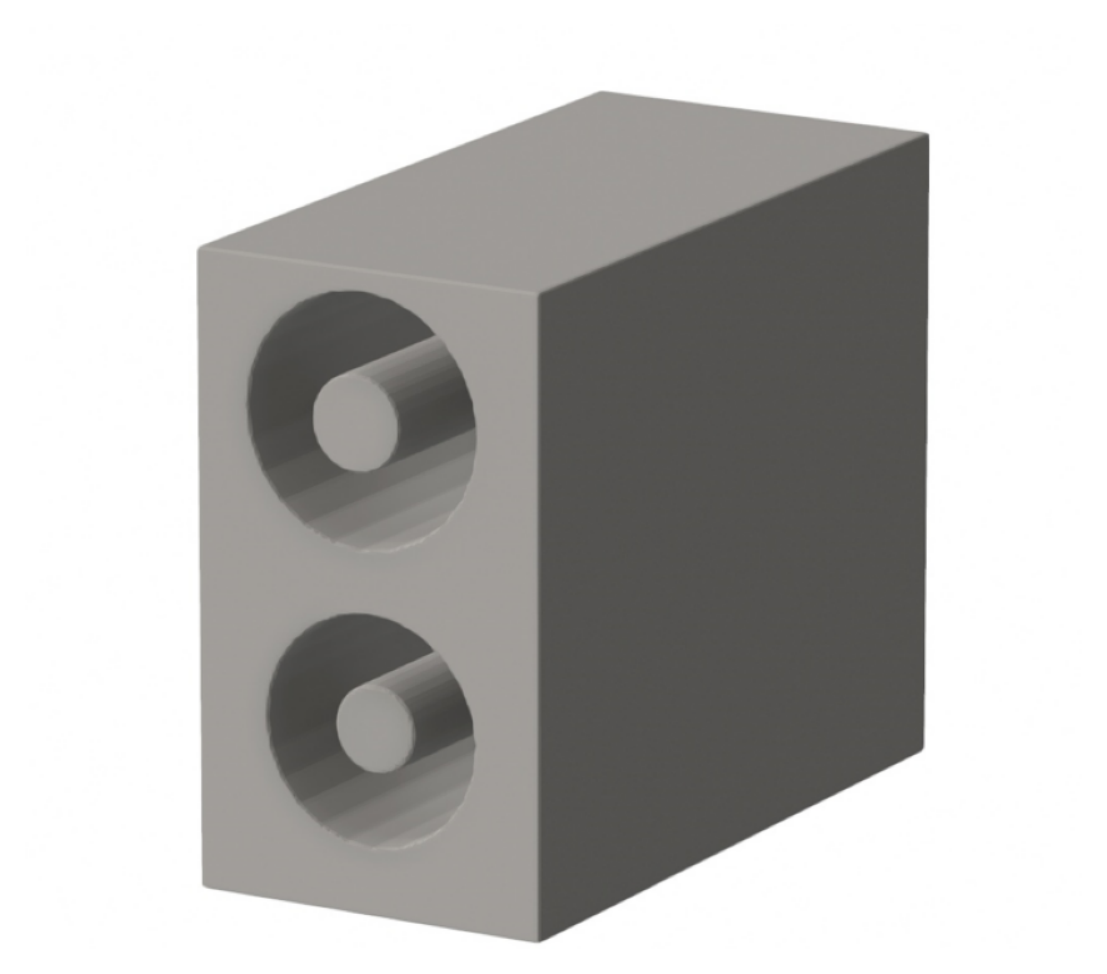}};
				\node[] (b) at (15/8,0) {\small (b) GT };
				
				\node[] (c) at (15/8*2,0.6) {\includegraphics[width=0.09\textwidth]{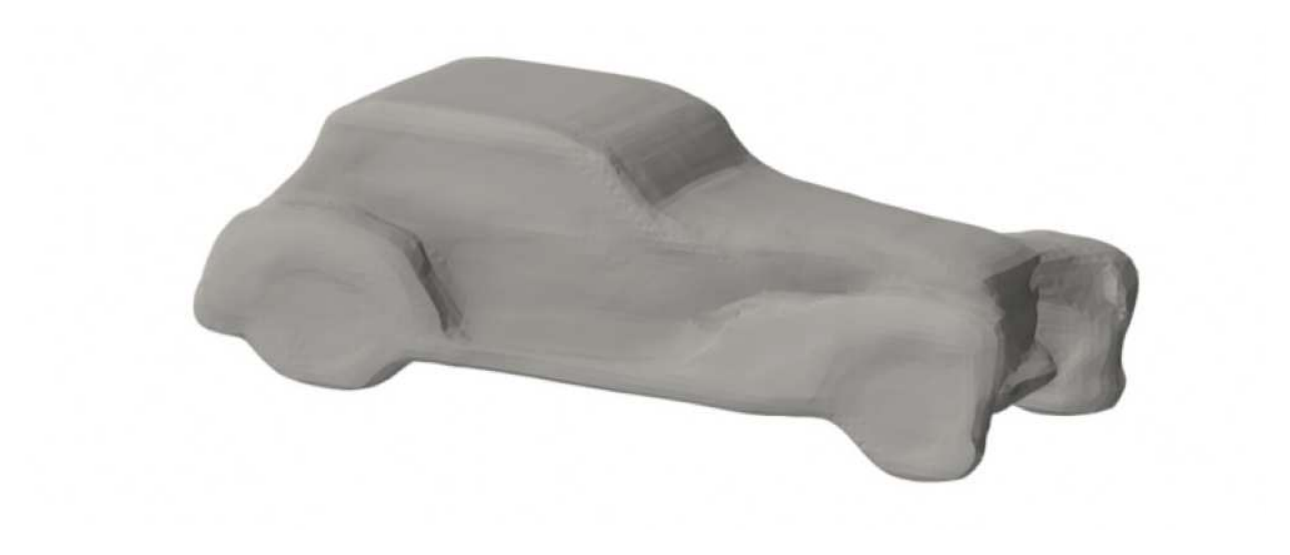}};
				\node[] (c) at (15/8*2,1.4) {\includegraphics[width=0.09\textwidth]{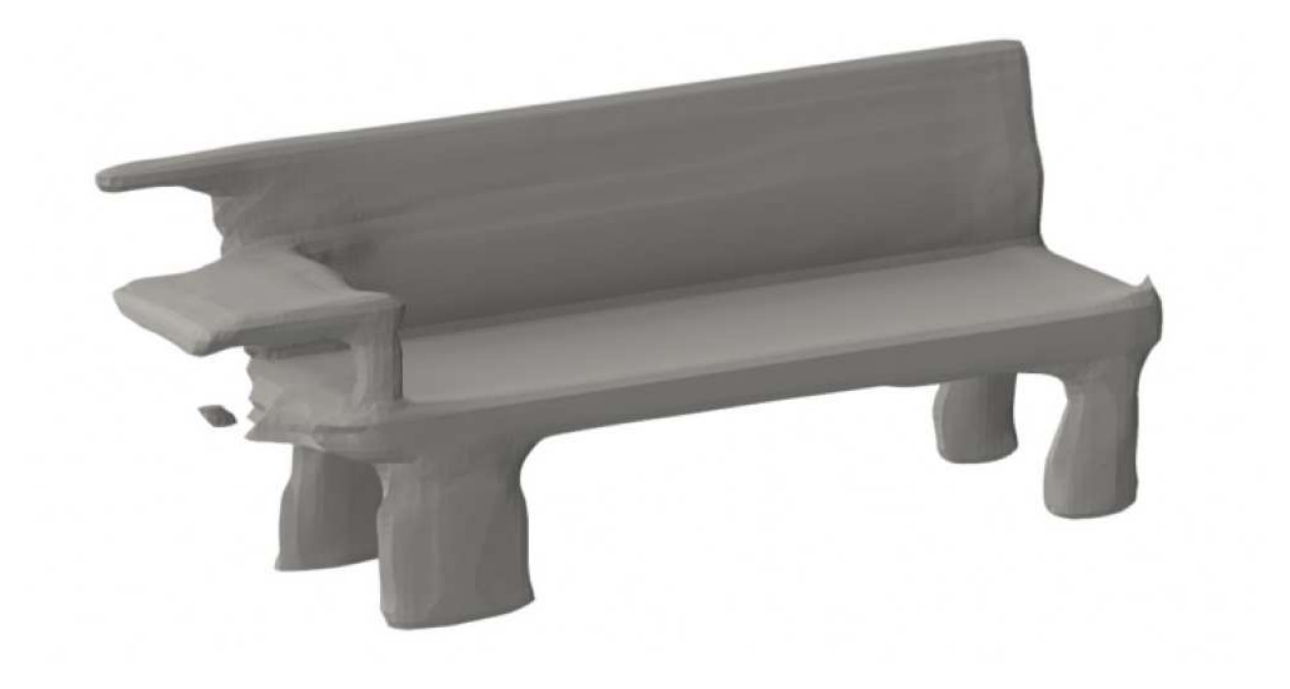}};
				\node[] (c) at (15/8*2,2.6) {\includegraphics[width=0.09\textwidth]{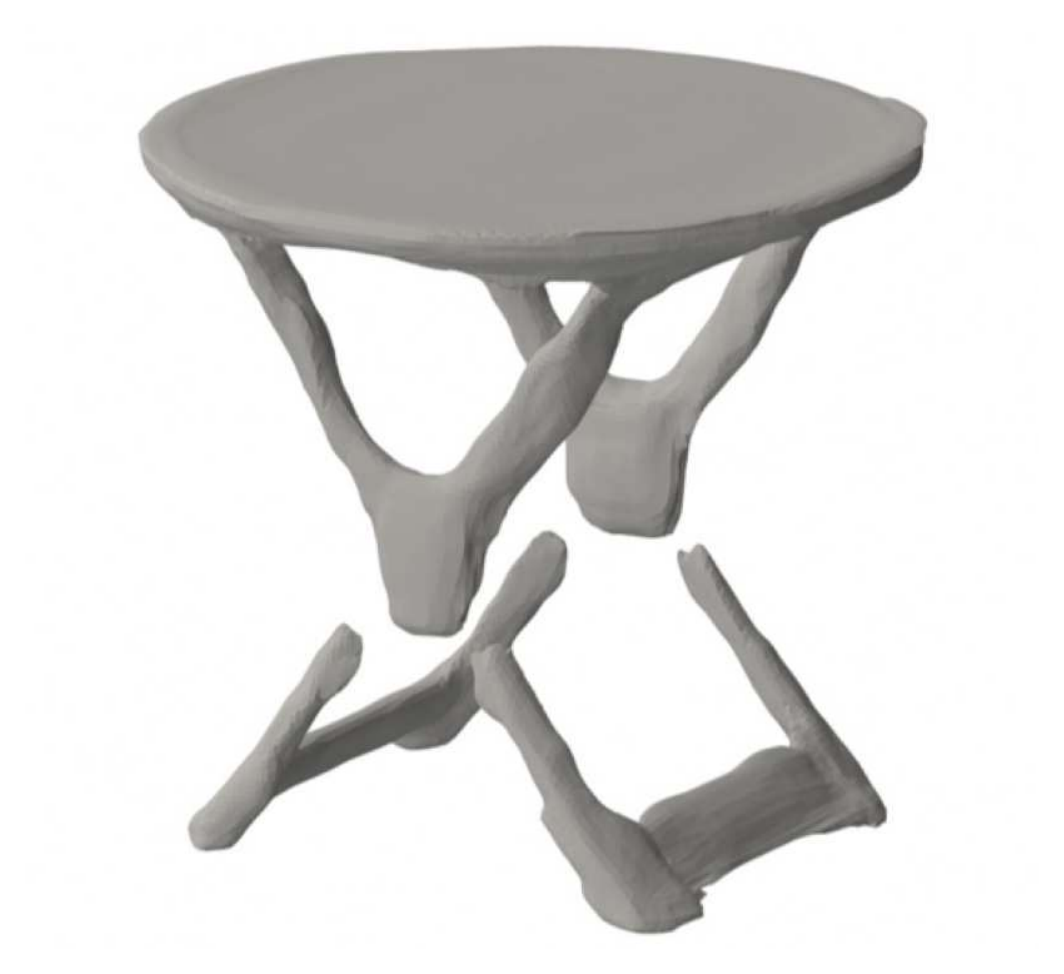} };
				\node[] (c) at (15/8*2,4) {\includegraphics[width=0.09\textwidth]{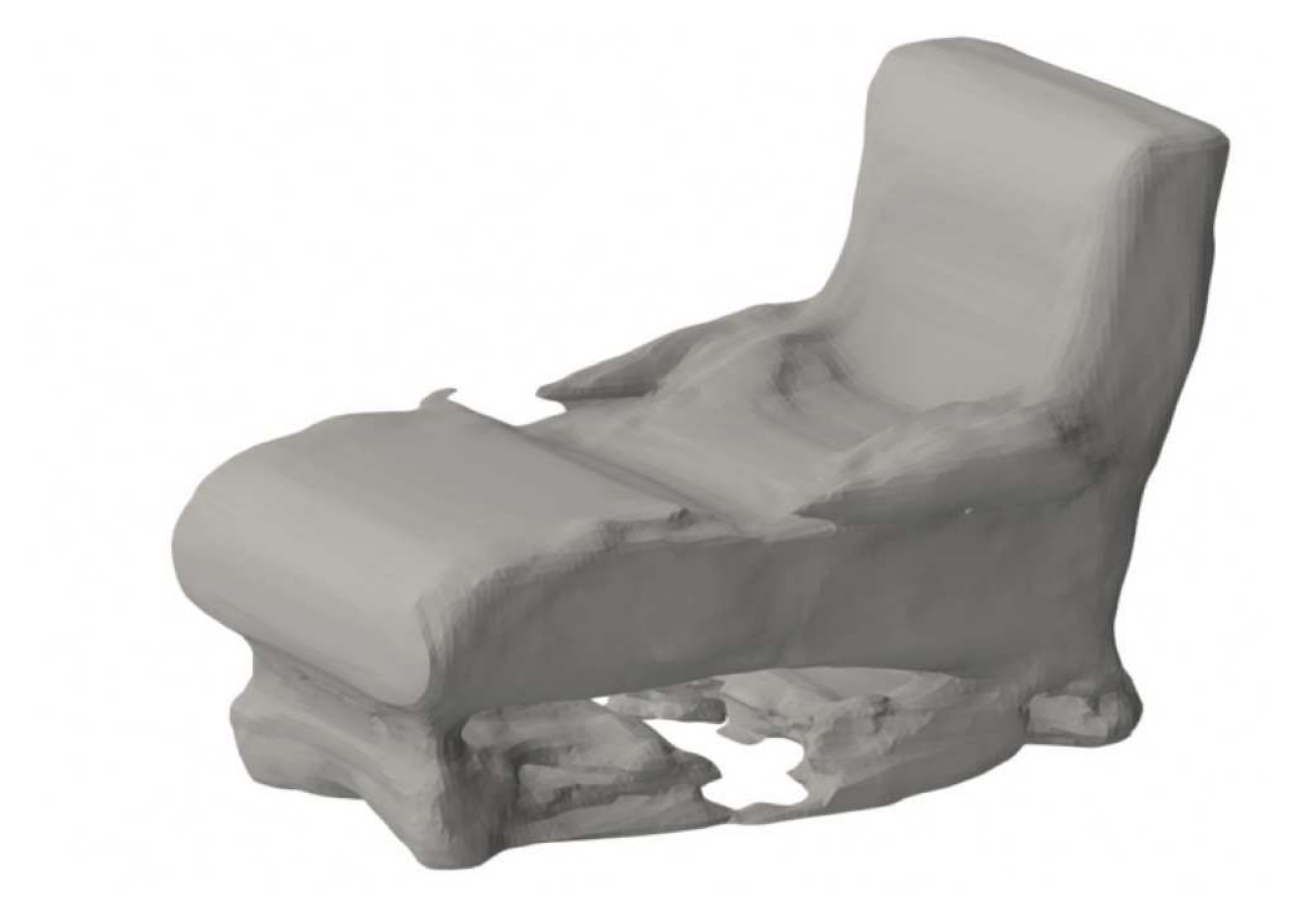} };
				\node[] (c) at (15/8*2,5.4) {\includegraphics[width=0.09\textwidth]{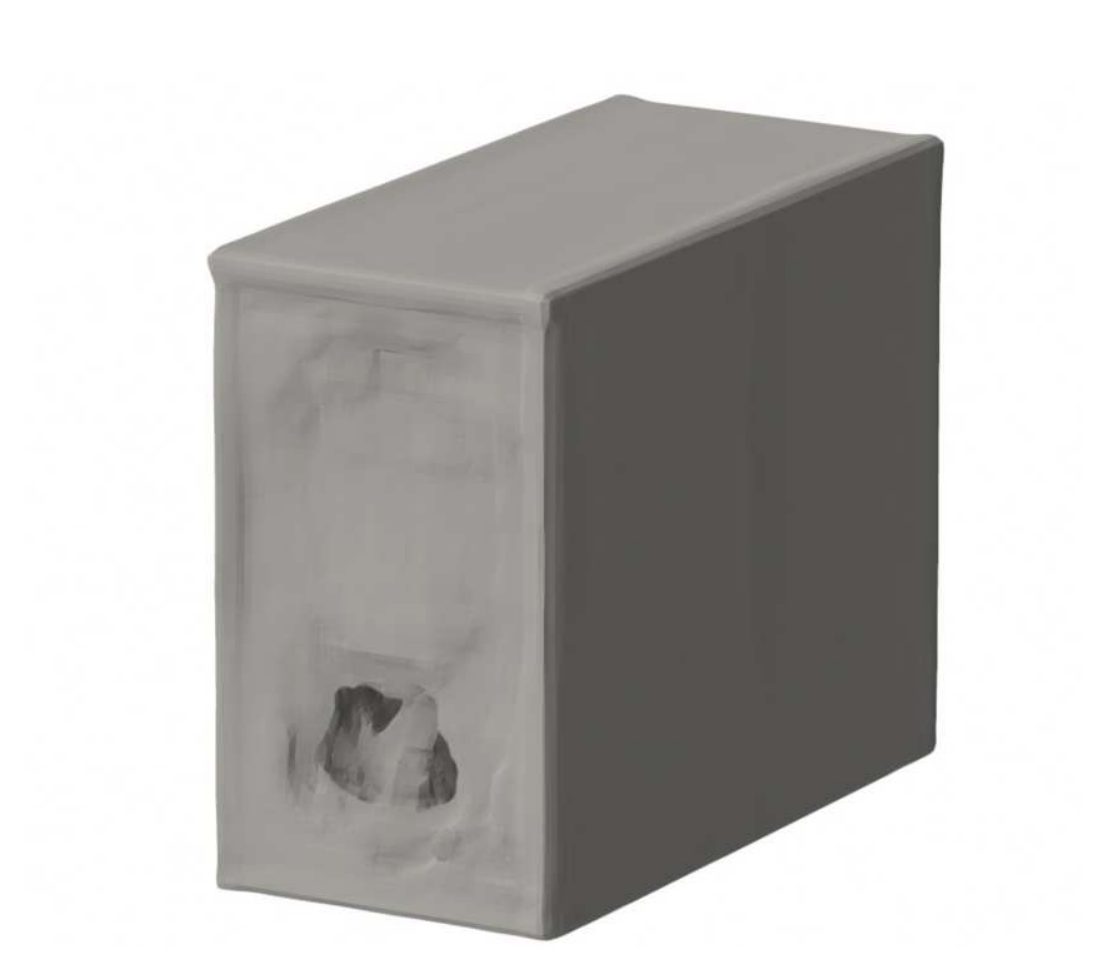} };
				\node[] (c) at (15/8*2,0) {\small (c) ONet \cite{OCCNET} };
				
				\node[] (d) at (15/8*3,0.6) {\includegraphics[width=0.09\textwidth]{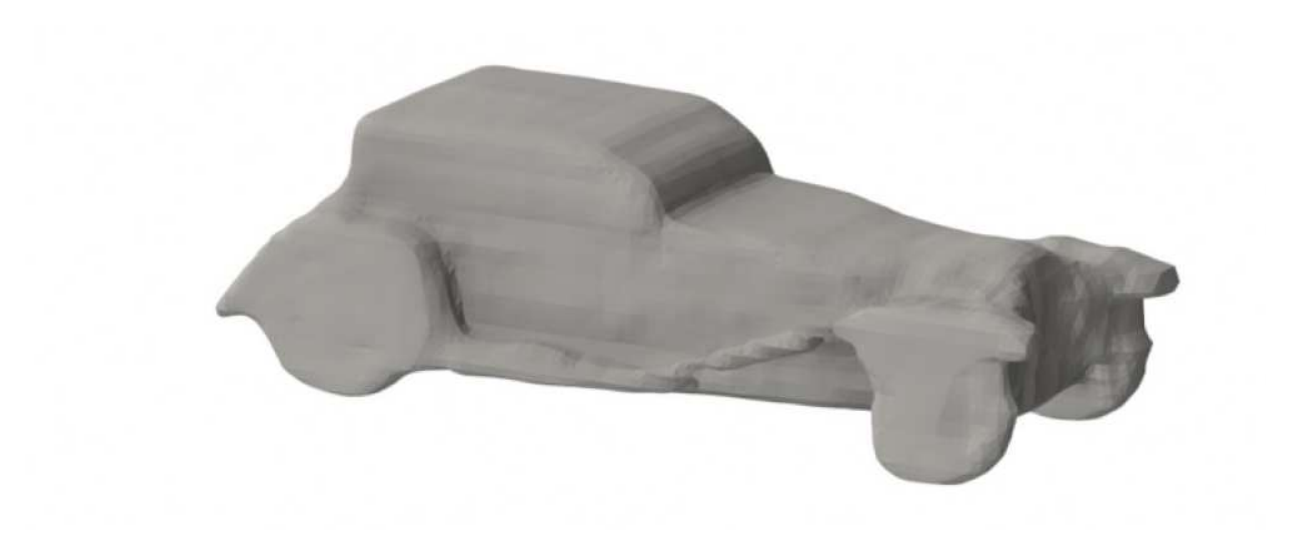}};
				\node[] (d) at (15/8*3,1.4) {\includegraphics[width=0.09\textwidth]{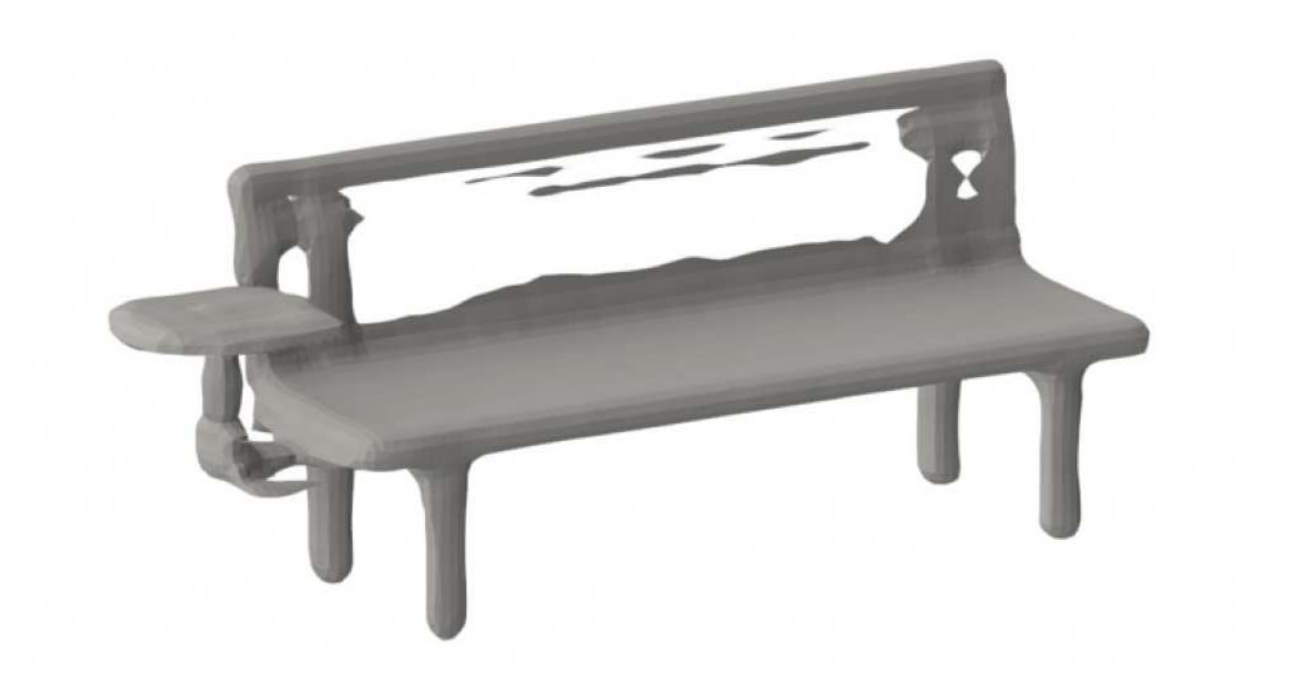}};
				\node[] (d) at (15/8*3,2.6) {\includegraphics[width=0.09\textwidth]{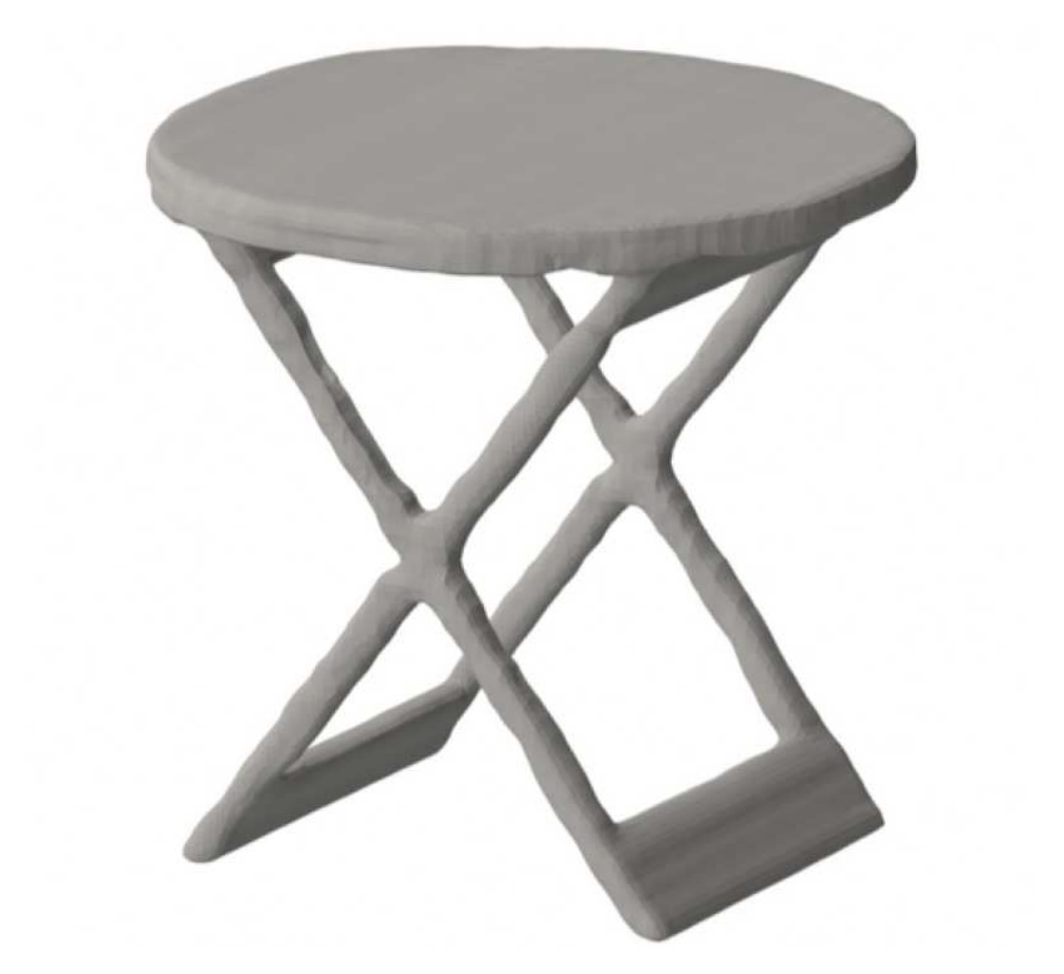} };
				\node[] (d) at (15/8*3,4) {\includegraphics[width=0.09\textwidth]{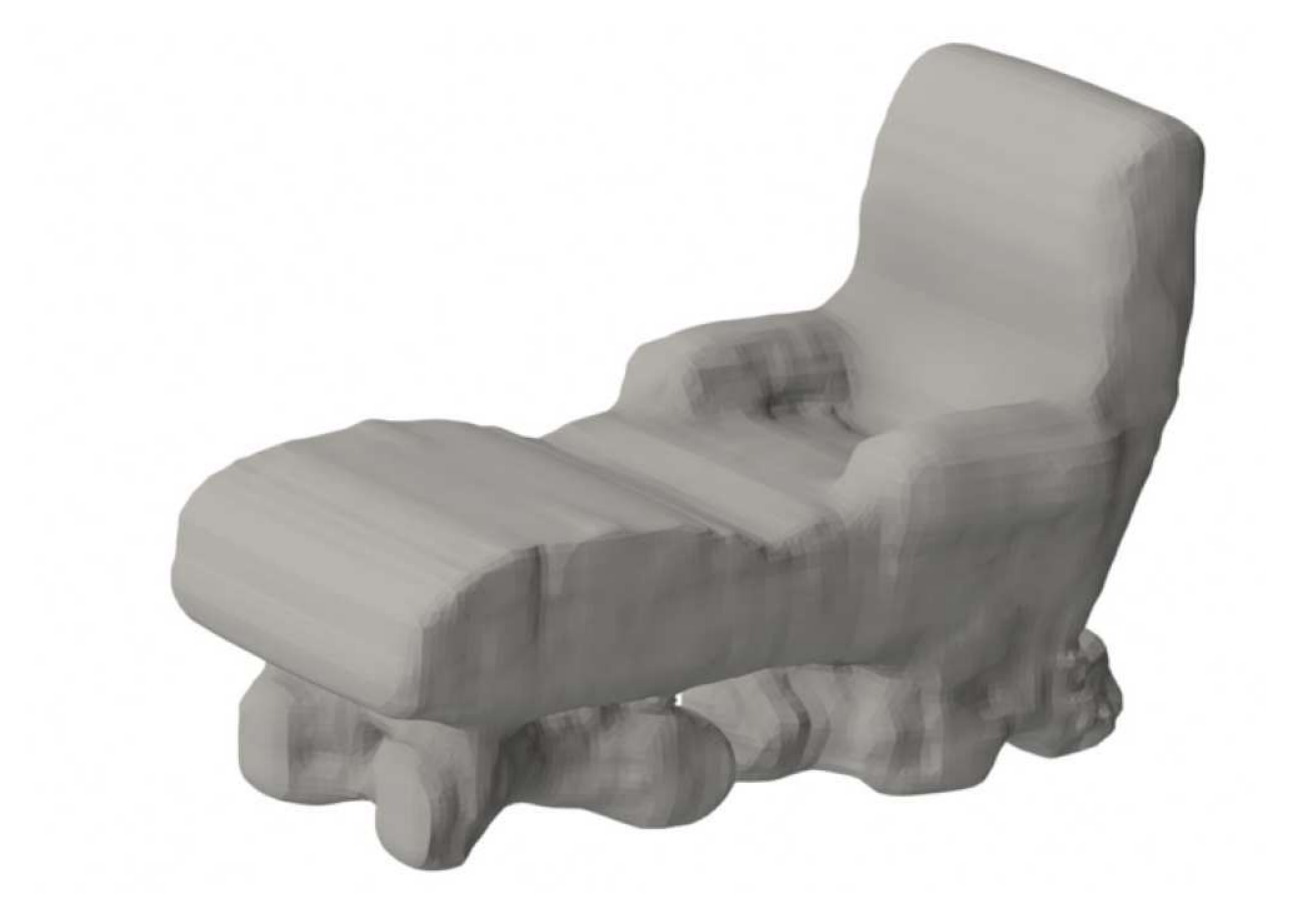} };
				\node[] (d) at (15/8*3,5.4) {\includegraphics[width=0.09\textwidth]{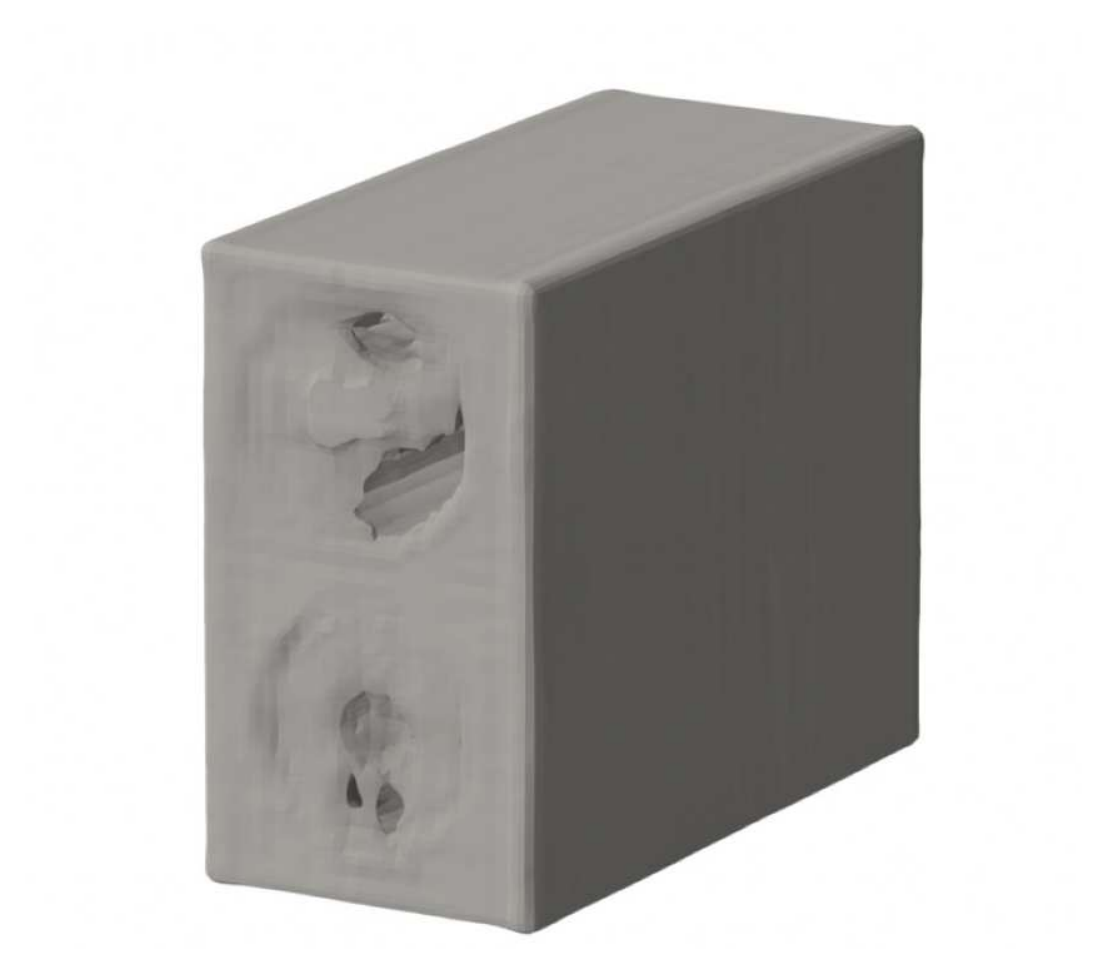} };
				\node[] (d) at (15/8*3,0) {\small (d) CONet \cite{CONVOCCNET} };
				
				\node[] (e) at (15/8*4,0.6) {\includegraphics[width=0.09\textwidth]{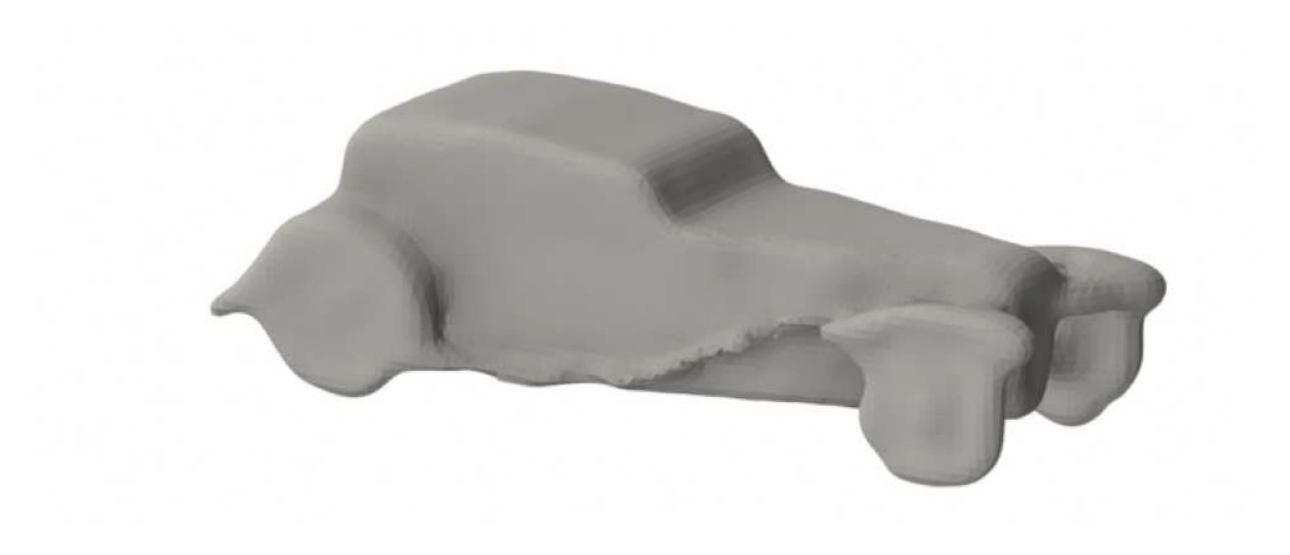}};
				\node[] (e) at (15/8*4,1.4) {\includegraphics[width=0.09\textwidth]{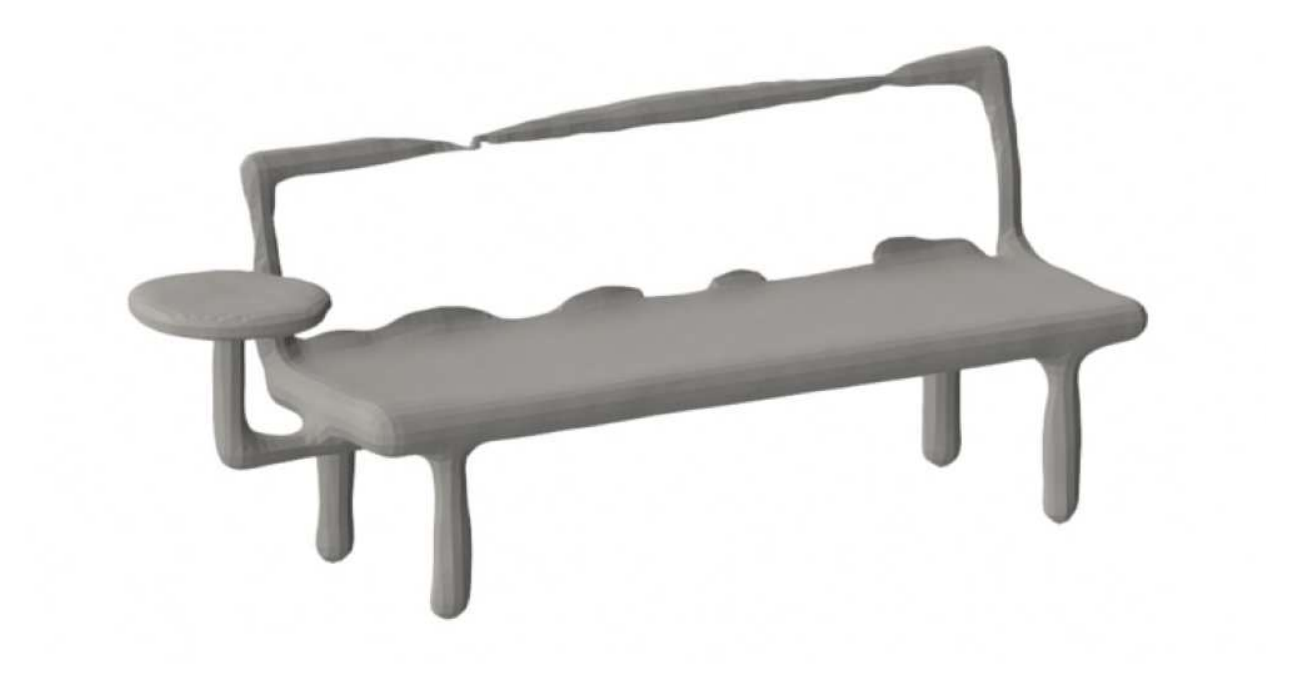}};
				\node[] (e) at (15/8*4,2.6) {\includegraphics[width=0.09\textwidth]{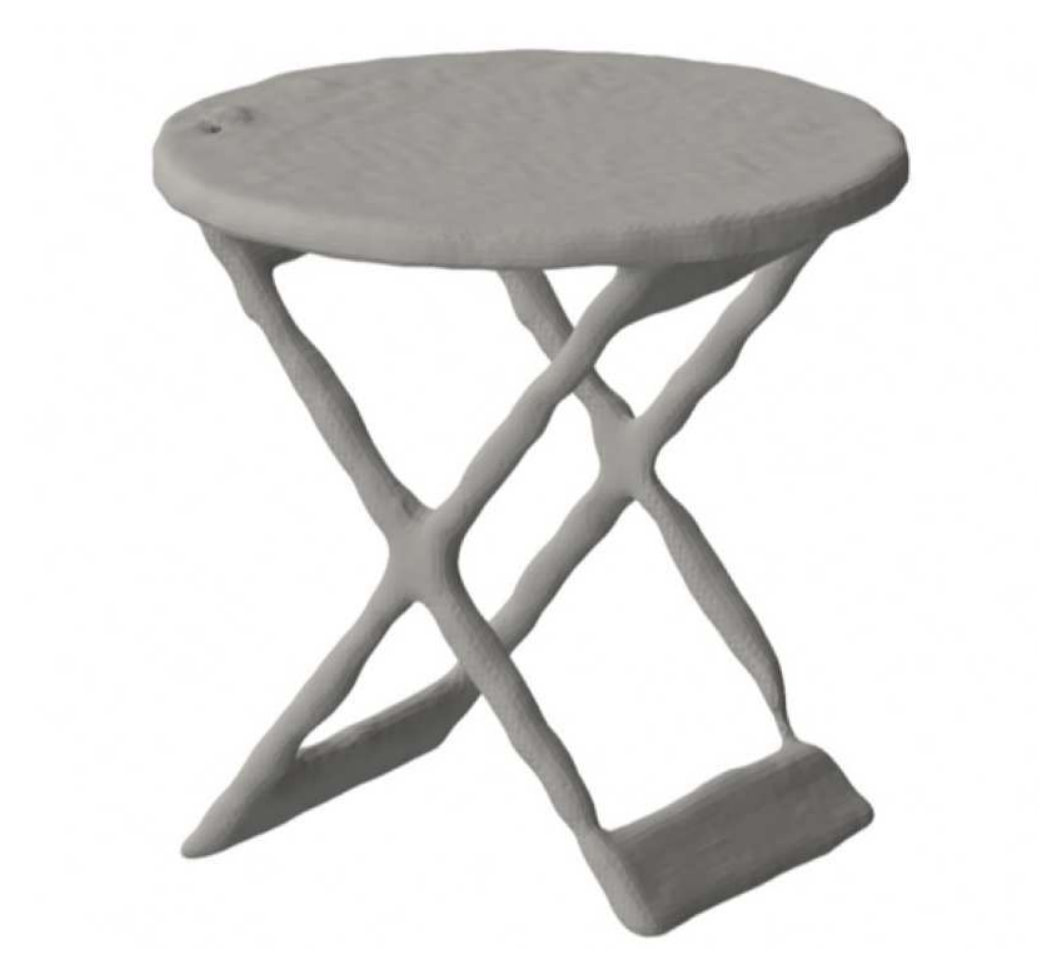} };
				\node[] (e) at (15/8*4,4) {\includegraphics[width=0.09\textwidth]{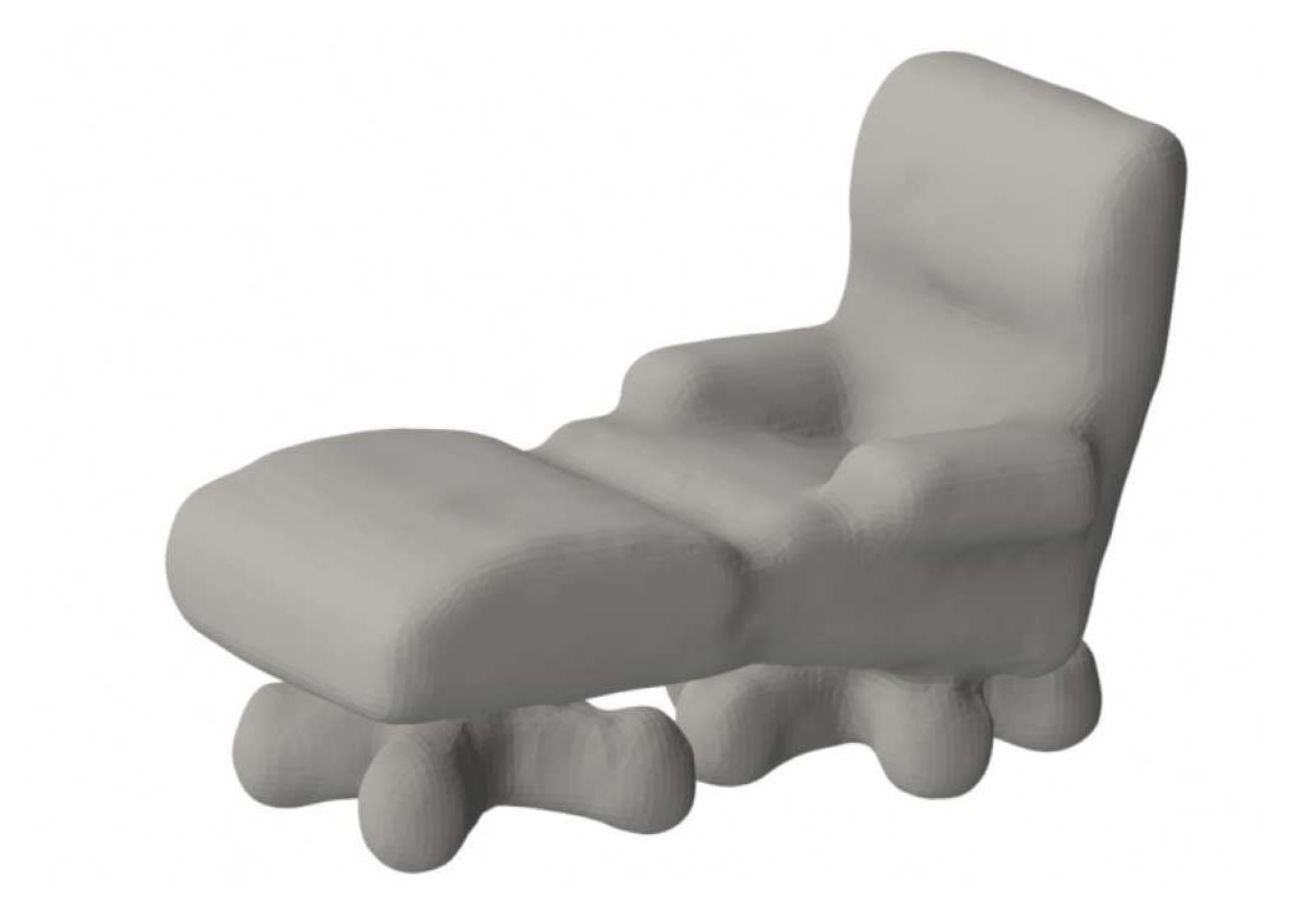} };
				\node[] (e) at (15/8*4,5.4) {\includegraphics[width=0.09\textwidth]{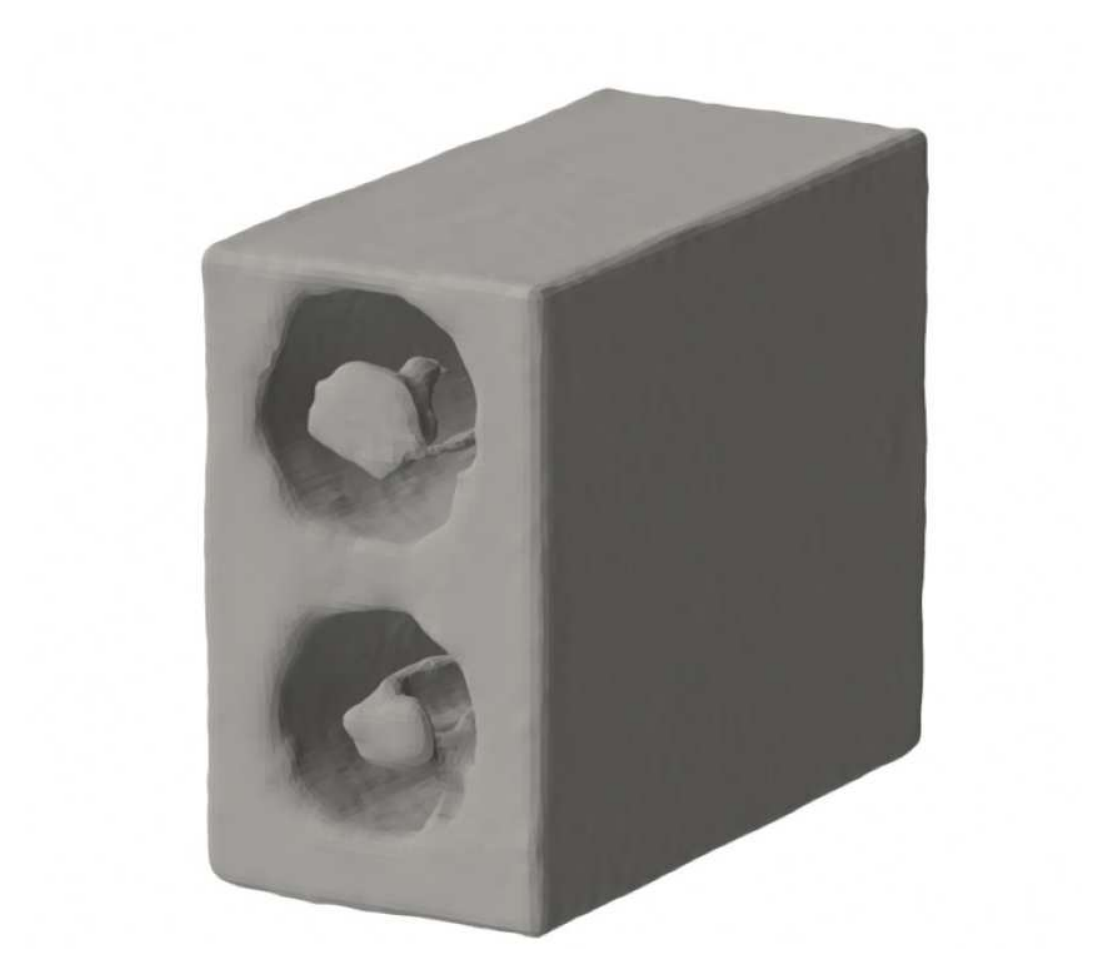} };
				\node[] (e) at (15/8*4,0) {\small (e) SAP \cite{SAP} };
				
				\node[] (f) at (15/8*5,0.6) {\includegraphics[width=0.09\textwidth]{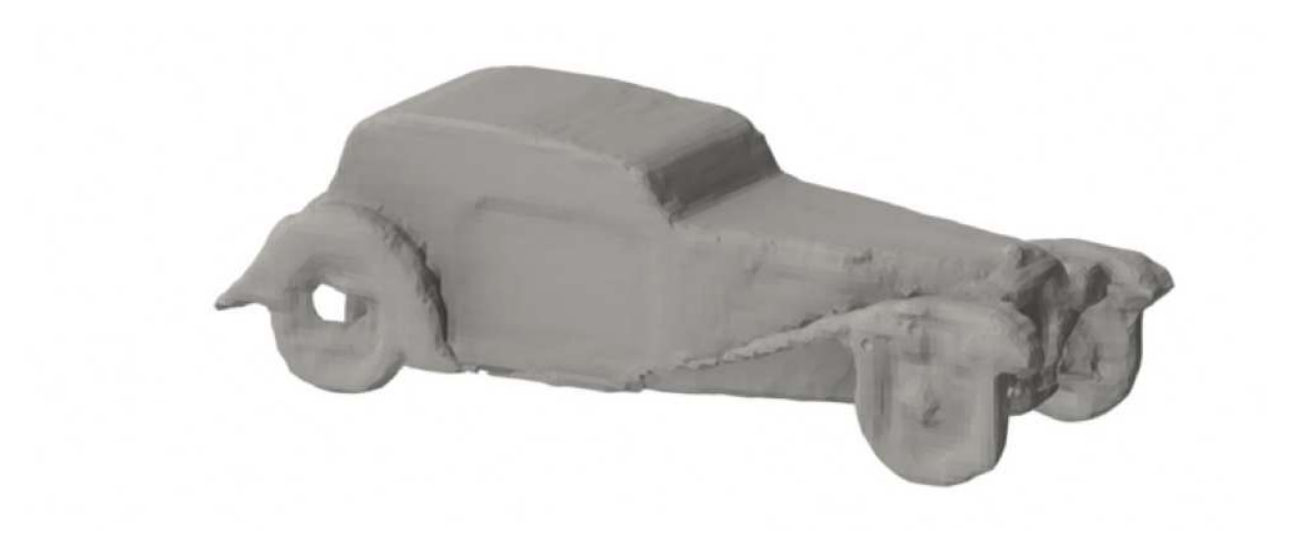}};
				\node[] (f) at (15/8*5,1.4) {\includegraphics[width=0.09\textwidth]{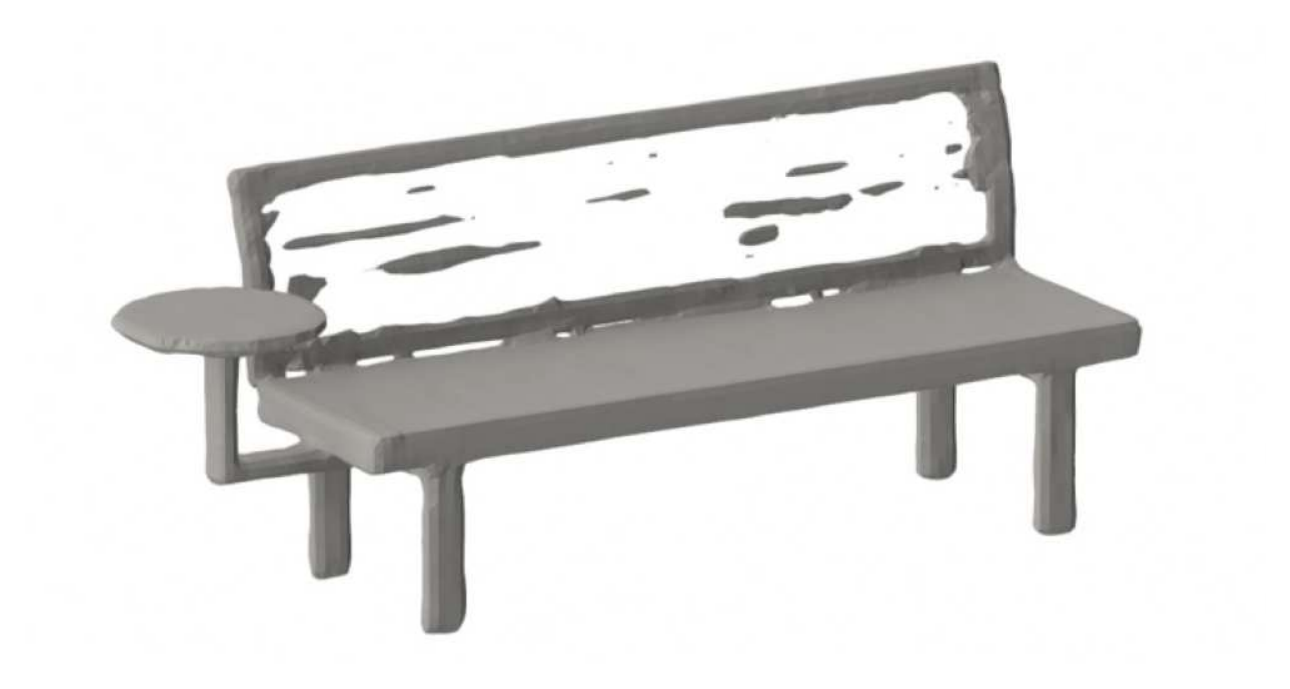}};
				\node[] (f) at (15/8*5,2.6) {\includegraphics[width=0.09\textwidth]{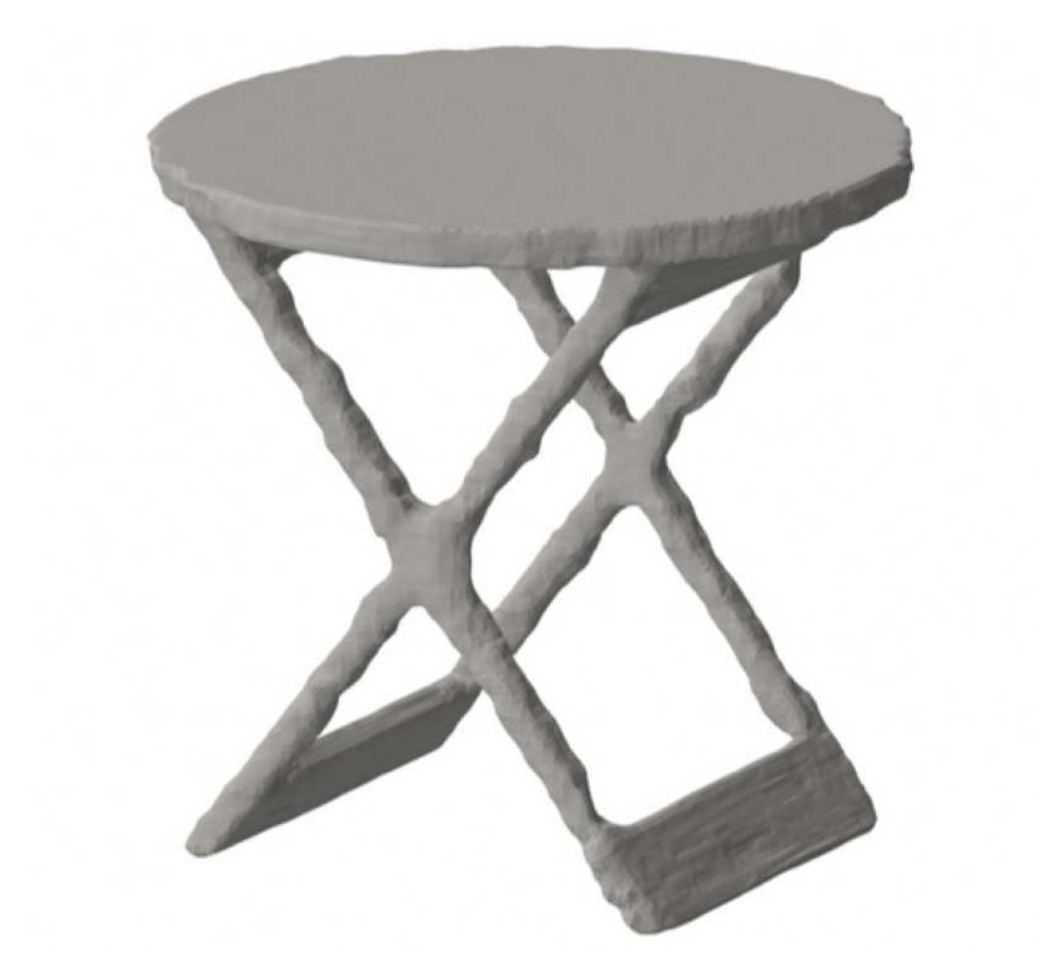} };
				\node[] (f) at (15/8*5,4) {\includegraphics[width=0.09\textwidth]{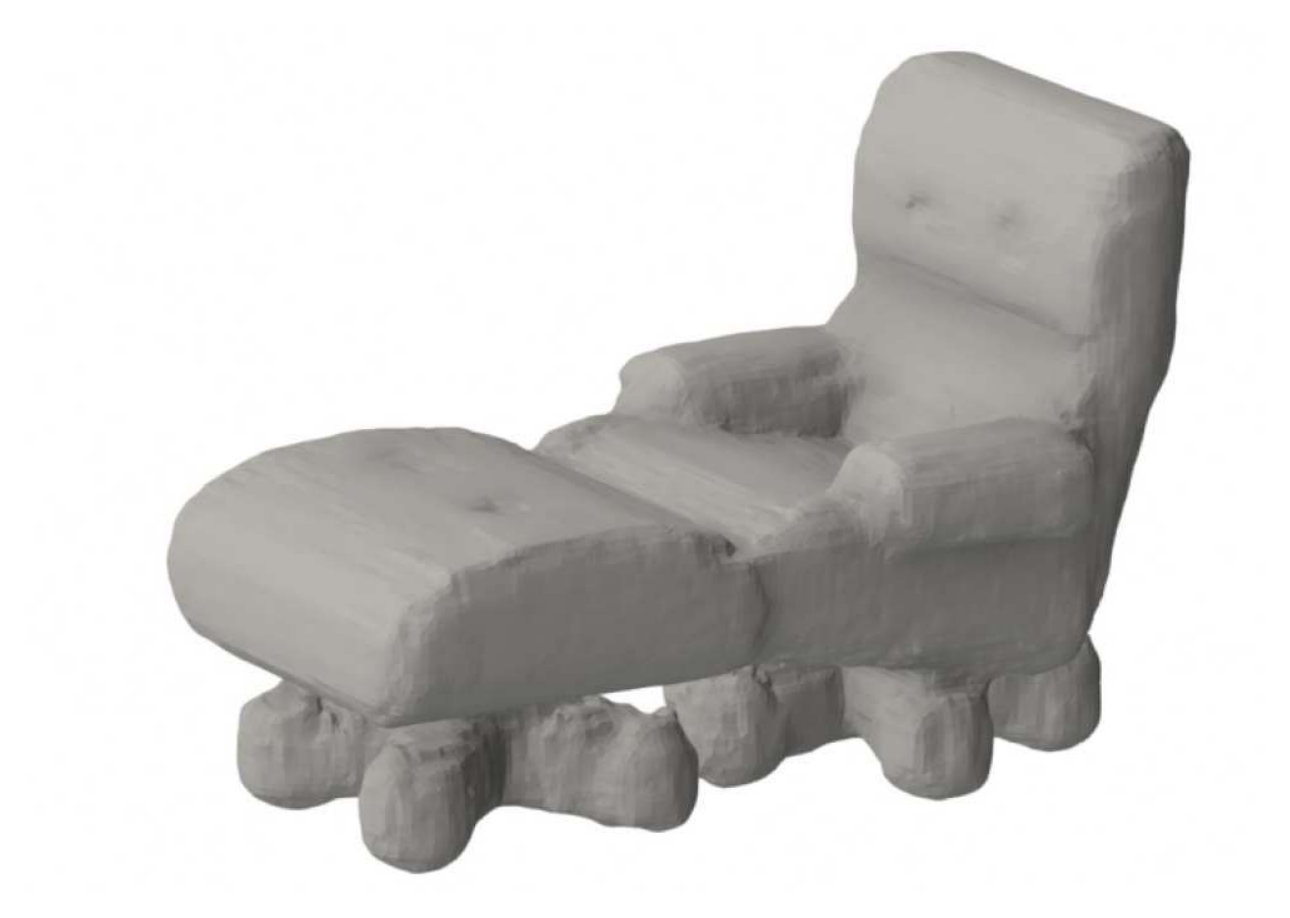} };
				\node[] (f) at (15/8*5,5.4) {\includegraphics[width=0.09\textwidth]{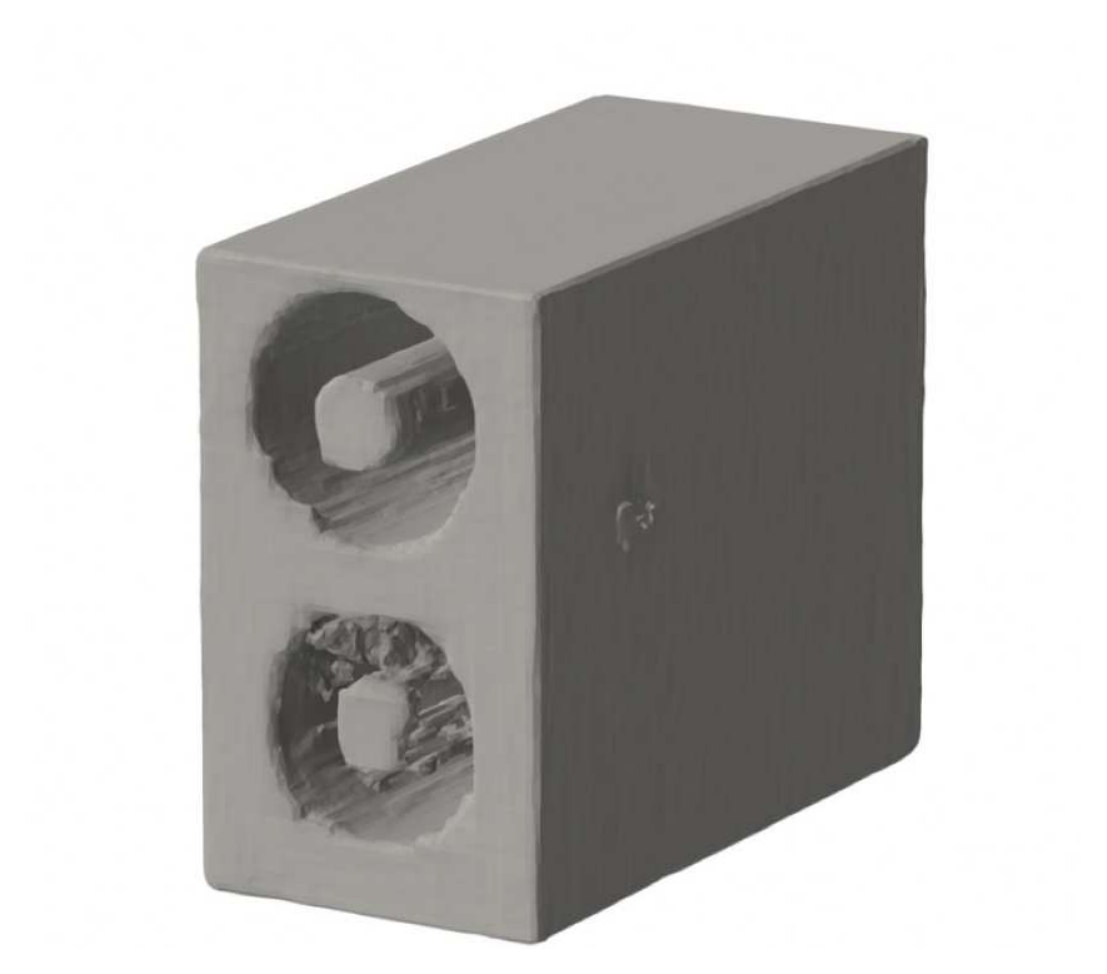} };
				\node[] (f) at (15/8*5,0) {\small (f) POCO \cite{POCO} };
				
				\node[] (g) at (15/8*6,0.6) {\includegraphics[width=0.09\textwidth]{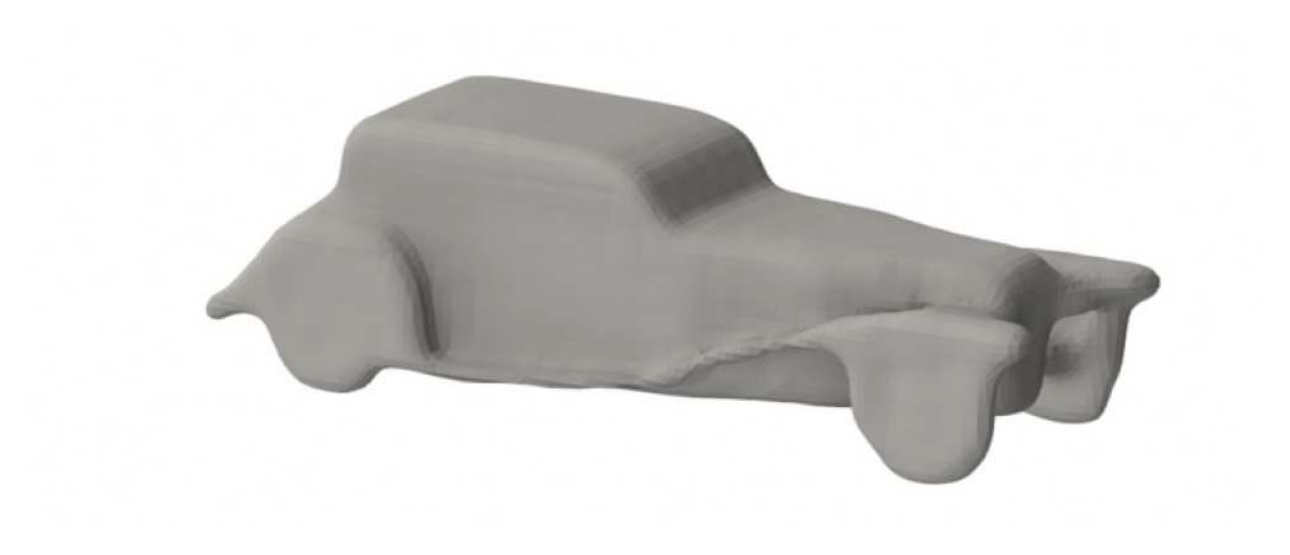}};
				\node[] (g) at (15/8*6,1.4) {\includegraphics[width=0.09\textwidth]{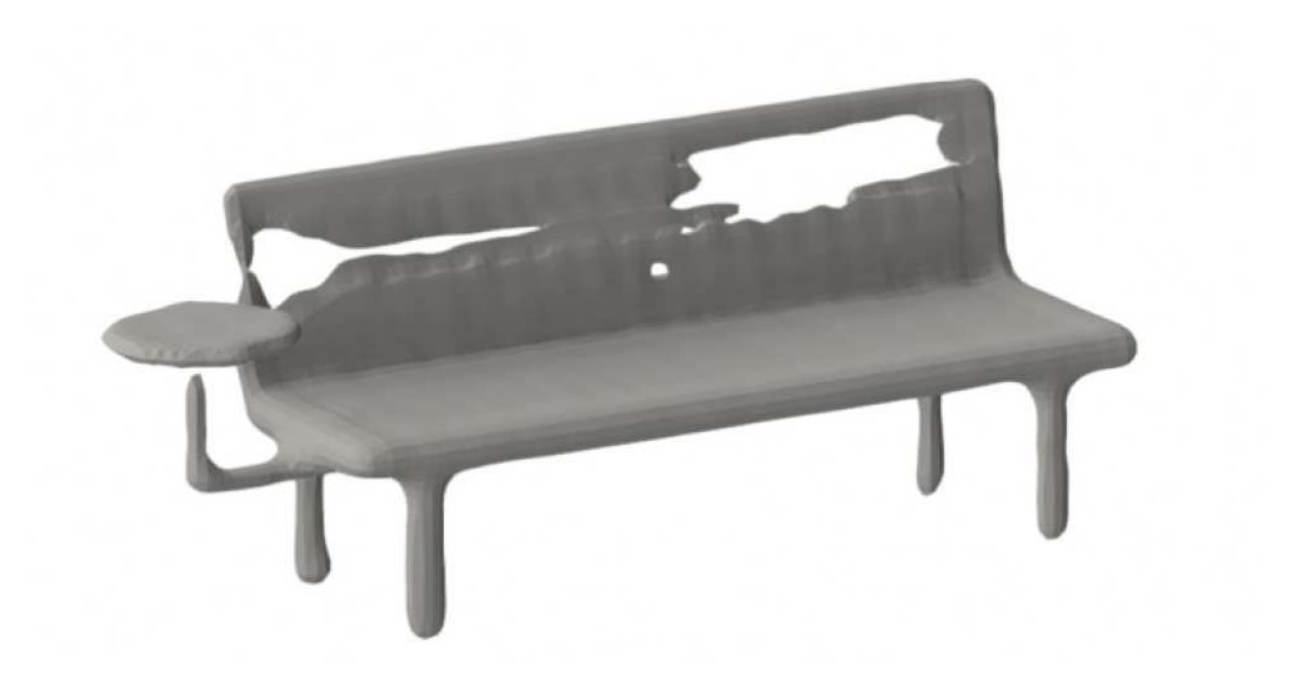}};
				\node[] (g) at (15/8*6,2.6) {\includegraphics[width=0.09\textwidth]{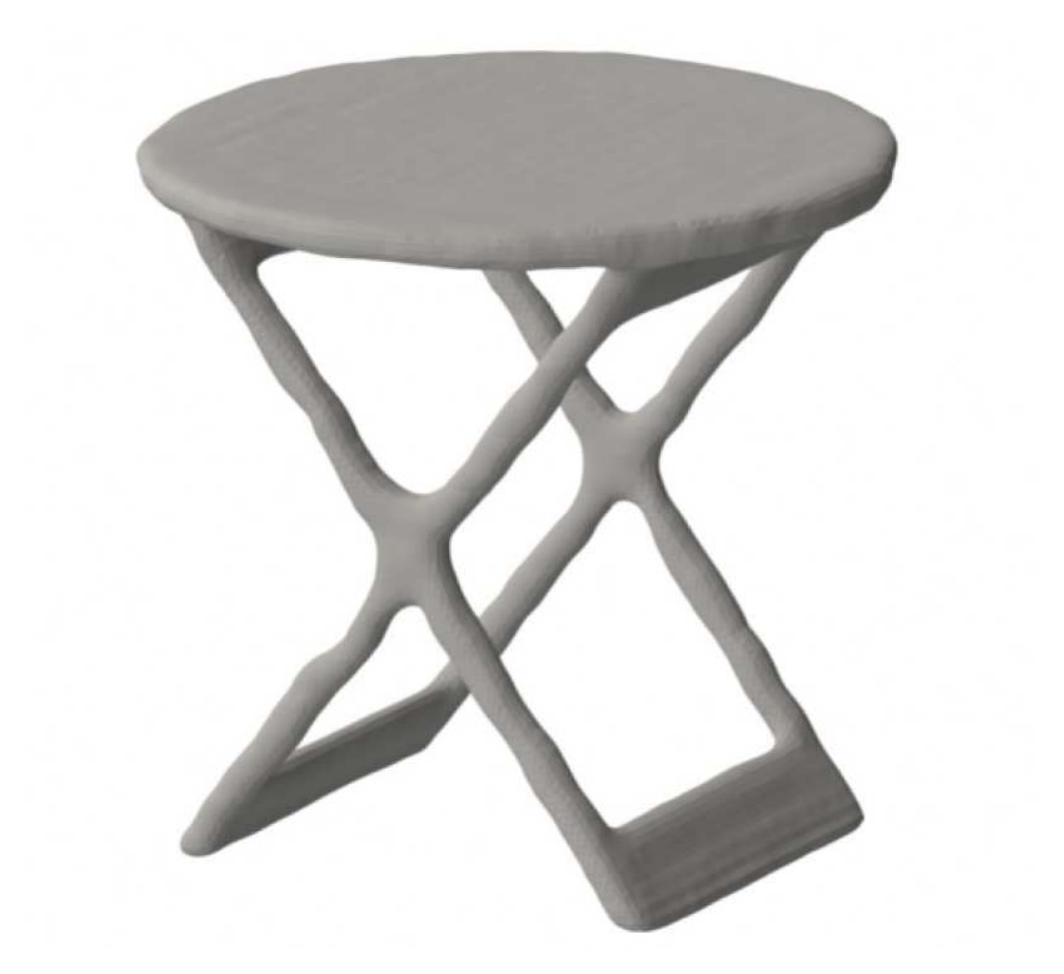} };
				\node[] (g) at (15/8*6,4) {\includegraphics[width=0.09\textwidth]{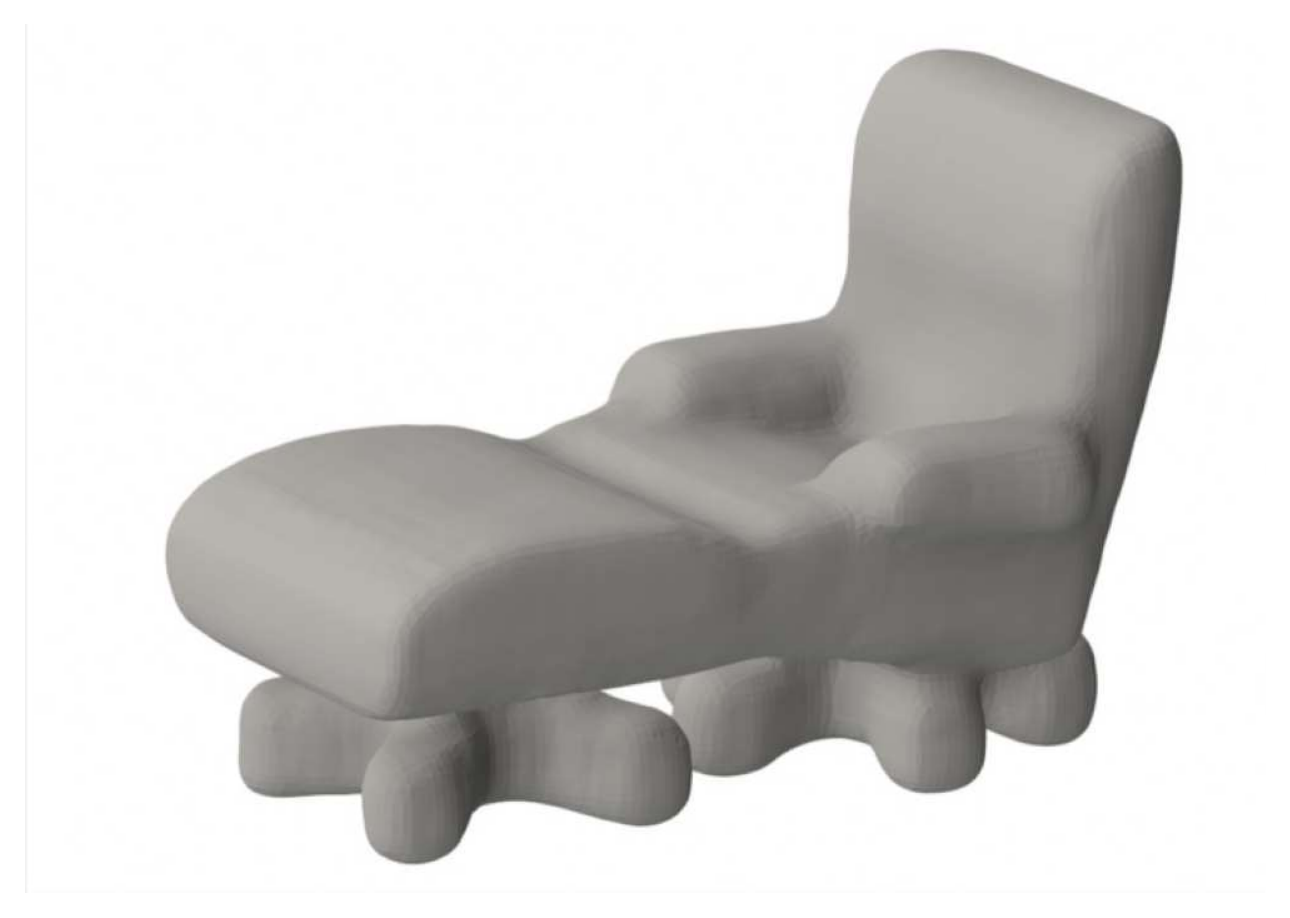} };
				\node[] (g) at (15/8*6,5.4) {\includegraphics[width=0.09\textwidth]{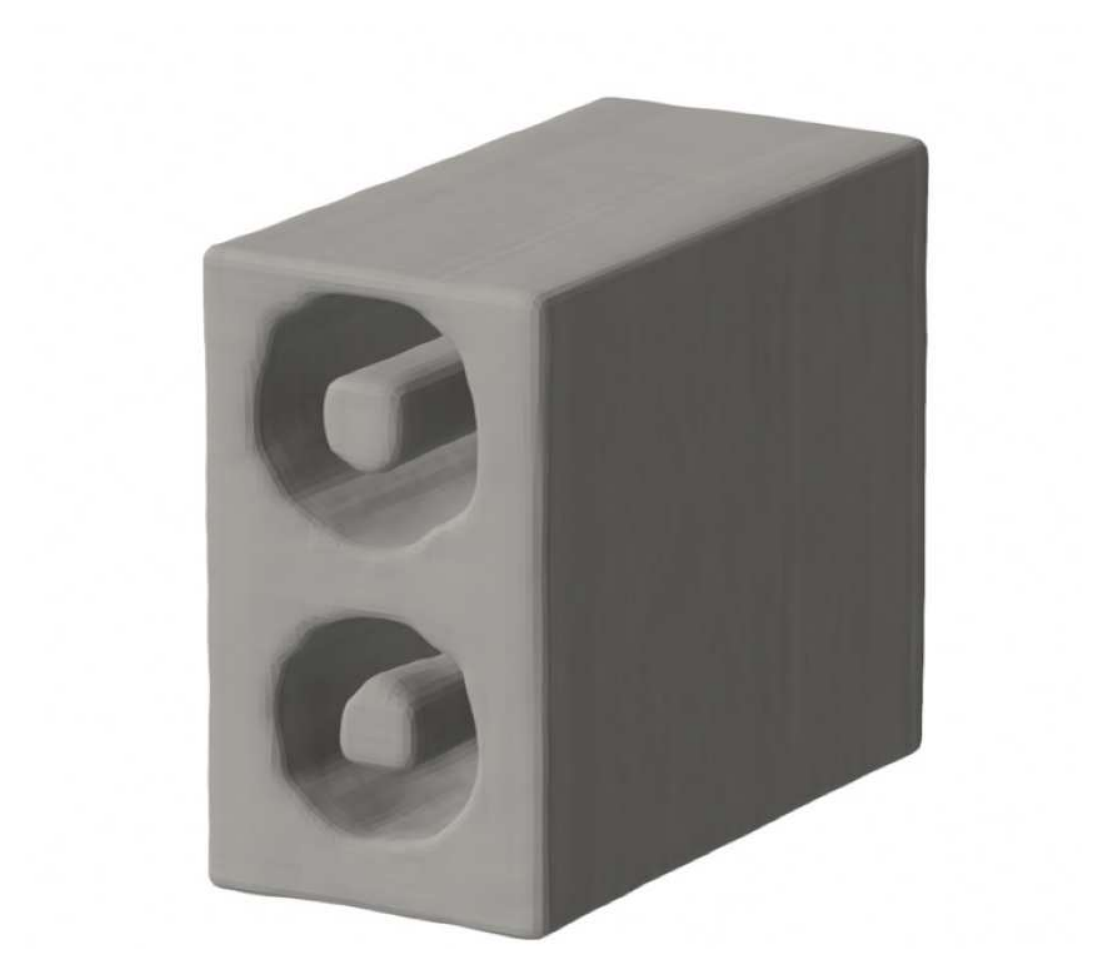} };
				\node[] (g) at (15/8*6,0) {\small (g) {DOG \cite{DOG}}};

				\node[] (h) at (15/8*7,0.6) {\includegraphics[width=0.09\textwidth]{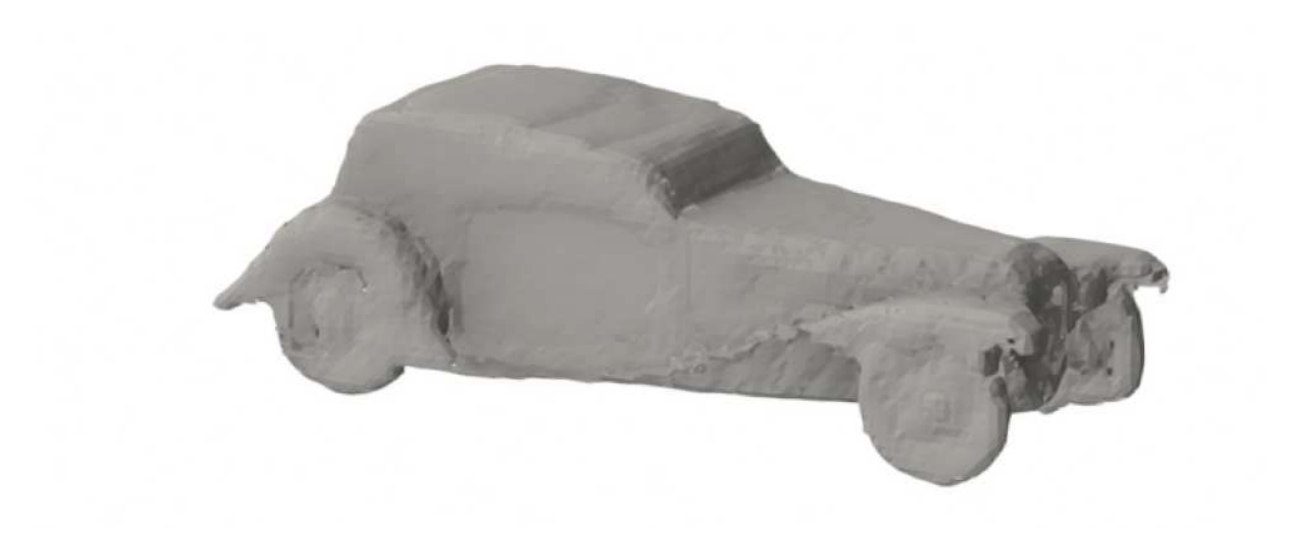}};
				\node[] (h) at (15/8*7,1.4) {\includegraphics[width=0.09\textwidth]{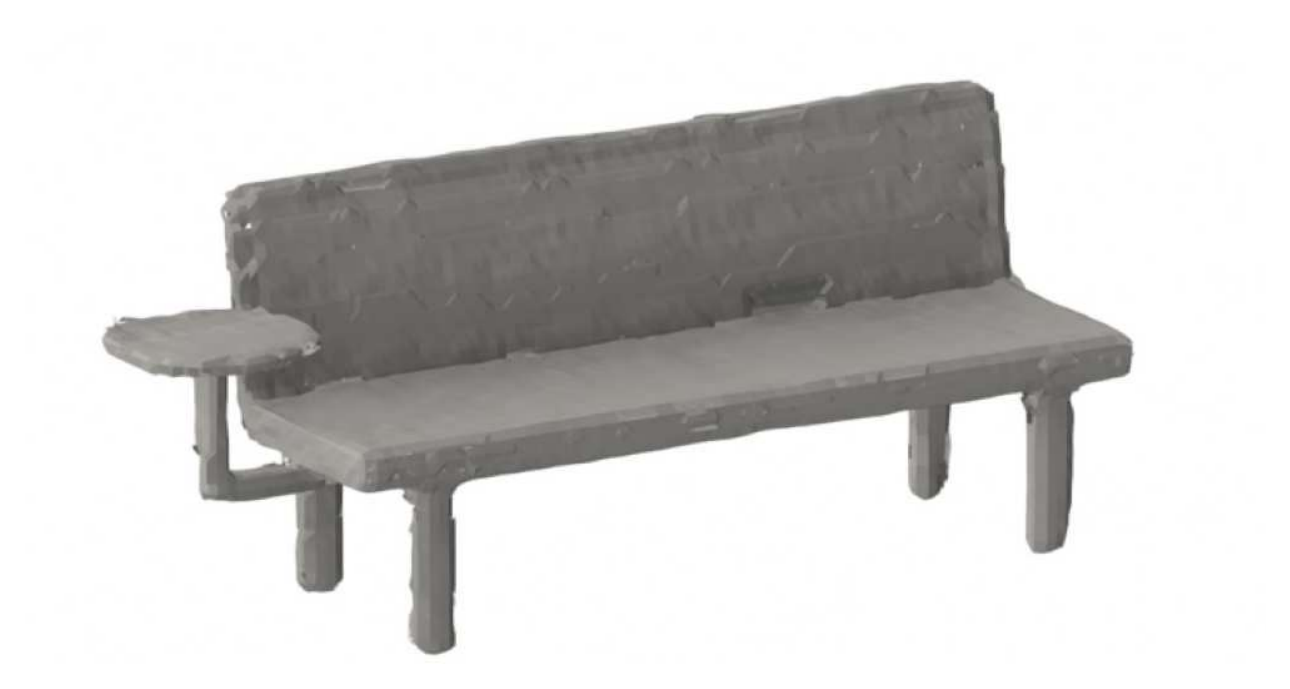}};
				\node[] (h) at (15/8*7,2.6) {\includegraphics[width=0.09\textwidth]{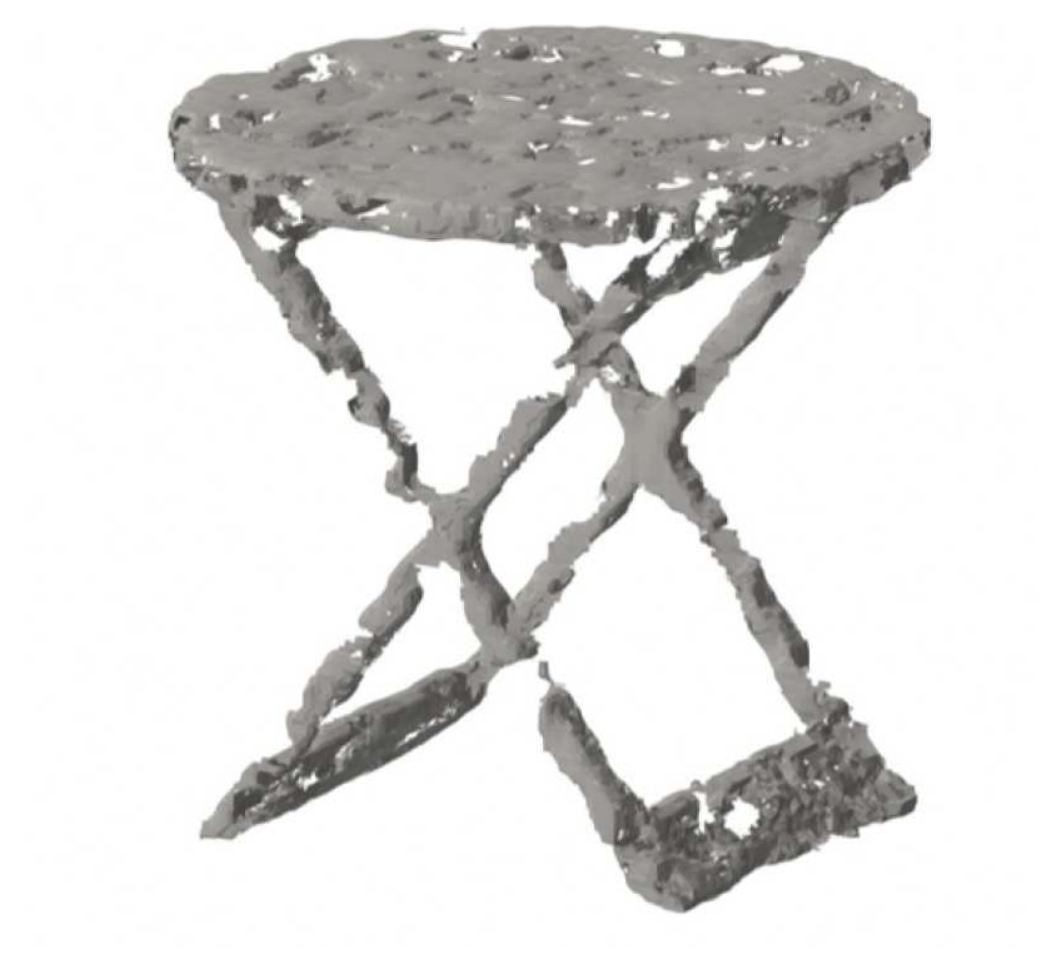} };
				\node[] (h) at (15/8*7,4) {\includegraphics[width=0.09\textwidth]{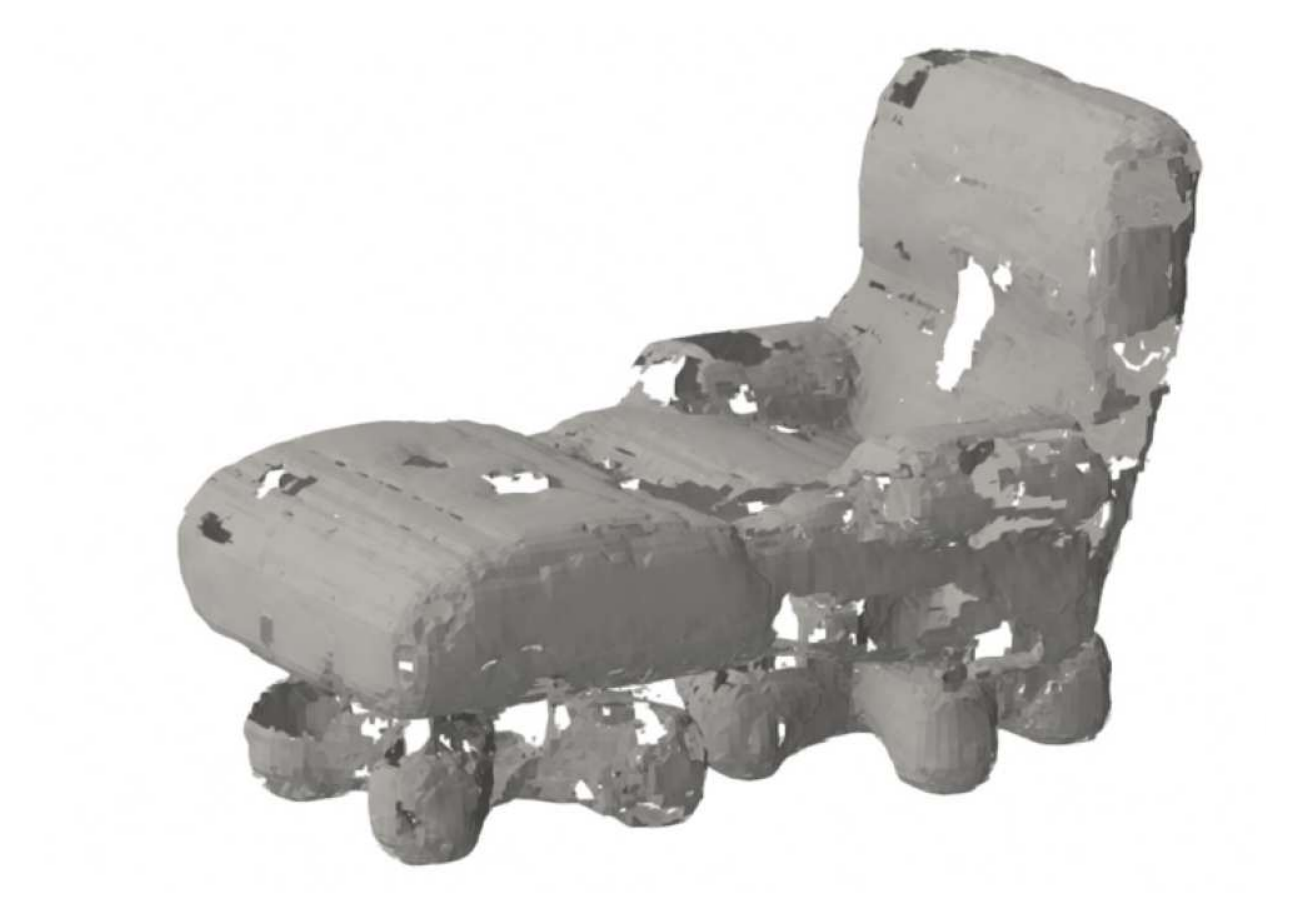} };
				\node[] (h) at (15/8*7,5.4) {\includegraphics[width=0.09\textwidth]{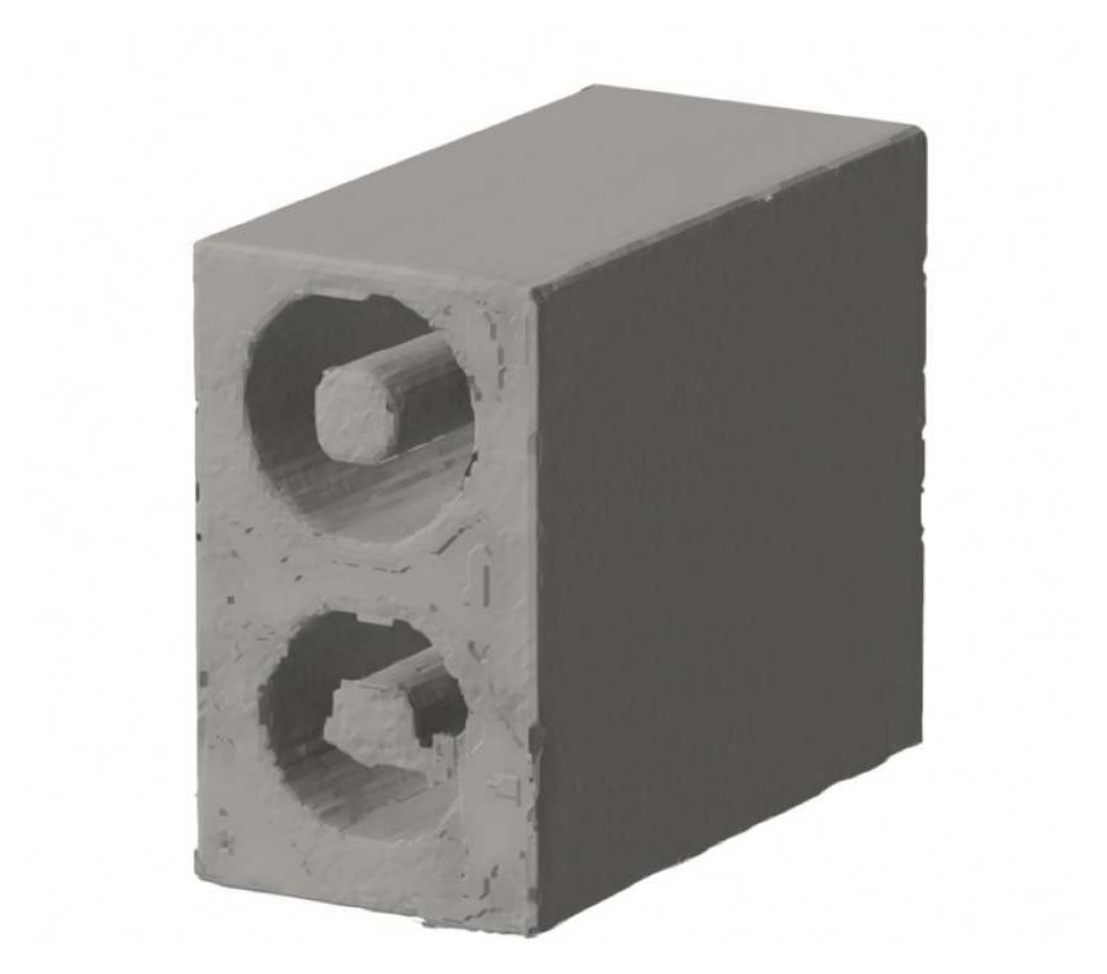} };
				\node[] (h) at (15/8*7,0) {\small (h) GIFS \cite{GIFS} };

				\node[] (i) at (15,0.6) {\includegraphics[width=0.09\textwidth]{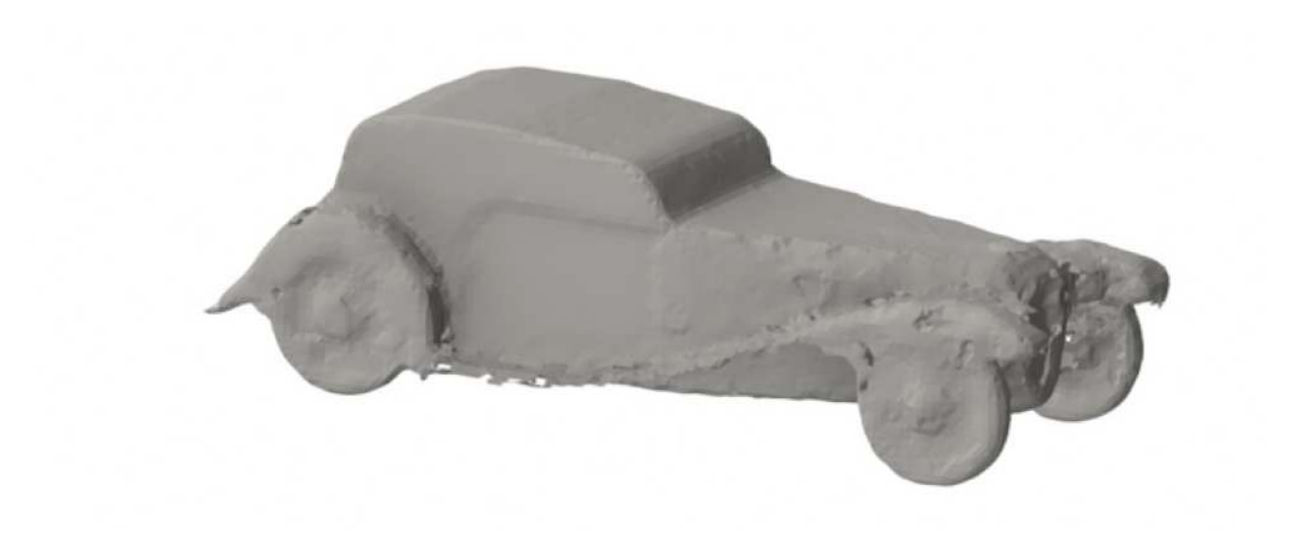}};
				\node[] (i) at (15,1.4) {\includegraphics[width=0.09\textwidth]{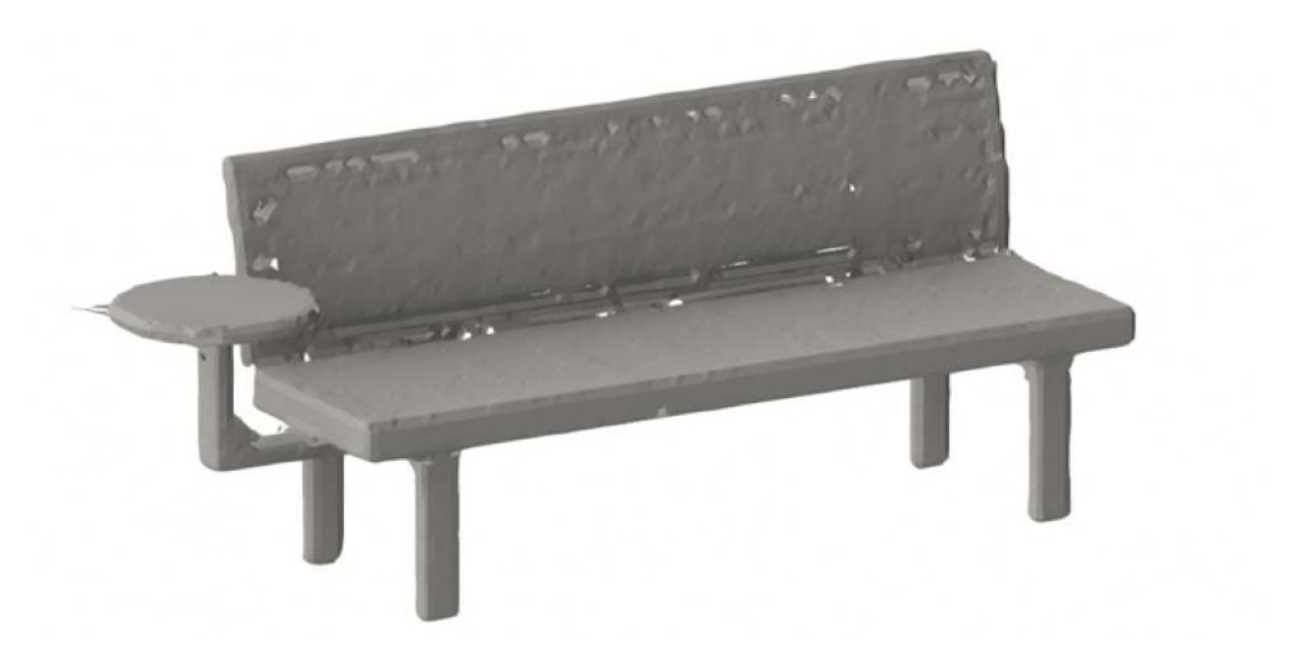}};
				\node[] (i) at (15,2.6) {\includegraphics[width=0.09\textwidth]{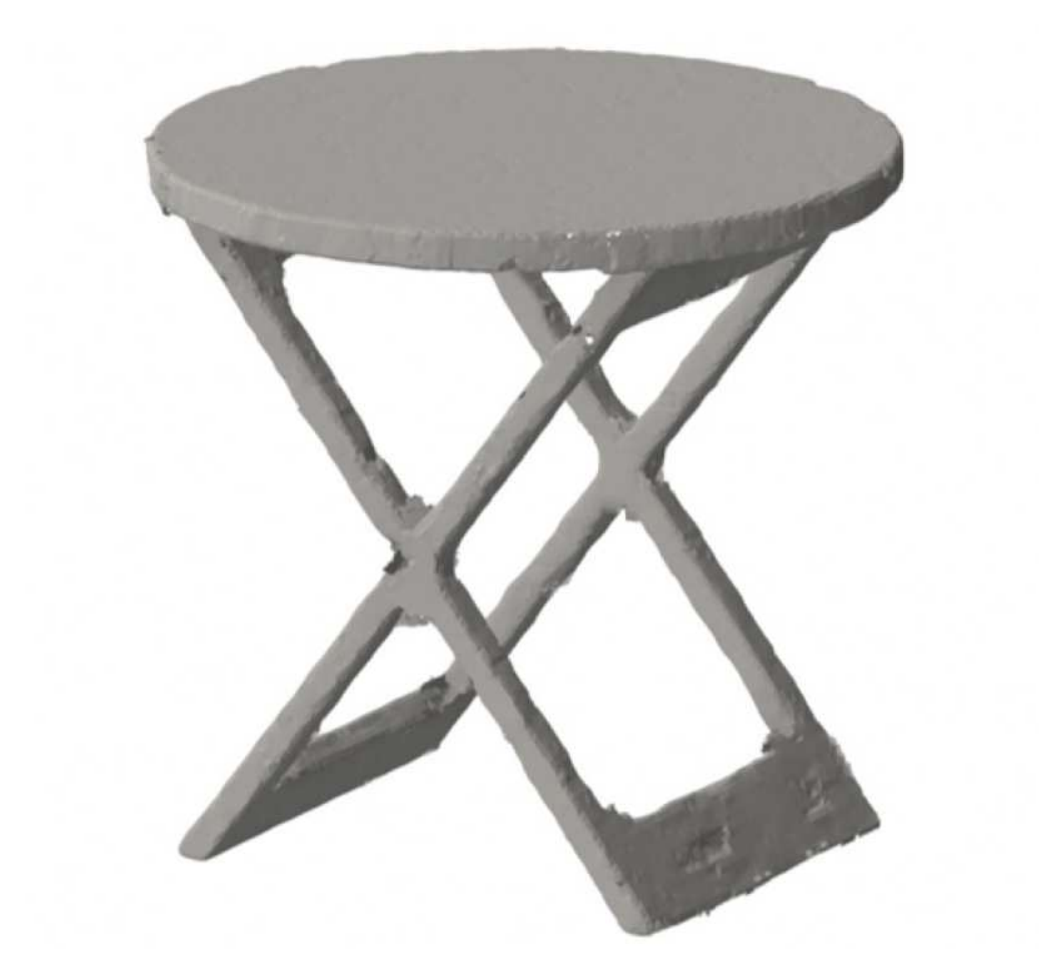}};
				\node[] (i) at (15,4) {\includegraphics[width=0.09\textwidth]{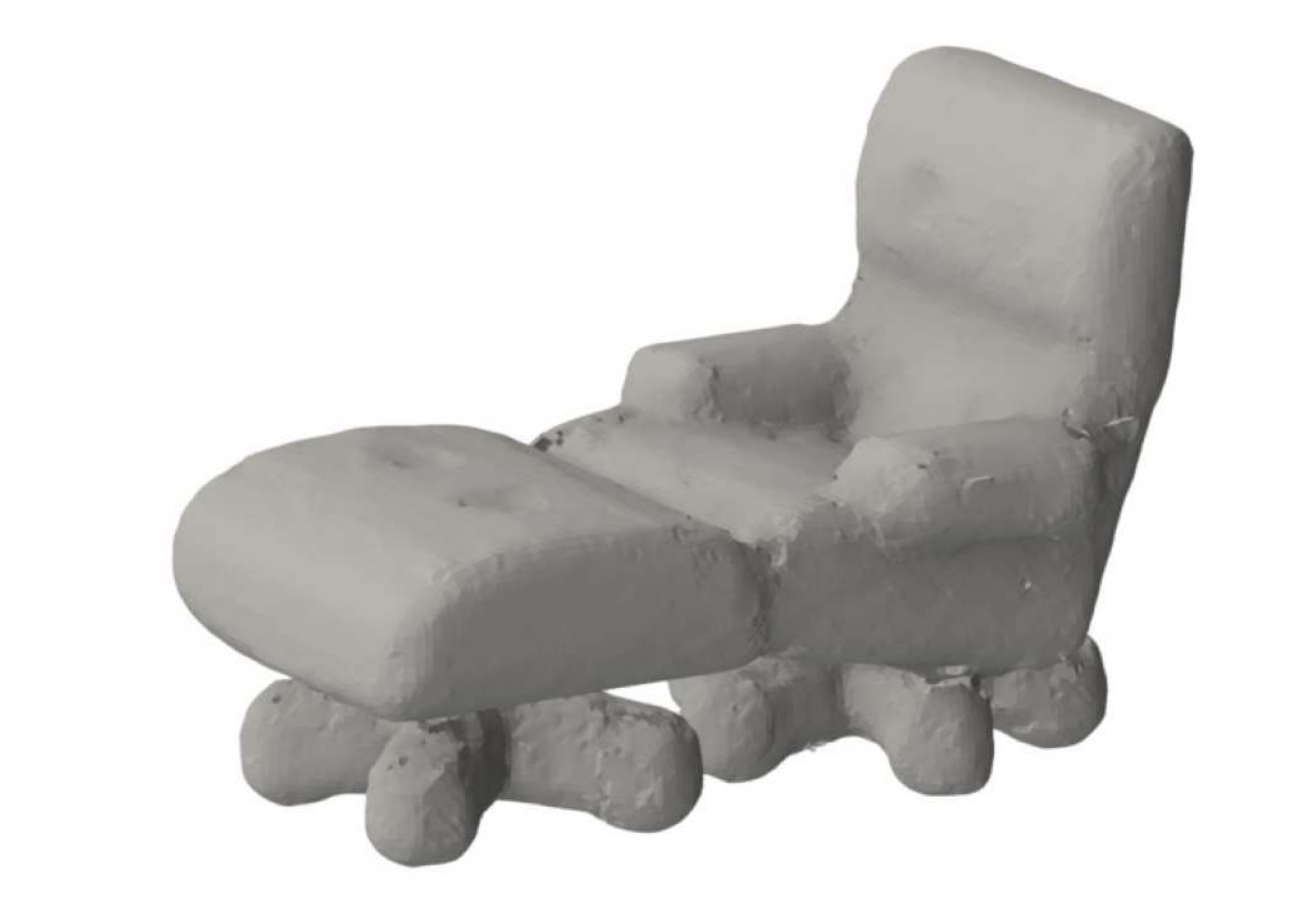}};
				\node[] (i) at (15,5.4) {\includegraphics[width=0.09\textwidth]{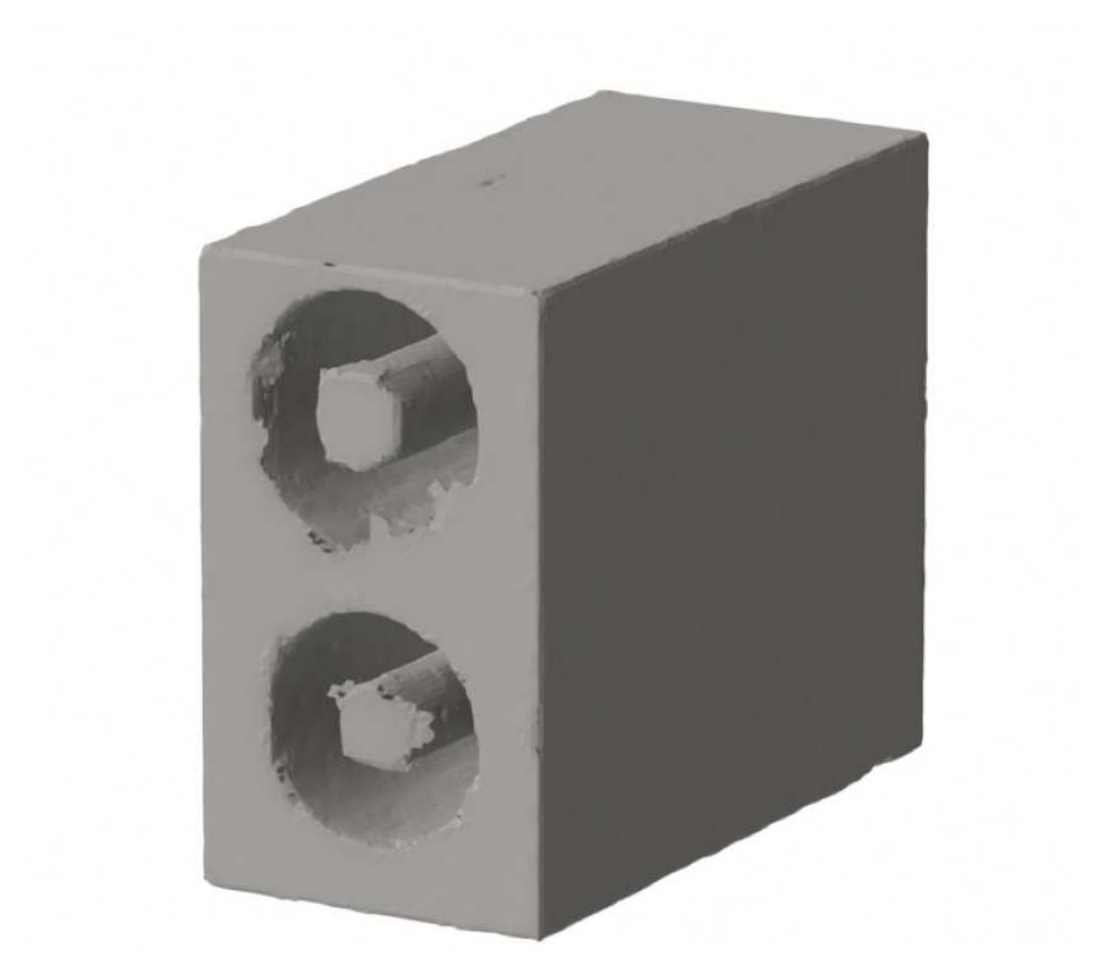}};
				\node[] (i) at (15,0) {\small (i) Ours };
				
			\end{tikzpicture}
		}
		\setlength{\abovecaptionskip}{-0.05cm} 
		\caption{Visual comparisons 
			on the ShapeNet dataset \cite{SHAPENET}. We refer readers to the \textit{supplementary material} for more visual results.}
		\label{SHAPENET:VIS} \vspace{-0.3cm}
	\end{figure*}
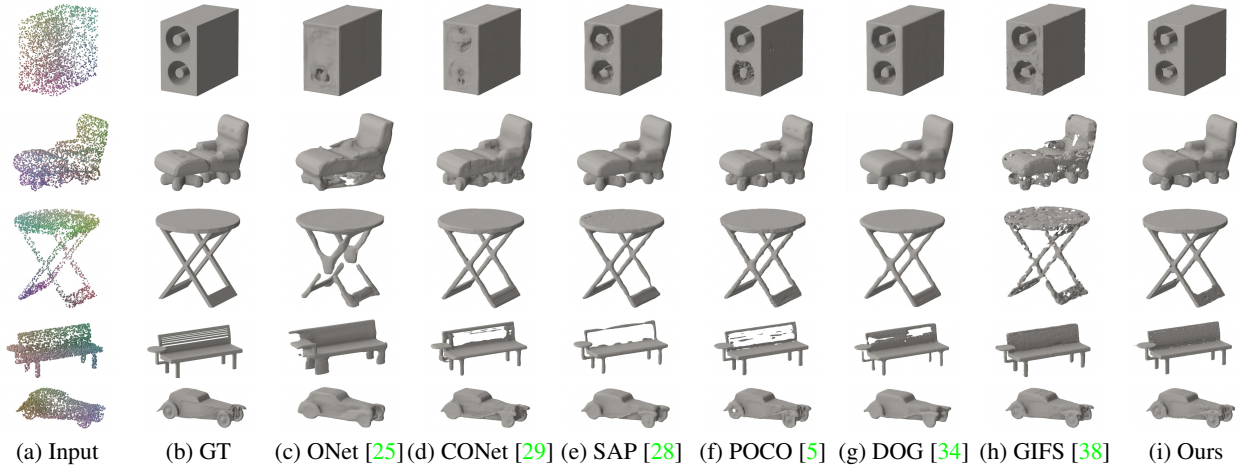

	\subsection{Watertight Surface Reconstruction}\label{subsec:eva_watertight}  
	\noindent 
	\textbf{Experiment settings.} 
	Following previous works \cite{OCCNET,CONVOCCNET,SAP,POCO}, 
	we adopted the ShapeNet Core dataset \cite{SHAPENET} processed by DISN \cite{DISN}, removing the non-manifold and interior structures of the original shapes.
	We split the data into the train/val/test sets according to 3D-R2N2 \cite{3DR2N2}. 
	For each shape, we randomly sampled 3000 points from the surface as the input sparse point cloud. We conducted experiments under two settings, i.e, clean data and noisy data derived by adding 
	the Gaussian noise of standard deviation 0.005 to the clean data. 
	We followed \cite{OCCNET, CONVOCCNET, SAP, DOG} 
	to set the resolution of Marching Cube as 128 for all methods for fair comparisons. 
	To measure reconstruction quality quantitatively, we followed GIFS \cite{GIFS}, a UDF-based method, to sample $10^5$ points from each reconstructed surface to calculate 
	CD and F-Score with the threshold of 0.5\% and 1\% with respect to ground-truths.
	
	\noindent \textbf{Comparisons.} We compared our GeoUDF with state-of-the-art methods, including ONet \cite{OCCNET}, CONet \cite{CONVOCCNET}, SAP \cite{SAP}, POCO \cite{POCO}, {DOG \cite{DOG}}, NDF \cite{NDF}, and GIFS \cite{GIFS}. 
	On noisy data, we directly applied the pre-trained networks of ONet, CONet, SAP, POCO and DOG released by the authors, which share the same training settings as ours. However, as their pre-trained networks on clean data were not publicly available\footnote{Despite POCO releases the pre-trained network on clean data, it was trained with GT normals.}, we used their offical codes to retrain them with the same settings as ours for fair comparisons. 
	Besides, as the pre-trained networks of NDF and GIFS were trained only with the ShapeNet car dataset rather than the whole dataset, we retrained them on the whole dataset for fair comparisons. 
	
	We quantitatively compared those methods in Table \ref{SHAPENET:TAB}, 
	where it can be seen that our method outperforms the other methods in terms of all metrics, even POCO and GIFS with larger Marching Cube resolution. 
	Besides, as illustrated
	in Fig. \ref{SHAPENET:VIS}, our GeoUDF can reconstruct surfaces with sharp edges and correct structures that are closer to GTs. 
	
	
	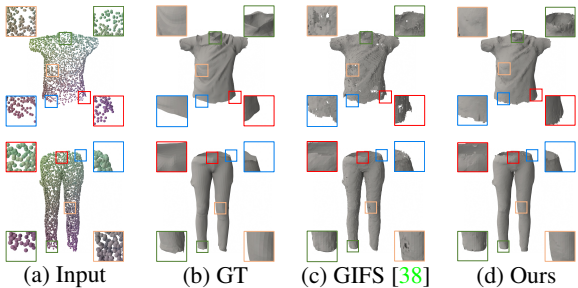
\begin{figure}[t]
		\centering
		{
			\begin{tikzpicture}[]
				
				\node[] (d) at (4, 2.8) {\includegraphics[width=0.09\textwidth]{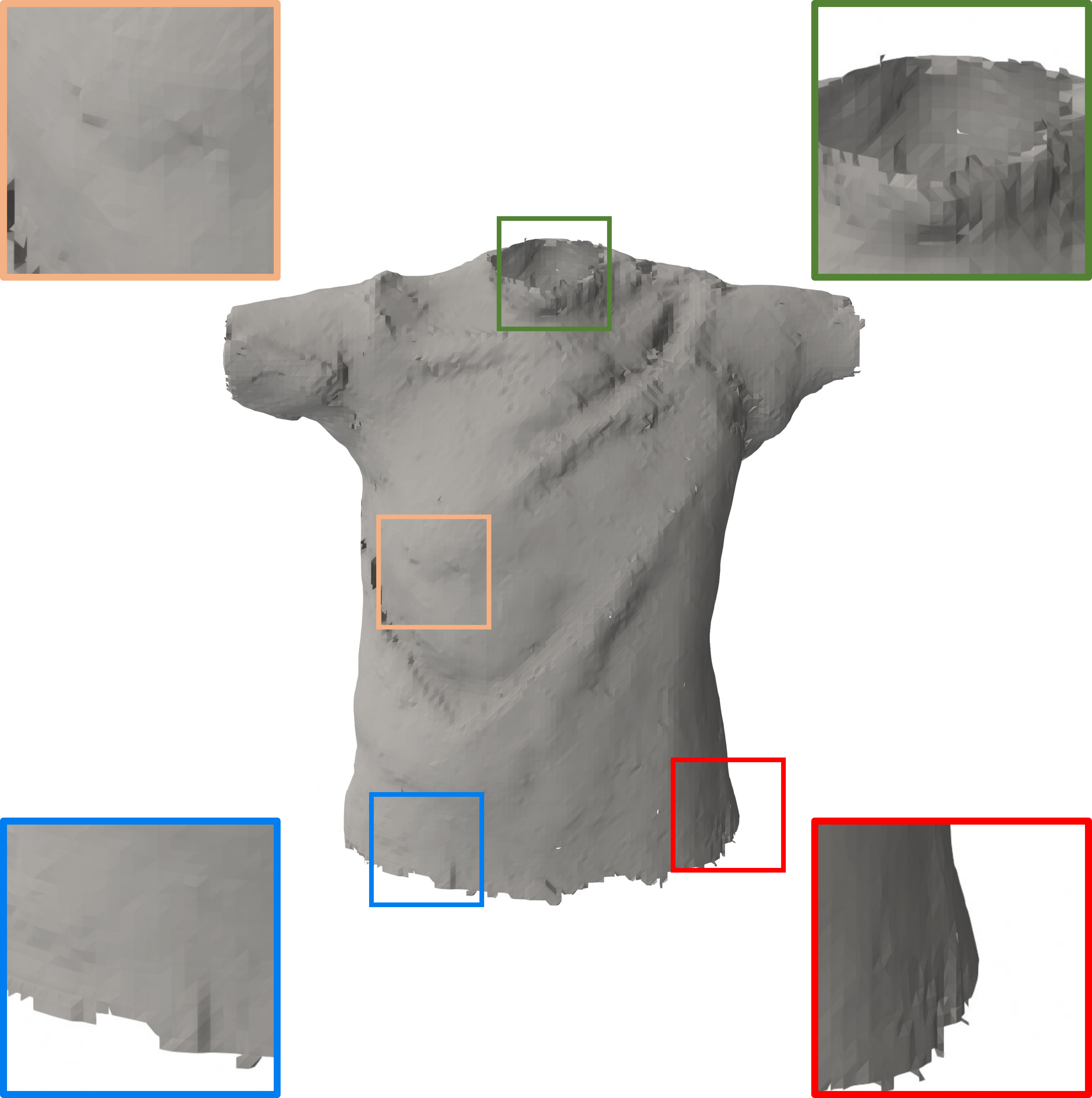}};
				\node[] (d) at (4, 1) {\includegraphics[width=0.09\textwidth]{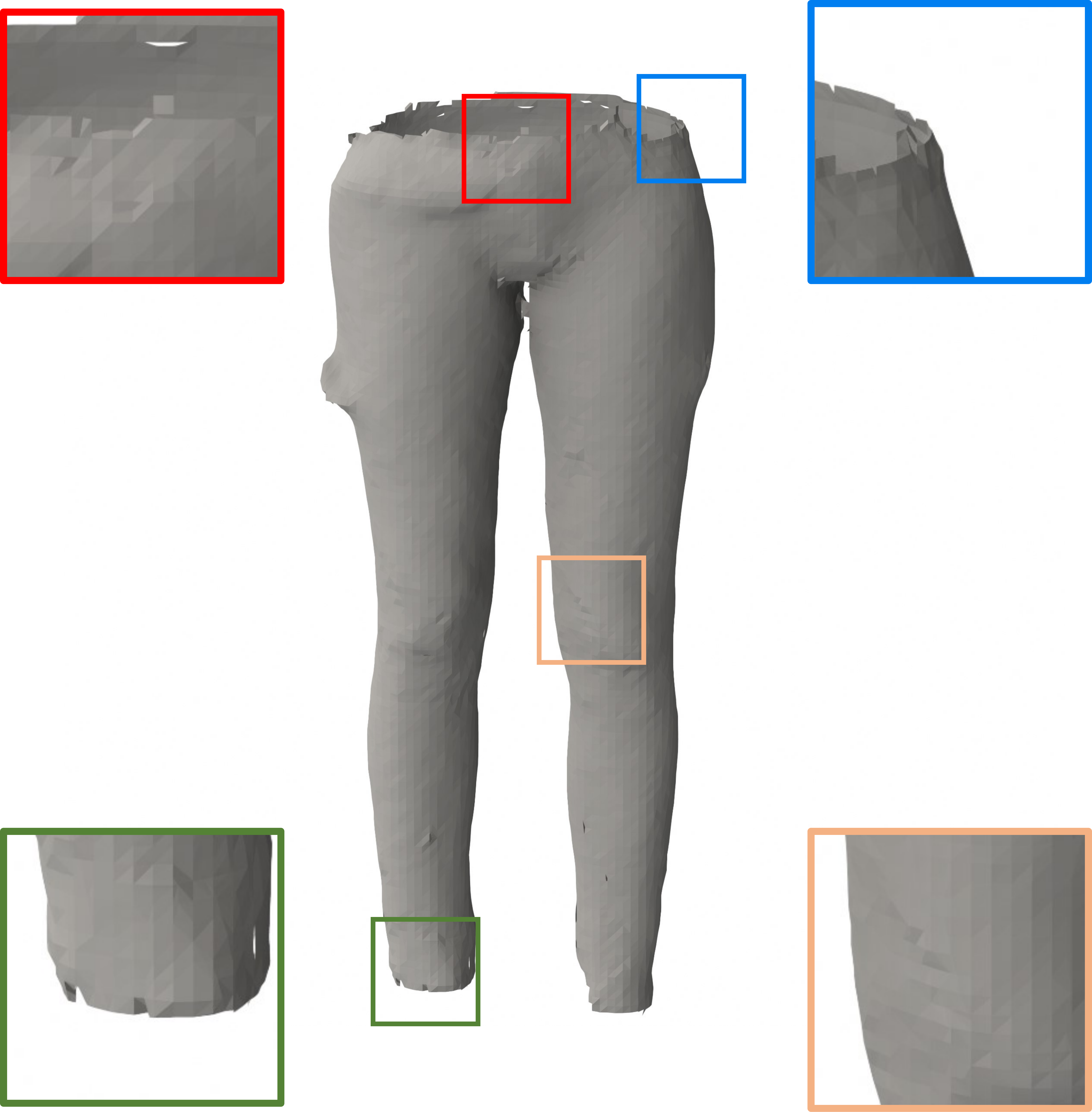}};
				\node[] (d) at (4, 0) {\small{(d) Ours} };
				
				\node[] (c) at (2, 2.8) {\includegraphics[width=0.09\textwidth]{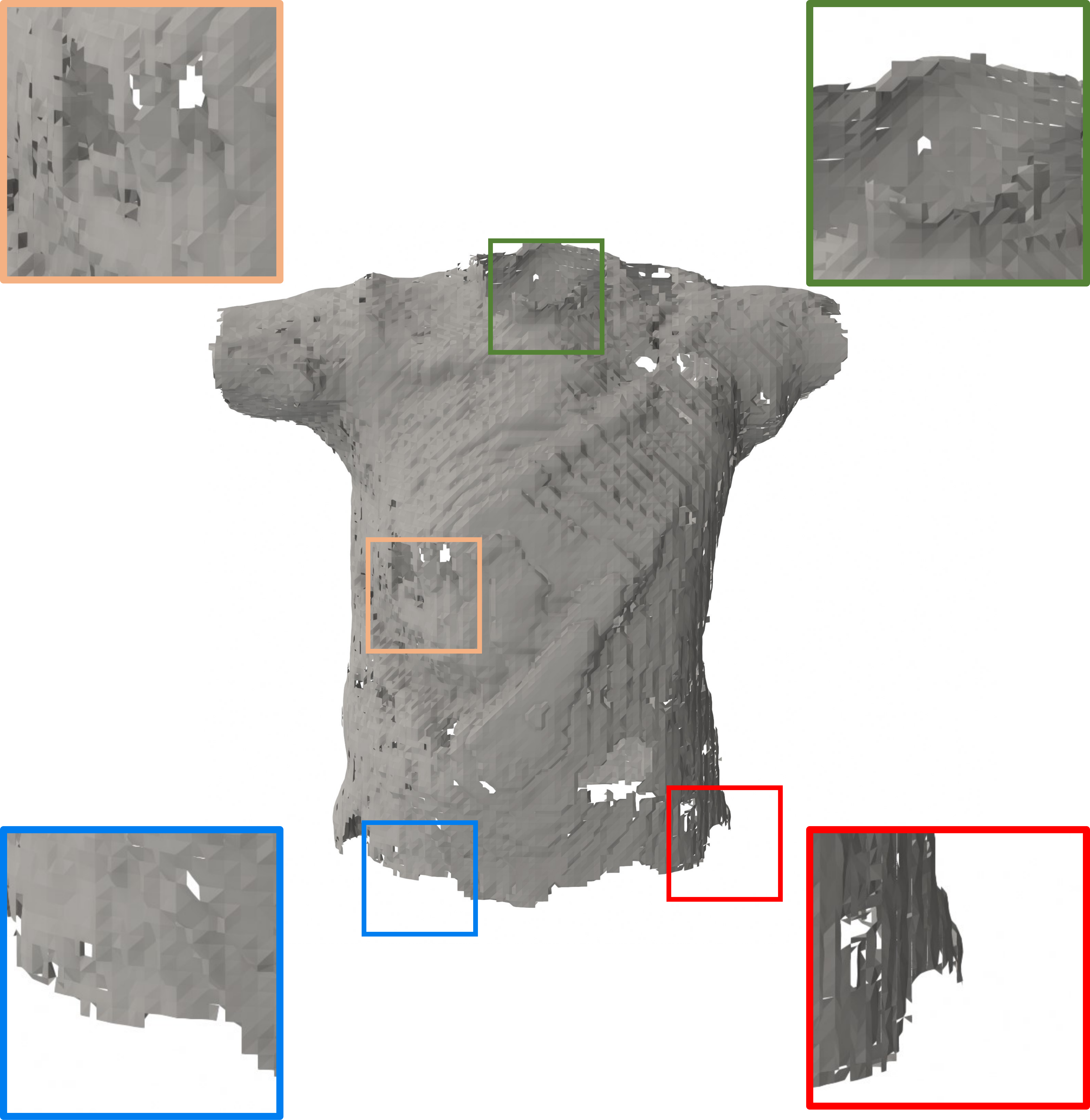}};
				\node[] (c) at (2, 1) {\includegraphics[width=0.09\textwidth]{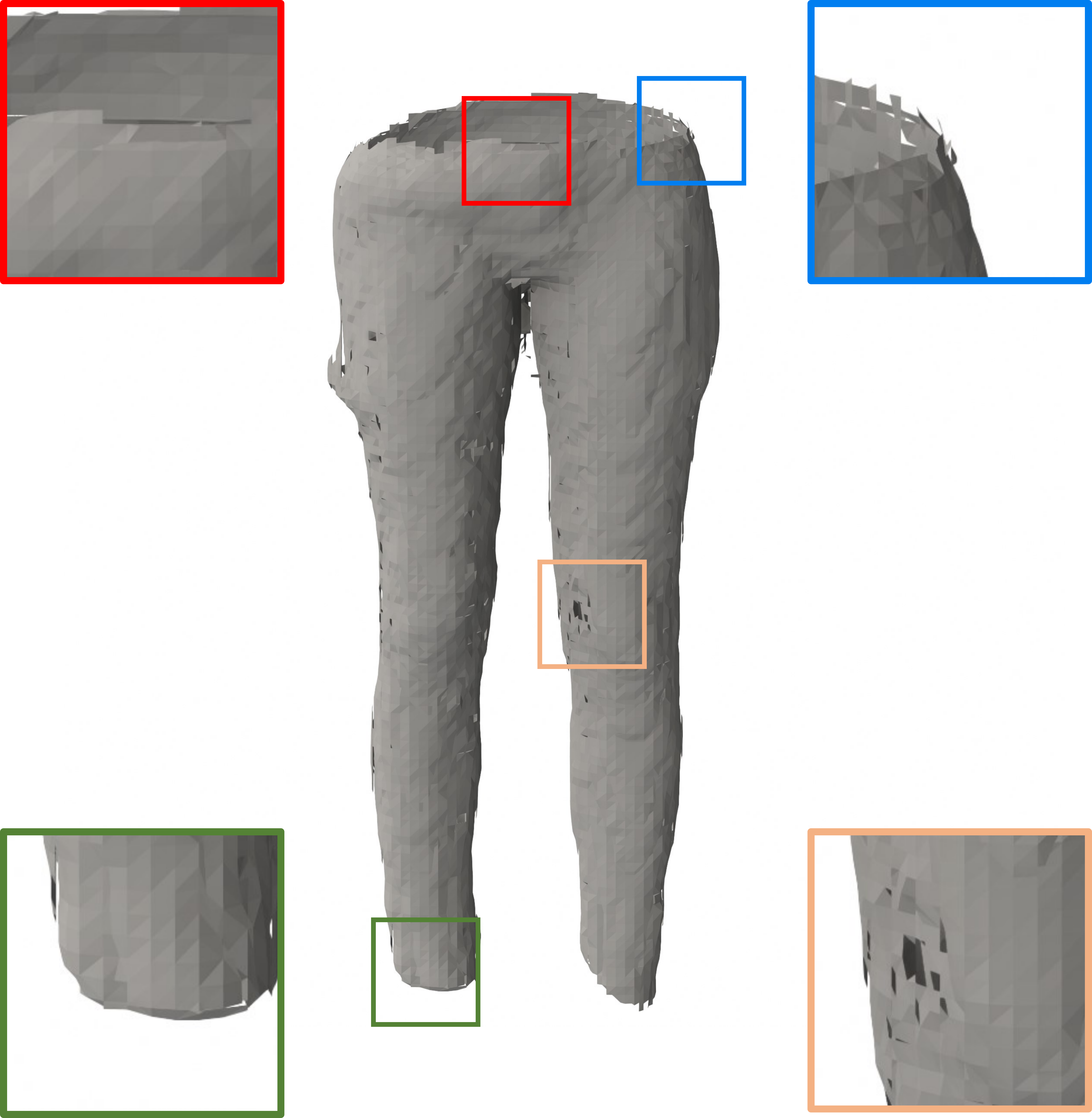}};
				\node[] (c) at (2, 0) {\small{(c) GIFS \cite{GIFS}} };
				
				\node[] (b) at (0, 2.8) {\includegraphics[width=0.09\textwidth]{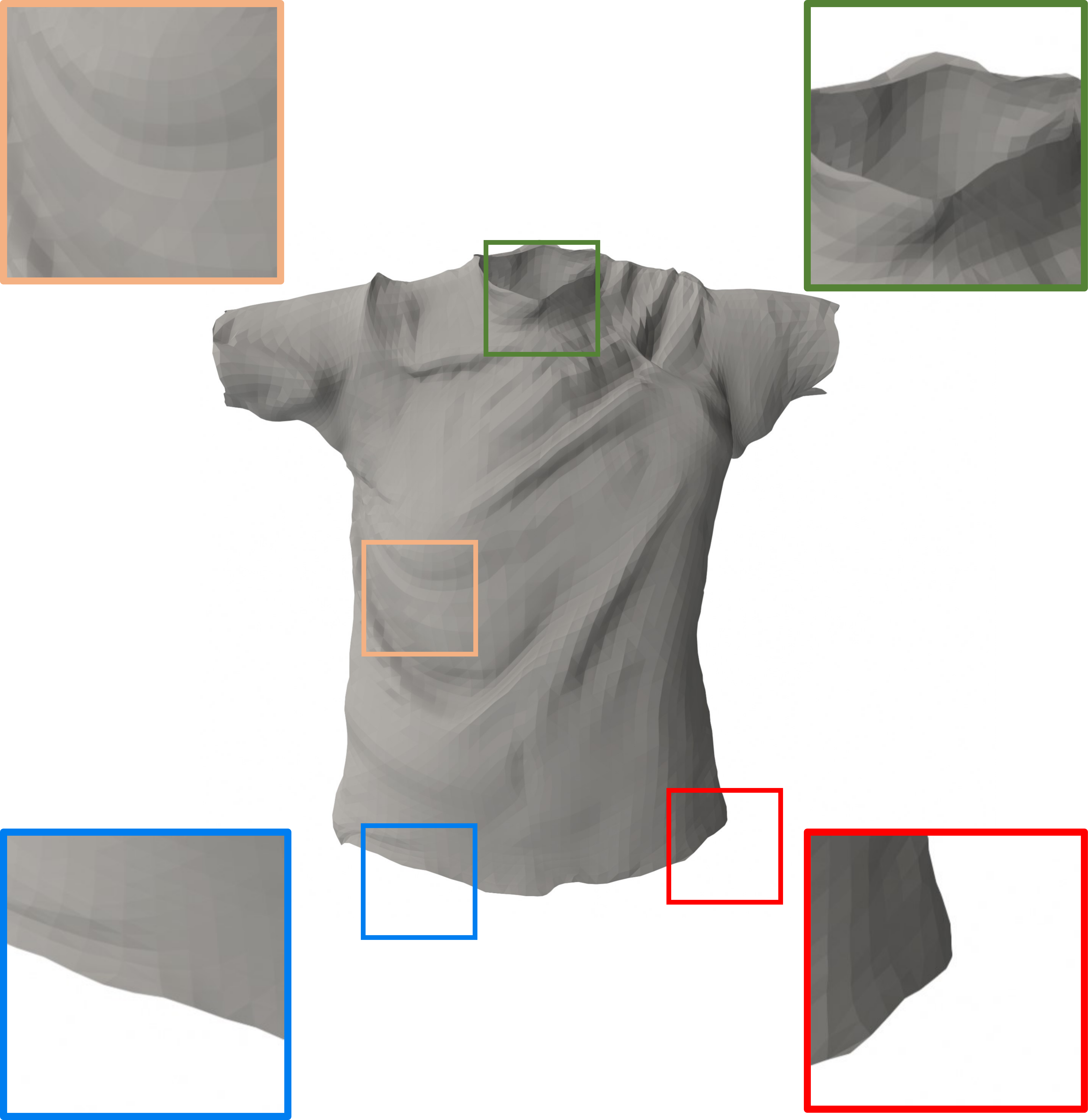}};
				\node[] (b) at (0, 1) {\includegraphics[width=0.09\textwidth]{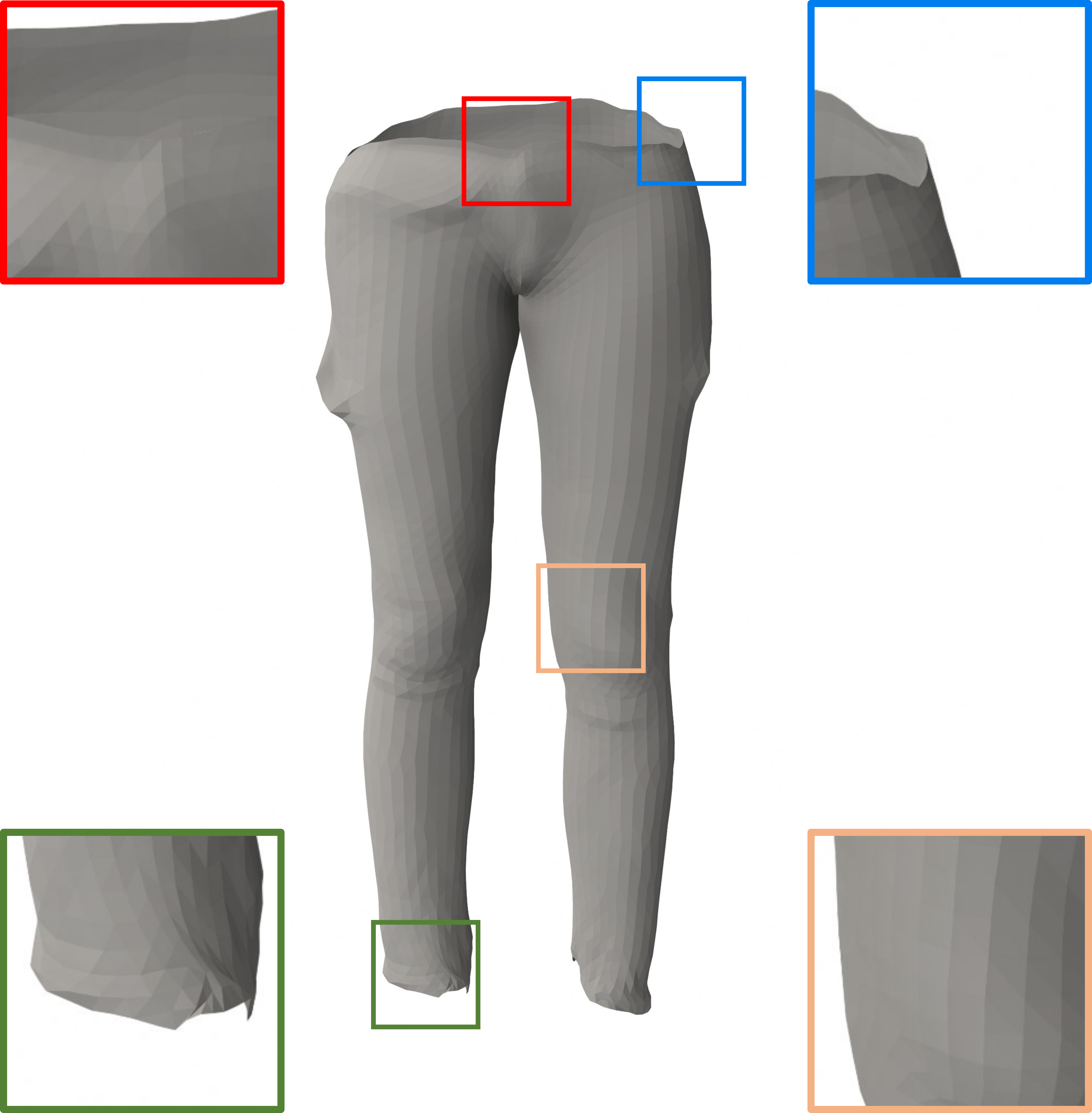}};
				\node[] (b) at (0, 0) {\small{(b) GT} };
				
				\node[] (a) at (-2, 2.8) {\includegraphics[width=0.09\textwidth]{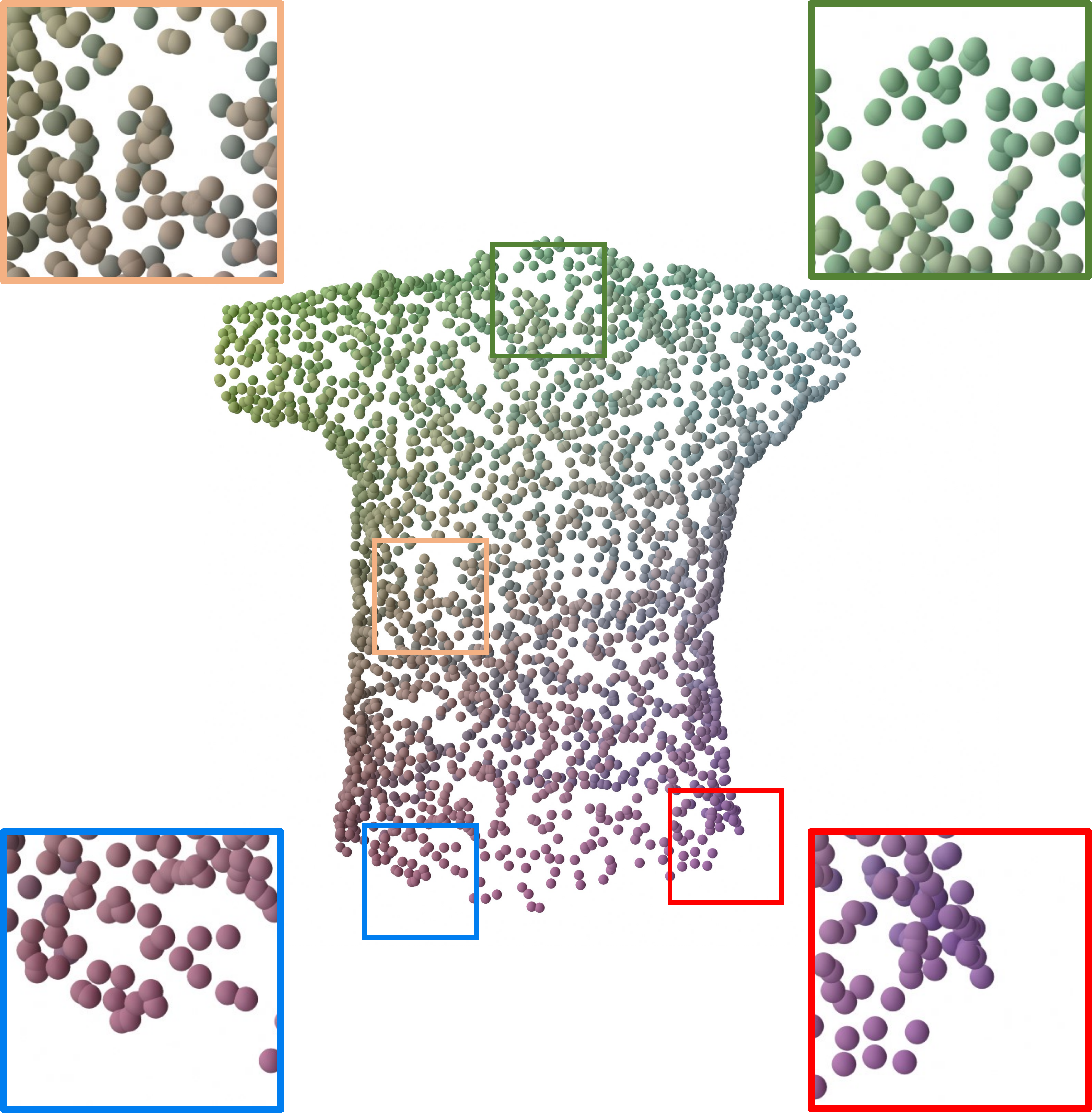} };
				\node[] (a) at (-2, 1) {\includegraphics[width=0.09\textwidth]{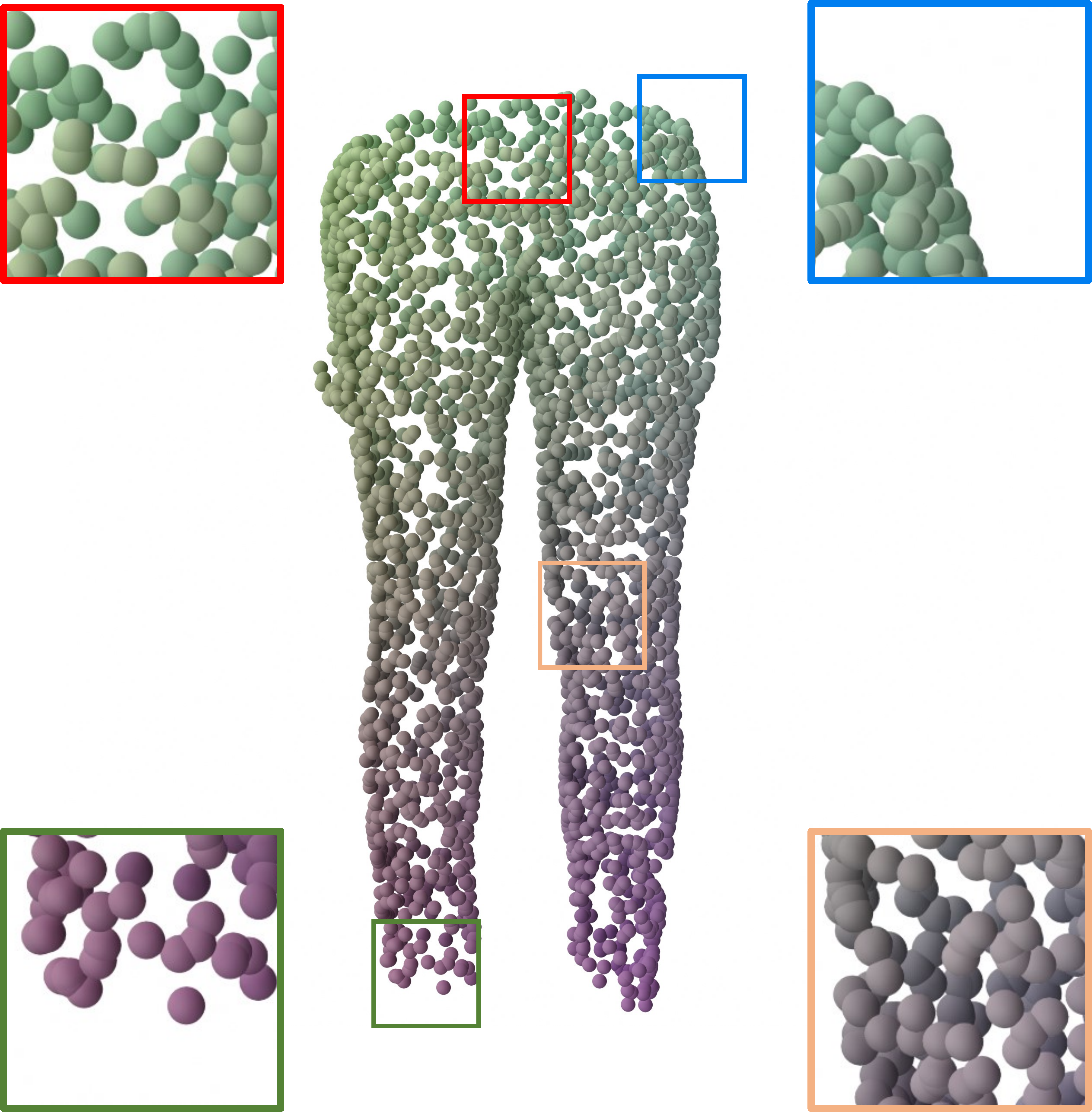} };
				\node[] (a) at (-2, 0) {\small{(a) Input} };
				
			\end{tikzpicture}
		}
		\setlength{\abovecaptionskip}{-0.1cm}
		\caption{{Visual comparisons on the MGN dataset \cite{MGN}.} \color{blue}{\faSearch~} Zoom in to see details.}
		\label{MGN:VIS} \vspace{-0.5cm}
	\end{figure}
	
	\begin{figure*}[tbp]
		\centering
		{
			\begin{tikzpicture}[]
				\node[] (f) at (12+0.2,3.3) {\includegraphics[width=0.15\textwidth]{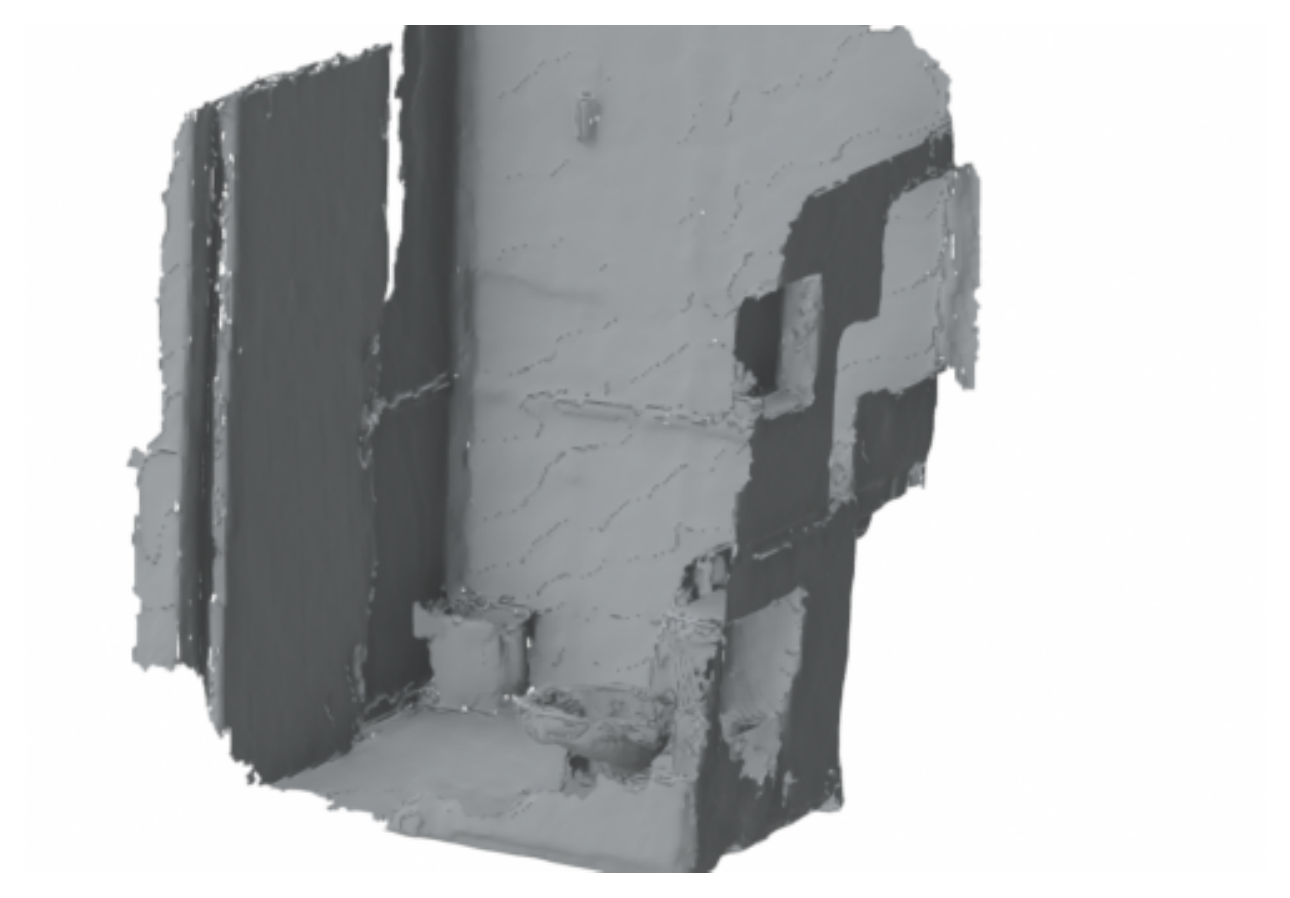}};
				\node[] (f) at (12,1.4) {\includegraphics[width=0.15\textwidth]{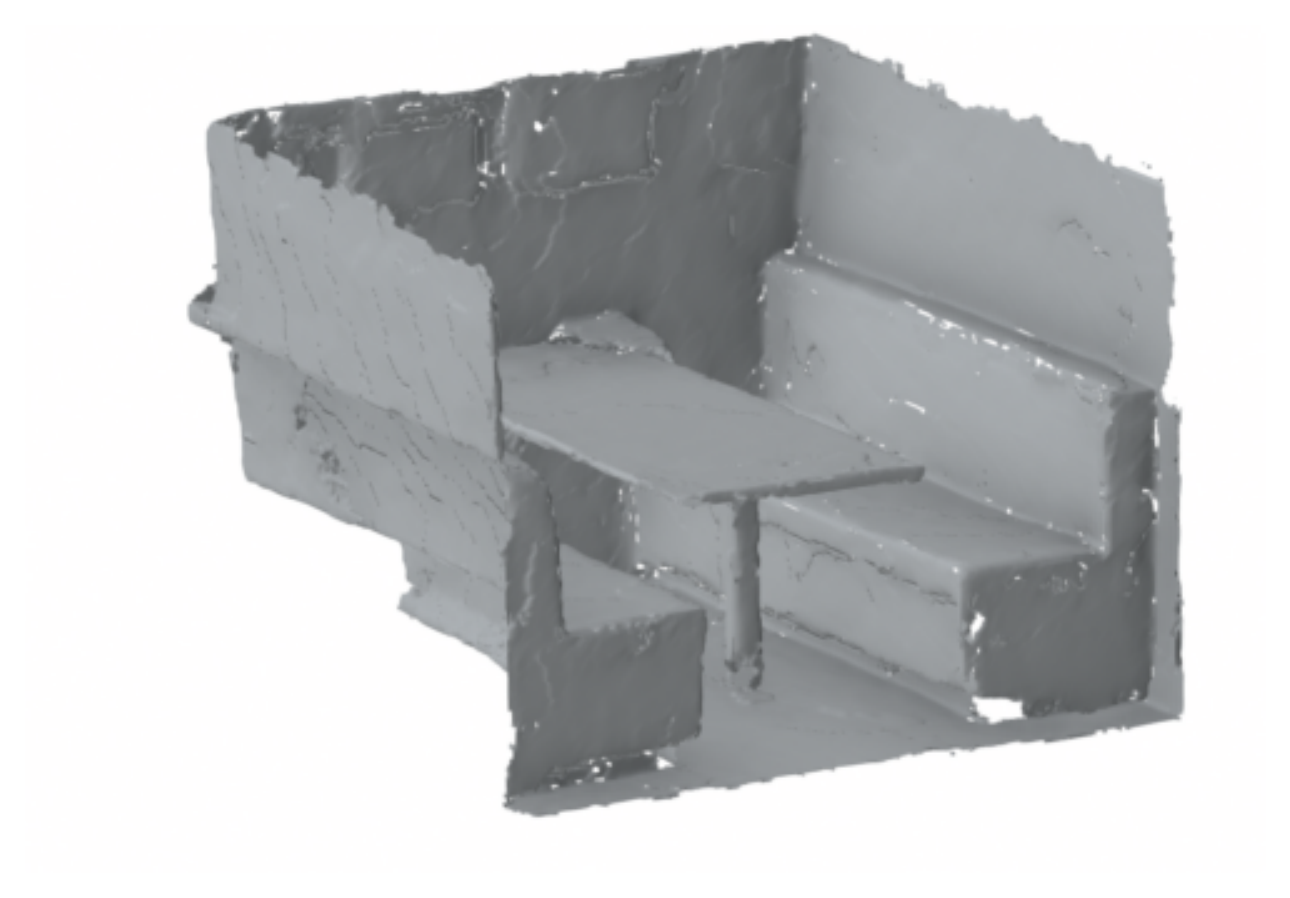}};
				\node[] (f) at (12,0.3) {\small{(f) Ours} };
				
				\node[] (e) at (12/5*4+0.2,3.3) {\includegraphics[width=0.15\textwidth]{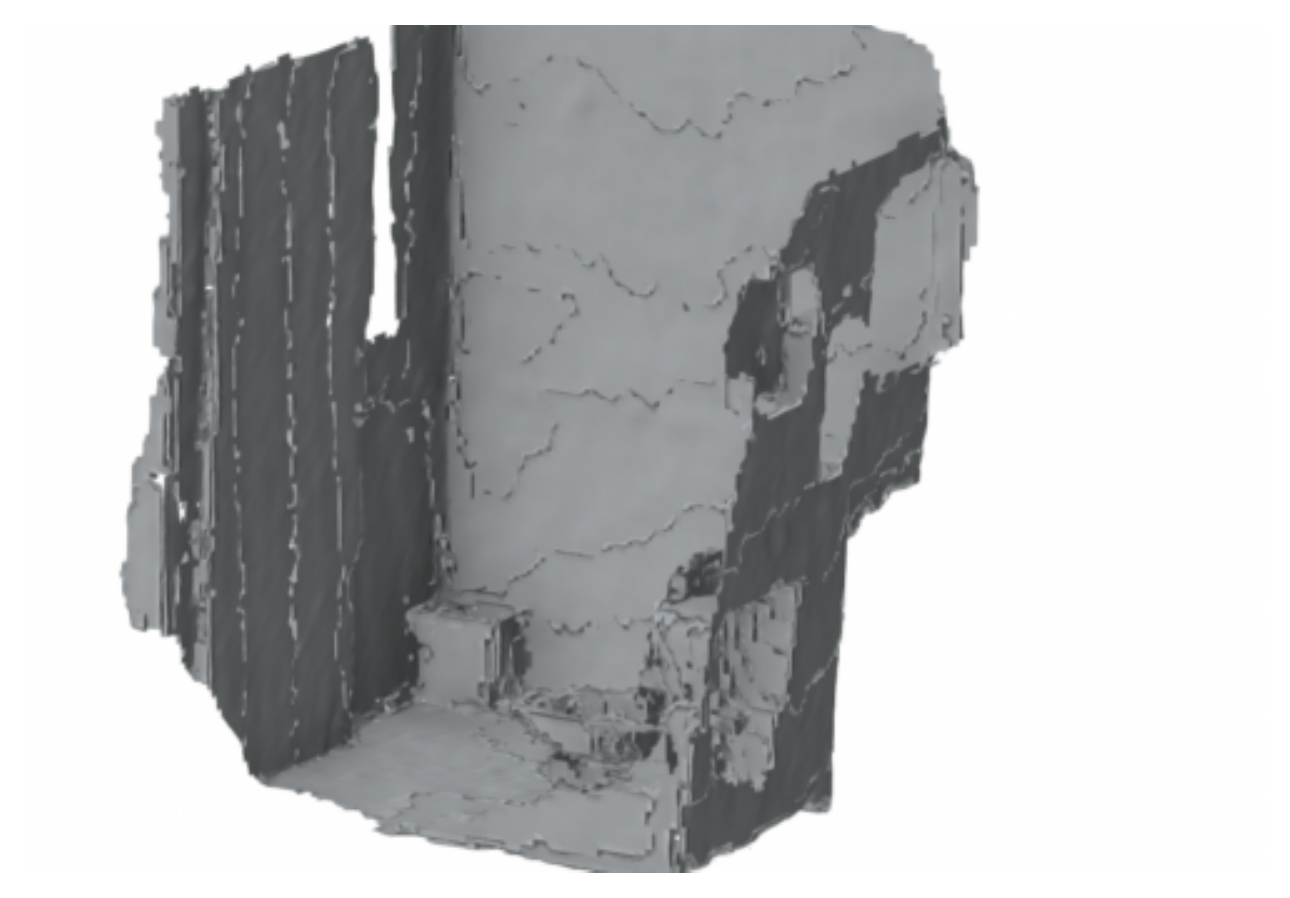}};
				\node[] (e) at (12/5*4,1.4) {\includegraphics[width=0.15\textwidth]{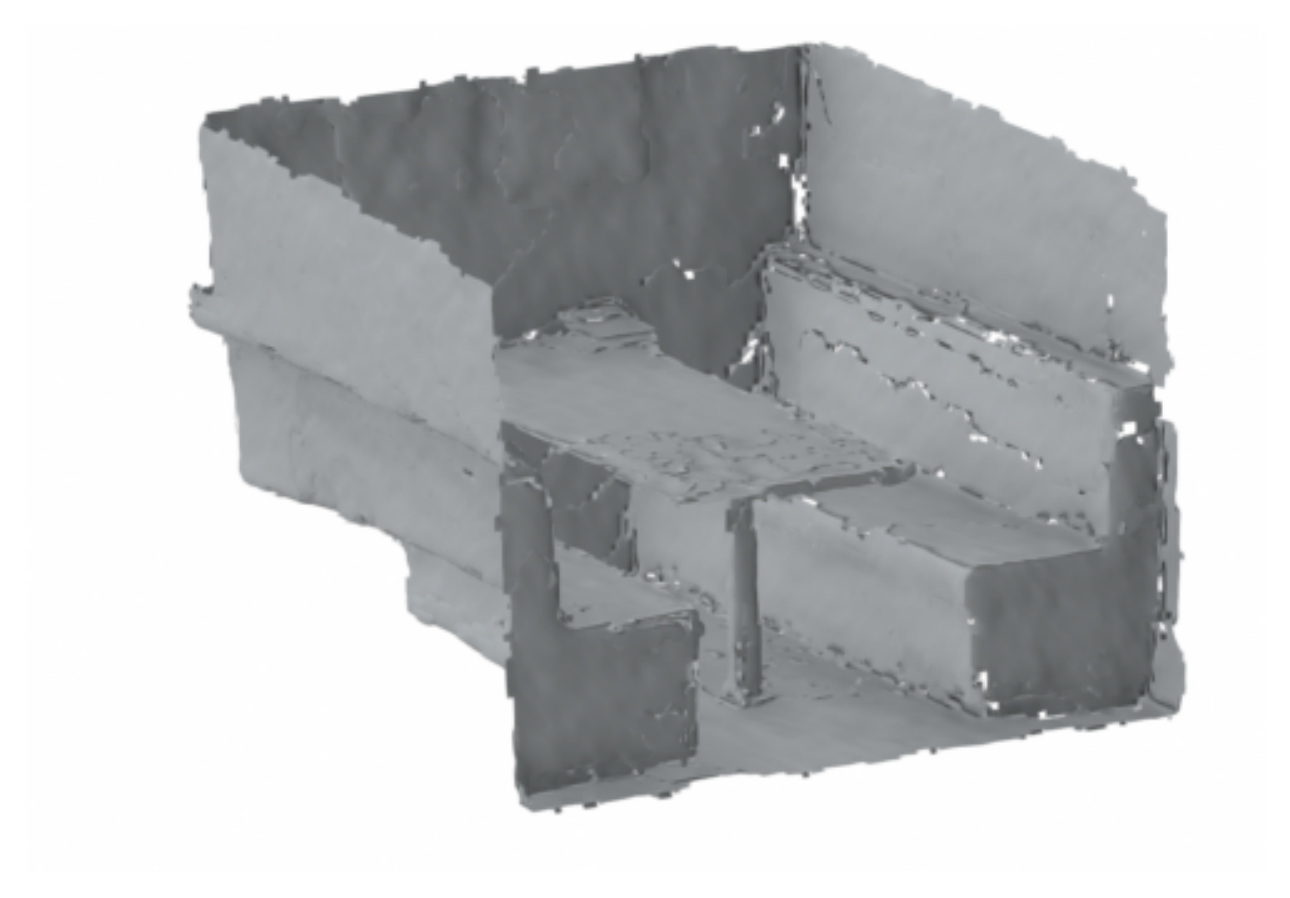}};
				\node[] (e) at (12/5*4,0.3) {\small{(e) GIFS \cite{GIFS}} };
				
				\node[] (d) at (12/5*3+0.2,3.3) {\includegraphics[width=0.15\textwidth]{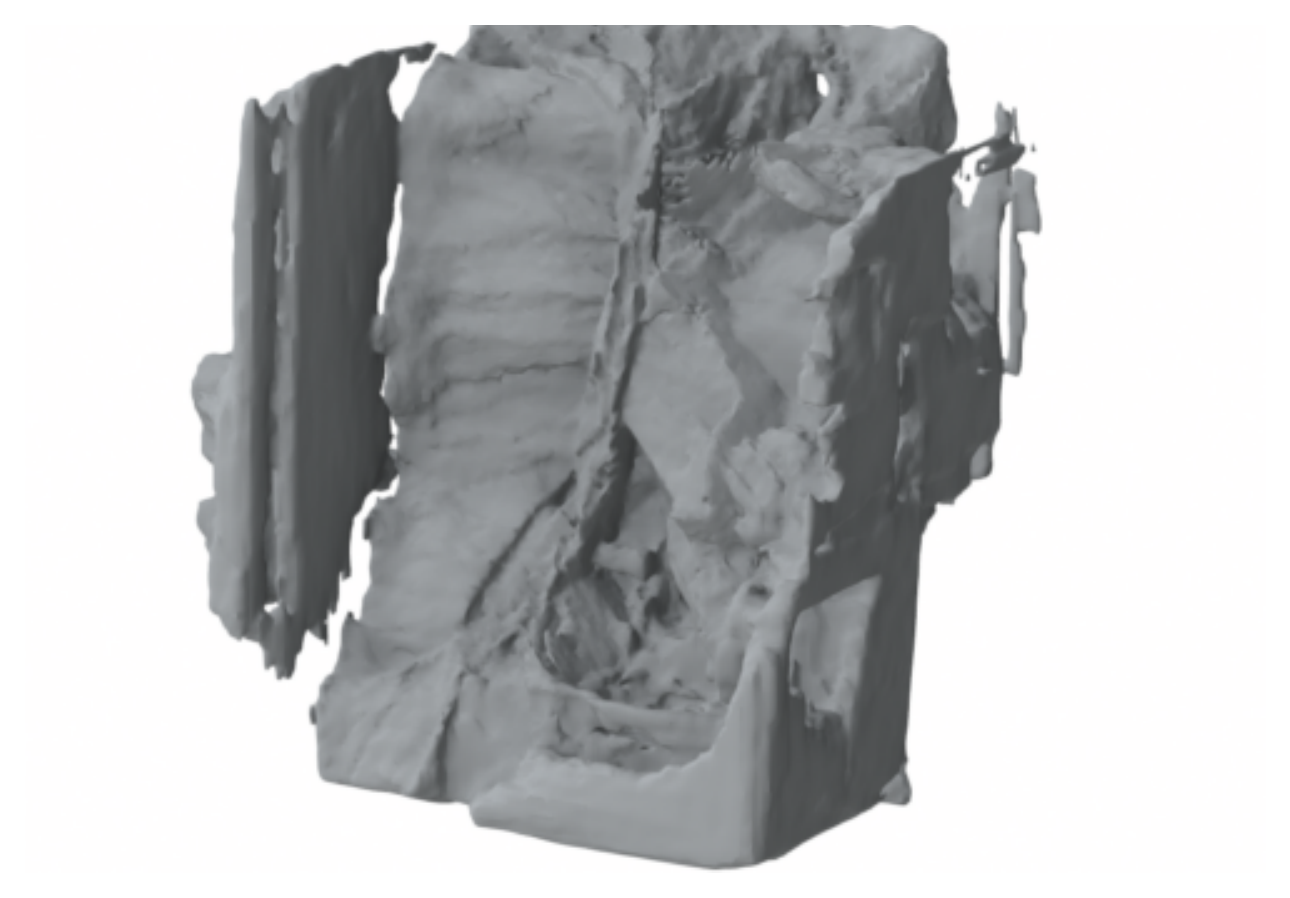}};
				\node[] (d) at (12/5*3,1.4) {\includegraphics[width=0.15\textwidth]{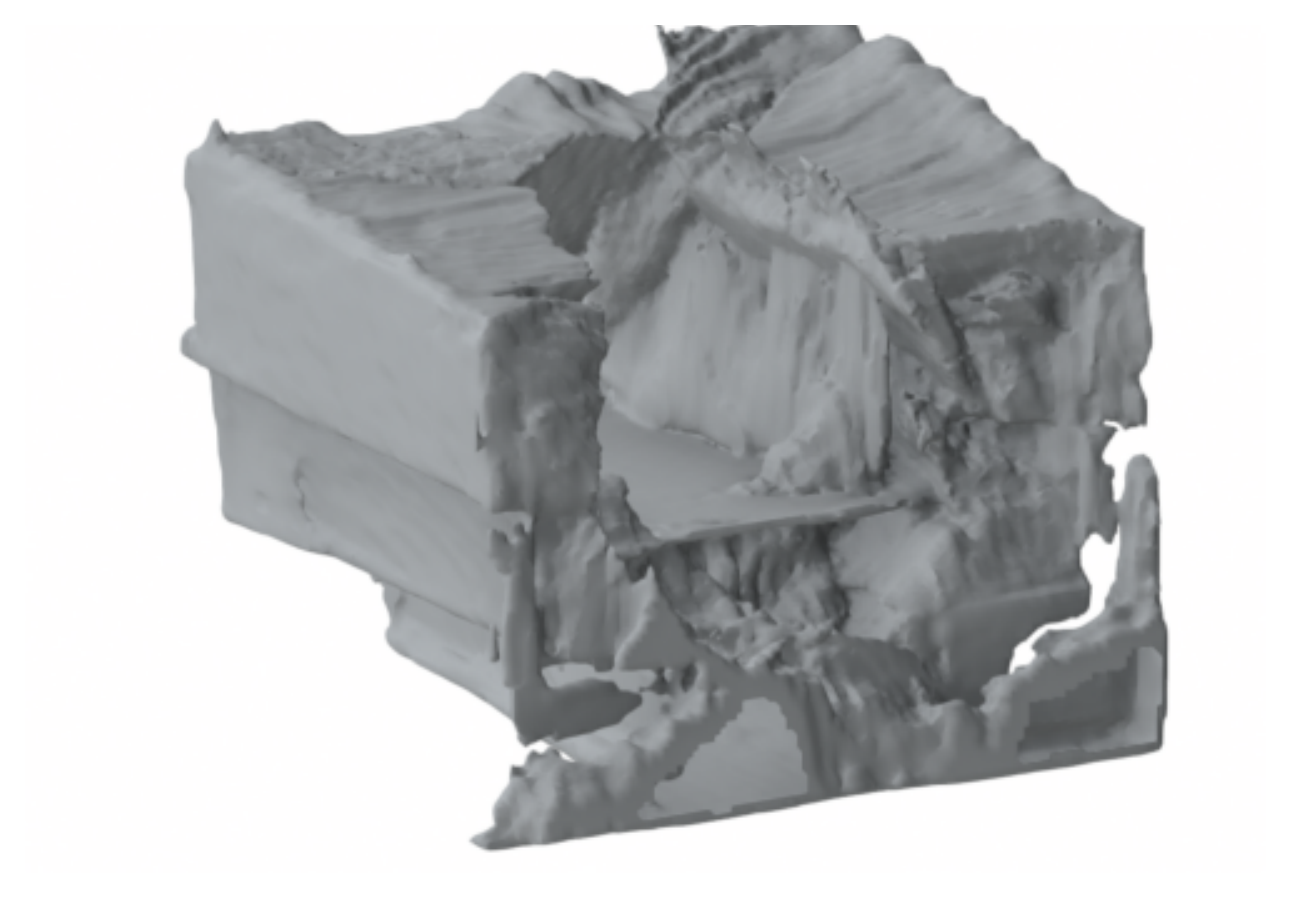}};
				\node[] (d) at (12/5*3,0.3) {\small{(d) POCO \cite{POCO}} };
				
				\node[] (c) at (12/5*2+0.2,3.3) {\includegraphics[width=0.15\textwidth]{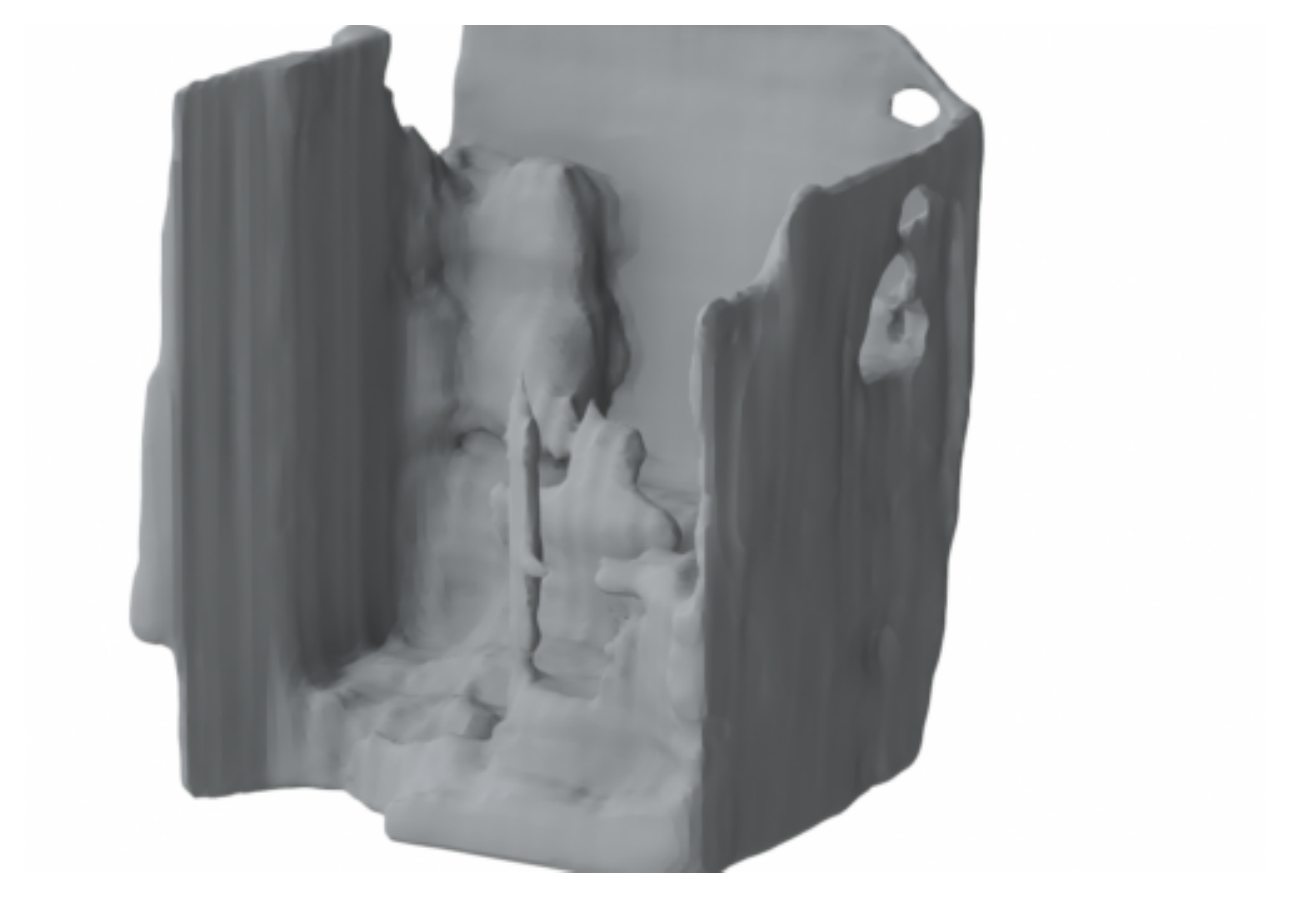}};
				\node[] (c) at (12/5*2,1.4) {\includegraphics[width=0.15\textwidth]{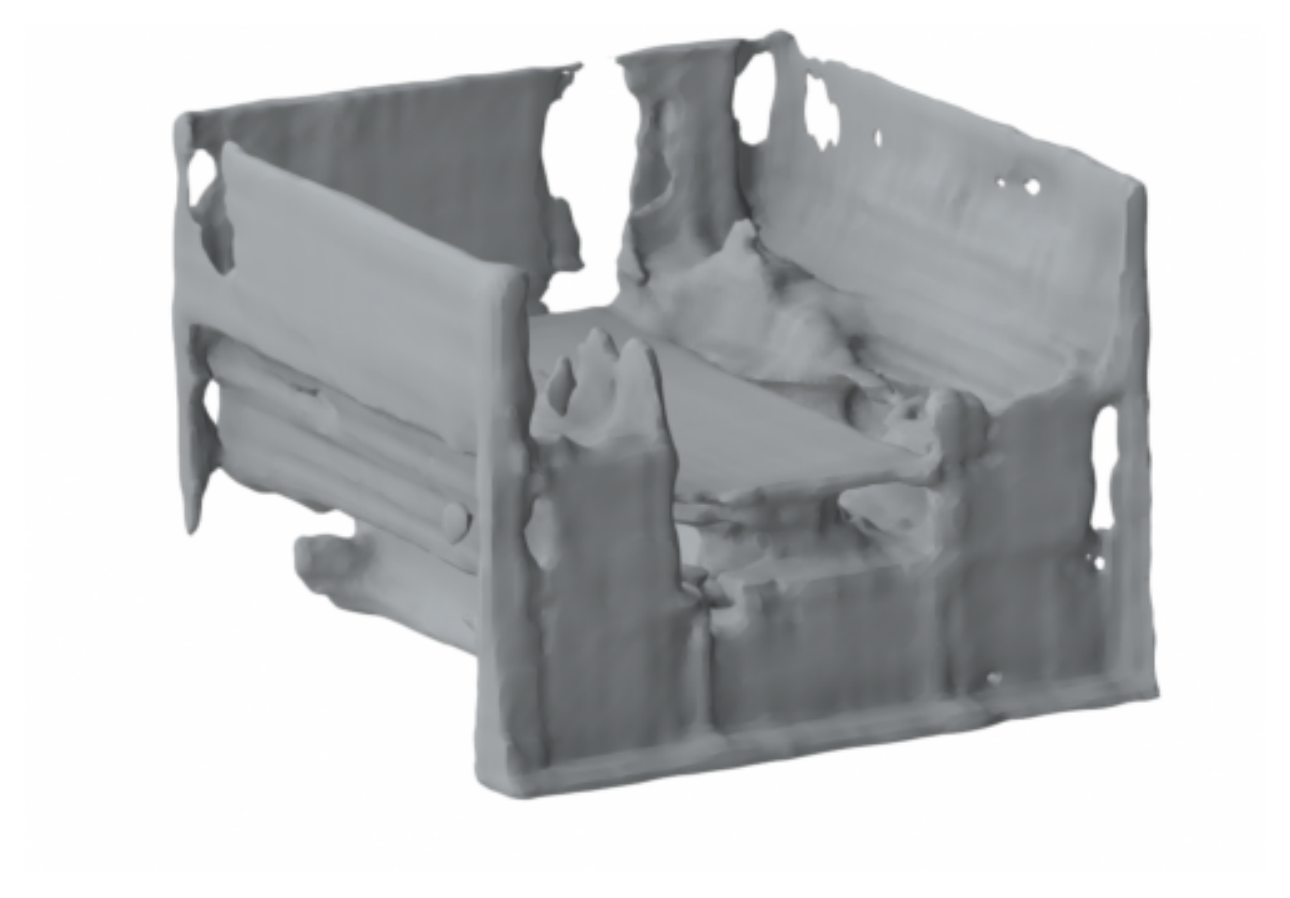}};
				\node[] (c) at (12/5*2,0.3) {\small{(c) CONet \cite{CONVOCCNET} }};
				
				\node[] (b) at (12/5+0.2,3.3) {\includegraphics[width=0.15\textwidth]{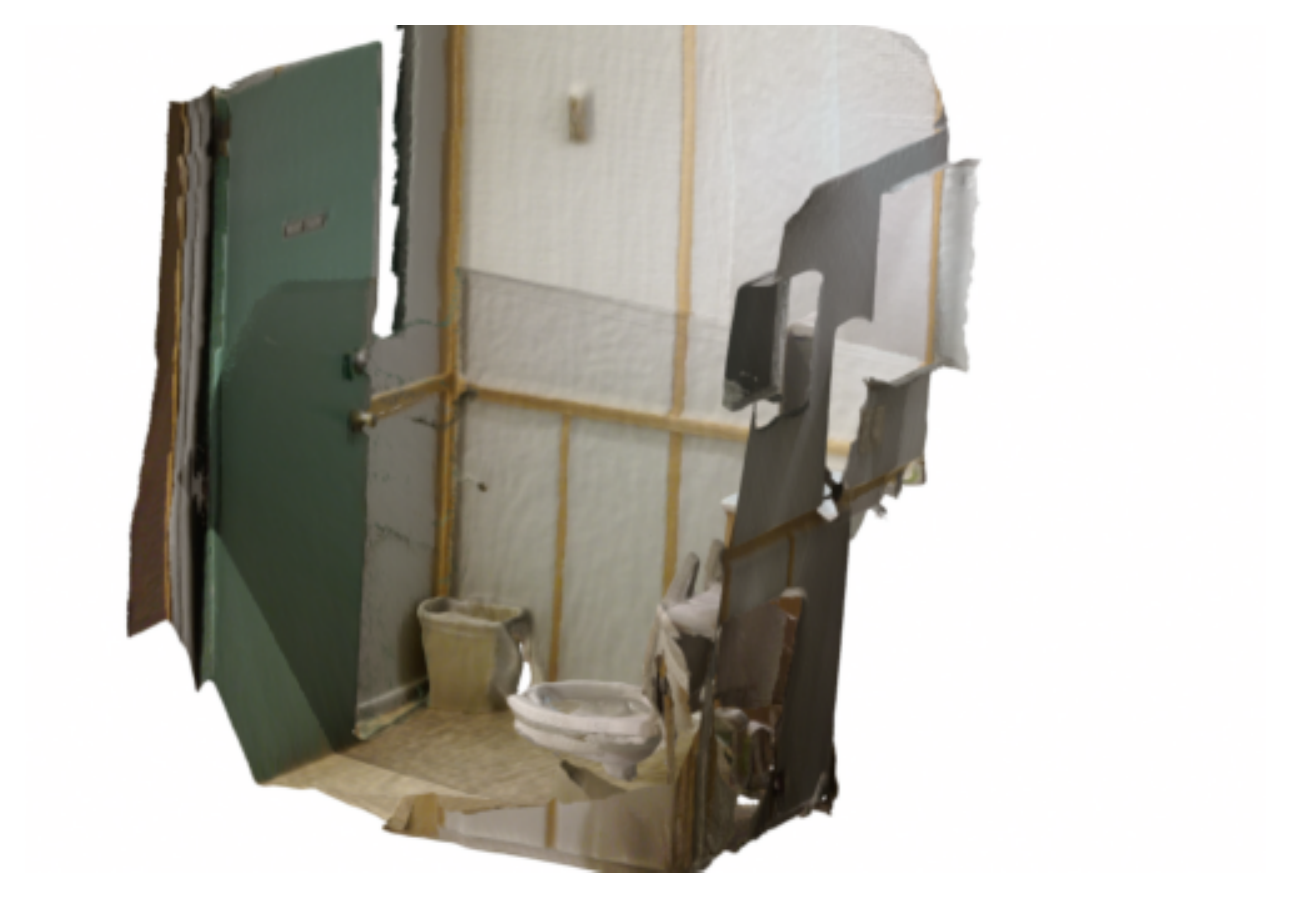}};
				\node[] (b) at (12/5,1.4) {\includegraphics[width=0.15\textwidth]{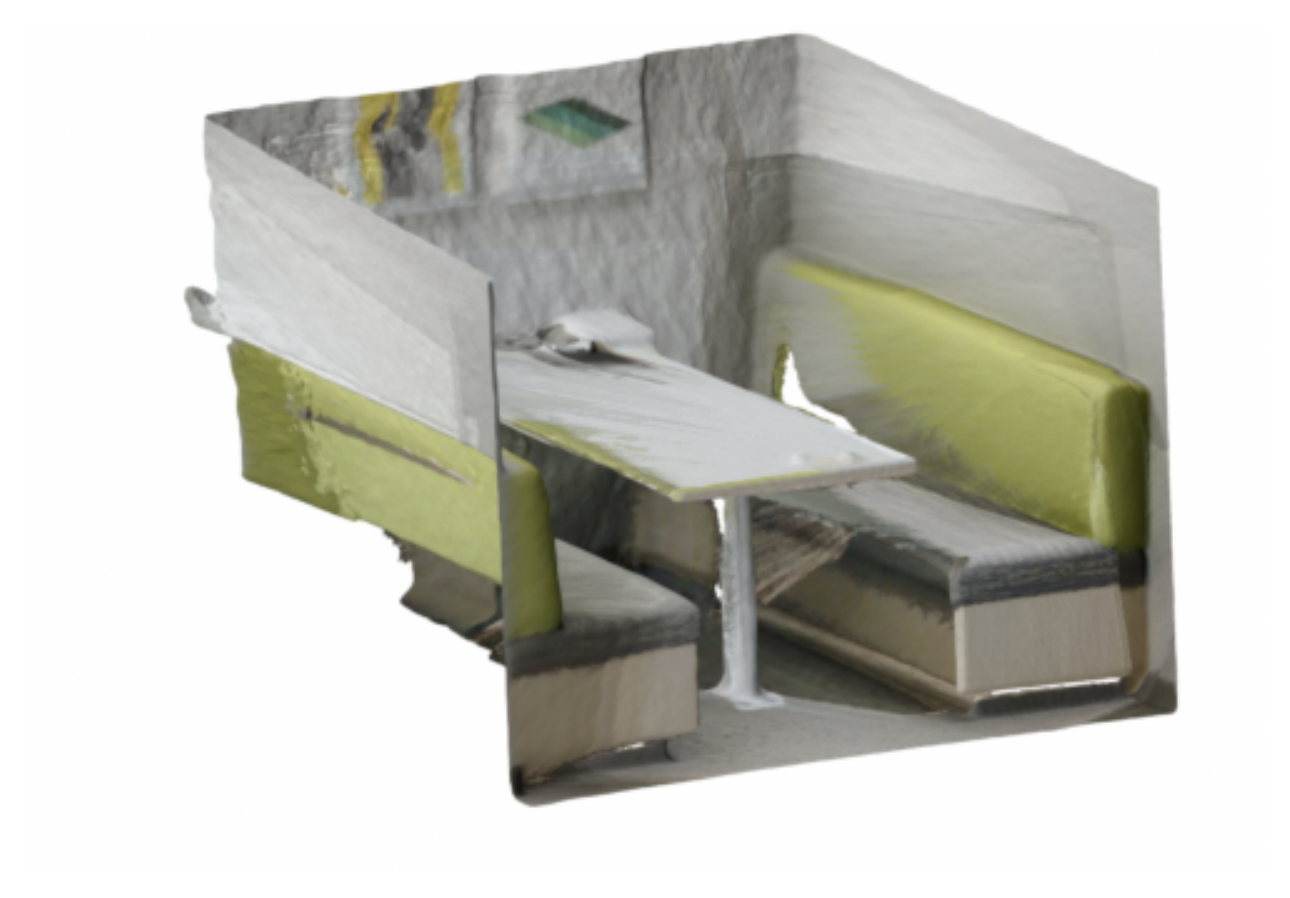}};
				\node[] (b) at (12/5,0.3) {\small{(b) GT} };
				
				\node[] (a) at (0.2,3.3) {\includegraphics[width=0.15\textwidth]{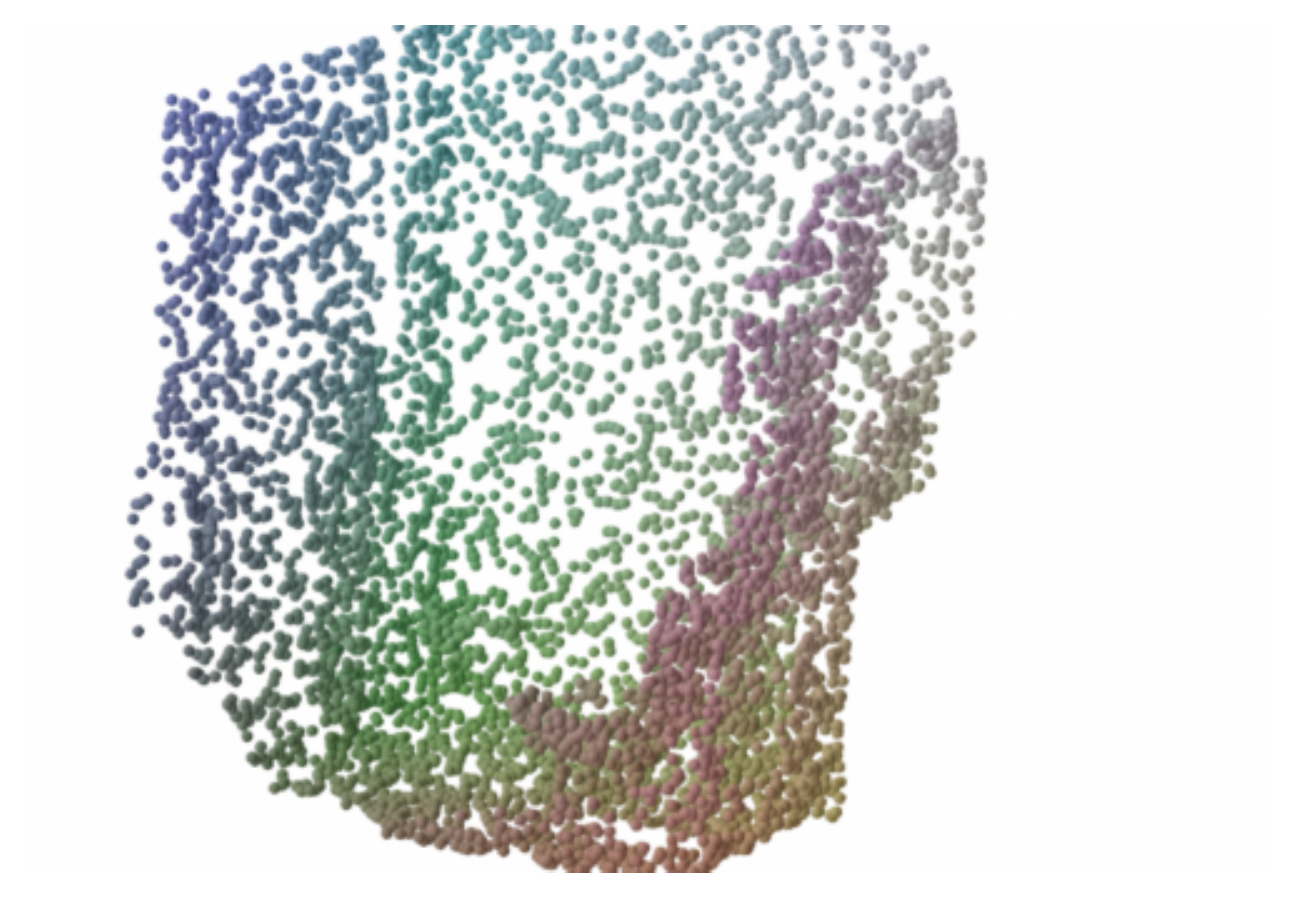}};
				\node[] (a) at (0,1.4) {\includegraphics[width=0.15\textwidth]{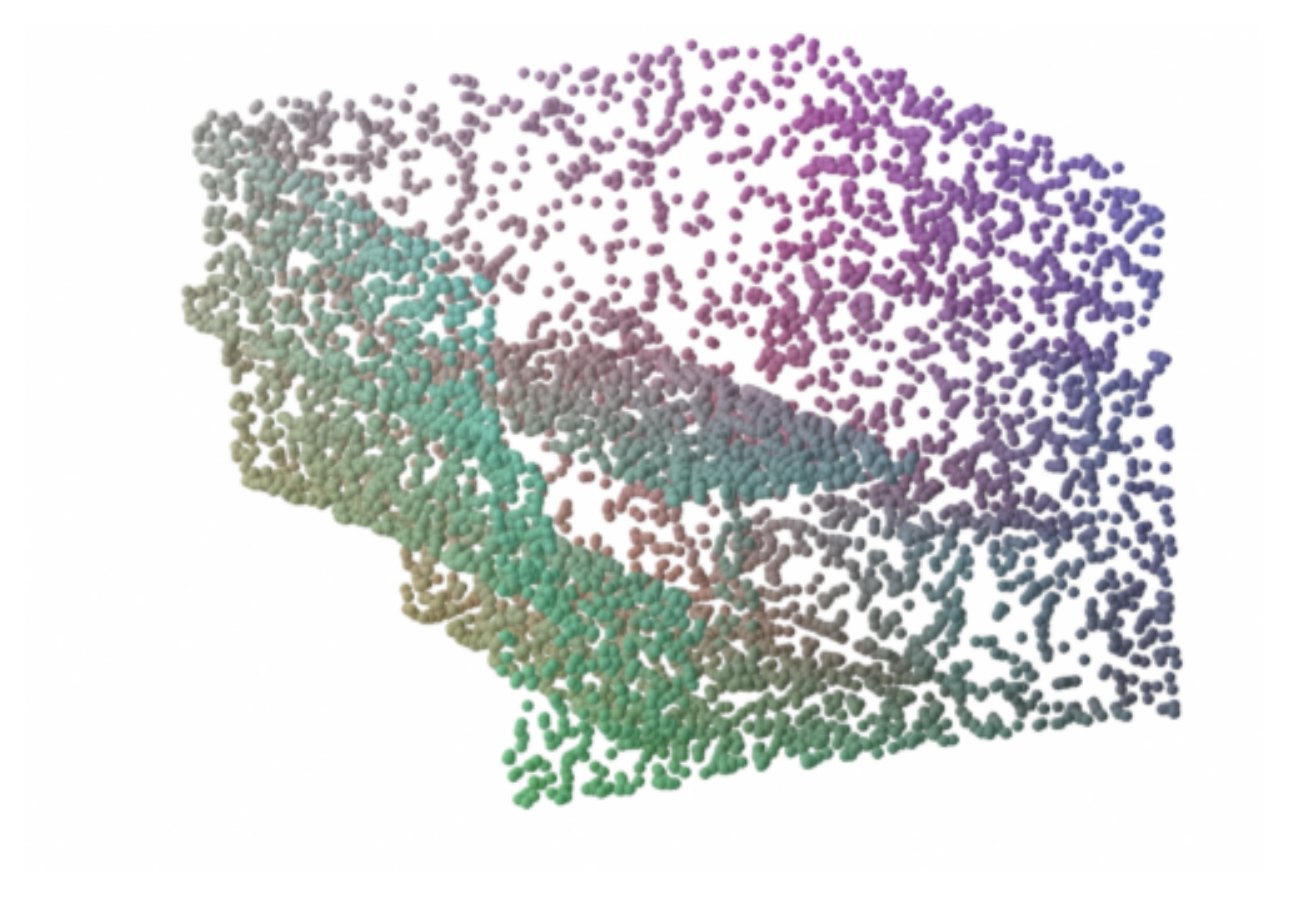}};
				\node[] (a) at (0,0.3) {\small{(a) Input}};
				
			\end{tikzpicture}
		}
		\setlength{\abovecaptionskip}{-0.15cm}
		\caption{Visual comparisons of different surface reconstruction methods on the ScanNet dataset  \cite{SCANNET}.}
		\label{SCANNET:VIS}\vspace{-0.5cm}
	\end{figure*}
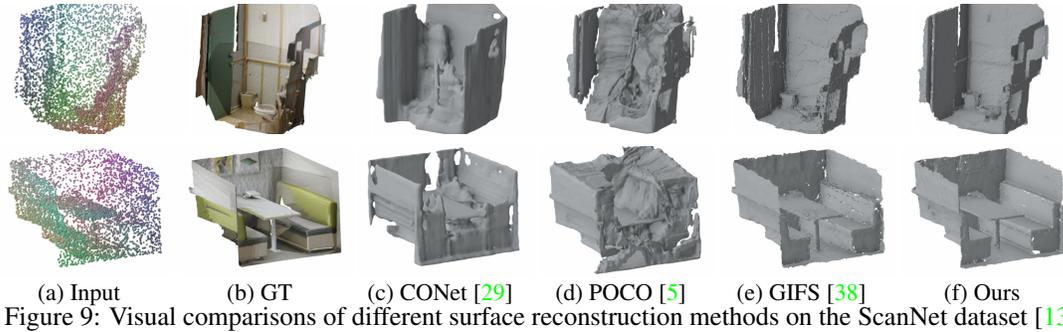
	
	\subsection{Unseen Non-watertight Surface Reconstruction}
	\noindent\textbf{Dataset and Metrics.}  Following GIFS \cite{GIFS}, we evaluated our method on the MGN \cite{MGN} and the raw ShapeNet car\footnote{These raw data were not used in the training of the experiment in Sec. \ref{subsec:eva_watertight}.} \cite{SHAPENET} datasets, containing the shapes with boundaries and interior structures. 
	Besides, we also evaluated our method on the ScanNet dataset \cite{SCANNET}, in which the point clouds were collected through an RGB-D camera from the real scenes. For the MGN dataset, we randomly sampled 3000 points from the surface as the input.
	As for the shapes and scenes in the ShapeNet car and ScanNet dataset, we randomly sampled 6000 points from the surface or the dense point cloud as the input.
	
\begin{figure}[tbp]
		\centering
		{
			\begin{tikzpicture}[]
				\node[] (a) at (-2, 2.7) {\includegraphics[width=0.12\textwidth]{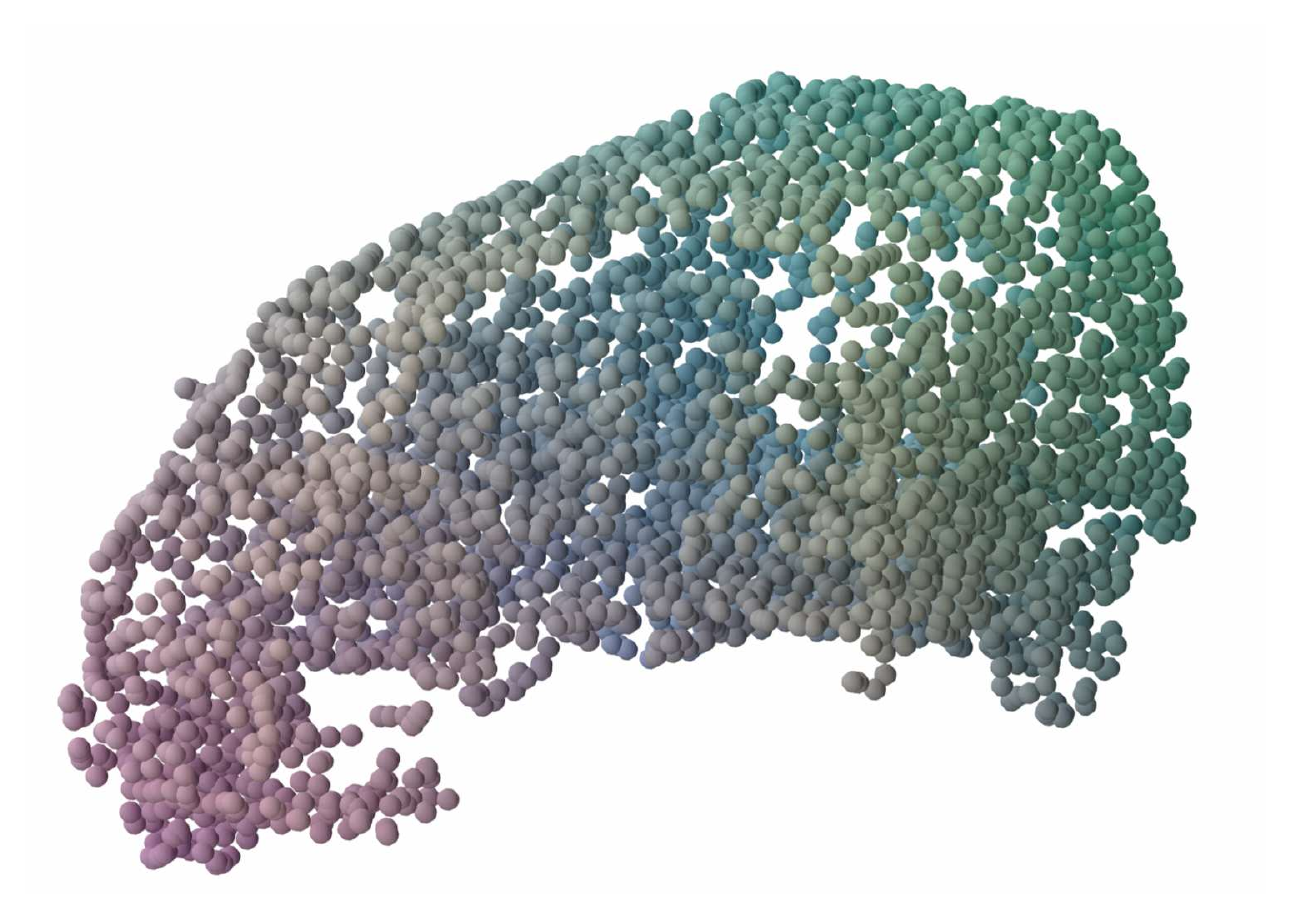} };
				\node[] (a) at (-2, 1.2) {\includegraphics[width=0.12\textwidth]{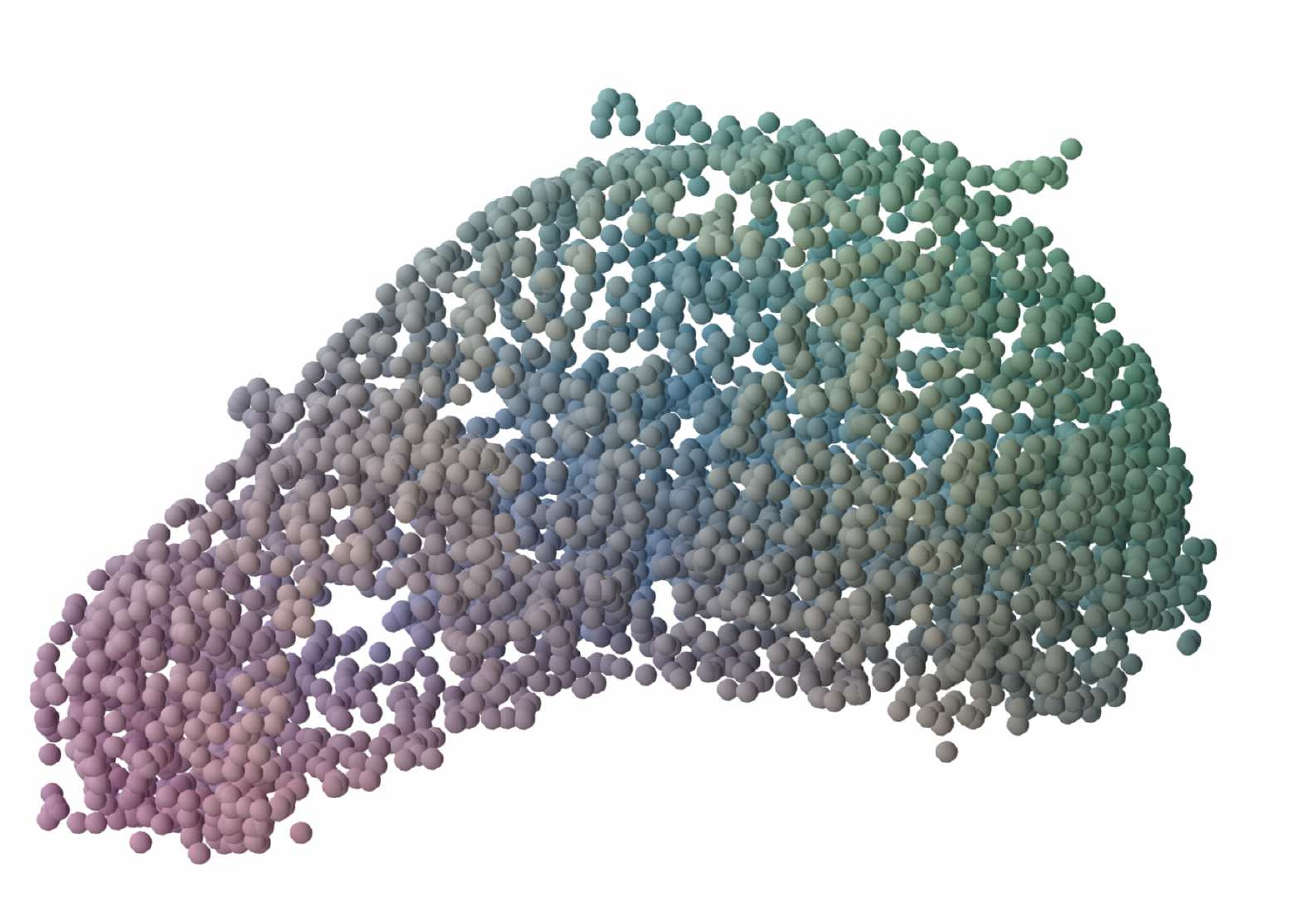} };
				\node[] (a) at (-2, 0.3) {\small{(a) Input} };
				
				\node[] (b) at (0, 2.7) {\includegraphics[width=0.12\textwidth]{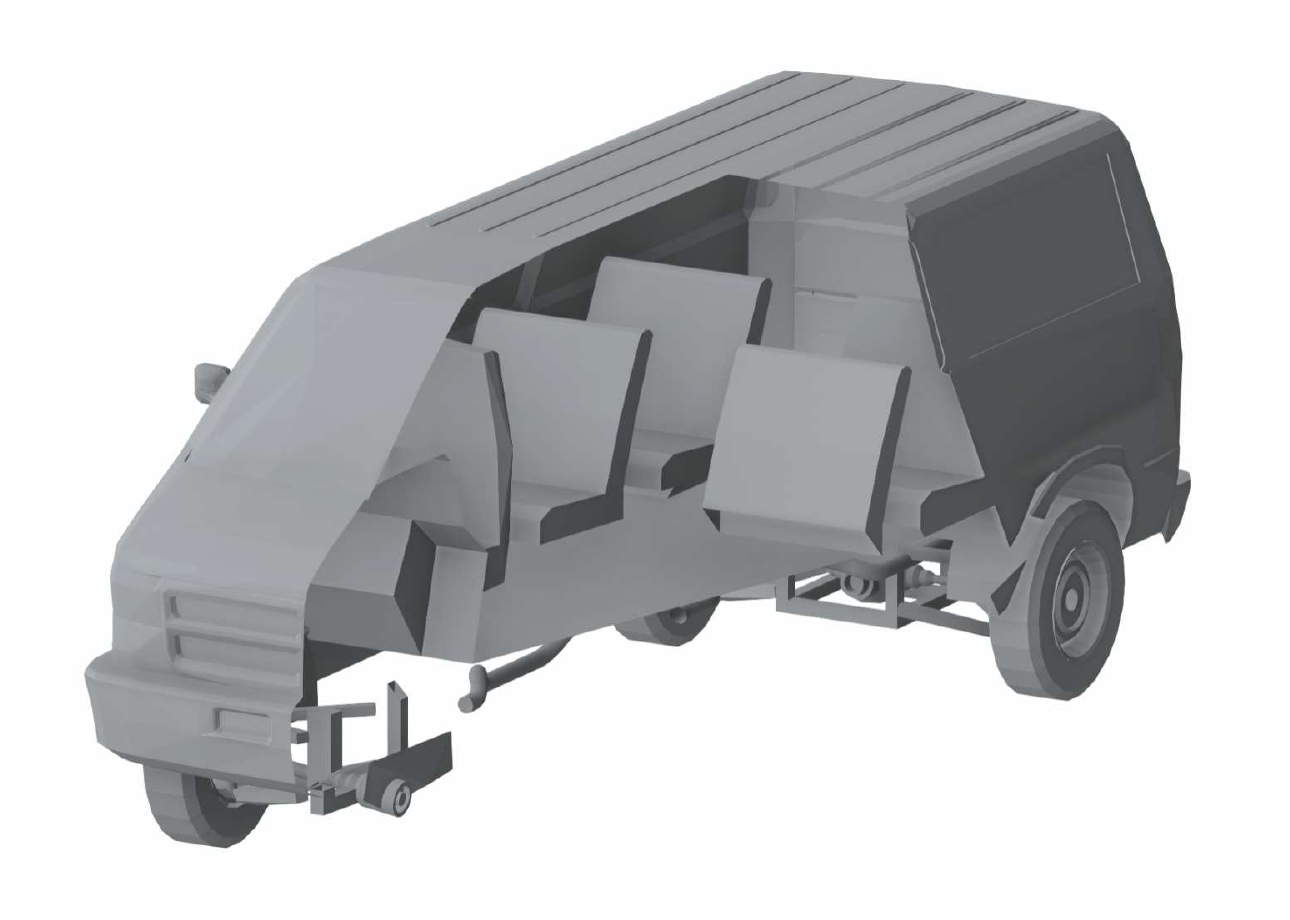}};
				\node[] (b) at (0, 1.2) {\includegraphics[width=0.12\textwidth]{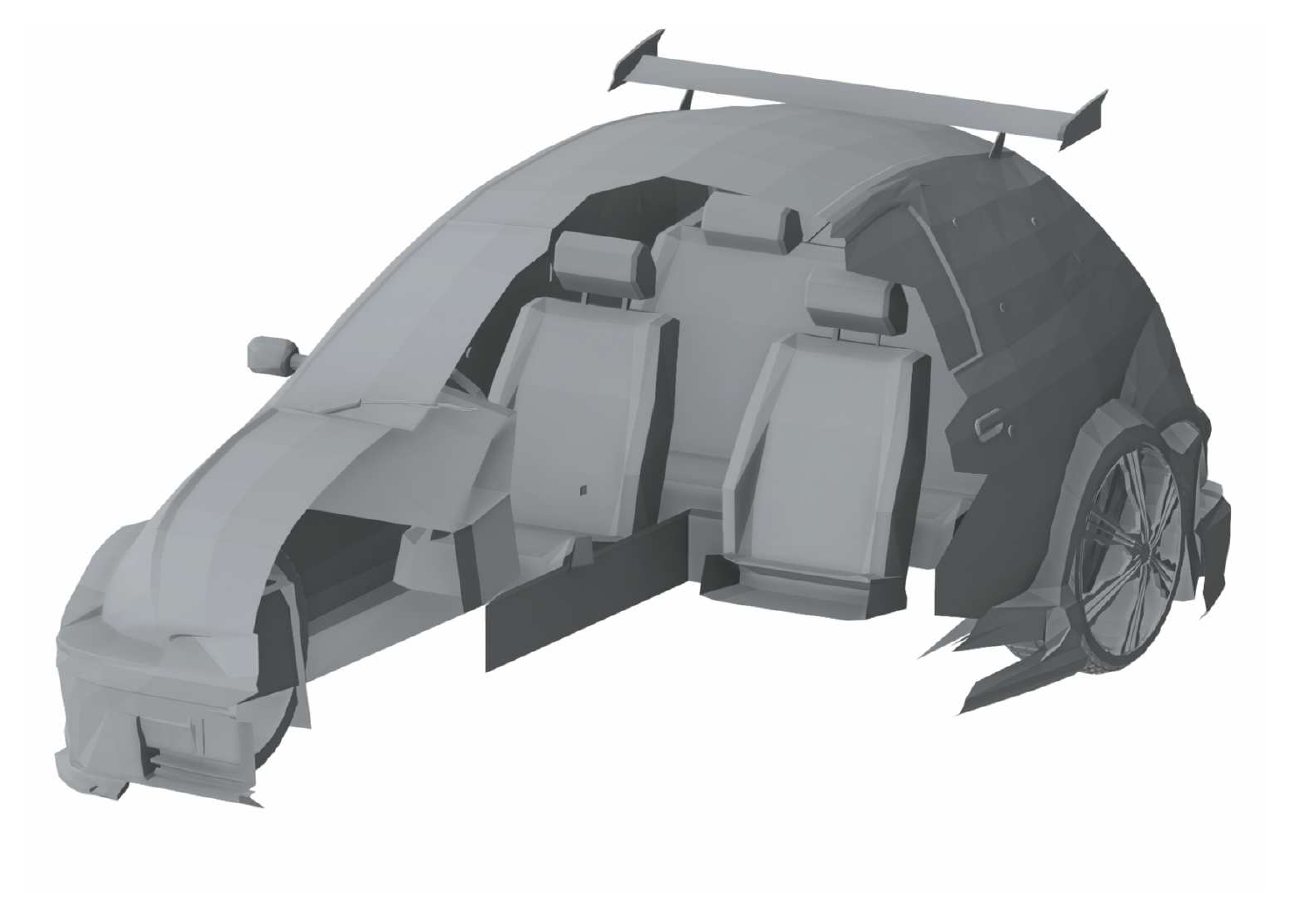}};
				\node[] (b) at (0, 0.3) {\small{(b) GT} };
				
				\node[] (c) at (2, 2.7) {\includegraphics[width=0.12\textwidth]{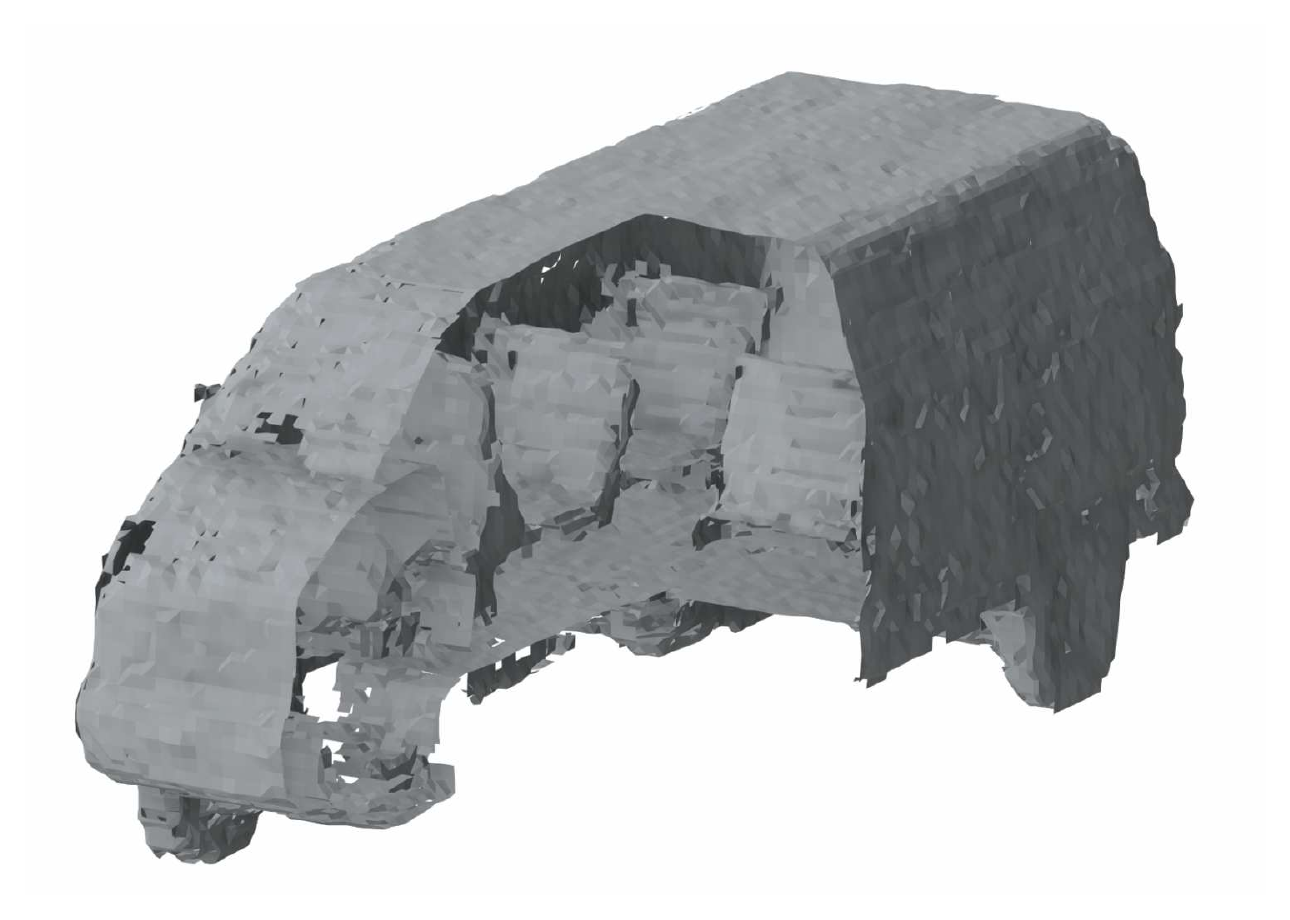}};
				\node[] (c) at (2, 1.2) {\includegraphics[width=0.12\textwidth]{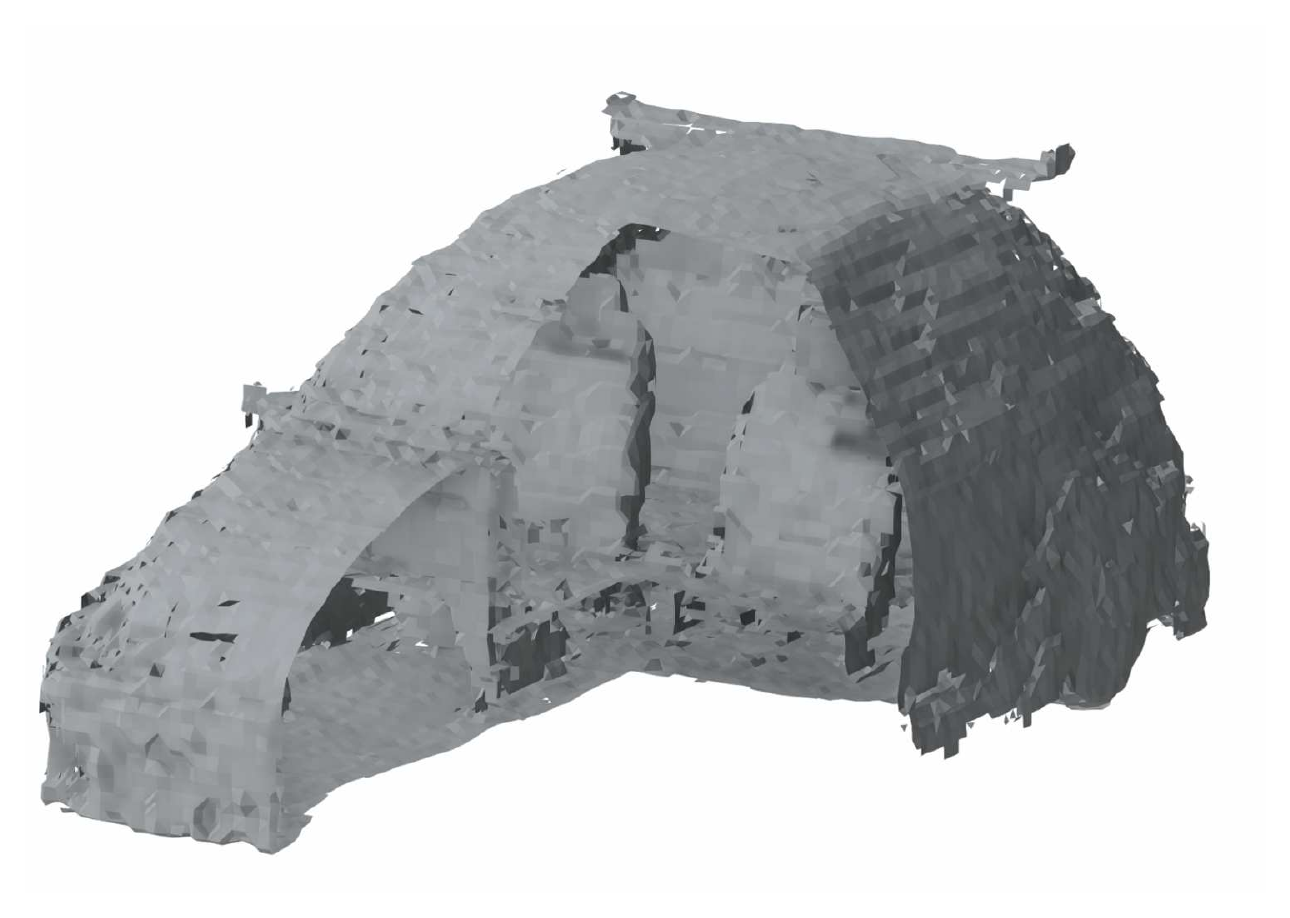}};
				\node[] (c) at (2, 0.3) {\small{(c) GIFS \cite{GIFS}} };
				
				\node[] (d) at (4, 2.7) {\includegraphics[width=0.12\textwidth]{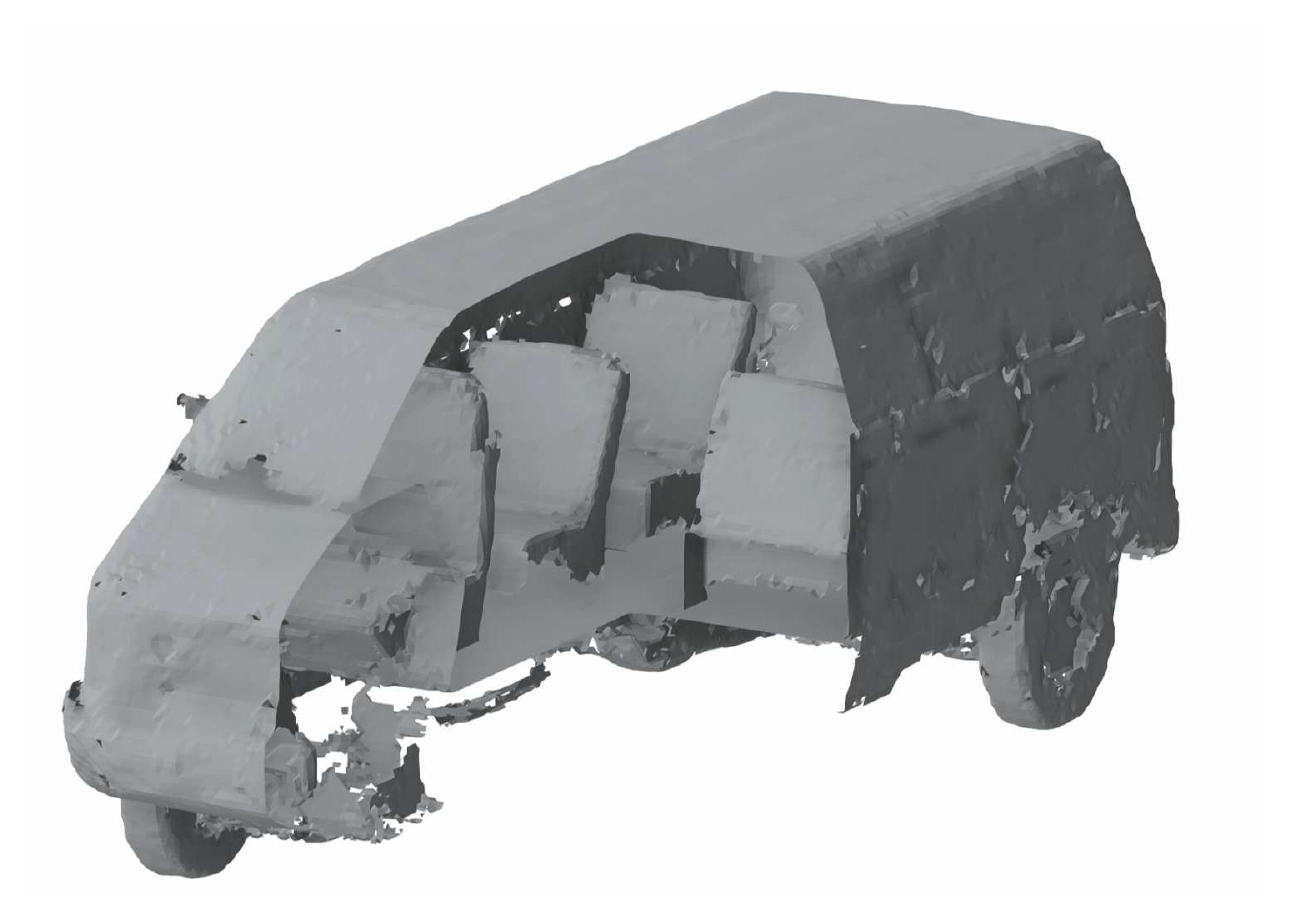}};
				\node[] (d) at (4, 1.2) {\includegraphics[width=0.12\textwidth]{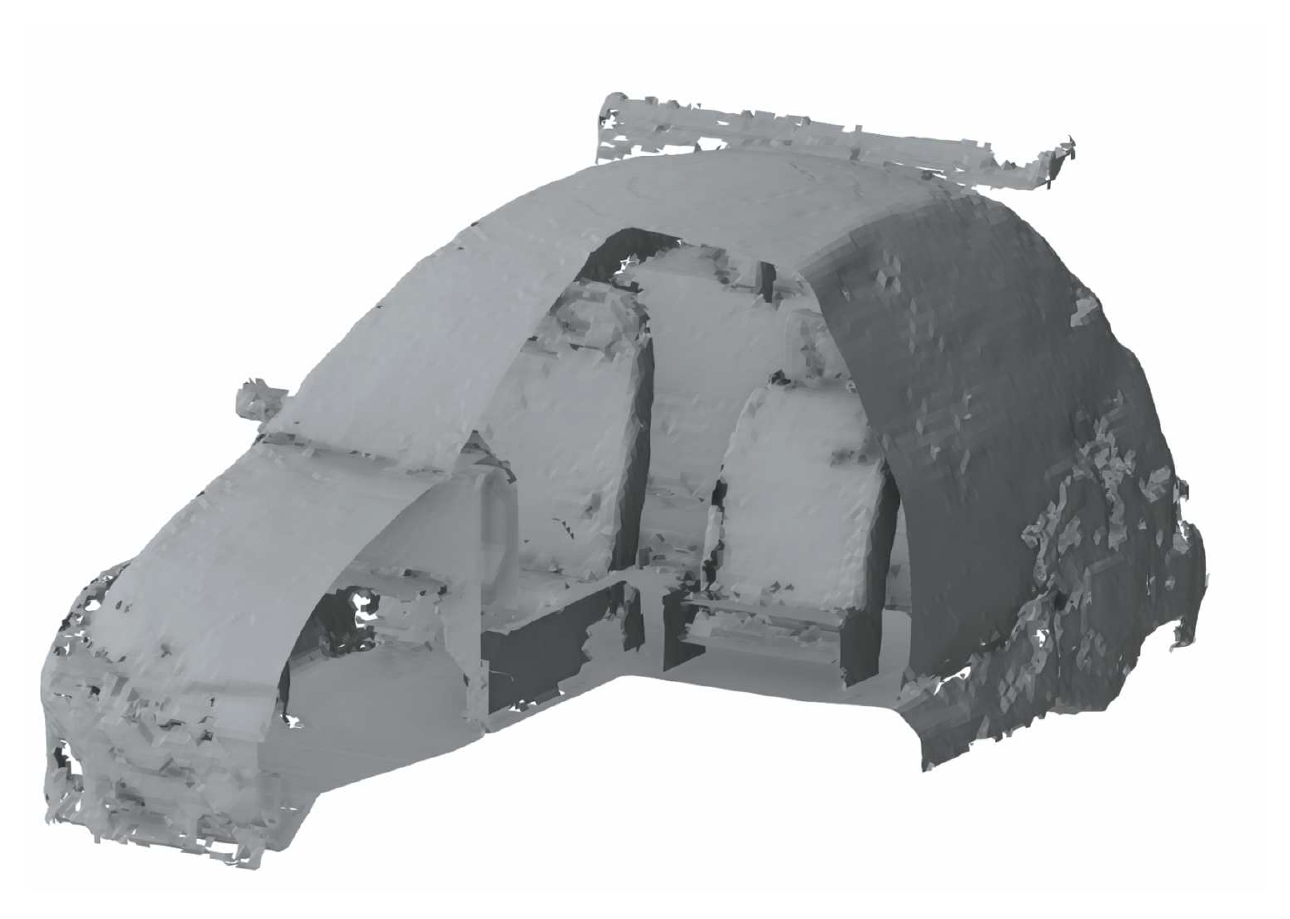}};
				\node[] (d) at (4, 0.3) {\small{(d) Ours} };
			\end{tikzpicture}
		}
		\setlength{\abovecaptionskip}{-0.45cm}
		\caption{Visual comparisons on the ShapeNet cars.}
		\label{car:VIS} \vspace{-0.35cm}
	\end{figure}
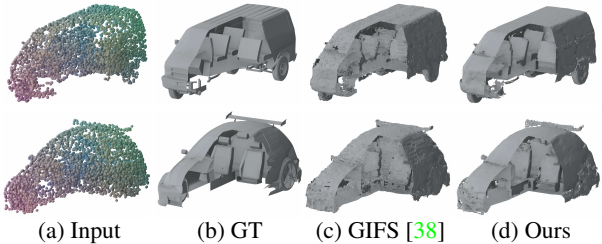

	\noindent \textbf{Comparisons.}  To verify the \textit{generalizability}, for all methods, we used the trained models  on the ShapeNet dataset (watertight shapes) to perform inference directly.
	As shown in Table \ref{MGN:TAB}, Figs. \ref{MGN:VIS} and \ref{car:VIS}, our GeoUDF exceeds GIFS \cite{GIFS} both quantitatively and visually on the MGN and ShapeNet car datasets.
	Besides, 
	Fig. \ref{SCANNET:VIS} visualizes the results of different methods on the ScanNet dataset, where it can be seen that CONet \cite{CONVOCCNET}  and POCO \cite{POCO} fail to reconstruct the whole scene; GIFS can reconstruct the objects in the scene, but the details are not well preserved. By contrast, the surfaces by our GeoUDF contain more details and are closer to GT ones.
	
	
	\begin{table}[tbp] 
		\centering
		\caption{Quantitative comparisons 
			on the MGN and ShapeNet car dataset. 
		}\vspace{-0.3cm}
		\resizebox{.45\textwidth}{!}{
			\begin{tabular}{l|l|c c|c c}
				\toprule[1.2pt]
				\multirow{2}{0.1\textwidth}{Dataset} & \multirow{2}{0.1\textwidth}{Method} & \multicolumn{2}{c|}{CD ($\times 10^{-2}$) $\downarrow$} &  \multicolumn{2}{c}{F-Score $\uparrow$} \\
				\cline{3-6}
				& & Mean & Media   & $\text{F1}^{0.5\%}$ & $\text{F1}^{1\%}$ \\
				\hline
				\multirow{2}{0.1\textwidth}{MGN}   &  GIFS \cite{GIFS} & $0.248$ & $0.234$ & $0.951$ & $0.991$ \\
				& Ours & $\pmb{0.194}$ & $\pmb{0.190}$ & $\pmb{0.975}$ & $\pmb{0.996}$ \\
				\hline
				\multirow{2}{0.1\textwidth}{ShapeNet Car}   &  GIFS \cite{GIFS} & $0.408$ & $0.346$ & $0.758$ & $0.954$ \\
				& Ours & $\pmb{0.310}$ & $\pmb{0.271}$ & $\pmb{0.870}$ & $\pmb{0.986}$ \\
				\bottomrule[1.2pt]
		\end{tabular}}
		\label{MGN:TAB}\label{SHAPENET:CAR:TAB}
	\end{table}

	\subsection{Ablation Study}\label{SEC:ABLATION}
	We conducted comprehensive ablation studies on the ShapeNet dataset to better understand our framework. 
	
\begin{figure}[tbp]
		\centering
		{
			\begin{tikzpicture}[]
				\node[] (a) at (0, 2.5) {\includegraphics[width=0.12\textwidth]{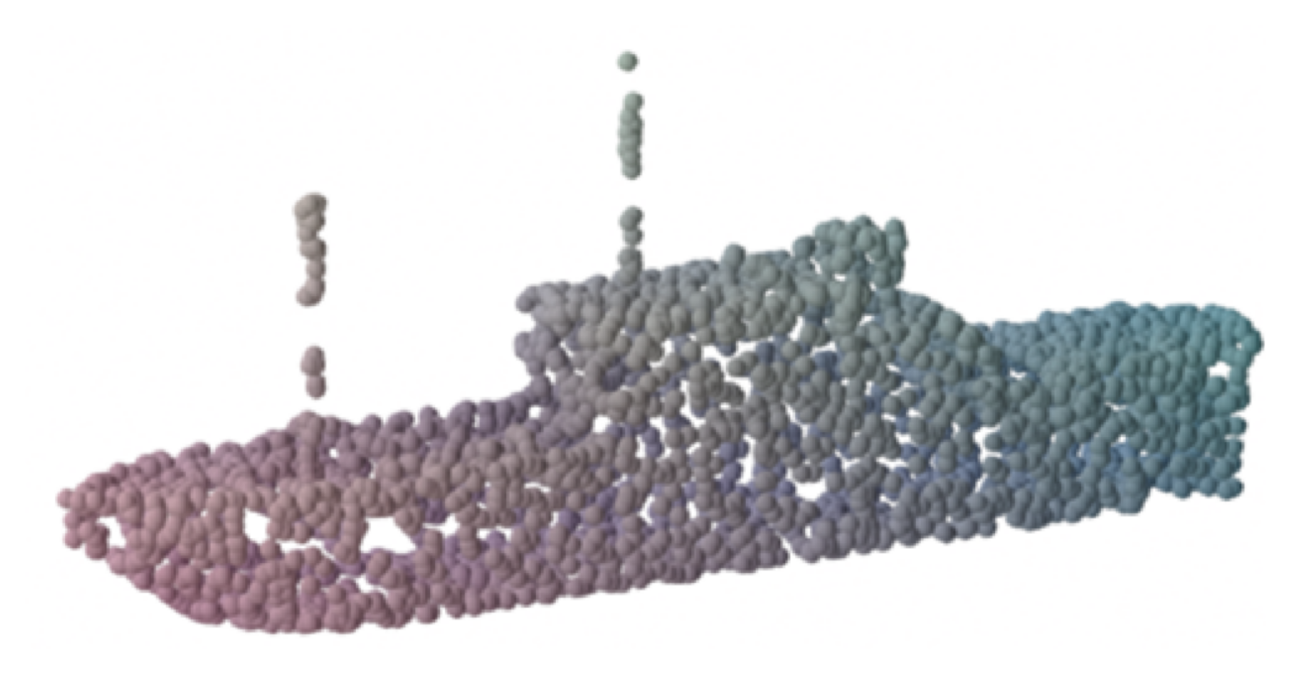} };
				\node[] (a) at (0, 1.7) {\small{(a) Input} };
				\node[] (a) at (0, 1) {\includegraphics[width=0.12\textwidth]{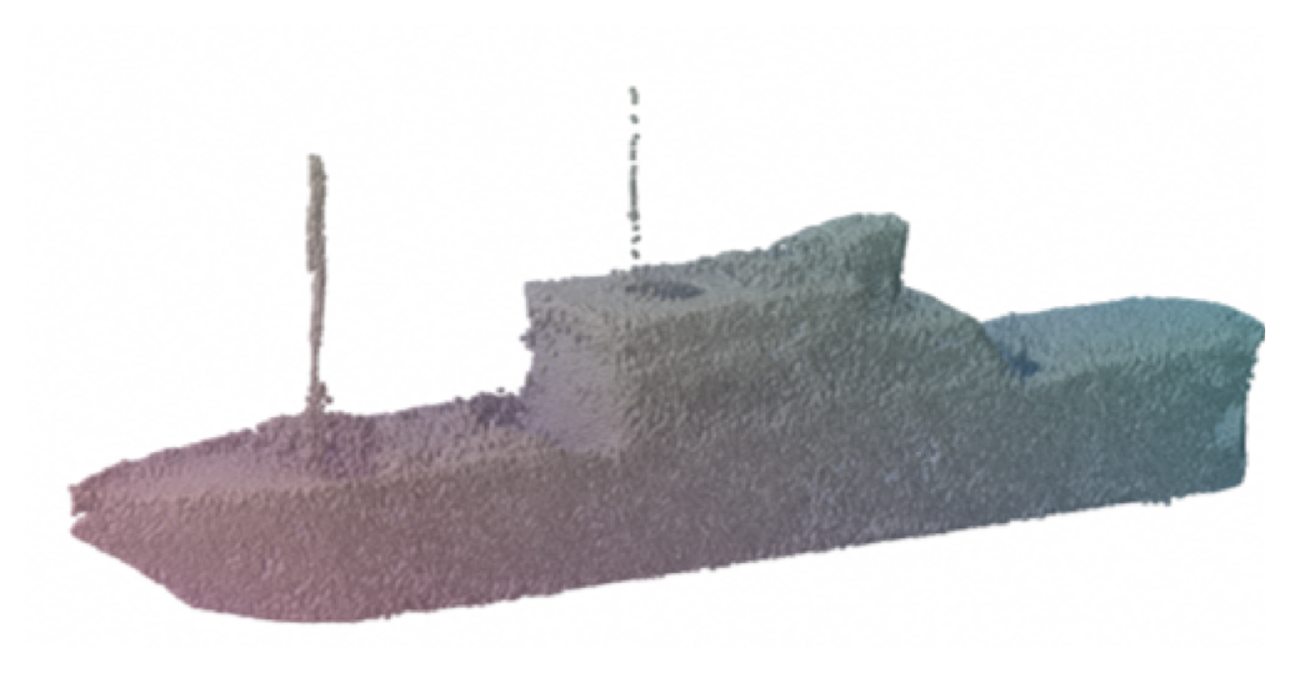} };
				\node[] (a) at (0, 0.2) {\small{(d) MAFU \cite{DMFU}} };
				
				\node[] (b) at (2.5, 2.5) {\includegraphics[width=0.12\textwidth]{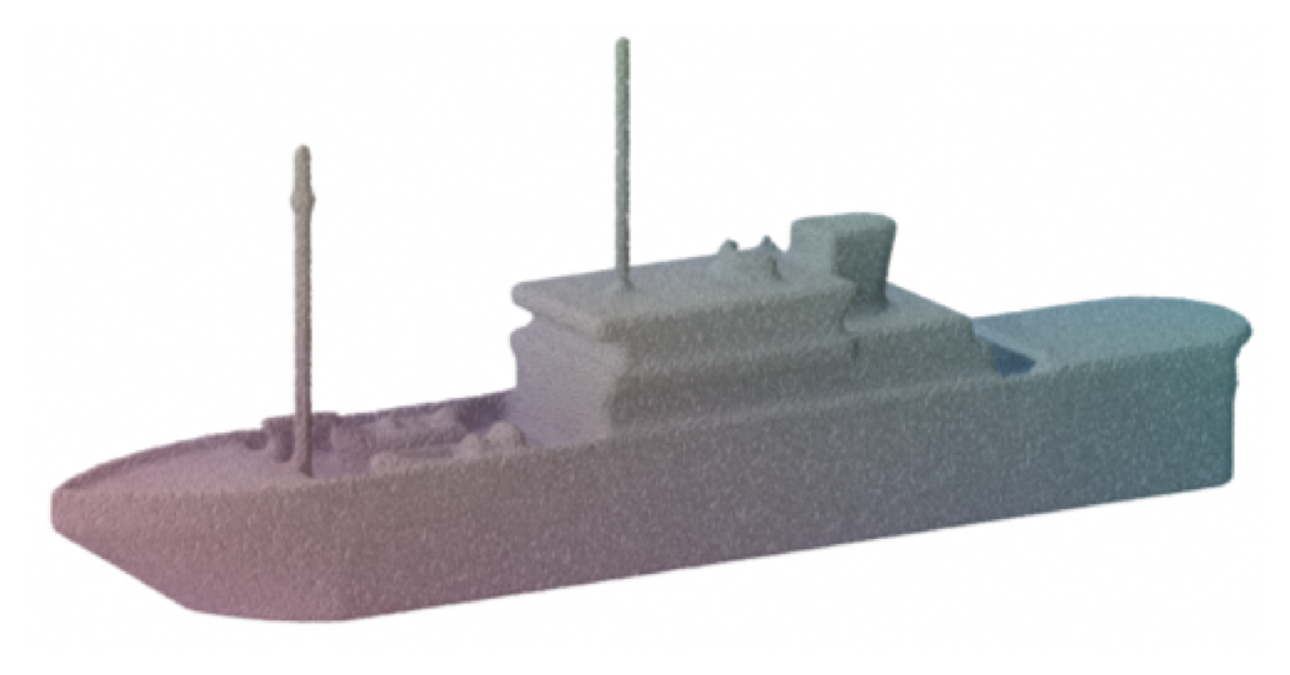}};
				\node[] (b) at (2.5, 1.7) {\small{(b) GT} };
				\node[] (b) at (2.5, 1) {\includegraphics[width=0.12\textwidth]{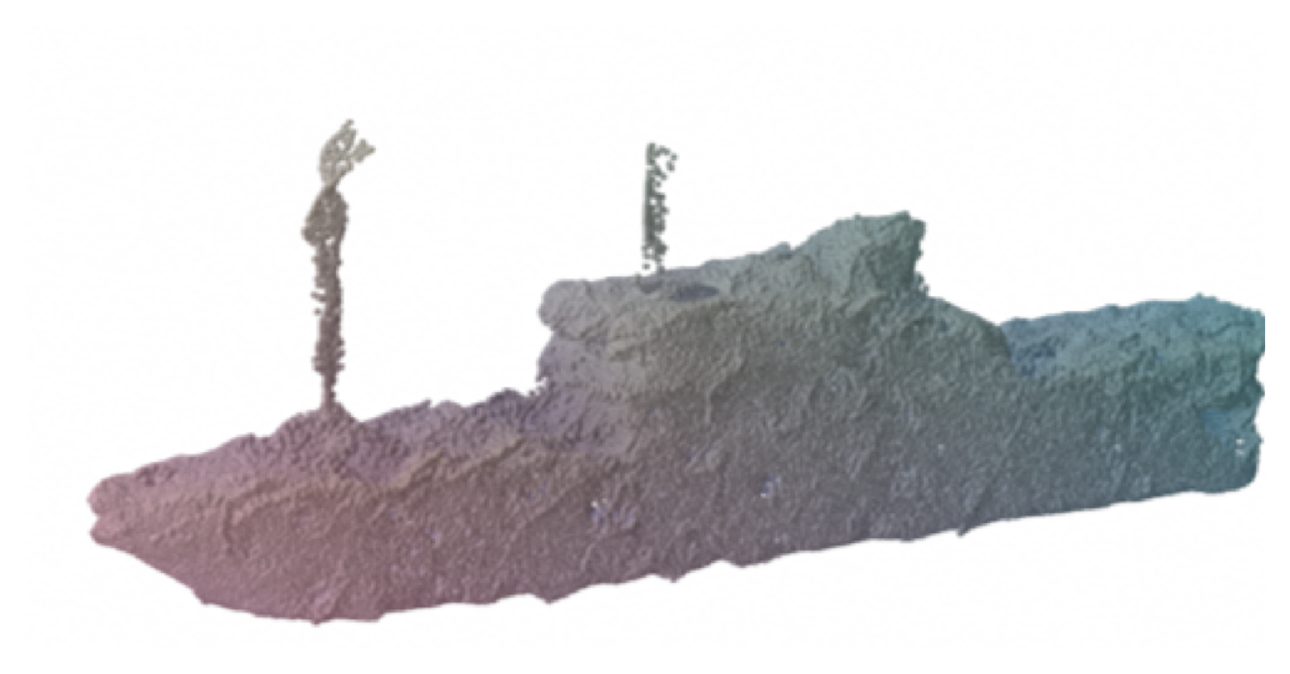}};
				\node[] (b) at (2.5, 0.2) {\small{(e) NP \cite{NP}} };
				
				\node[] (c) at (5, 2.5) {\includegraphics[width=0.12\textwidth]{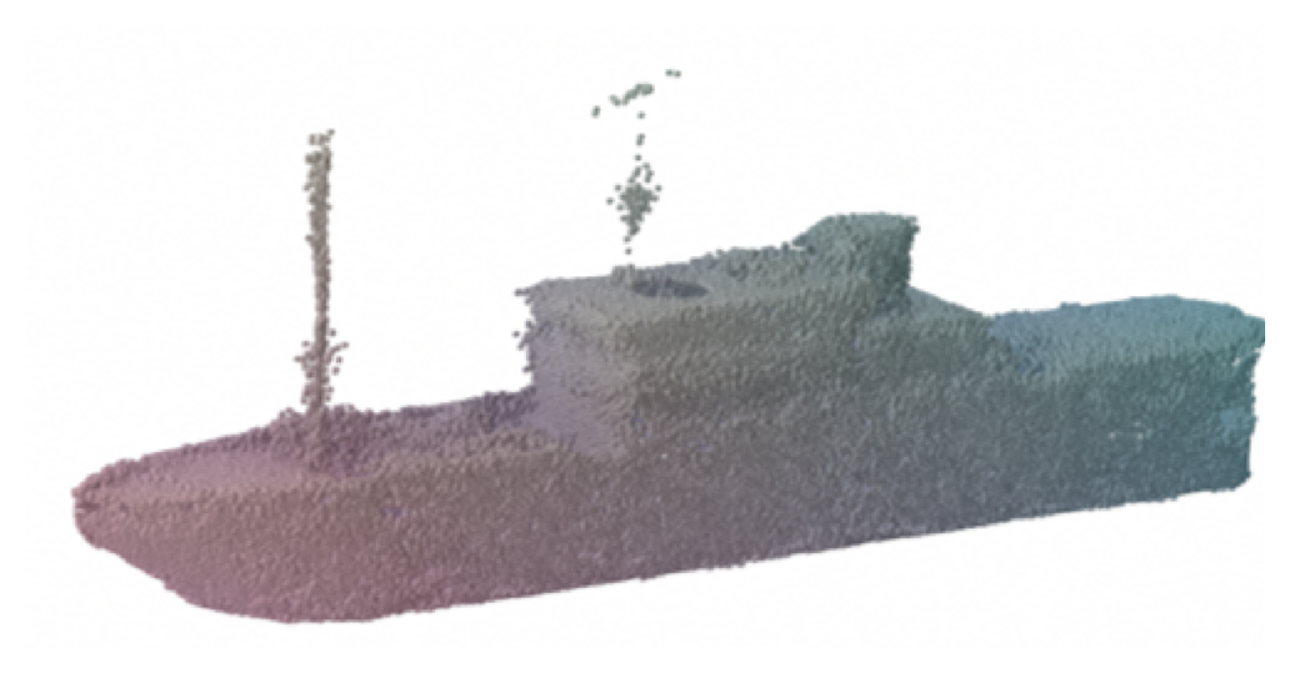}};
				\node[] (c) at (5, 1.7) {\small{(c) PUGeo \cite{PUGEO}} };
				\node[] (c) at (5, 1) {\includegraphics[width=0.12\textwidth]{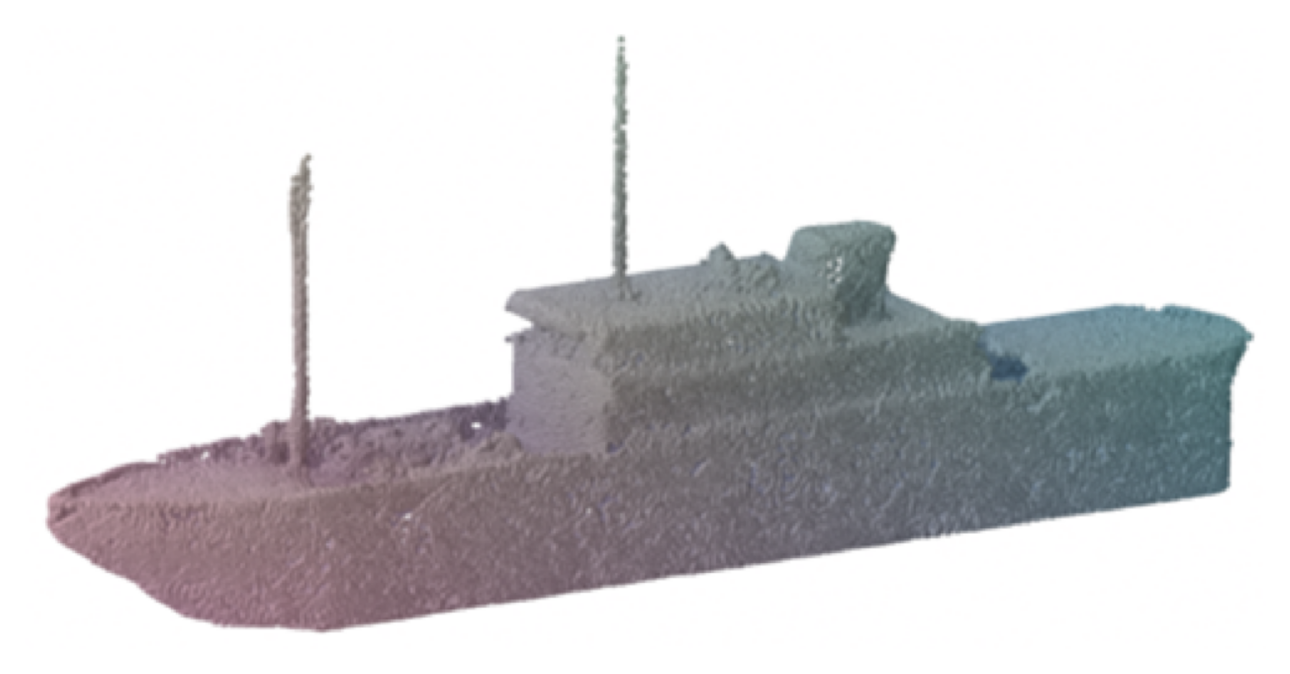}};
				\node[] (c) at (5, 0.2) {\small{(f) ours} };
			\end{tikzpicture}
		}
		\setlength{\abovecaptionskip}{-0.05cm}
		\caption{Visual comparison of the results by different upsampling methods  on the ShapeNet dataset \cite{SHAPENET}.}
		\label{PU:VIS} \vspace{-0.4cm}
	\end{figure}
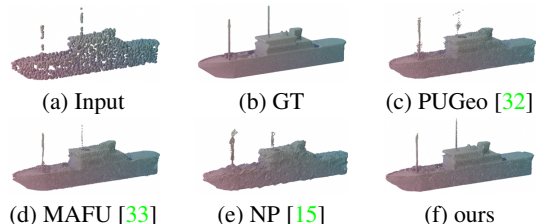

	\noindent\textbf{Advantage of LGR.} We evaluated the upsampling function of our LGR module
	on the ShapeNet \cite{SHAPENET} dataset with
	$M=16$  and compared with 
	recent representative PU methods, i.e., PUGeo-Net \cite{PUGEO}, MAFU \cite{DMFU}, and NP \cite{NP}. For a fair comparison, we retrained the official codes of the compared methods with the same training data as ours.
	As listed in Table \ref{PU:TAB},  it can be seen that our method with the fewest number of parameters achieves the best performance. 
	NP is much worse than others because its modeling manner makes it hard to deal with very sparse point clouds. 
	Besides,  Fig. \ref{PU:VIS} visually compares the results of different methods, further demonstrating the advantage of our LRG. We also refer readers to the \textit{supplementary material} for the results of LGR achieved by the 3$^{rd}$-order polynomial.
	
	\begin{table}[tbp] 
		\centering
		\caption{Comparisons of different PU methods ($M=16$). P2F: point-to-face distance.}\vspace{-0.3cm}
		\resizebox{.4\textwidth}{!}{
			\setlength{\tabcolsep}{1.8mm}{
				\begin{tabular}{l|c | c| c c  }
					\toprule[1.2pt]
					Methods  &         \# Param & \makecell[c]{Normal \\ Supervision} & \makecell[c]{CD ($\downarrow$) \\ ($\times 10^{-2}$)} &  \makecell[c]{P2F ($\downarrow$) \\ ($\times 10^{-3}$)}  \\
					\hline
					PUGeo \cite{PUGEO} &  1.287M    & Yes   & $0.287$         & $1.309$  \\
					MAFU \cite{DMFU}  &   1.390M   & No    & $0.289$           & $1.220$  \\
					NP \cite{NP}      &  0.637M    & Yes   &  $0.493$    & $4.572$ \\
					\hline
					Ours             &   0.522M    & No    & $\pmb{0.245}$    & $\pmb{0.512}$ \\
					\bottomrule[1.2pt]
		\end{tabular}}}\vspace{-0.3cm}
		\label{PU:TAB}
		\vspace{-0.3cm}
	\end{table}
	
\begin{figure}[tbp]
		\centering
		{
			\begin{tikzpicture}[]
				
				\node[] (d) at (-2, 1) {\includegraphics[width=0.09\textwidth]{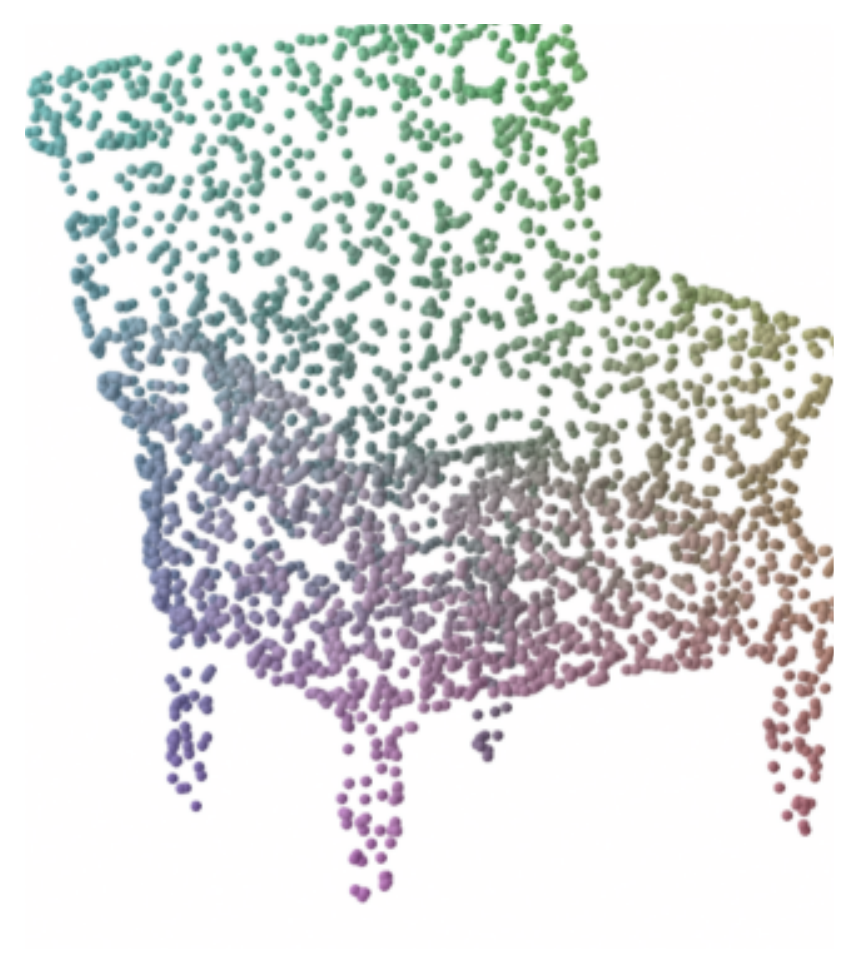}};
				\node[] (d) at (-2, 0) {\small{(a) Input} };
				
				\node[] (c) at (0, 1) {\includegraphics[width=0.09\textwidth]{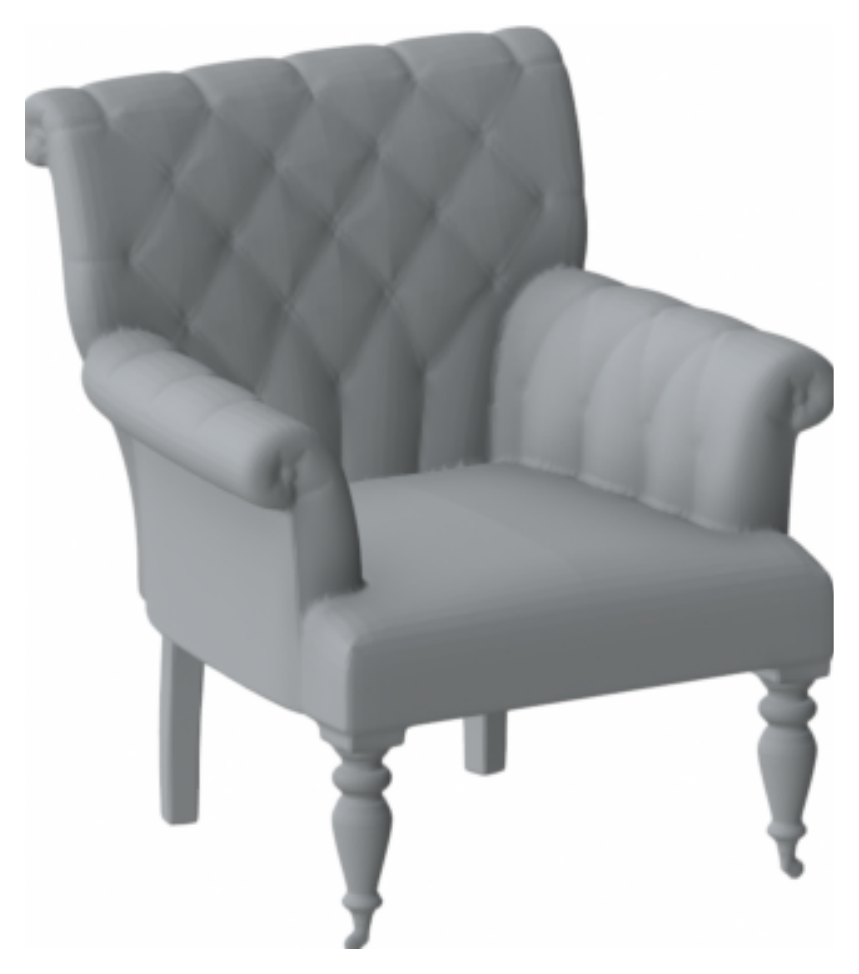}};
				\node[] (c) at (0, 0) {\small{(b) GT } };
				
				\node[] (b) at (2, 1) {\includegraphics[width=0.09\textwidth]{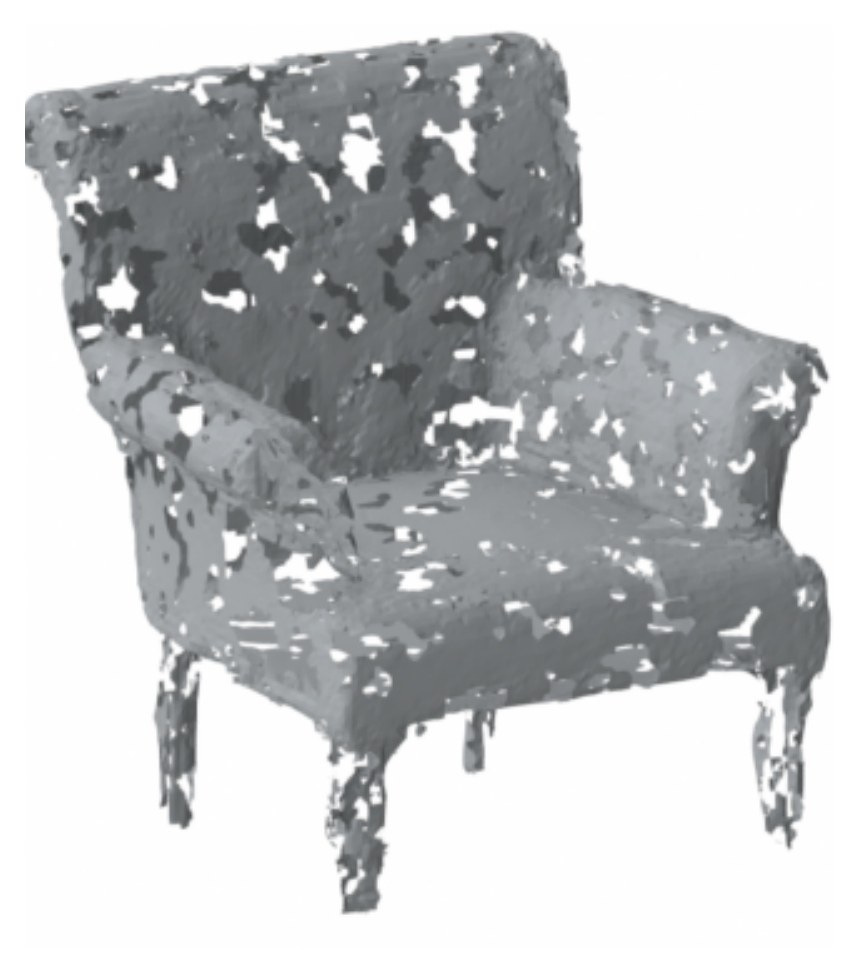}};
				\node[] (b) at (2, 0) {\small{(c) w/o LGR} };
				
				\node[] (a) at (4, 1) {\includegraphics[width=0.09\textwidth]{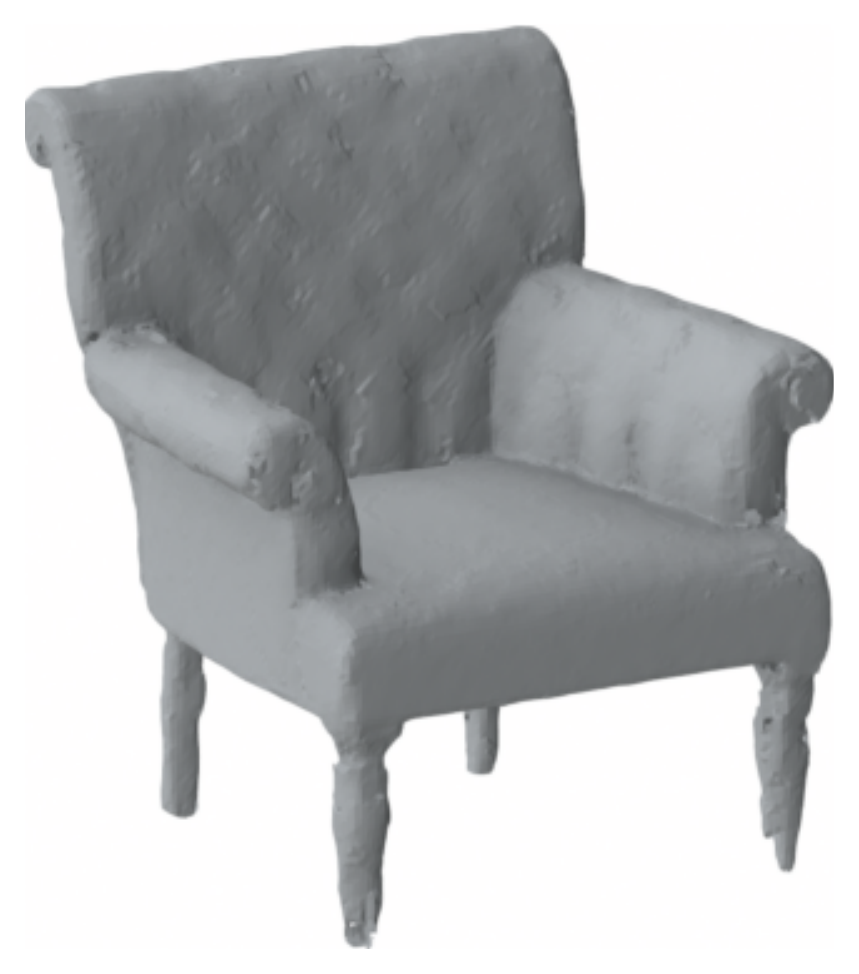} };
				\node[] (a) at (4, 0) {\small{(d) w/ LGR} };
				
			\end{tikzpicture}
		}
		\setlength{\abovecaptionskip}{-0.1cm}
		\caption{{Visual results of our method w/ and w/o LGR.}}
		\label{ABLATION:PU:VIS} \vspace{-0.3cm}
	\end{figure}
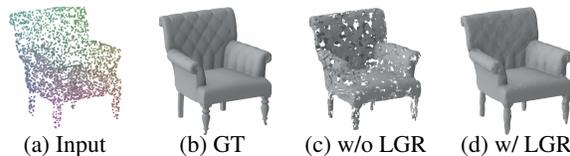

	\noindent\textbf{Necessity of LGR.} 
	We removed LGR from our GeoUDF to reconstruct surfaces from input sparse point clouds directly. Under this scenario, we estimated the normal vectors 
	by using the principal component analysis (PCA). Besides, we also equipped GIFS \cite{GIFS} with our LGR, i.e., input point clouds were upsampled by our LGR before voxelization.
	From Table \ref{ABLATION:PU:RATIO},  
	it can be seen that LRG can boost the reconstruction accuracy of both GIFS and ours, validating its effectiveness. 
	Besides, as visualized in Fig. \ref{ABLATION:PU:VIS},
	without LGR, the reconstructed surfaces are much worse, i.e., there are many holes caused by the sparsity of the input point cloud.

	\begin{table}[tbp]
		\centering
		\caption{The reconstruction accuracy of GIFS \cite{GIFS} and our method \textit{with} and \textit{without} LGR. 
		}\vspace{-0.3cm}
		\resizebox{.4\textwidth}{!}{
			\setlength{\tabcolsep}{2.0mm}{
				\begin{tabular}{l|c c|c c}
					\toprule[1.2pt]
					\multirow{2}{0.12\textwidth}{Method}  & \multicolumn{2}{c|}{L1-CD $(\times 10^{-2})$ $\downarrow$} &  \multicolumn{2}{c}{F-Score $\uparrow$}    \\
					\cline{2-5}
					& Mean & Median & $\text{F1}^{0.5\%}$ & $\text{F1}^{1\%}$ \\
					\hline
					GIFS              & 0.328 & 0.276 & 0.860 & 0.974     \\
					GIFS w/ LGR       & 0.316 & 0.292 & 0.894 & 0.981     \\
					\hline
					Ours   w/o LGR     & 0.284 & 0.264  & 0.882 & 0.967    \\
					Ours              & 0.234 & 0.226  & 0.938 & 0.992    \\
					\bottomrule[1.2pt]
		\end{tabular} }}\vspace{-0.3cm}
		\label{ABLATION:PU:RATIO}
	\end{table} 

	\noindent\textbf{Accuracy of UDF and its gradient estimation}.
	The accuracy of reconstructed surfaces cannot exactly reflect
	that of UDFs and corresponding gradients due to the discrete cubes of MC.
	Thus, we directly compared the accuracy of UDFs and gradients estimated by different methods.
	In addition to NDF \cite{NDF} and GIFS \cite{GIFS} for comparison, we also set another three baselines (named \textit{Regress}, \textit{P2P} and \textit{BP}). \textit{Regress} replaces Eqs. \eqref{EQ:UDF} and \eqref{EQ:UDF:GRAD} of our GUE with two separated MLPs to regress UDFs and gradients, and \textit{P2P} utilizes Eq. \eqref{EQ:UDF:DIST} to calculate UDF as described before. \textit{BP} utilizes the back propagation of the network to calculate the gradient. 
As listed in Table \ref{ABLATION:UDF}, our GUE produces the most accurate UDFs and gradients. moreover, the error is much smaller than the edge length of the cube of E-MC (1/127).
Besides, BP can obtain gradients that are slightly worse than those of our GUE, but it is much slower.
Finally, we validated that the shape encoding (SE) of $\Omega(\mathbf{q})$ is essential to our GUE, i.e., GUE w/o SE produces UDFs and gradients with much larger error than GUE.

\begin{table}[tbp] \small
	\centering
	\caption{UDF accuracy of different methods. Note that GIFS does not rely on the gradients of UDFs. 
	}\vspace{-0.3cm}
	\resizebox{.4\textwidth}{!}{
		\begin{tabular}{l |c c| c}
			\toprule[1.2pt]
			\multirow{2}{0.005\textwidth}{UDF Estimation}  &  \multicolumn{2}{c|}{UDF Error $\downarrow$} & \multirow{2}{0.08\textwidth}{\makecell[c]{Time $\downarrow$ \\ ($\upmu$s/point)}}\\
			\cline{2-3}
			& UDF $(\times 10 ^{-3})$   & Grad $(^\circ)$   &             \\
			\hline
			NDF \cite{NDF}      & 2.536                   & 13.930              & 1.173 \\
			GIFS \cite{GIFS}    & 2.439                   & -                   & 4.228 \\
			\hline
			Regress             & 0.994                   & 8.371               & 1.762 \\
			P2P         & 1.240                   & 7.238               & 3.497 \\
			BP                  & 0.615                   & 7.877               & 12.125 \\
			\hline
			GUE w/o SE          & 1.453                   & 10.707              & 1.271 \\
			GUE                 & 0.615                   & 7.237               & 3.405 \\
			\bottomrule[1.2pt]
	\end{tabular}}
	\label{ABLATION:UDF} 
\end{table}

\noindent \textbf{Size of $\Omega(\mathbf{q})$}. From Table \ref{KNNPATCH:SIZE}, we can conclude that our GeoUDF is robust to the size of neighborhood used in Eqs. \eqref{EQ:UDF} and \eqref{EQ:UDF:GRAD}.
This is credited to the manner of learning adaptive weights, i.e., very smaller weights would be predicted for not important neighbours.

\begin{table}[tbp] 
	\centering
	\caption{The reconstruction accuracy of different $K$-NN sizes.
	}\vspace{-0.3cm}
	\resizebox{.35\textwidth}{!}{
		\setlength{\tabcolsep}{2mm}{
			\begin{tabular}{c|c c|c c}
				\toprule[1.2pt]
				\multirow{2}{0.09\textwidth}{\makecell*[c]{$K$}} & \multicolumn{2}{c|}{CD ($\times 10^{-2}$) $\downarrow$} &  \multicolumn{2}{c}{F-Score $\uparrow$}  \\
				\cline{2-5}
				& Mean      & Median        & $\text{F1}^{0.5\%}$       & $\text{F1}^{1\%}$ \\
				\hline
				5       & $0.235$   & $0.227$       & $0.937$                   & $0.992$ \\
				10      & $0.234$   & $0.226$       & $0.938$                   & $0.992$ \\
				20      & $0.232$   & $0.224$       & $0.939$                   & $0.993$ \\
				\bottomrule[1.2pt]
	\end{tabular}}}\vspace{-0.3cm}
	\label{KNNPATCH:SIZE}
\end{table}

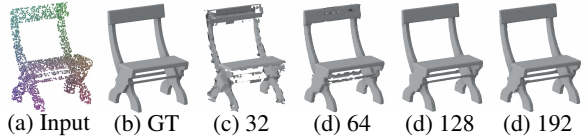
\begin{figure}[tbp] \small
	\centering
	{
		\begin{tikzpicture}[]
			\node[] (a) at (0, 0.9) {\includegraphics[width=0.08\textwidth]{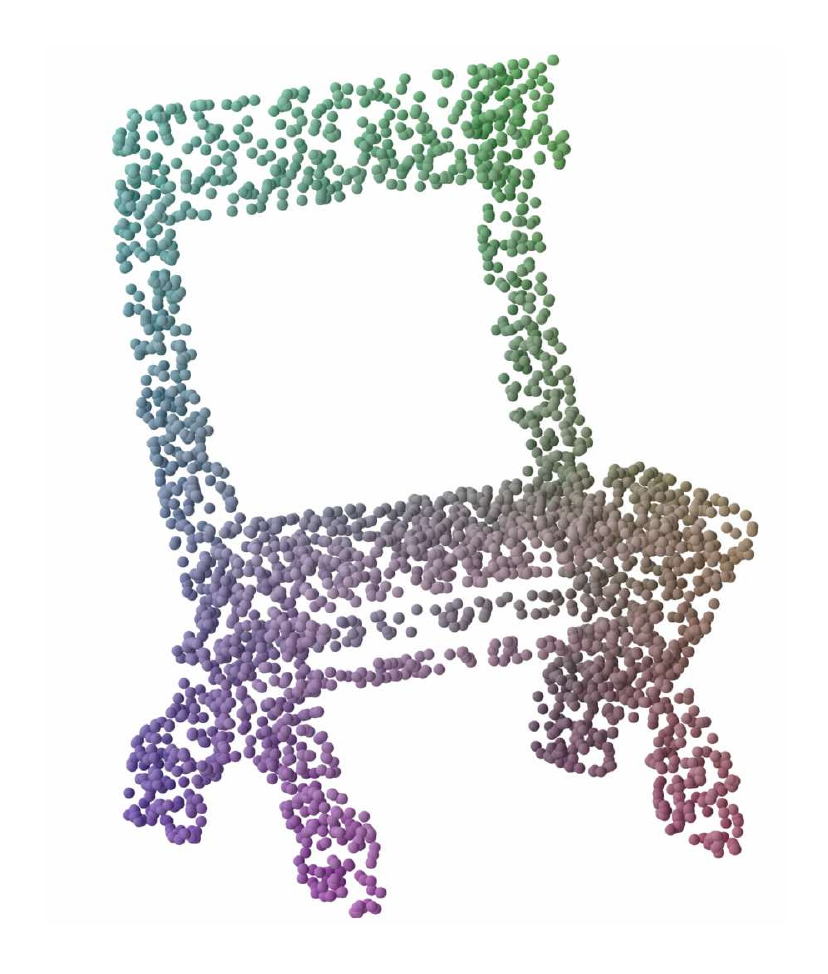} };
			\node[] (a) at (0, 0) {\small{(a) Input} };
			
			\node[] (b) at (1.3, 0.9) {\includegraphics[width=0.08\textwidth]{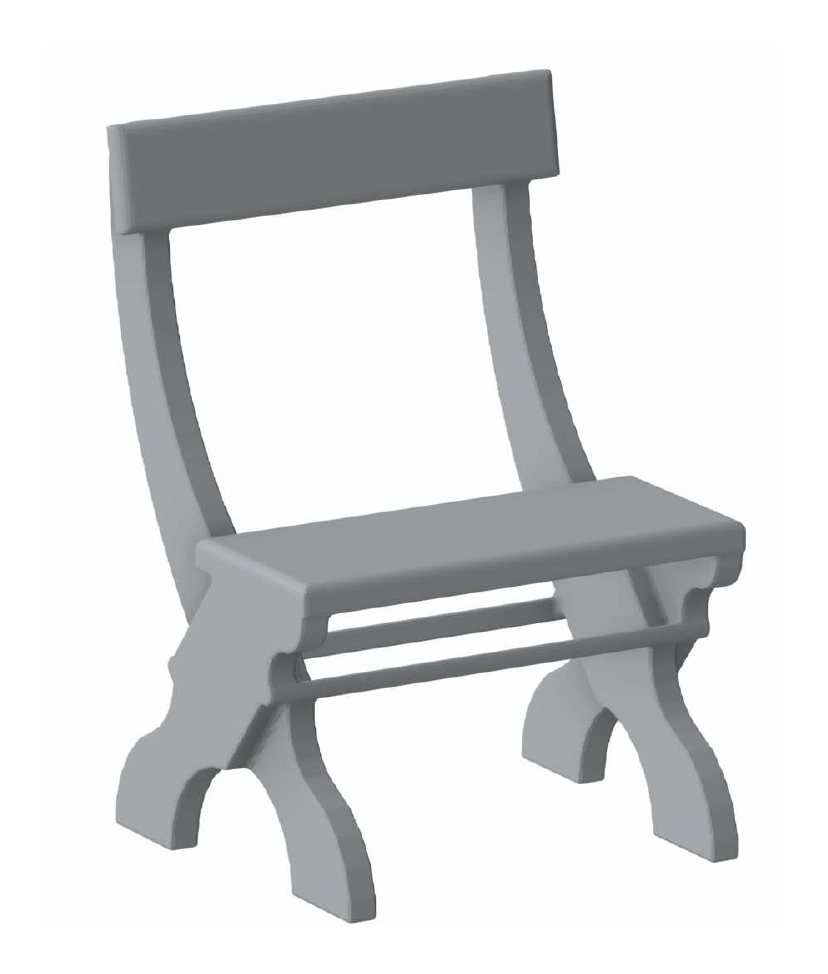}};
			\node[] (b) at (1.3,0) {\small{(b) GT} };
			
			\node[] (c) at (2.6, 0.9) {\includegraphics[width=0.08\textwidth]{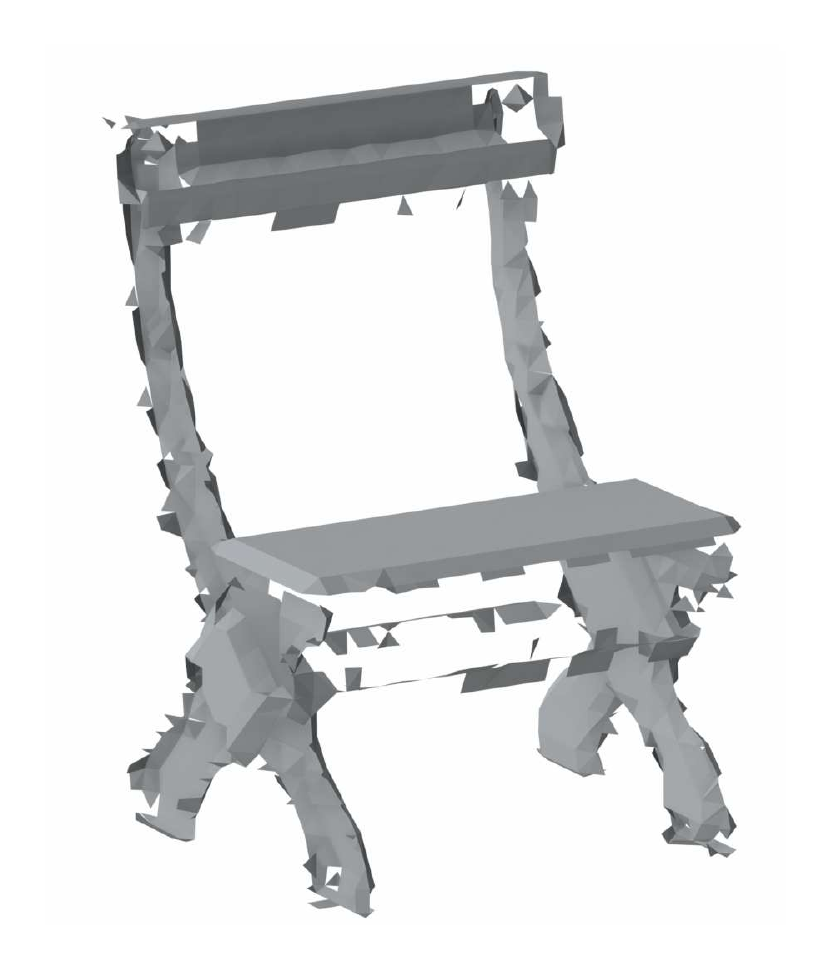}};
			\node[] (c) at (2.6, 0) {\small{(c) 32 } };
			
			\node[] (d) at (3.9, 0.9) {\includegraphics[width=0.08\textwidth]{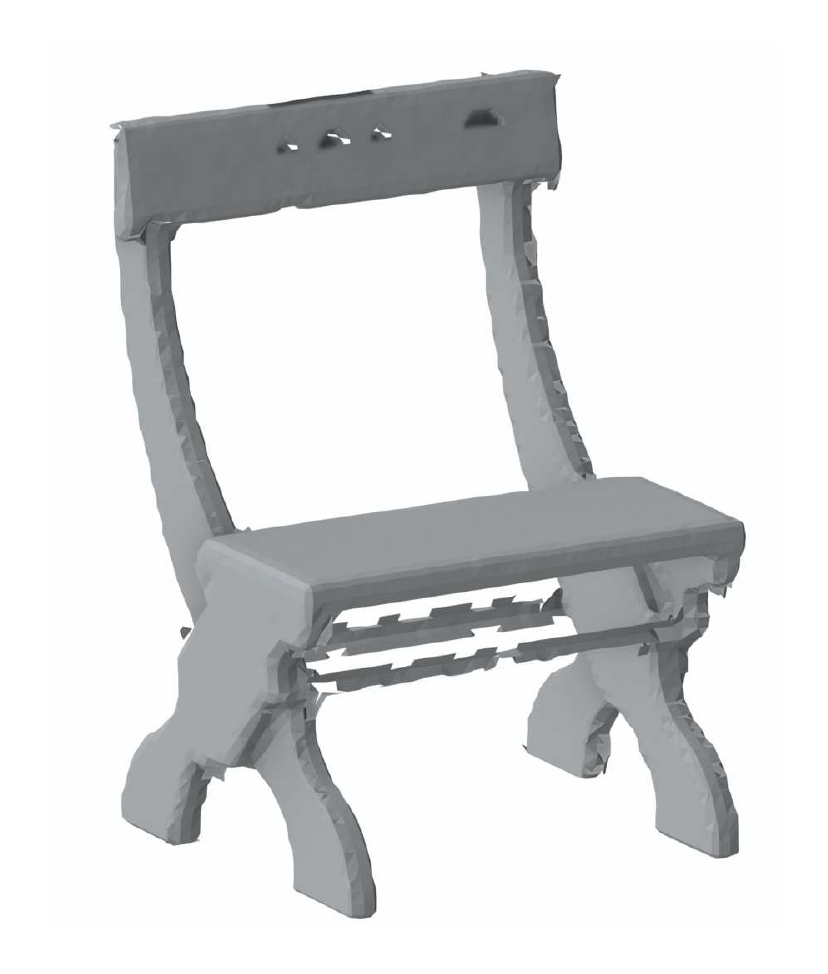}};
			\node[] (d) at (3.9,0) {\small{(d) 64} };
			
			\node[] (d) at (5.2, 0.9) {\includegraphics[width=0.08\textwidth]{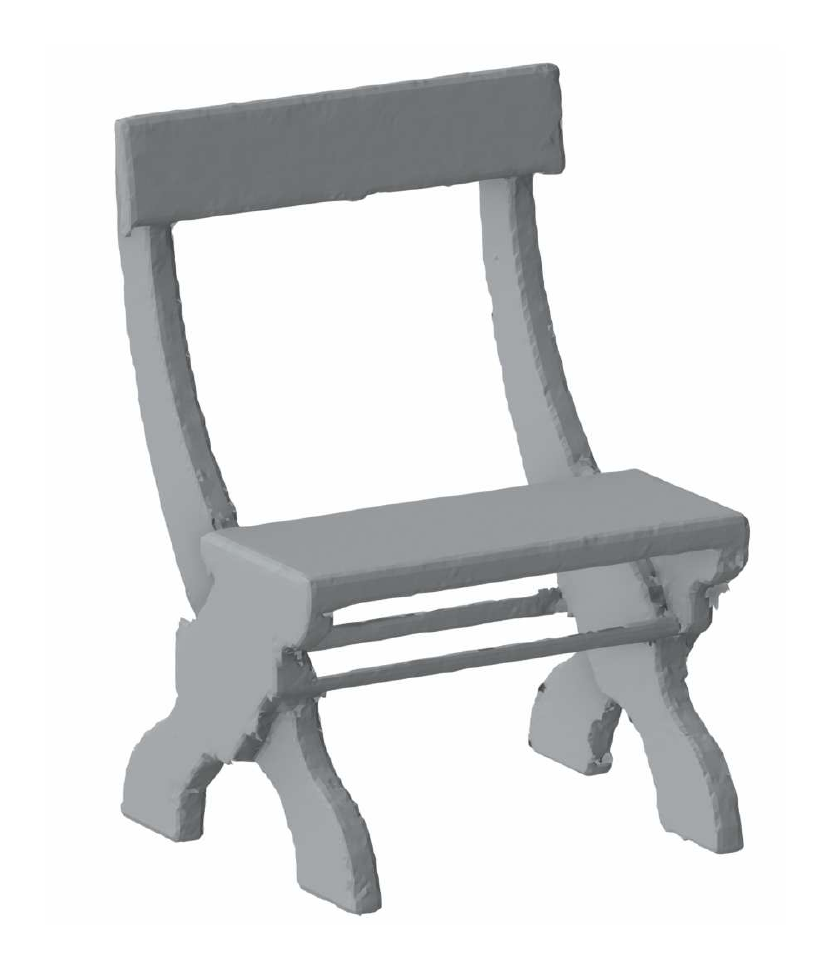}};
			\node[] (d) at (5.2, 0) {\small{(d) 128} };
			
			\node[] (d) at (6.5, 0.9) {\includegraphics[width=0.08\textwidth]{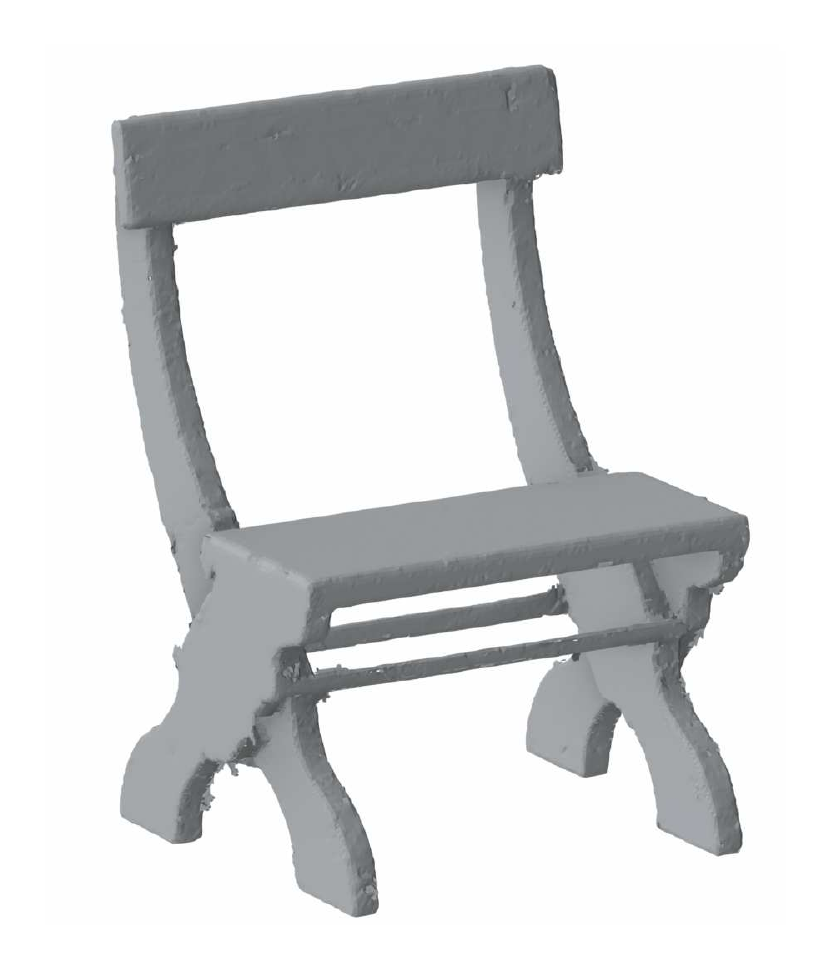}};
			\node[] (d) at (6.5, 0) {\small{(d) 192} };
			
		\end{tikzpicture}
	}
	\setlength{\abovecaptionskip}{-0.1cm}
	\caption{Visual illustration of the effect of the resolution of E-MC on the reconstructed surface. 
	}
	\label{ABLATION:MC:RES:VIS}
	\vspace{-0.3cm}
\end{figure}

\noindent \textbf{Superiority of E-MC.} We also compared our E-MC with MeshUDF \cite{MESHUDF}, the SOTA method for extracting triangle meshes from unsigned distance fields. For a fair comparison, we fed the two methods with identical unsigned distance fields estimated by our GUE.
From Table \ref{ABLATION:MC}, it can be seen that our E-MC can extract triangle meshes with higher quality than MeshUDF.
Besides, we also studied how the resolution of the 3D grid affects reconstruction accuracy. From Table \ref{ABLATION:RES}, we can see that the reconstruction accuracy gradually improves with the resolution increasing, which is consistent with the visual results in Fig. \ref{ABLATION:MC:RES:VIS}, but more times are consumed. \textit{Such an observation is fundamentally credited to the highly-accurate UDFs by our method}. 

\begin{table}[tbp] 
	\centering
	\caption{Quantitative comparison of our E-MC  with  MeshUDF \cite{MESHUDF}. \vspace{-0.4cm}
	}
	\resizebox{.4\textwidth}{!}{
		\setlength{\tabcolsep}{2.5mm}{
			\begin{tabular}{l|c c|c c}
				\toprule[1.2pt]
				\multirow{2}{0.1\textwidth}{Method} & \multicolumn{2}{c|}{CD ($\times 10^{-2}$) $\downarrow$} &  \multicolumn{2}{c}{F-Score $\uparrow$}  \\
				\cline{2-5}
				& Mean & Median   & $\text{F1}^{0.5\%}$ & $\text{F1}^{1\%}$ \\
				\hline
				GUE+MeshUDF \cite{MESHUDF}    & $0.266$ & $0.251$ & $0.916$ & $0.978$ \\
				GUE+E-MC & $\pmb{0.234}$ & $\pmb{0.226}$ & $\pmb{0.938}$ & $\pmb{0.992}$ \\
				\bottomrule[1.2pt]
	\end{tabular}}}\vspace{-0.3cm}
	\label{ABLATION:MC}
\end{table} 

\begin{table}[tbp] 
	\centering
	\caption{The reconstruction accuracy under various 3D grid resolutions used in E-MC.}\vspace{-0.3cm}
	\resizebox{.4\textwidth}{!}{
		\setlength{\tabcolsep}{1.4mm}{
			\begin{tabular}{c|c c|c c|c c}
				\toprule[1.2pt]
				\multirow{2}{0.05\textwidth}{\makecell[c]{Res.}} & \multicolumn{2}{c|}{CD ($\times 10^{-2}$) $\downarrow$} &  \multicolumn{2}{c|}{F-Score $\uparrow$} & \multicolumn{2}{c}{Time (s) $\downarrow$} \\
				\cline{2-7}
				& Mean      & Median     & $\text{F1}^{0.5\%}$   & $\text{F1}^{1\%}$   & UDF     & E-MC  \\
				\hline
				32       & $0.584$ & $0.554$     & $0.646$               & $0.824$           &0.051     & 3.295  \\
				64       & $0.312$ & $0.285$     & $0.860$               & $0.958$           &0.138     & 4.744  \\
				128      & $0.234$ & $0.226$     & $0.938$               & $0.992$           &0.759     & 15.282 \\
				192      & $0.223$ & $0.218$     & $0.949$               & $0.995$           &2.264     & 32.052 \\
				\bottomrule[1.2pt]
	\end{tabular}}}\vspace{-0.3cm}
	\label{ABLATION:RES} 
\end{table}

\subsection{Efficiency Analysis}
We also evaluated the efficiency of our GeoUDF on the ShapeNet dataset.  As listed in Table \ref{EFFICIENCY}, our method has the fewest number of parameters while achieving the highest reconstruction accuracy among all methods, which is credited to our explicit and elegant formulations to this problem.  
We also evaluated the time consumption of our method,
demonstrating that all modules before surface extraction are very efficient.
As for the surface extraction process, due to the global optimization in our E-MC, it is slower than 
the traditional Marching Cube methods, but faster than GIFS.

\begin{table}[tbp] \small
	\centering
	\caption{Efficiency comparisons. The time of ``Inference"   refers to the overall time minus the time for iso-surface extraction. 
	}\vspace{-0.3cm}
	\resizebox{.4\textwidth}{!}{
		\begin{tabular}{l |c|c c}
			\toprule[1.2pt]
			\multirow{2}{0.005\textwidth}{Method}   & \multirow{2}{0.08\textwidth}{\# Param}    &  \multicolumn{2}{c}{Time (s) $\downarrow$}        \\
			\cline{3-4}
			&                  & Inference      & Surface Extraction      \\
			\hline
			ONet \cite{OCCNET}          &       10.373M    & 0.653          & 0.219                   \\
			CONet \cite{CONVOCCNET}     &       1.978M     & 0.941          & 0.589                   \\
			SAP \cite{SAP}              &       1.085M     & 0.247          & 0.384                   \\
			POCO \cite{POCO}            &       12.790M    & 4.652          & 8.560                   \\
			{DOG \cite{DOG} }     &       2.182M     & 0.302          & 0.228 \\
			\hline
			GIFS \cite{GIFS}            &       3.682M     & 27.771         & 26.915                  \\
			Ours                        &       0.775M     & 0.759          & 15.282                  \\
			\bottomrule[1.2pt]
	\end{tabular}}\vspace{-0.3cm}
	\vspace{-0.4cm}
	\label{EFFICIENCY}
\end{table}

\section{Conclusion}
GeoUDF is a novel learning-based framework for reconstructing surfaces from sparse 3D point clouds. It has several advantages, including being lightweight, efficient, accurate, explainable, and generalizable, which have been validated through extensive experiments and ablation studies. The various advantages of our framework are fundamentally credited to the explicit and elegant formulations of the problems of UDF and its gradient estimation and local structure representation from the geometric perspective, as well as the edge-based marching cube.

{\small
	\bibliographystyle{ieee_fullname}
	\bibliography{egbib}
}

\end{document}


	\title{GeoUDF: Surface Reconstruction from 3D Point Clouds via \\Geometry-guided Distance Representation\\(Supplementary Materials)}
	
	\author{Siyu Ren$^{1,2}$ \quad Junhui Hou$^{1*}$  \quad Xiaodong Chen$^2$ \quad Ying He$^3$ \quad Wenping Wang$^4$  \\ 
		$^1$City University of Hong Kong \quad $^2$Tianjin University \\ $^3$ Nanyang Technological University \quad $^4$Texas A\&M University \\
		{\tt\small siyuren2-c@my.cityu.edu.hk, jh.hou@cityu.edu.hk, xdchen@tju.edu.cn,} \\ {\tt\small yhe@ntu.edu.sg, wenping@cs.hku.hk}
	}
	
	\maketitle
	
	\setcounter{equation}{8}
	\setcounter{figure}{13}
	
	In this supplementary material, we will provide more details of our framework GeoUDF in Section \ref{DETAIL:ARC}, and more details of experiment settings and more results in Section \ref{EXP}. 
	We also refer the readers to the \href{https://rsy6318.github.io/GeoUDF.html}{\textit{project page}}. 
	
	\section{More Details of GueUDF} \label{DETAIL:ARC}
	\subsection{Local Geometry Representation}
	
\begin{figure}[h]
		\centering 
		\subfloat[]{\includegraphics[width=0.7\textwidth]{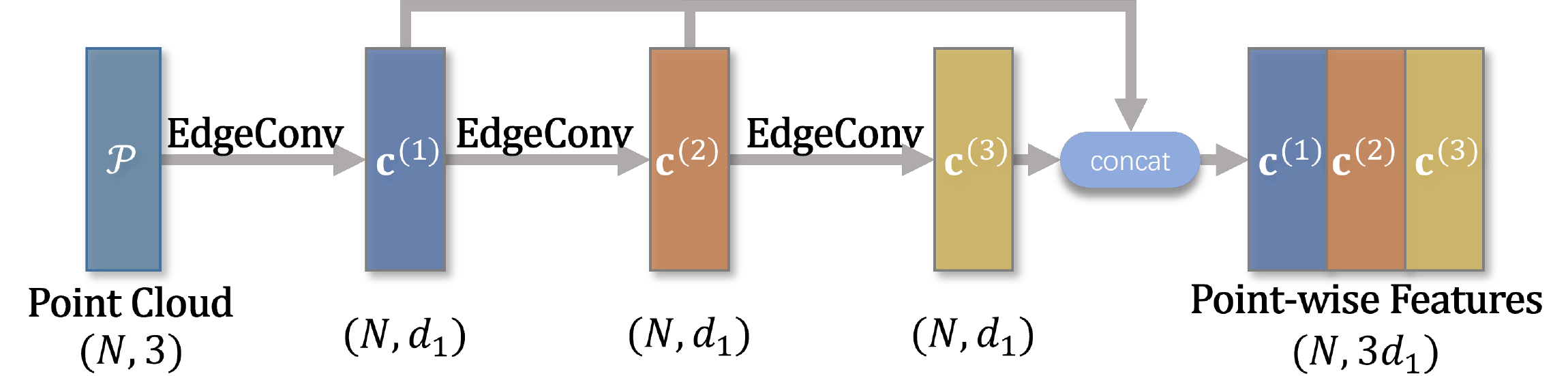} \label{feature:extraction}} 
		\subfloat[]{\includegraphics[width=0.3\textwidth]{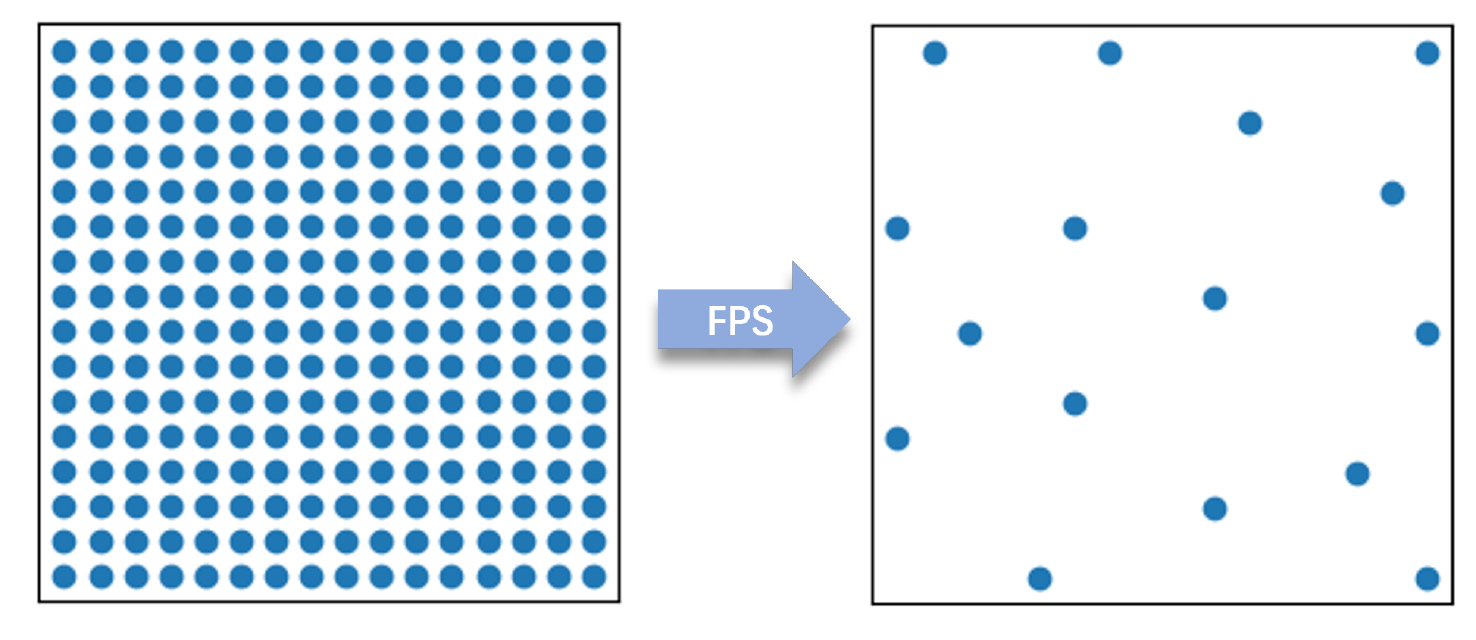}\label{uv} } 
		\vspace{-0.3cm}
		\caption{ (a) Feature Extraction. (b) Poisson Disk Sample in $\mathcal{D}$.}
	\end{figure}

	\noindent\textbf{Feature Extraction.} As shown in Fig. \ref{feature:extraction}, we employ 3-layer EdgeConv \cite{DGCNN} to embed $\mathcal{P}$ into a high-dimensional feature space, producing hierarchical features, $\{\mathbf{c}_i^{(l)}\in\mathbb{R}^{d_1}\}_{i=1}^{N}$, $l=1,2,3$, which are concatenated as the point-wise features of $\mathcal{P}$. \\
	
	\noindent\textbf{Poisson Disk Sampling over $\mathcal{D}$.}
	As shown in Fig. \ref{uv},  we apply the farthest point sampling (FPS) on the uniform 2D grids to approximate Poisson Disk Sampling on the 2D area $\mathcal{D}$ for obtaining 2D coordinates.
	
	\noindent\textbf{Differential Properties.} From the explicit parameterization function in Eq. (1) of the manuscript, we can obtain the differential properties of the produced dense point cloud $\mathcal{P}_M$.
	Let $\mathbf{p}_{i,k}=f_i(\mathbf{u}_k)$ be one point of $\mathcal{P}_M$ and
	$\mathbf{J}_{i,k}=\left[\frac{\partial f_i}{\partial u_{1,k}},\frac{\partial f_i}{\partial u_{2,k}} \right]\in\mathbb{R}^{3\times 2}$ be the Jacobian matrix of $f_i$ at $\mathbf{p}_{i,k}$, then its un-oriented normal vector can be computed as  
	\begin{equation}
		\mathbf{n}_{i,k}=\frac{\frac{\partial f_i}{\partial u_{1,k}} \times \frac{\partial f_i}{\partial u_{2,k}} }{\|\frac{\partial f_i}{\partial u_{1,k}} \times \frac{\partial f_i}{\partial u_{2,k}}\|},
	\end{equation}
	where $\times$ is the cross product of two vectors.\\

	\subsection{Geometry-guided UDF Estimation}

	\noindent \textbf{Details of Learning the Weights.} We utilize an MLP $g_1(\cdot)$ followed by a MaxPooling operation $\texttt{MaxPool}(\cdot)$ to encode the shape of  $\mathbf{q}$'s $K$-NN, $\Omega(\mathbf{q})$, into $d_2$ dimensional point-wise and global features :
	\begin{equation}
		\begin{aligned}
			\label{PATCH:FEATURE}
			& \mathbf{F}_k(\mathbf{q})=g_1\left(\mathbf{\tilde{p}}_k{\oplus}\mathbf{\tilde{n}}_k\right),\\
			&\mathbf{F}(\mathbf{q})={\texttt{MaxPool}}\left(\{\mathbf{F}_k\}_{k=1}^K\right), 
		\end{aligned}
	\end{equation}
	where $\oplus$ stands for the concatenation operator.
	Then, we concatenate the point-wise and global features with $\mathbf{\tilde{p}}_k$ and $\mathbf{\tilde{n}}_k$, which are further fed into another two separated MLPs, denoted as $h_1(\cdot)$ and $h_2(\cdot)$, followed by the softmax operation $\texttt{SoftMax}(\cdot)$, to predict the weights $\{w_1(\mathbf{q},\mathbf{p}_k),w_2(\mathbf{q},\mathbf{p}_k)\}_{k=1}^{K}$:
	\begin{equation}
		\begin{aligned}
			&e_1\left(\mathbf{q},\mathbf{p}_k\right)=h_1\left(\mathbf{\tilde{p}}_k\oplus\mathbf{\tilde{n}}_k\oplus\mathbf{F}_k(\mathbf{q})\oplus\mathbf{F}(\mathbf{q})\right),\\
			&e_2\left(\mathbf{q},\mathbf{p}_k\right)=h_2\left(\mathbf{\tilde{p}}_k\oplus\mathbf{\tilde{n}}_k\oplus\mathbf{F}_k(\mathbf{q})\oplus\mathbf{F}(\mathbf{q})\right), \\
			&w_1\left(\mathbf{q},\mathbf{p}_k\right)=\texttt{SoftMax}(\{e_1(\mathbf{q},\mathbf{p}_k)\}_{k=1}^{K}),\\
			&w_2\left(\mathbf{q},\mathbf{p}_k\right)=\texttt{SoftMax}(\{e_2(\mathbf{q},\mathbf{p}_k)\}_{k=1}^{K}).
		\end{aligned}
	\end{equation}
	
	We also refer the reviewers to the \textit{code} for the detailed parameters of the LGR and GUE modules.\\
	
	\noindent  \textbf{Discussion on the differences from IMLS-based methods DeepIMLS \cite{DEEPIMLS} and DOG \cite{DOG}.} 
	In DeepIMLS \cite{DEEPIMLS} and DOG \cite{DOG}, the \textit{SDF} of a query point is formulated as the weighted averaging of the distances of the query point to the tangent planes of a set of \textit{MLS (moving least-squares) points}, i.e., Eq. (\textcolor{red}{2}) of DeepIMLS \cite{DEEPIMLS} and Eq. (\textcolor{red}{3})  of  DOG \cite{DOG}. In the GUE module of our method (i.e., Eq. (\textcolor{red}{4})), we formulate the \textit{UDF} of a query point as the weighted averaging of the distances of the query point to the tangent planes of a set of \textit{neighboring points on the surface}.  Although the IMLS-based methods DeepIMLS \cite{DEEPIMLS} and DOG \cite{DOG} and our method  adopt similar formulas, 
	they are \textit{significantly} different.
	\begin{compactitem}
		\item In DeepIMLS \cite{DEEPIMLS} and DOG \cite{DOG}, the MLS points are NOT distributed on surfaces, and the MLS points and their weights are \textit{simultaneously} learned  
		i.e., the MLP points are learned, which are then used to calculate their weights via a \textit{pre-defined} function with a learnable factor directly (see Eq. (\textcolor{red}{2}) of \cite{DEEPIMLS} or Eq. (\textcolor{red}{3}) of \cite{DOG}), resulting in that they may suffer from the \textit{coupling} issue. 
		\textit{Differently}, in the GUE module of our method (i.e., Eq. (\textcolor{red}{4})), \textit{only} the weights are learned. The points used to calculate the distances (i.e., $\{\textbf{p}_k\}$) are distributed \textit{on the surface} and obtained by an upsampling module (i.e., LGR) pre-trained (see Lines 491 - 497 of our manuscript) under the supervision of dense point clouds.  
		In other words, our method decouples the learning of the weights and the points.
		
		
		\item In DeepIMLS \cite{DEEPIMLS}, the gradient estimation of the SDF (i.e., Eq. (\textcolor{red}{7}) of \cite{DEEPIMLS}) is part of the real derivative of the SDF (i.e., Eq. (\textcolor{red}{2}) of \cite{DEEPIMLS}), thus coupled with SDF estimation and suffering from the inaccuracy of SDF estimation. DOG \cite{DOG} estimates the SDF gradients through the auto-differentiation of SDF (i.e., Eq. (\textcolor{red}{3}) of \cite{DOG}), which is time-consuming.
		\textit{Differently}, the GUE module of our method \textit{separates} the estimation of UDFs and gradients. Specifically, based on the continuity of a surface, GUE learns another set of weights to average the aligned normals of $K$-NNs to estimate UDF gradients, i.e., Eq. (\textcolor{red}{5}) of our manuscript. Thus, our UDF gradient can achieve \textit{bias-free}.  Note that the learned weights in Eqs. (\textcolor{red}{4}) and (\textcolor{red}{5}) of our manuscript are independent.
		
		\item  More importantly, we demonstrate the superiority of our method over DOG \cite{DOG} both quantitatively and qualitatively in Table \textcolor{red}{1} and Fig. \textcolor{red}{7} of the manuscript. Note that according to the results reported in the original paper of DOG \cite{DOG}, DOG \cite{DOG} achieves better performance than DeepIMLS \cite{DEEPIMLS}. \\
	\end{compactitem} 
	
	\subsection{Edge-based Marching Cube}
\begin{figure}[h]
		\centering
		\subfloat[]{\includegraphics[width=0.16\textwidth]{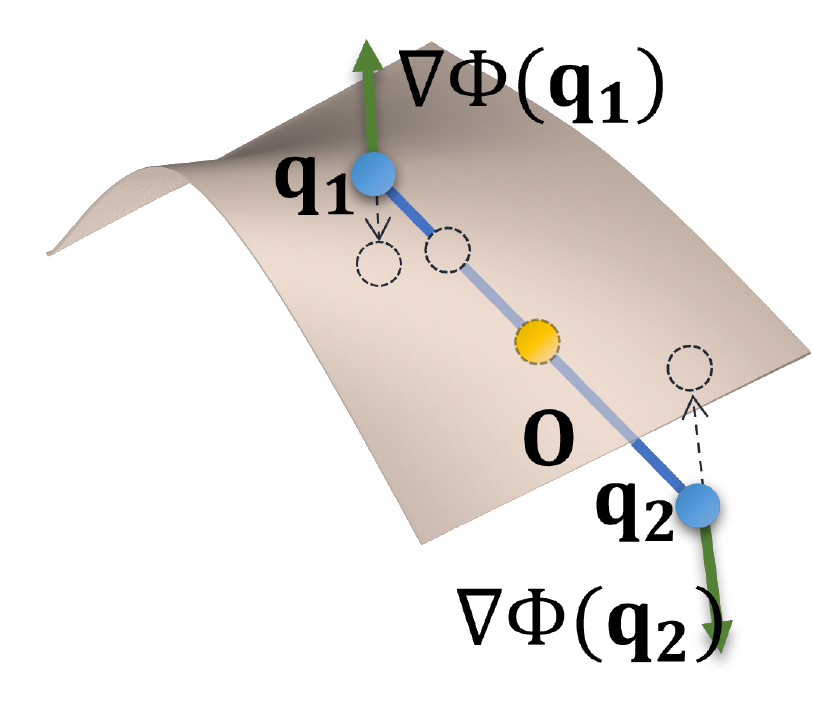} } 
		\subfloat[]{\includegraphics[width=0.16\textwidth]{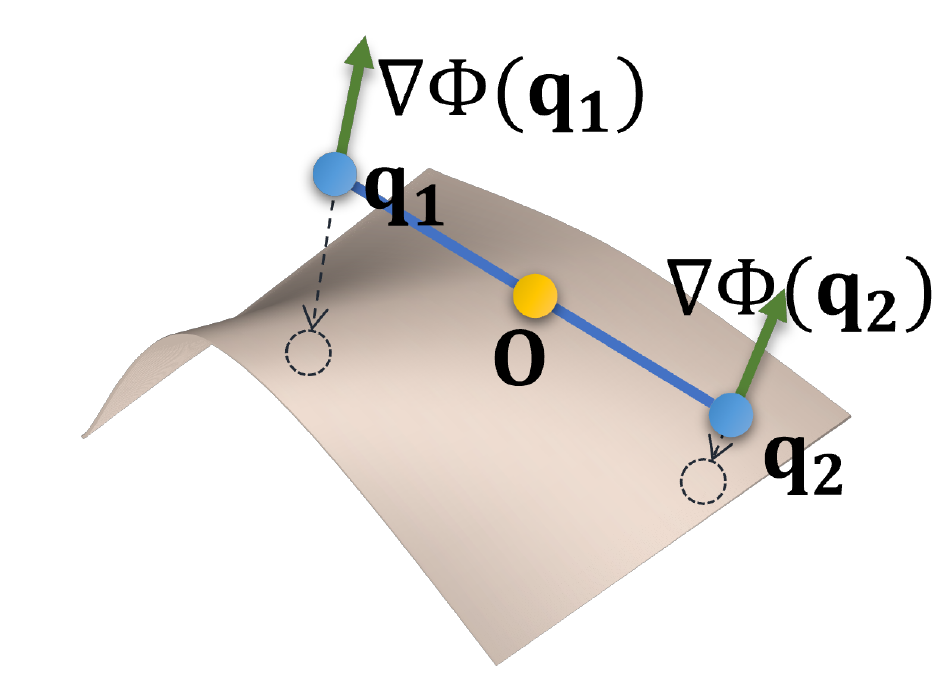} }    
		\subfloat[]{\includegraphics[width=0.16\textwidth]{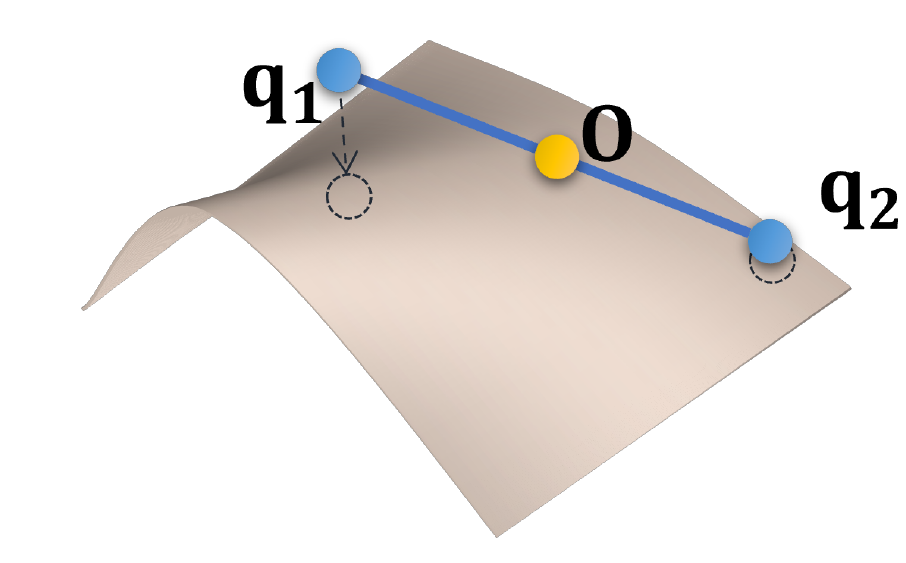} }  
		\subfloat[]{\includegraphics[width=0.16\textwidth]{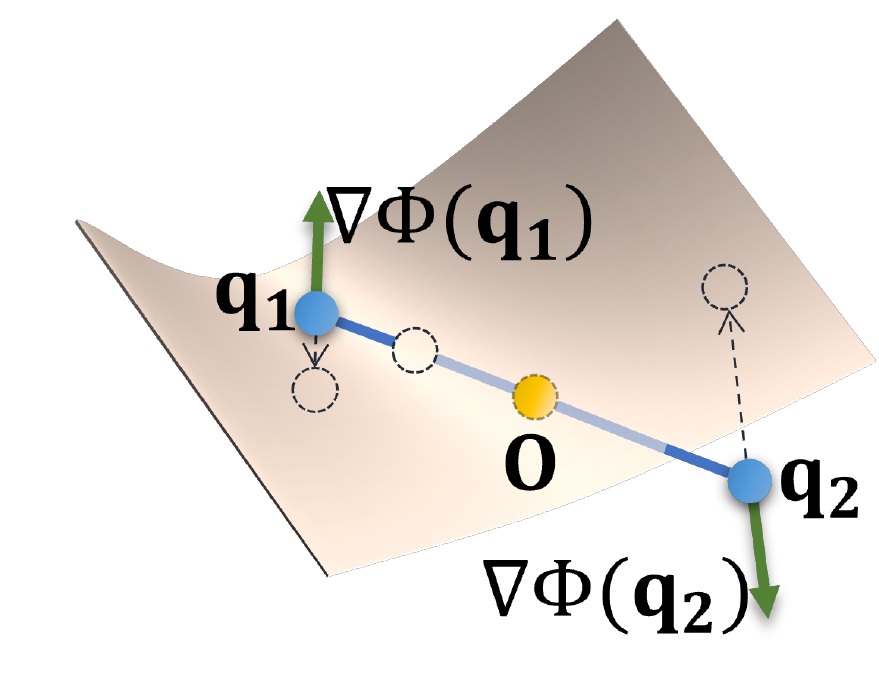} } 
		\subfloat[]{\includegraphics[width=0.16\textwidth]{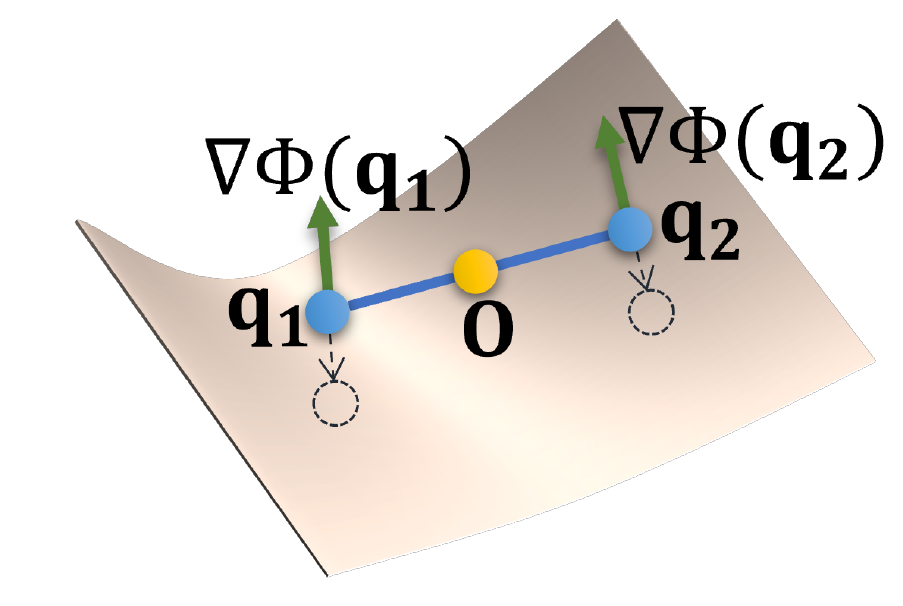} }  
		\subfloat[]{\includegraphics[width=0.16\textwidth]{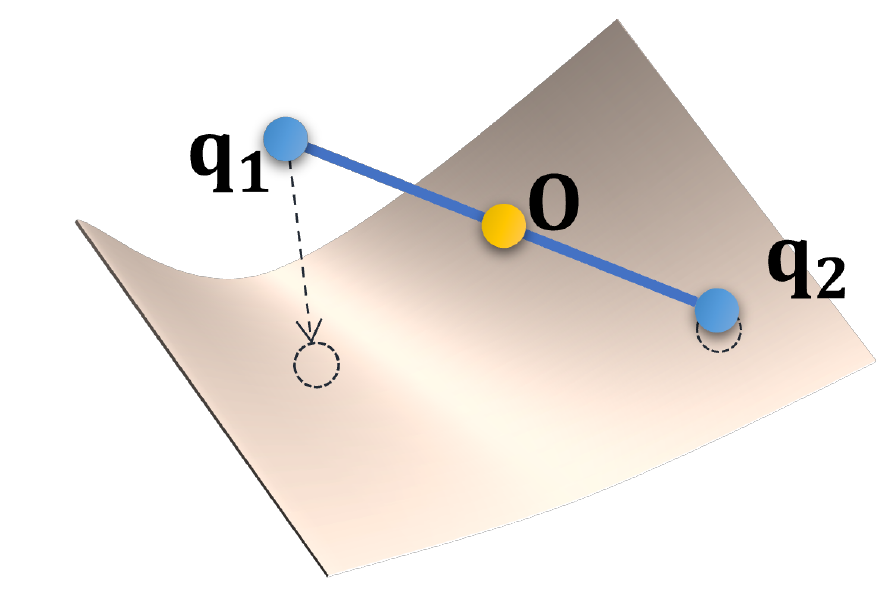} }  \\
		\subfloat[]{\includegraphics[width=0.16\textwidth]{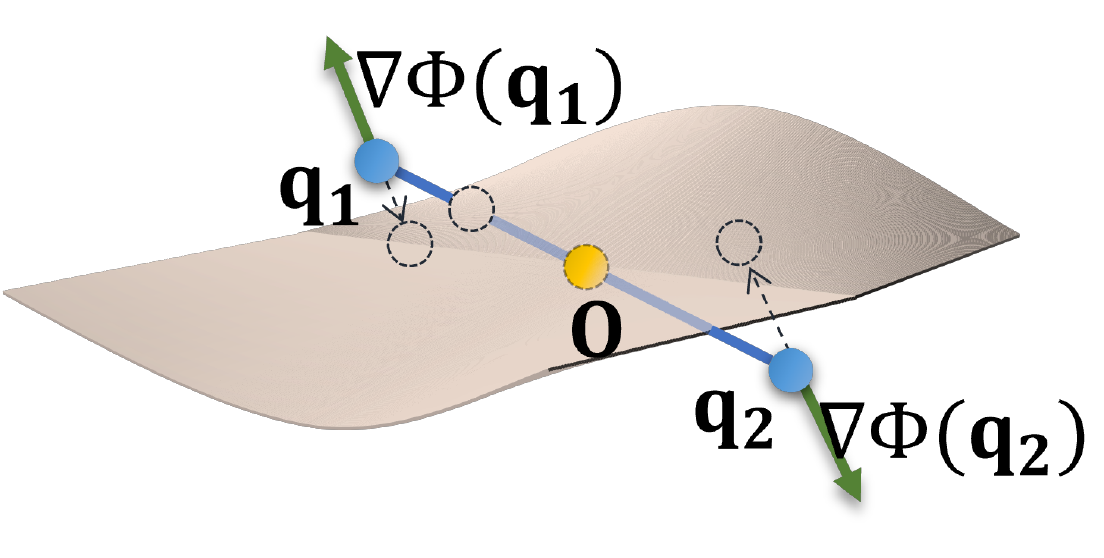} }  
		\subfloat[]{\includegraphics[width=0.16\textwidth]{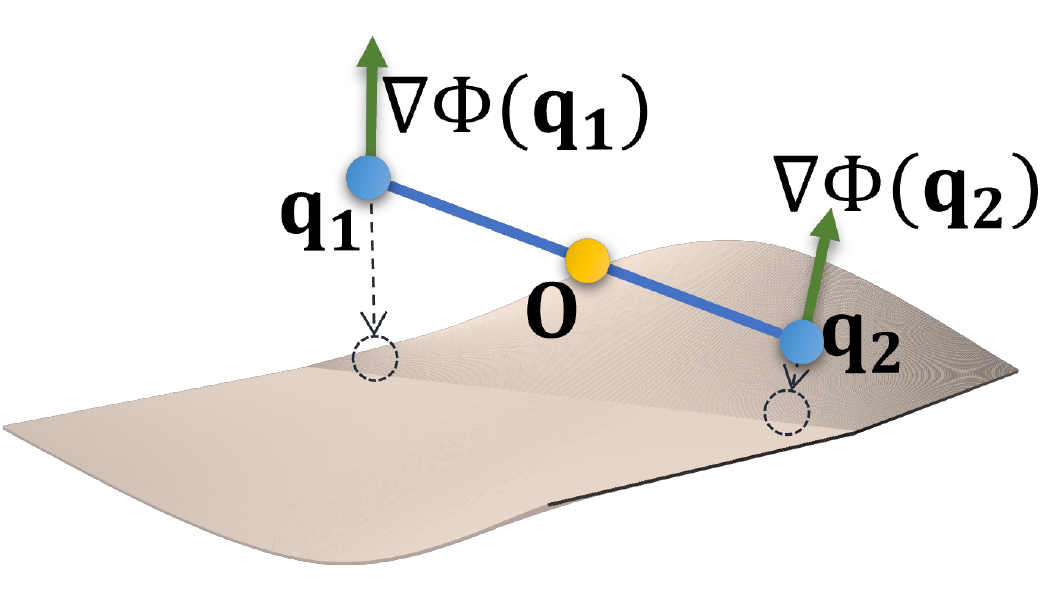} }  
		\subfloat[]{\includegraphics[width=0.16\textwidth]{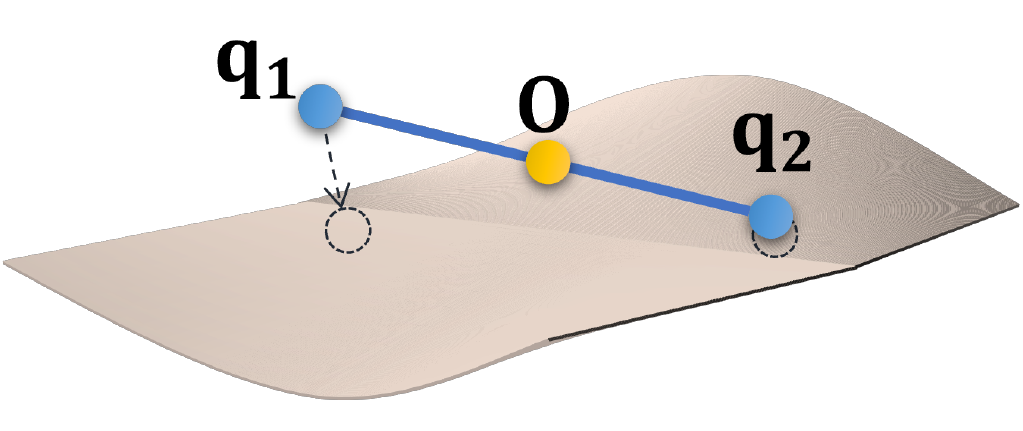} } 
		\subfloat[]{\includegraphics[width=0.16\textwidth]{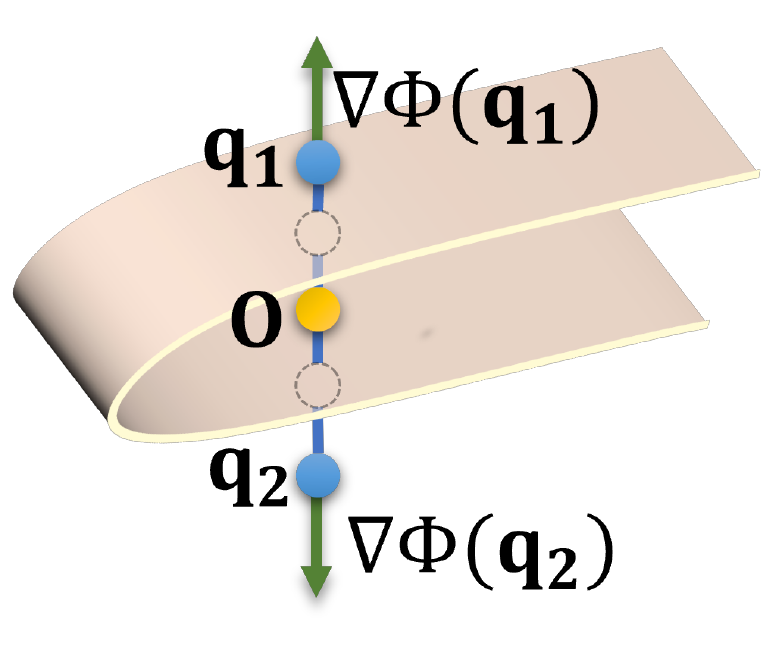} } 
		\subfloat[]{\includegraphics[width=0.16\textwidth]{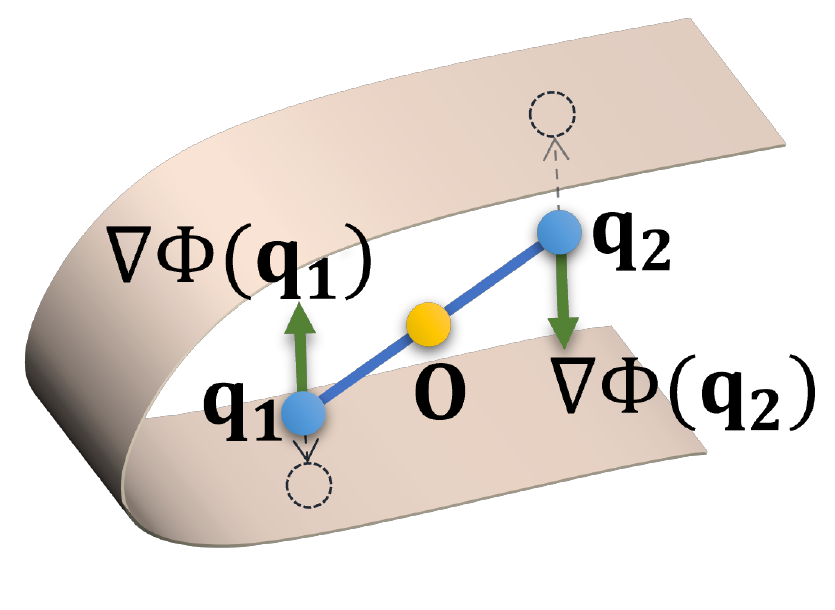} } 
		\subfloat[]{\includegraphics[width=0.16\textwidth]{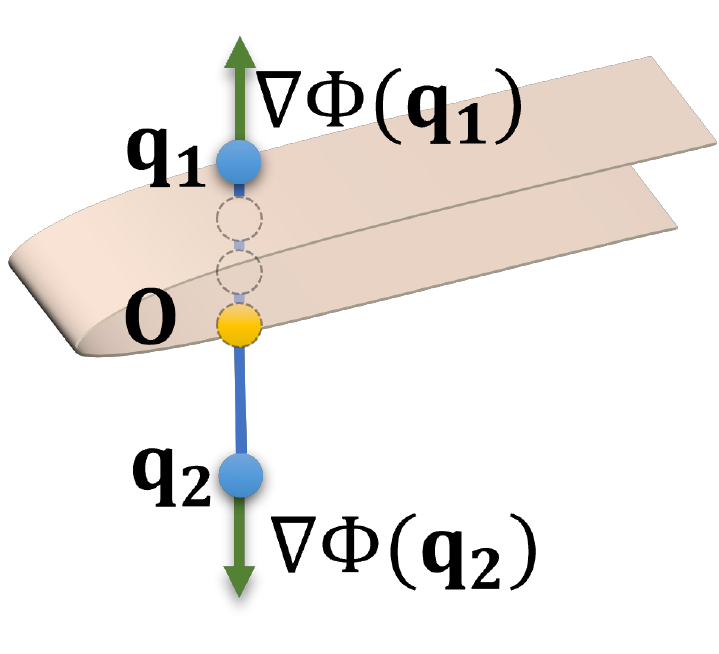} } 
		\caption{The position relationship between the surface and edge.}\label{EDGE:DETECTION} 
	\end{figure}

	\noindent\textbf{Edge Intersection Detection.} Fig. \ref{EDGE:DETECTION} shows all kinds of the position relationship between the surface and edge. \\
	
	\noindent\textbf{Triangle Extraction.} Under the assumption that the space of a small 3D cube could always be divided into two sides, we can classify its 8 vertices into 2 categories, i.e., inside or outside the surface. Accordingly, we can extract the triangles using the lookup table of Marching Cube \cite{MARCHINGCUBE}. Specifically, let $o_i\in\{0,~1\}$ be the occupancy of the $i$-th ($i=1,2,...,8$) vertex, and {$c_{ij}\in\{0,~1\}$} the intersection detection result on the connection between vertices $i$ and $j$. For each condition in the lookup table, we define its cost function as 
	\begin{equation}
		\mathcal{L}=\sum_{1\leq i<j\leq 8}\texttt{XOR}\left(c_{ij},\texttt{XOR}\left(o_i,o_j\right)\right),
	\end{equation}
	where $\texttt{XOR}(\cdot, \cdot)$ is the exclusive OR operation. From all the combinations in the lookup table, we select the one with the smallest cost as the current cube's vertex condition and extract the triangles through the lookup table.

	\subsection{Loss Function} 
	We use the Chamfer Distance (CD) to supervise the upsampled point clouds produced by the LGR module,
	\begin{equation}
		\mathcal{L}_{\rm PU}=\texttt{CD}(\mathcal{P}_M,~\mathcal{P}_M^{*}), 
	\end{equation}
	where $\mathcal{P}_M^{*}$ is the ground-truth dense point cloud, and 
	\begin{equation}
		\texttt{CD}(\mathcal{P}_M,~\mathcal{P}_M^{*})=\frac{1}{2NM}\left(\sum_{\mathbf{p}\in\mathcal{P}_M}\mathop{\text{min}}\limits_{\mathbf{p}^{'}\in\mathcal{P}_M^{*}}\|\mathbf{p}-\mathbf{p}'\|\right) +\frac{1}{2NM}\left(\sum_{\mathbf{p}'\in\mathcal{P}_M^{*}}\mathop{\text{min}}\limits_{\mathbf{p}\in\mathcal{P}_M}\|\mathbf{p}-\mathbf{p}'\|\right).  \nonumber
	\end{equation}
	
	As for the GUE module, we use the mean absolute error between the predicted and ground-truth UDF and the cosine distance between predicted and ground-truth gradients to supervise its learning,
	\begin{align}
		&\mathcal{L}_{\rm UDF}=\frac{1}{N'} \sum_{t=1}^{N'}\|\Phi(\mathbf{q}_t)-\texttt{U}(\mathbf{q}_t)\|, \\
		&\mathcal{L}_{\rm Grad}=\frac{1}{N'}\sum_{t=1}^{N'}\left(1-\langle\Theta(\mathbf{q}_t),\nabla\texttt{U}(\mathbf{q}_t)\rangle\right),
	\end{align}
	where $\{\mathbf{q}_t\}_{t=1}^{N'}$ are $N'$ sampled query points, and $\texttt{U}(\mathbf{q}_t)$ and $\nabla\texttt{U}(\mathbf{q}_t)$ are the ground-truth UDF and its gradient, respectively. 

	\section{Experiments} \label{EXP}
	\subsection{Implementation Details} 
	During training, we set the upsampling factor  $M=16$, and the boundary of $\mathcal{D}$  $\delta =0.1$. The ground-truth dense point clouds were obtained by sampling the corresponding mesh models through Poisson Disk Sampling \cite{POISSONSAMPLING}.
	For each shape, we randomly sampled $N'=2048$ points on the surface, then added Gaussian noise with the standard deviation of 0.05, 0.02, or 0.03 to move these points away from the mesh to generate query points in each training iteration. The feature dimensions were $d_1=384$ and $d_2=128$.
	We used the Adam \cite{ADAM} optimizer to train our network and set the batch size to  16. 
	
	During inference, the threshold for edge intersection detection in Sec. \textcolor{red}{3.3} of the manuscript was set to $\tau=5\times 10^{-4}$, and the resolution of E-MC was 128 for the shapes and garments in the ShapeNet and MGN datasets, and 192 for the shapes or scenes in the ShapeNet car and ScanNet datasets.
	
	\subsection{Evaluation Metric} 
	Following previous work \cite{OCCNET,CONVOCCNET,SAP,POCO,GIFS}, we used CD and F-Score to evaluate the reconstruction accuracy. We denote the reconstructed and ground-truth mesh by $\mathcal{S}_\text{REC}$ and $\mathcal{S}_\text{GT}$, on which we randomly sample a set of $10^5$ points, denoted as $\mathcal{P}_\text{REC}$ and $\mathcal{P}_\text{GT}$ and the CD of these two meshes is that of the two point sets, i.e., 
	\begin{equation}
		\texttt{CD}(\mathcal{S}_\text{REC},\mathcal{S}_\text{GT})=\texttt{CD}(\mathcal{P}_\text{REC},\mathcal{P}_\text{GT}).
	\end{equation}
	The F-Score is defined as the harmonic mean between the precision and the recall of points that lie within a certain distance threshold $\epsilon$ between $\mathcal{S}_\text{REC}$ and $\mathcal{S}_\text{GT}$,
	\begin{equation}
		\texttt{F-Score}(\mathcal{S}_\text{REC},\mathcal{S}_\text{GT},\epsilon)=\frac{2\cdot\texttt{Recall}\cdot\texttt{Precision}}{\texttt{Recall}+\texttt{Precision}},
	\end{equation}
	where
	\begin{align}
		\texttt{Recall}(\mathcal{S}_\text{REC},\mathcal{S}_\text{GT},\epsilon)&=\left| \left\{ \mathbf{p}_1\in\mathcal{P}_\text{REC},\ \text{s}.\text{t}. \ \mathop{\text{min}}_{\mathbf{p}_2\in\mathcal{P}_\text{GT}}\|\mathbf{p}_1-\mathbf{p}_2\|<\epsilon \right\} \right|, \nonumber \\
		\texttt{Precision}(\mathcal{S}_\text{REC},\mathcal{S}_\text{GT},\epsilon)&=\left| \left\{ \mathbf{p}_2\in\mathcal{P}_\text{GT},\ \text{s}.\text{t}. \ \mathop{\text{min}}_{\mathbf{p}_1\in\mathcal{P}_\text{REC}}\|\mathbf{p}_1-\mathbf{p}_2\|<\epsilon \right\} \right|. \nonumber
	\end{align}
	
	In the ablation study, we evaluate the accuracy of UDFs and their gradients. We first set uniform grids of $64^3$ in a unit cube, and then for each shape, we chose the grids near the surface, i.e., their UDFs are larger than $5\times 10^{-4}$ and smaller than 0.02. 
	We use the absolute error and angle error to measure the evaluated UDFs and their gradients of these query points.
	
	\subsection{Settings of the Experiments on Clean Data} 
	In our manuscript, the methods under comparison, including  ONet \cite{OCCNET}, CONet \cite{CONVOCCNET}, SAP \cite{SAP}, POCO \cite{POCO}, and DOG \cite{DOG}, only conducted experiments on the noisy data in their original papers. 
	And they did not release the pre-trained networks on clean data. Although POCO \cite{POCO} released the pre-trained network on clean data, it requires ground-truth normals. Due to the different distributions of clean and noisy data, directly applying their pre-trained networks with noisy data to clean data would result in an obvious performance drop. Thus, for a fair comparison, we retrained their models on the clean data using their official codes.
	The released pre-trained networks of NDF \cite{NDF} and GIFS \cite{GIFS} were only trained on the ShapeNet car dataset, rather than the whole 13 classes, and thus, we retrained them on both clean and noisy data for fair comparisons.
	
	From the results in Table \textcolor{red}{1} of our manuscript, it can be seen that some methods, including  ONet, CONet, SAP, and DOG, achieve better performance on noisy data than clean data. The possible reasons are: (\textbf{1}) with slight perturbation, some thin structures could be extracted through OCC or SDF; and (\textbf{2}) the noisy data enhance the robustness of the models during training.
	
	\subsection{More Experimental Results}
	
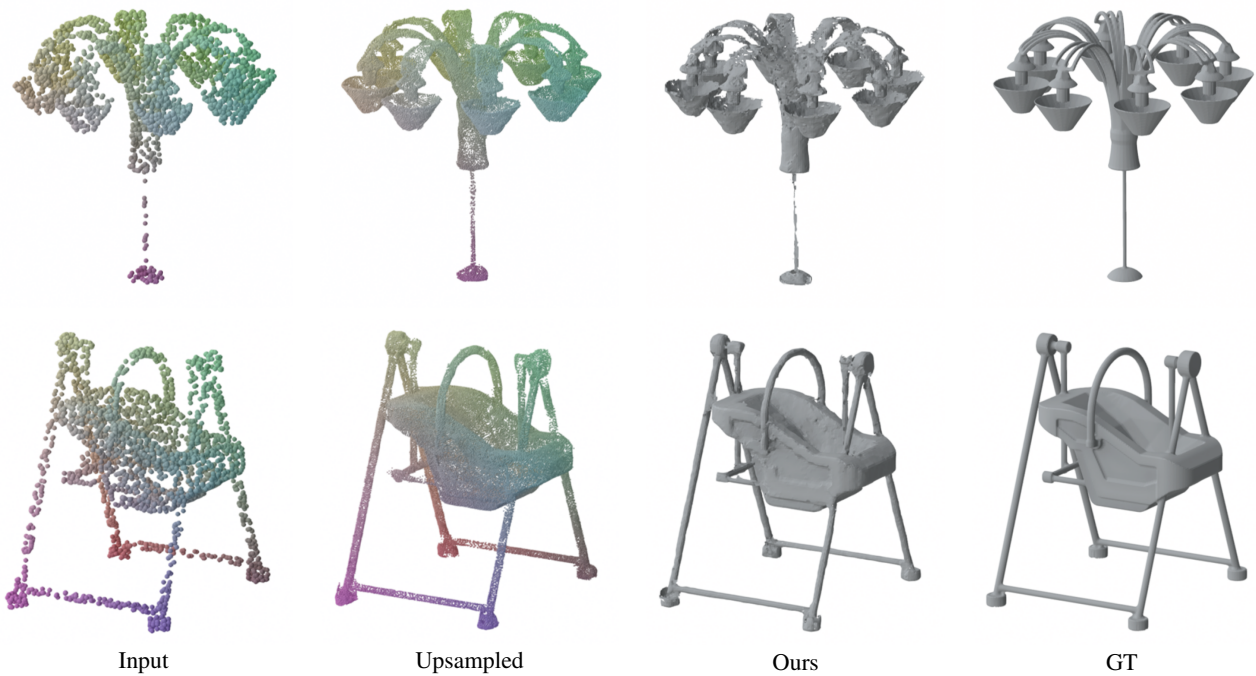
\begin{figure}[h]
		\centering
		{
			\begin{tikzpicture}[]
				\node[] (a) at (-6.5,0.6) {\includegraphics[width=0.24\textwidth]{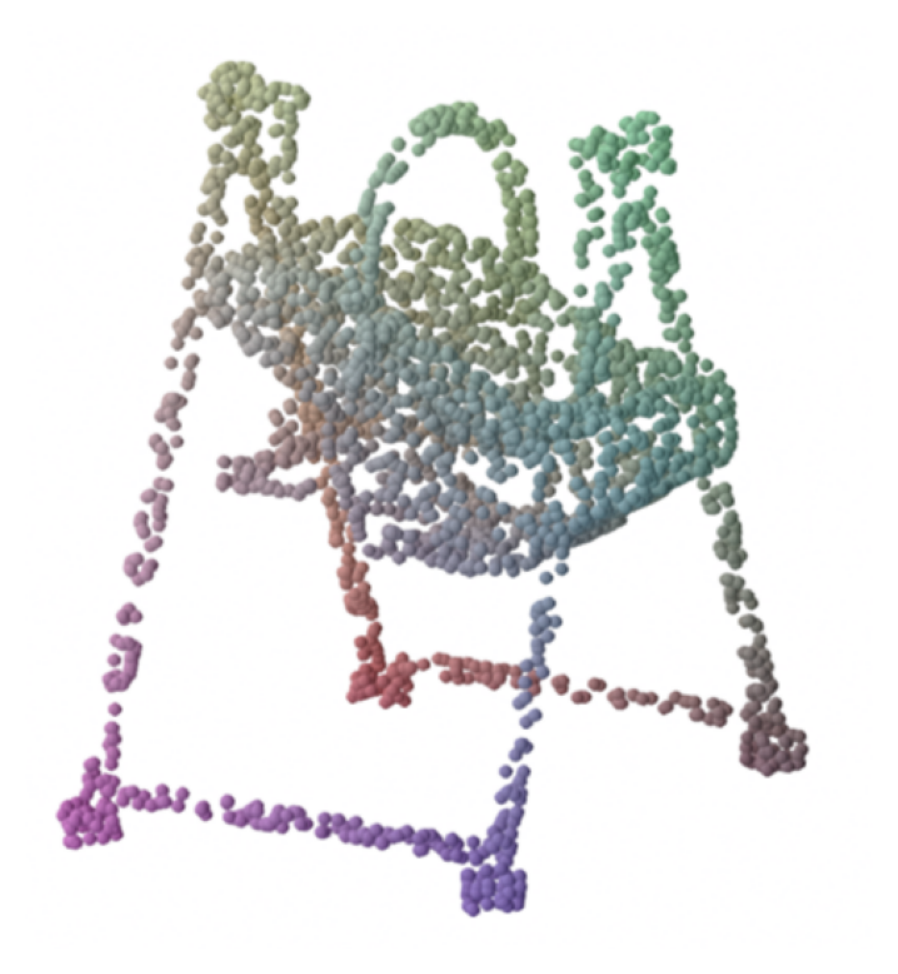} };
				\node[] (a) at (-6.5,5) {\includegraphics[width=0.24\textwidth]{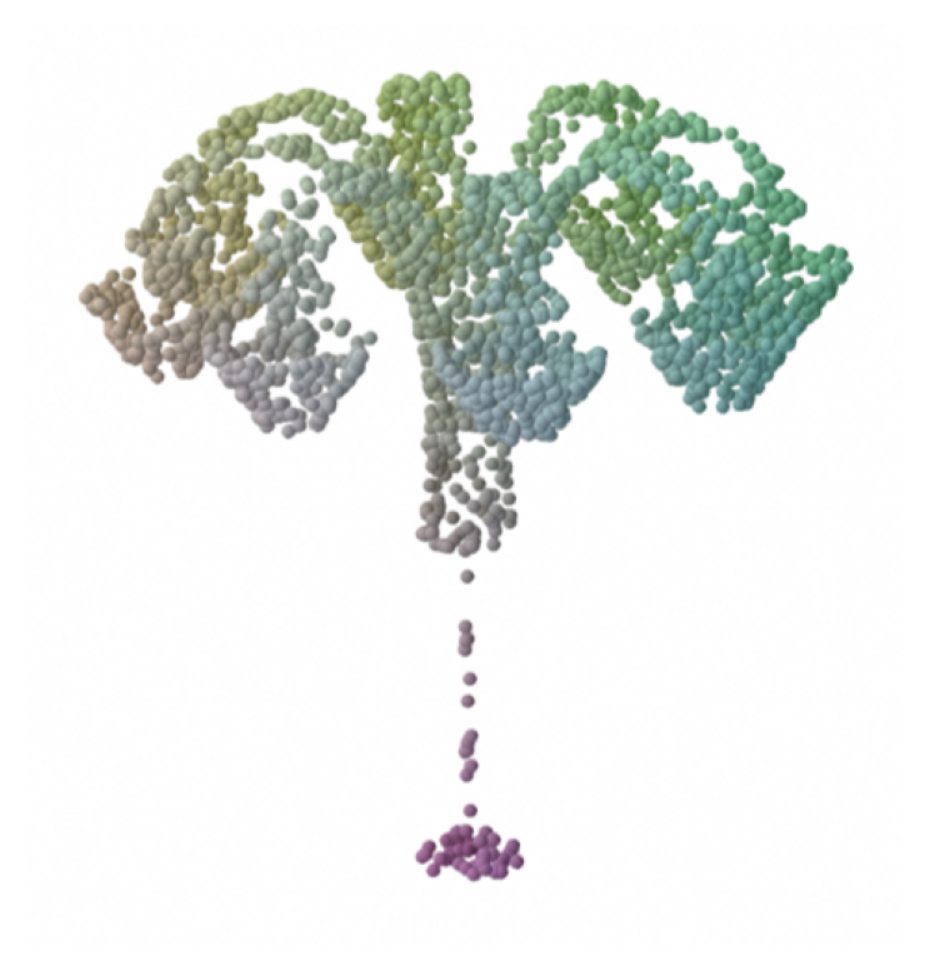} };
				\node[] (a) at (-6.5,-1.8) {\small  Input};
				
				\node[] (b) at (-6.5+13/3,0.6) {\includegraphics[width=0.24\textwidth]{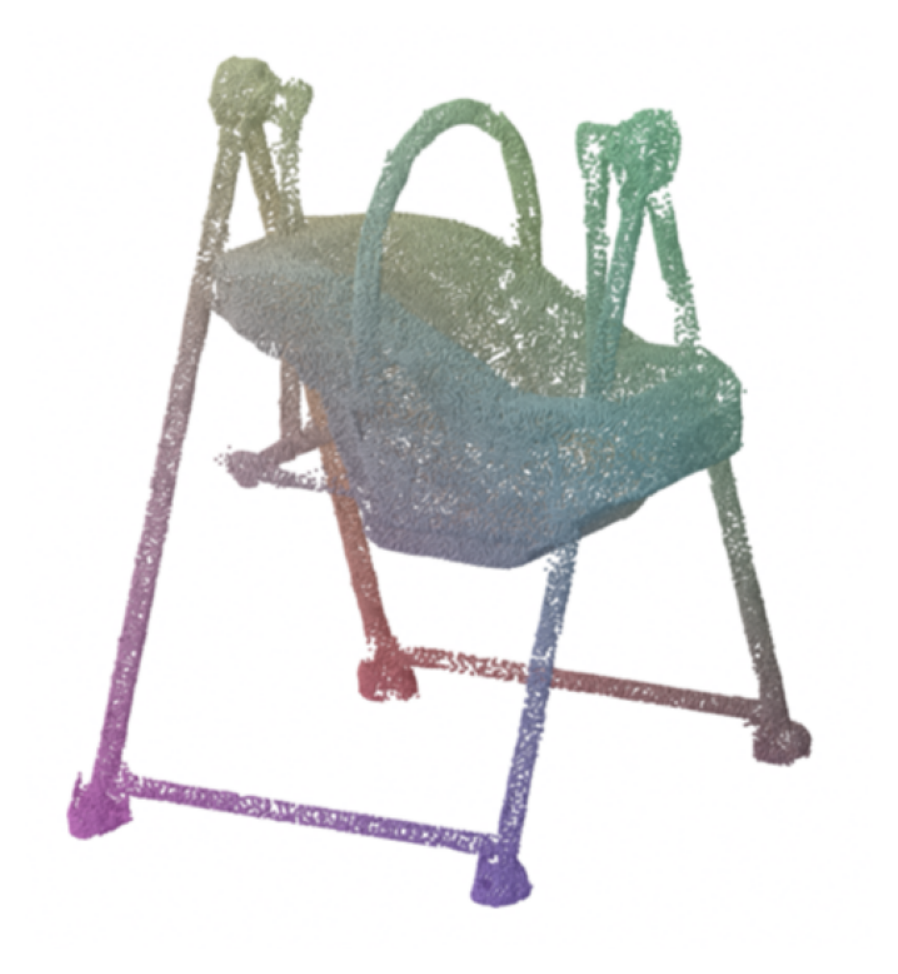}};
				\node[] (b) at (-6.5+13/3,5) {\includegraphics[width=0.24\textwidth]{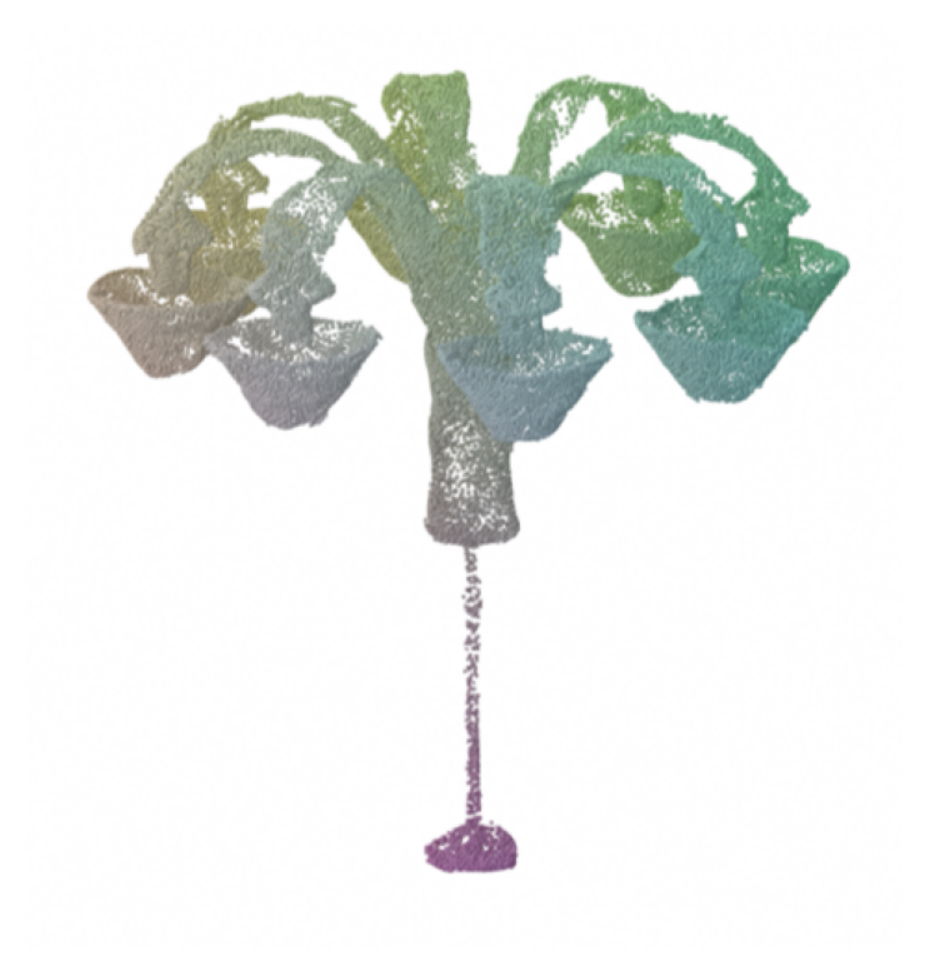}};
				\node[] (b) at (-6.5+13/3,-1.8) {\small  Upsampled };
				
				\node[] (c) at (-6.5+13/3*2,0.6) {\includegraphics[width=0.24\textwidth]{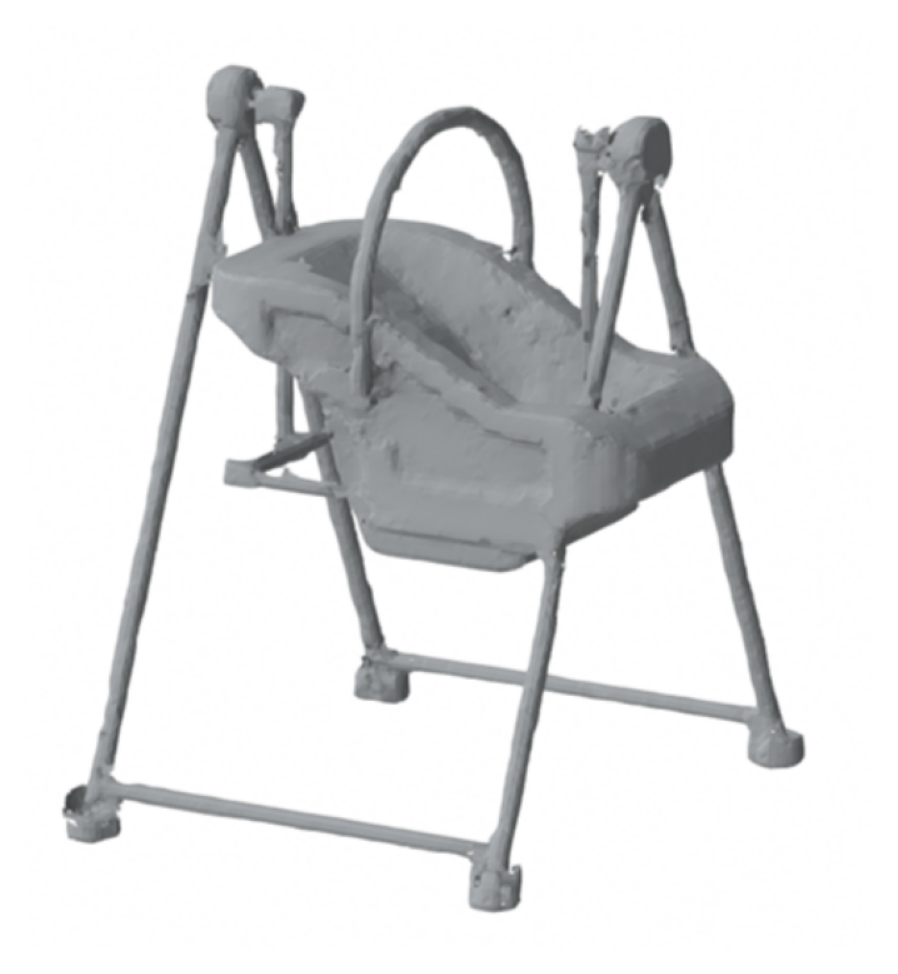}};
				\node[] (c) at (-6.5+13/3*2,5) {\includegraphics[width=0.24\textwidth]{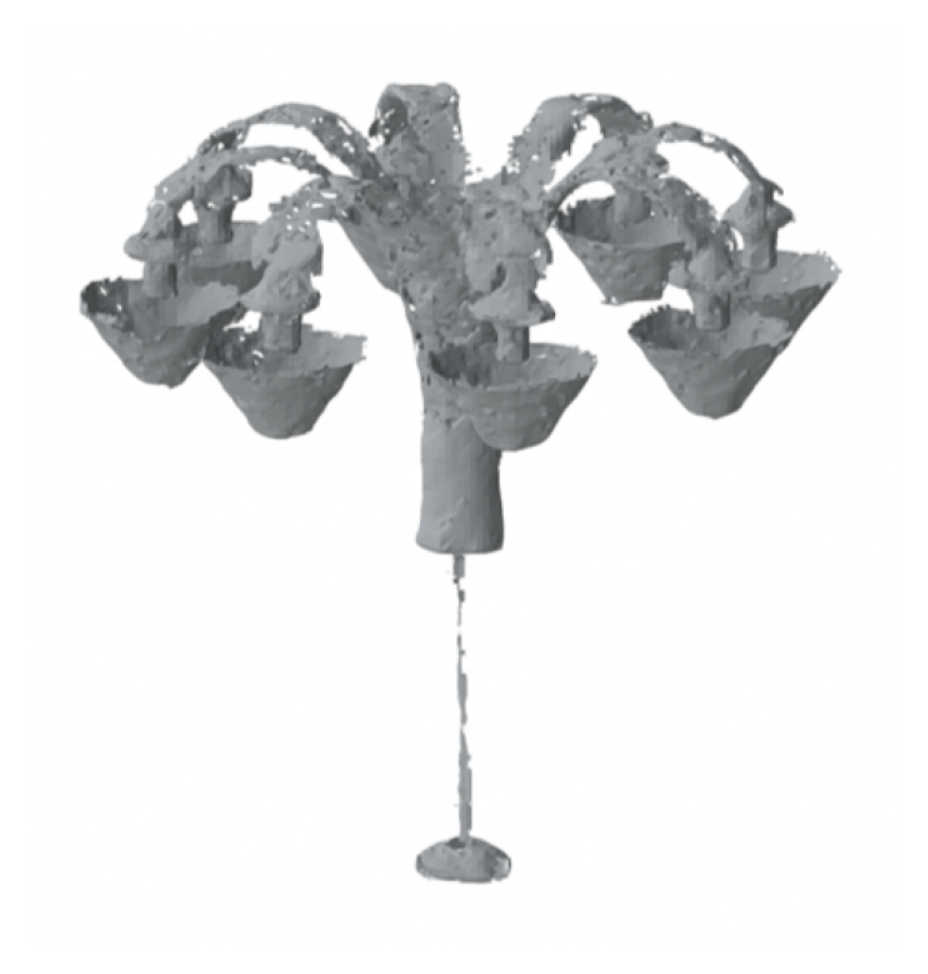}};
				\node[] (c) at (-6.5+13/3*2,-1.8) {\small  Ours  };
				
				\node[] (d) at (-6.5+13/3*3,0.6) {\includegraphics[width=0.24\textwidth]{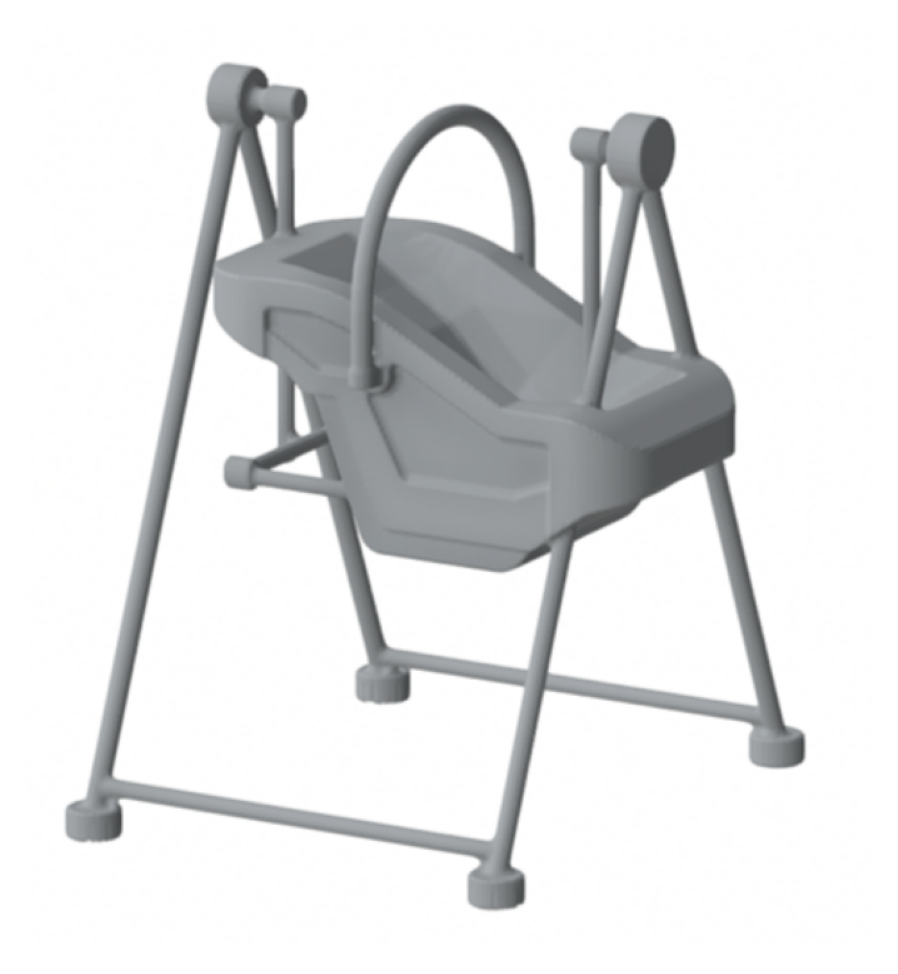}};
				\node[] (d) at (-6.5+13/3*3,5) {\includegraphics[width=0.24\textwidth]{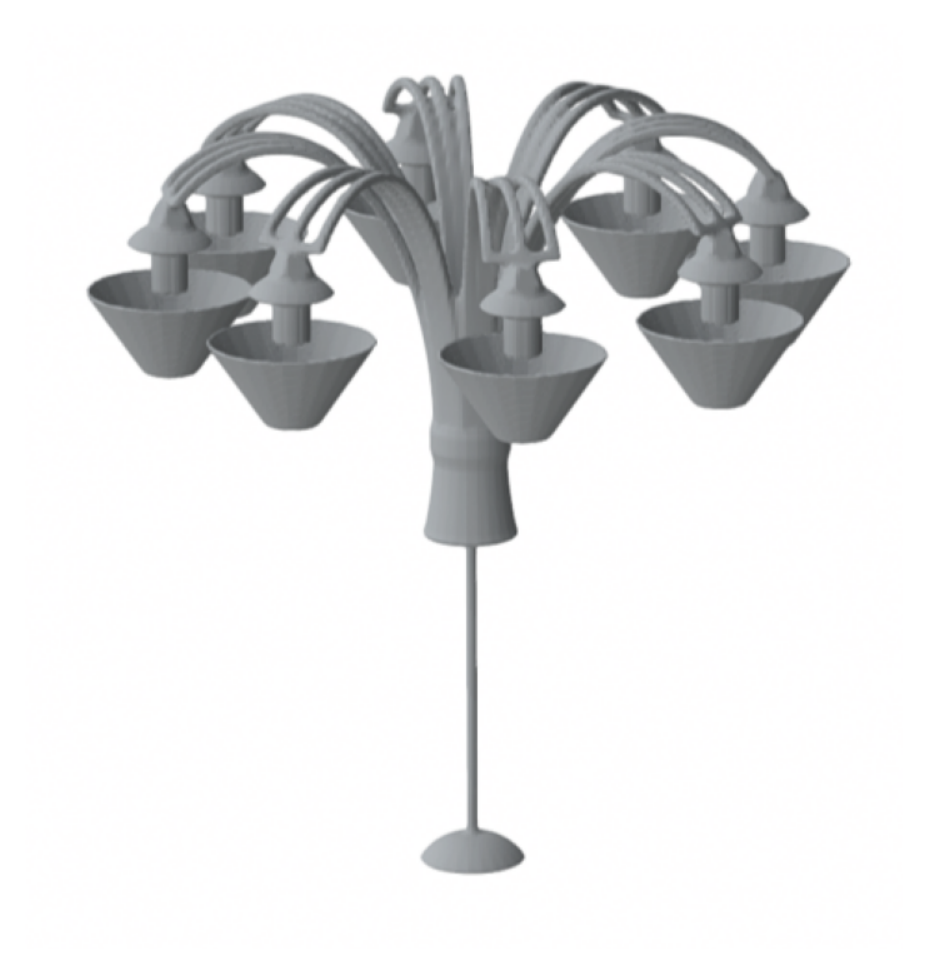}};
				\node[] (d) at (-6.5+13/3*3,-1.8) {\small  GT };
				
			\end{tikzpicture}
		}
		\setlength{\abovecaptionskip}{-0.35cm} \vspace{0.4cm}
		\caption{{Reconstruction results of complex shapes in the ShapeNet dataset. } }
		\label{COMPLEX:SHAPENET}
	\end{figure}

	\noindent{\textbf{Complex Shapes.} Fig. \ref{COMPLEX:SHAPENET} shows the reconstruction results of complex shapes in the ShapeNet dataset.}\\
	
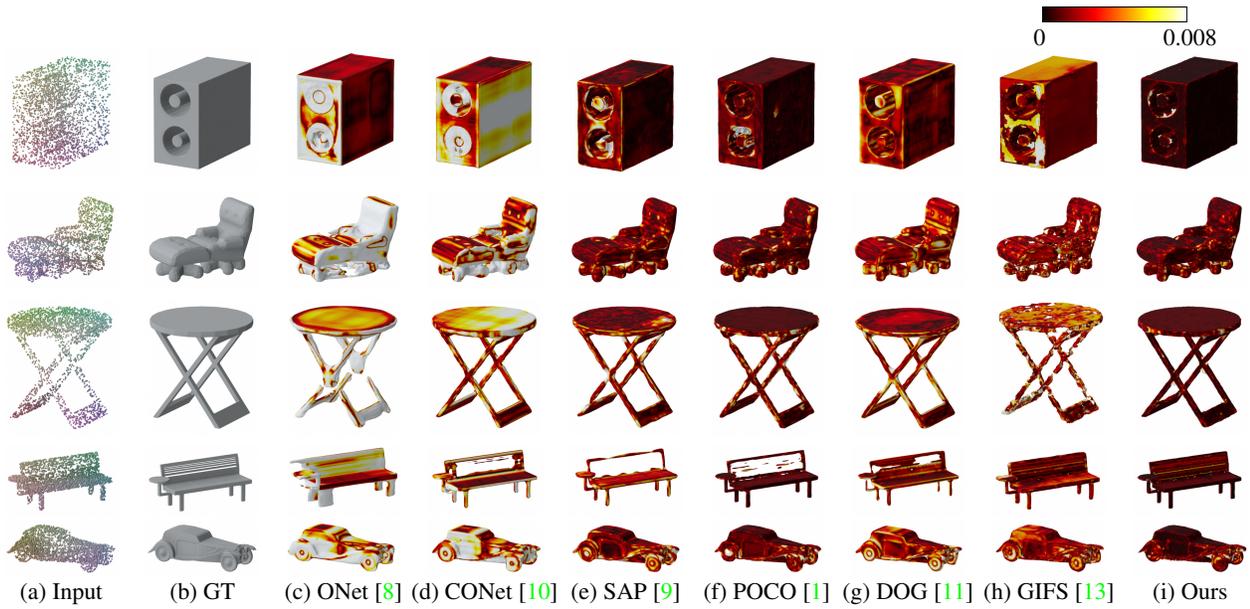
\begin{figure}[h]
		\centering
		{
			\begin{tikzpicture}[]
				\node[] (a) at (0,0.6) {\includegraphics[width=0.09\textwidth]{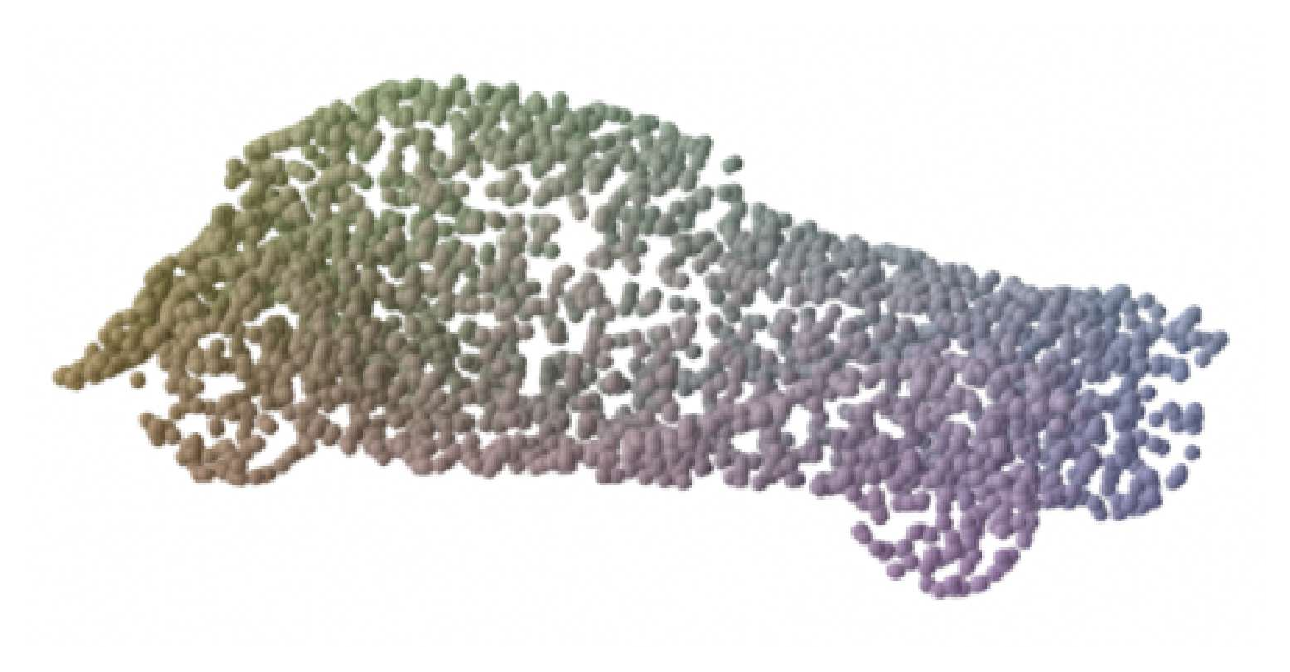} };
				\node[] (a) at (0,1.5) {\includegraphics[width=0.09\textwidth]{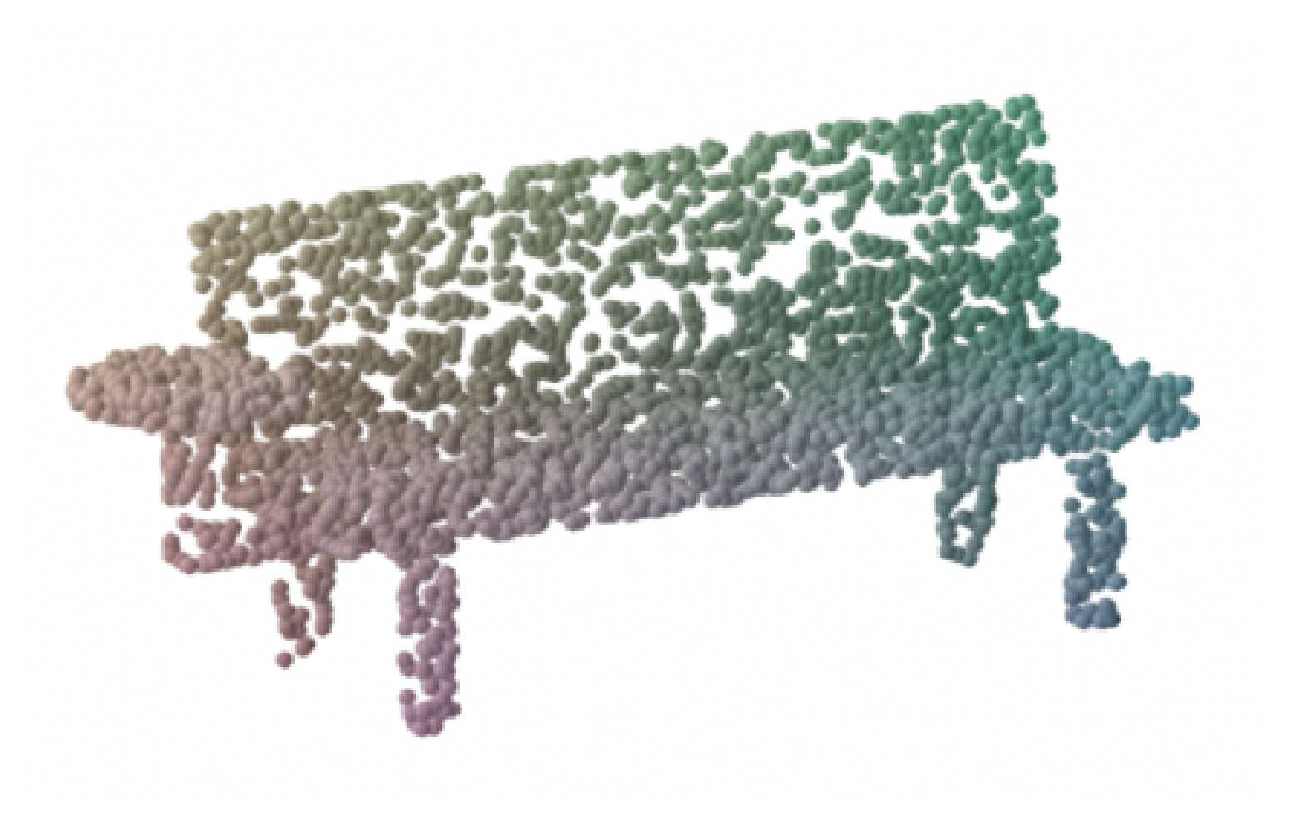} };
				\node[] (a) at (0,3) {\includegraphics[width=0.09\textwidth]{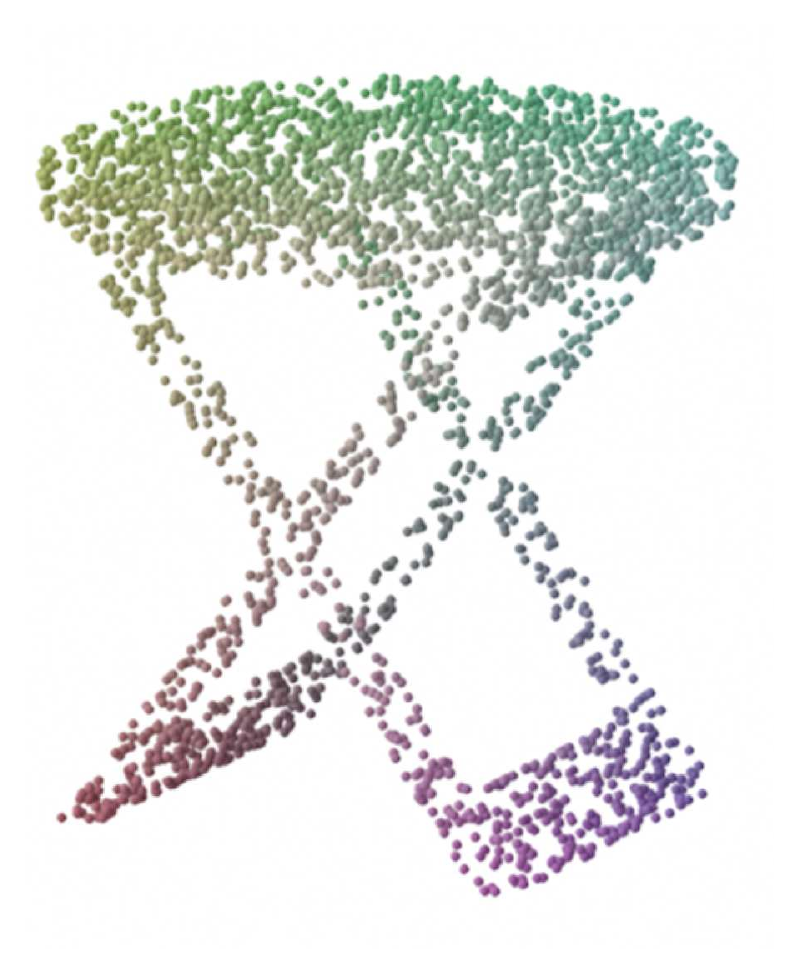} };
				\node[] (a) at (0,4.7) {\includegraphics[width=0.09\textwidth]{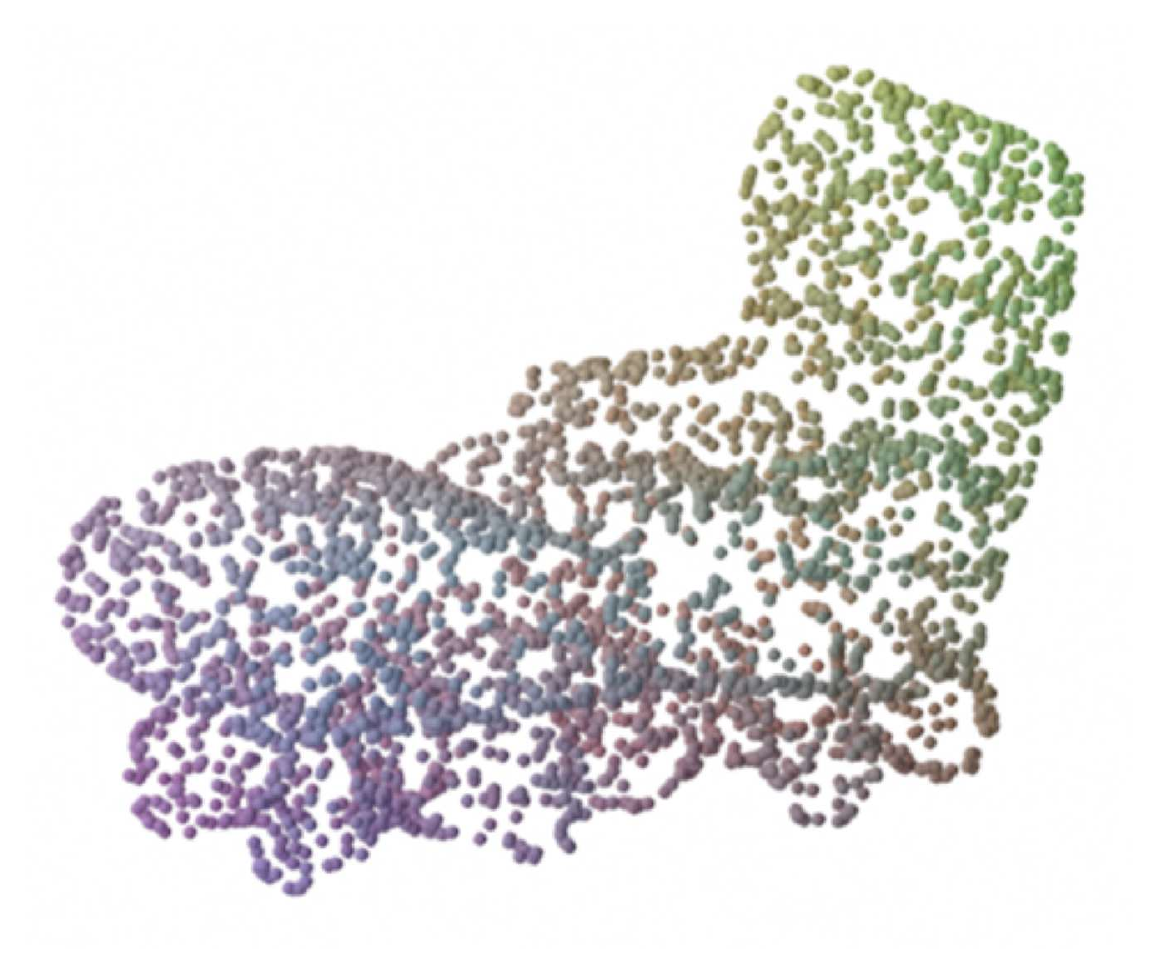} };
				\node[] (a) at (0,6.4) {\includegraphics[width=0.09\textwidth]{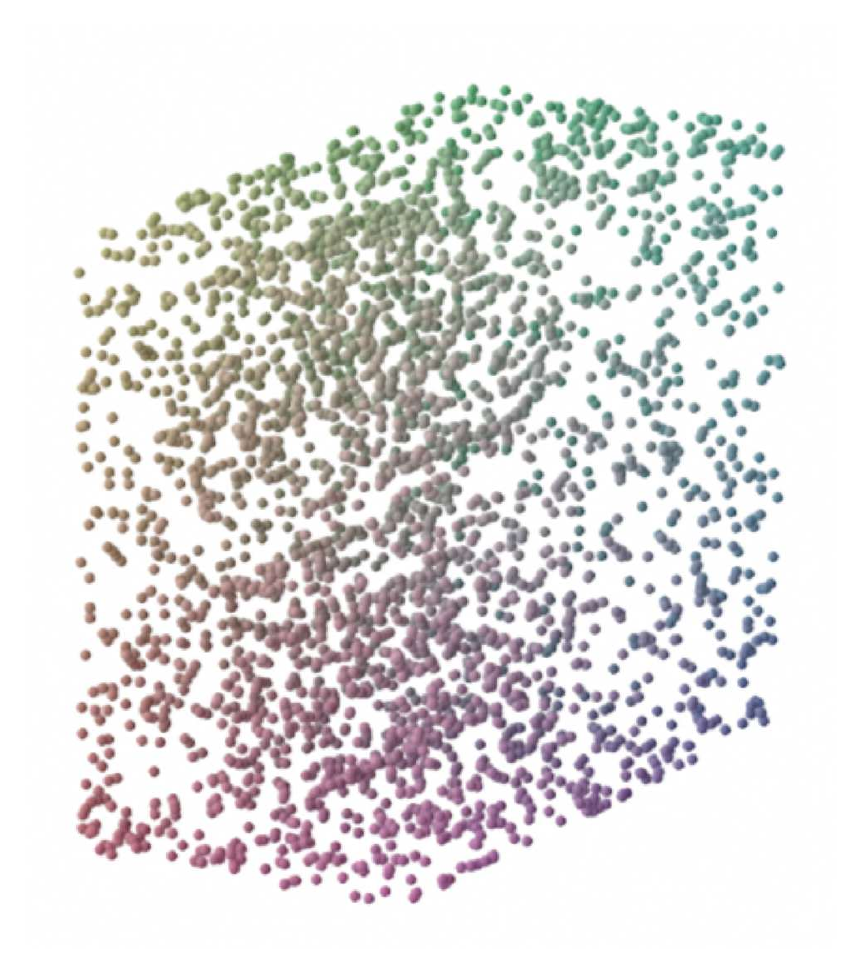} };
				\node[] (a) at (0,0) {\small (a) Input};
				
				\node[] (b) at (15/8,0.6) {\includegraphics[width=0.09\textwidth]{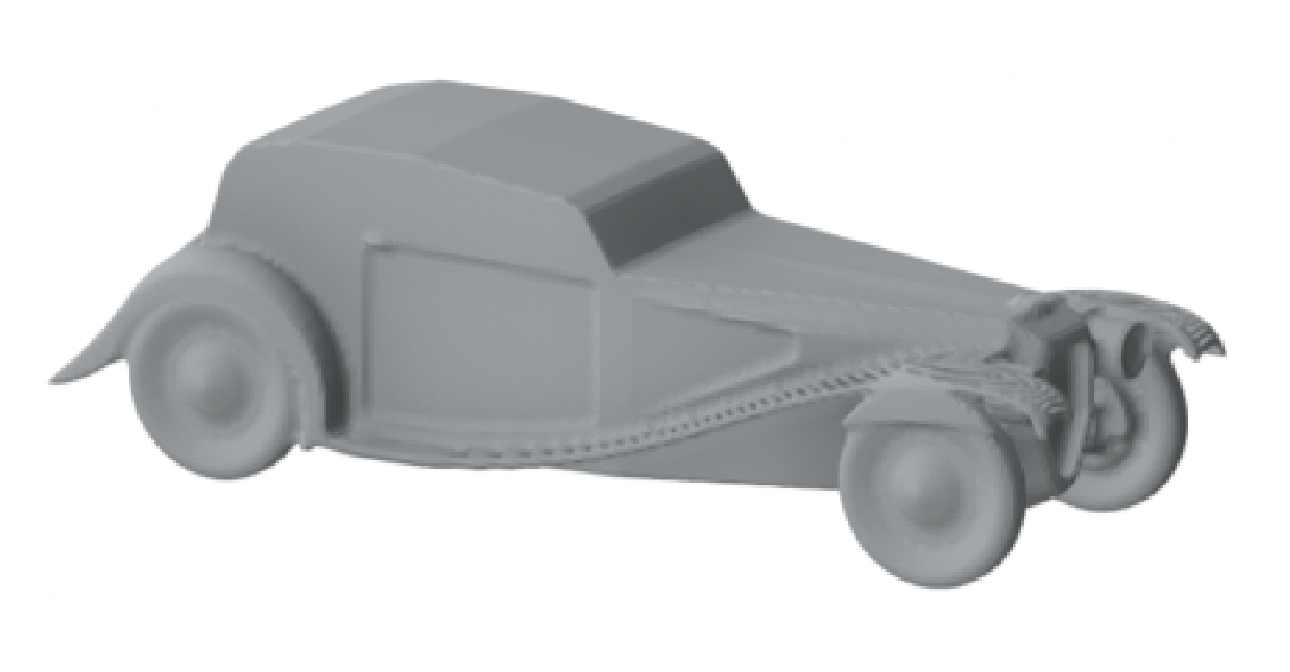}};
				\node[] (b) at (15/8,1.5) {\includegraphics[width=0.09\textwidth]{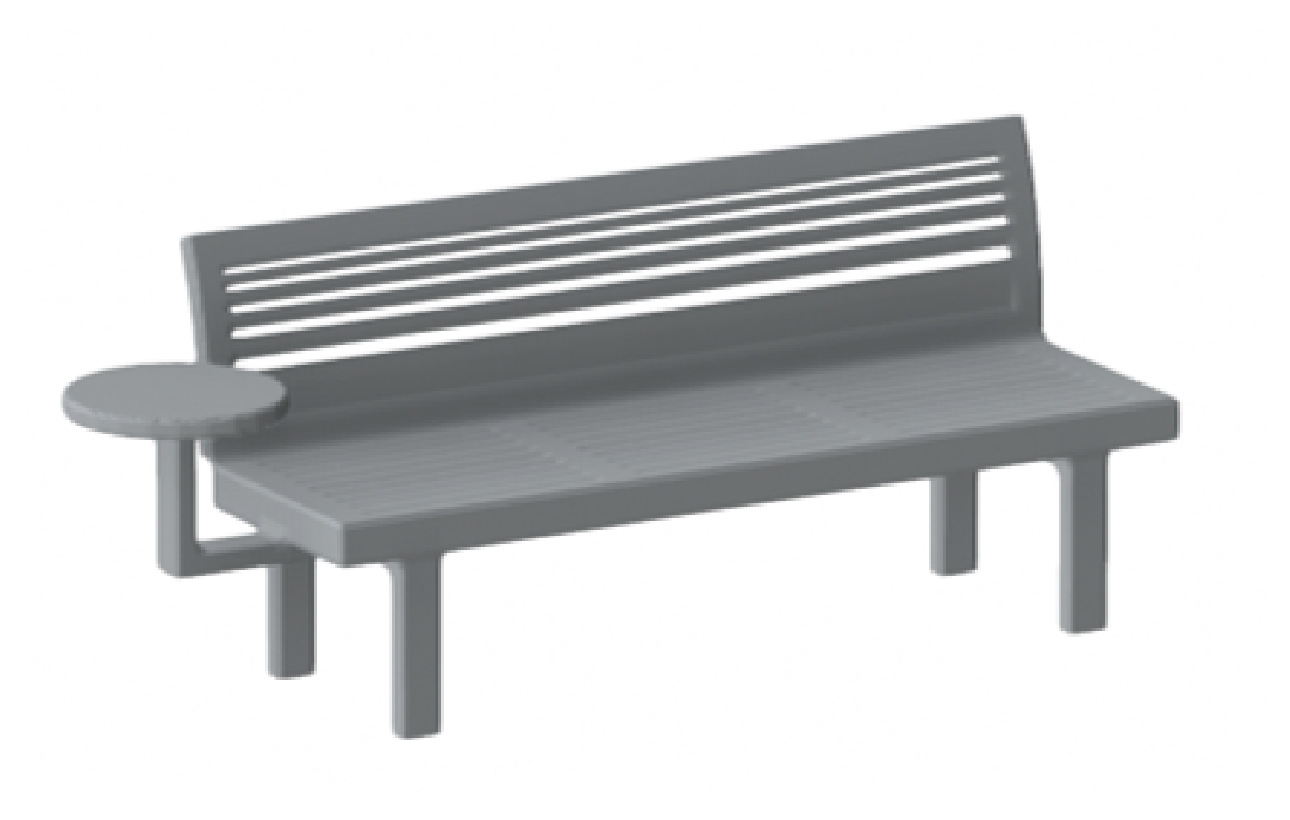}};
				\node[] (b) at (15/8,3) {\includegraphics[width=0.09\textwidth]{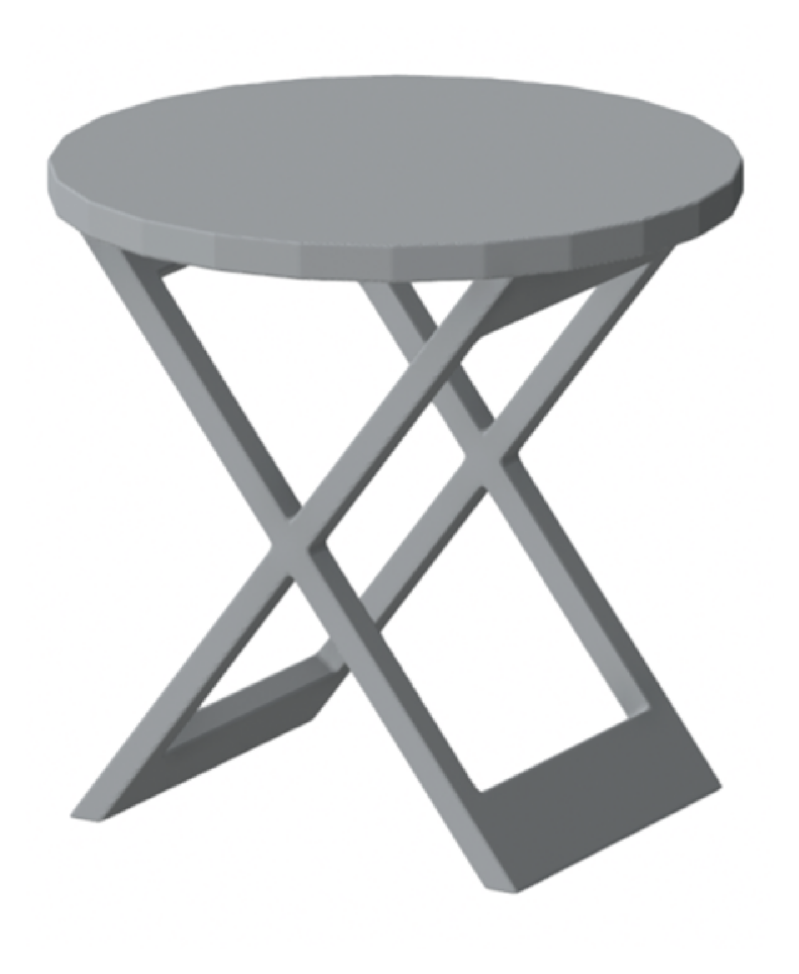}};
				\node[] (b) at (15/8,4.7) {\includegraphics[width=0.09\textwidth]{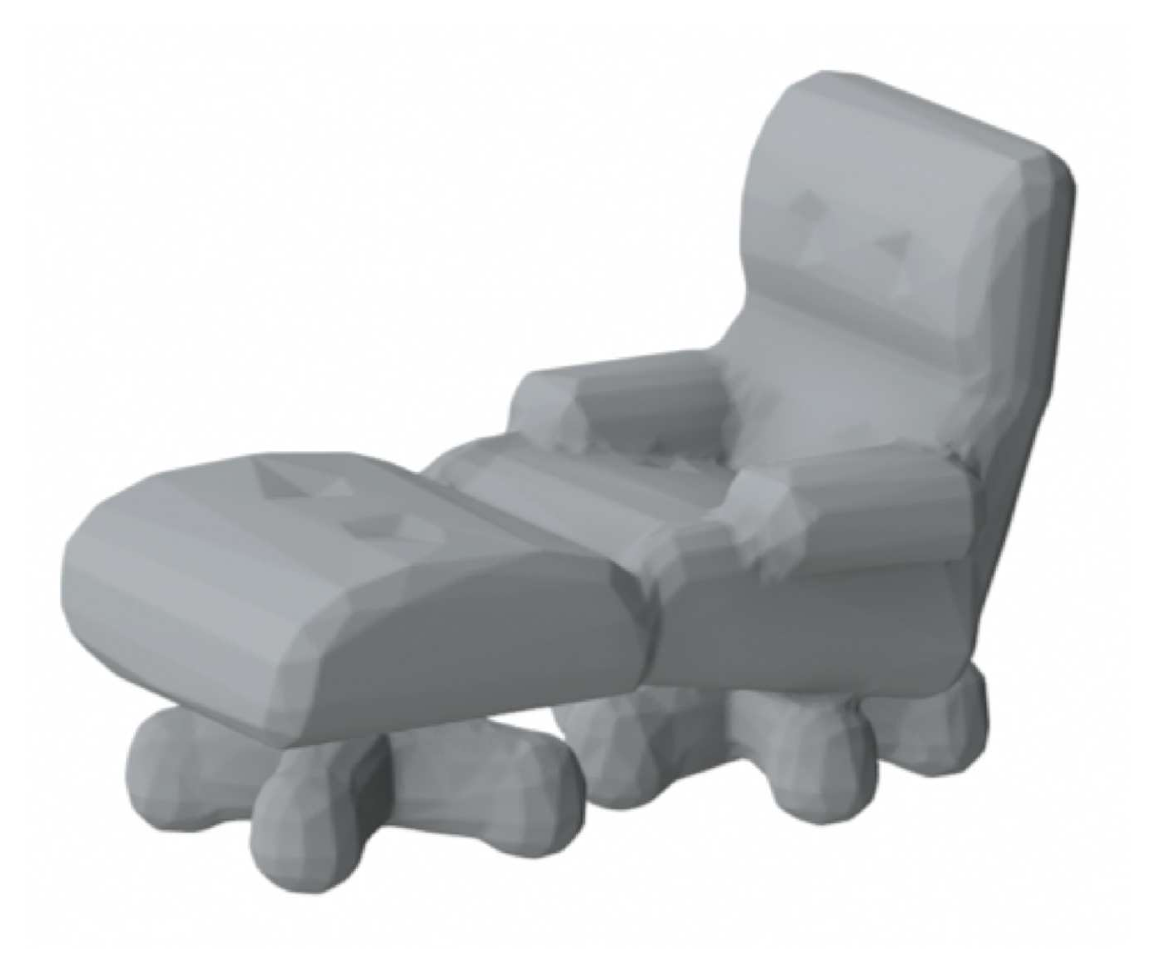}};
				\node[] (b) at (15/8,6.4) {\includegraphics[width=0.09\textwidth]{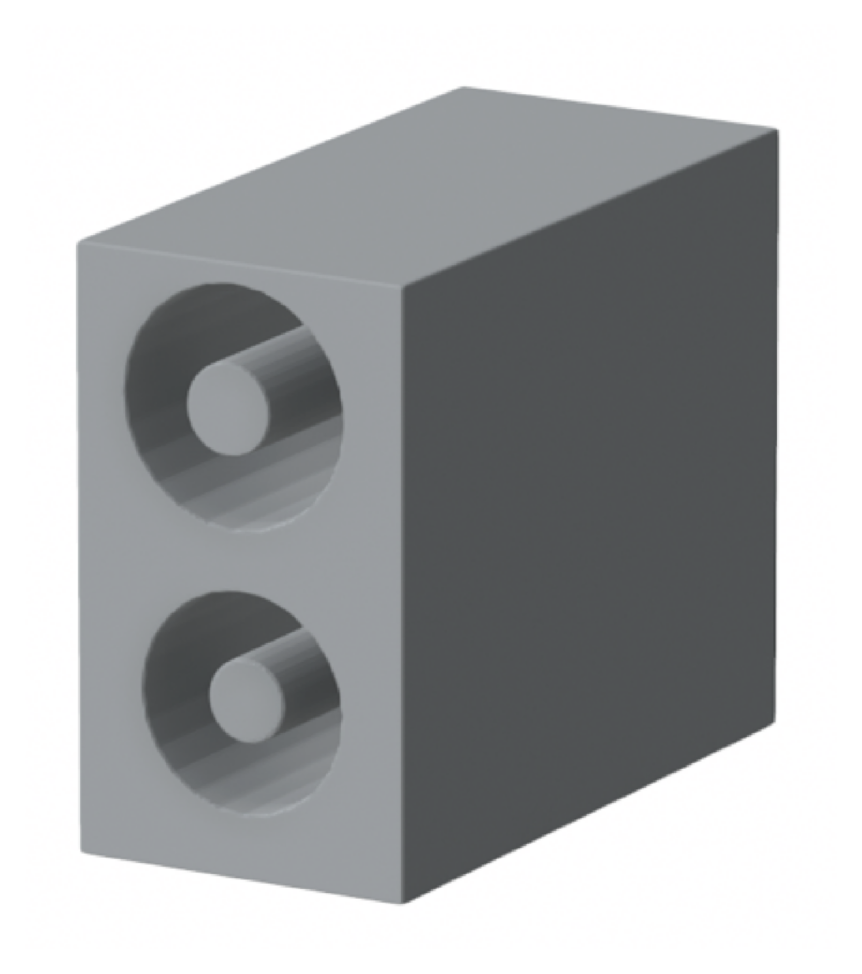}};
				\node[] (b) at (15/8,0) {\small (b) GT };
				
				\node[] (c) at (15/8*2,0.6) {\includegraphics[width=0.09\textwidth]{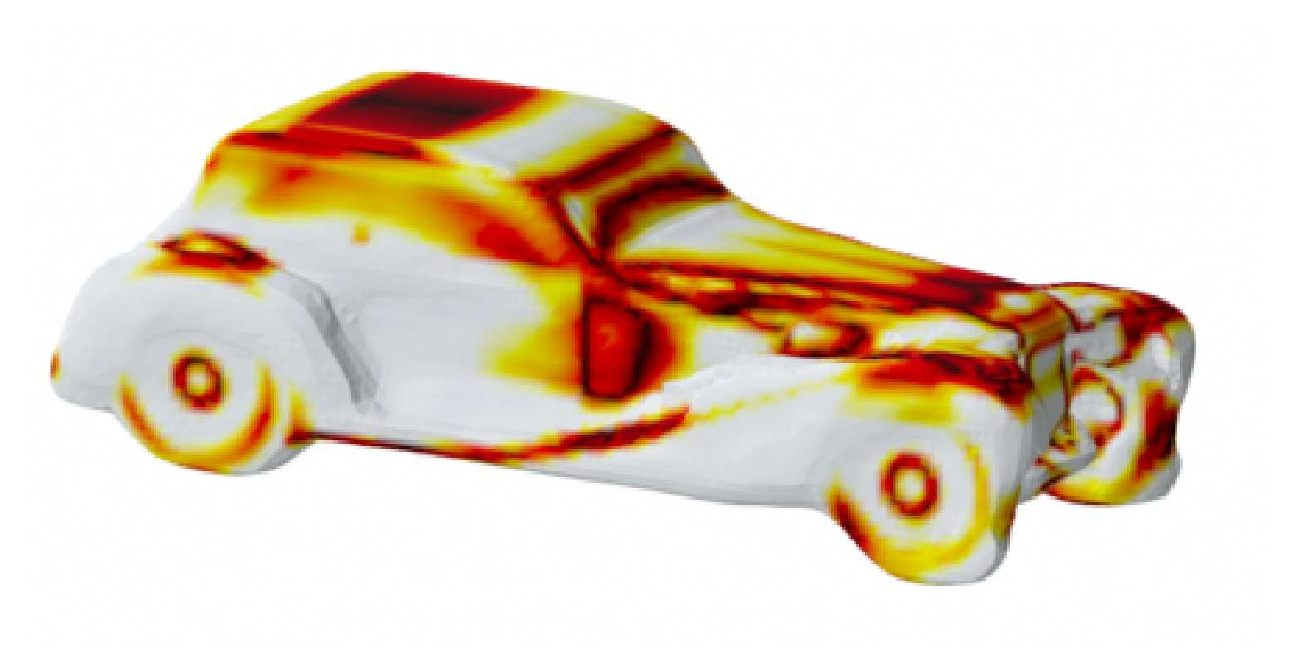}};
				\node[] (c) at (15/8*2,1.5) {\includegraphics[width=0.09\textwidth]{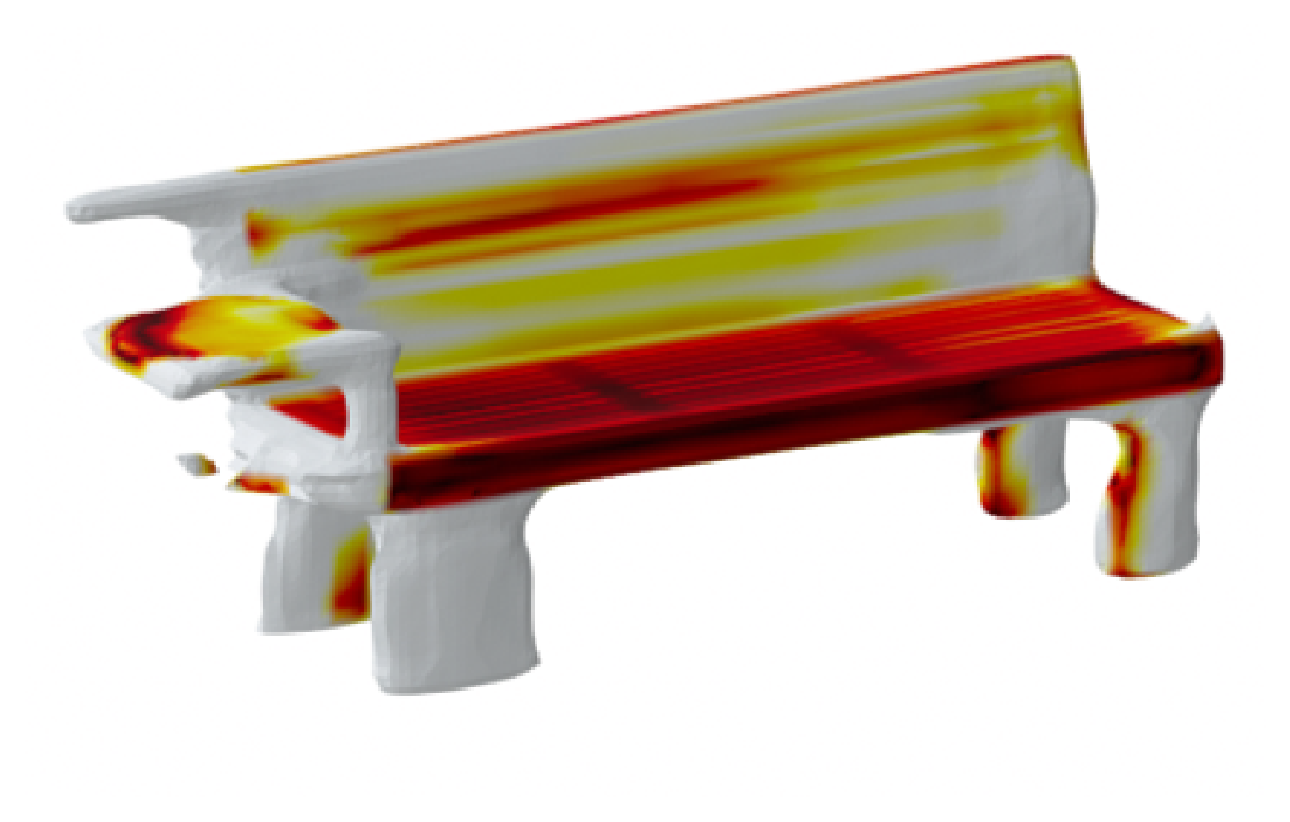}};
				\node[] (c) at (15/8*2,3) {\includegraphics[width=0.09\textwidth]{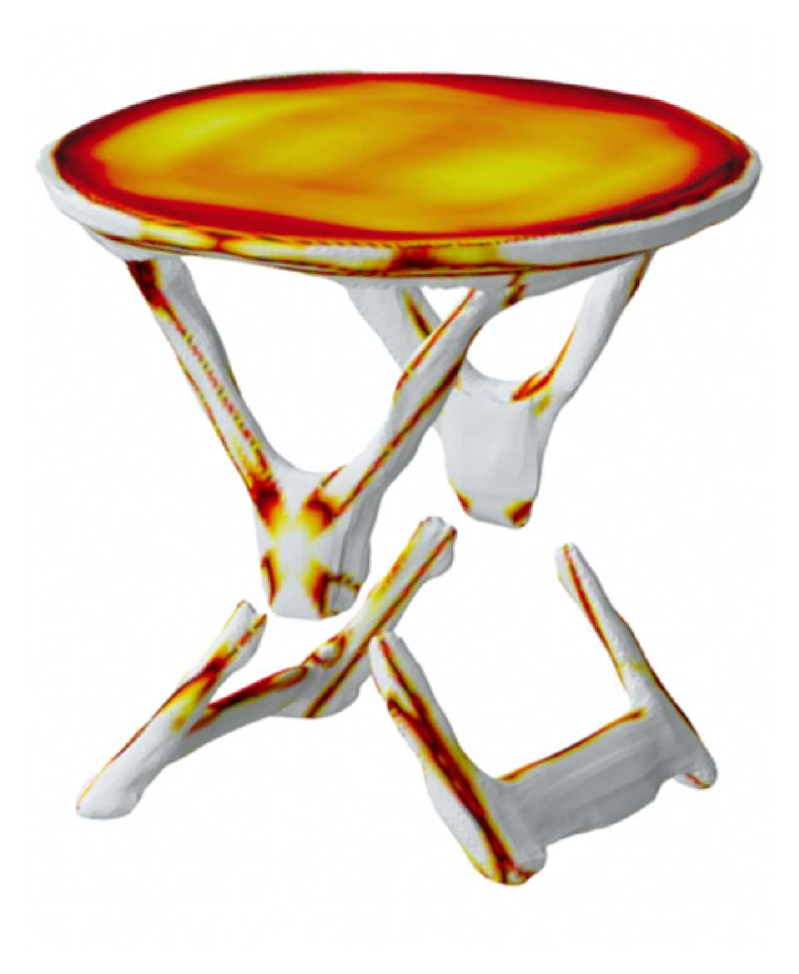} };
				\node[] (c) at (15/8*2,4.7) {\includegraphics[width=0.09\textwidth]{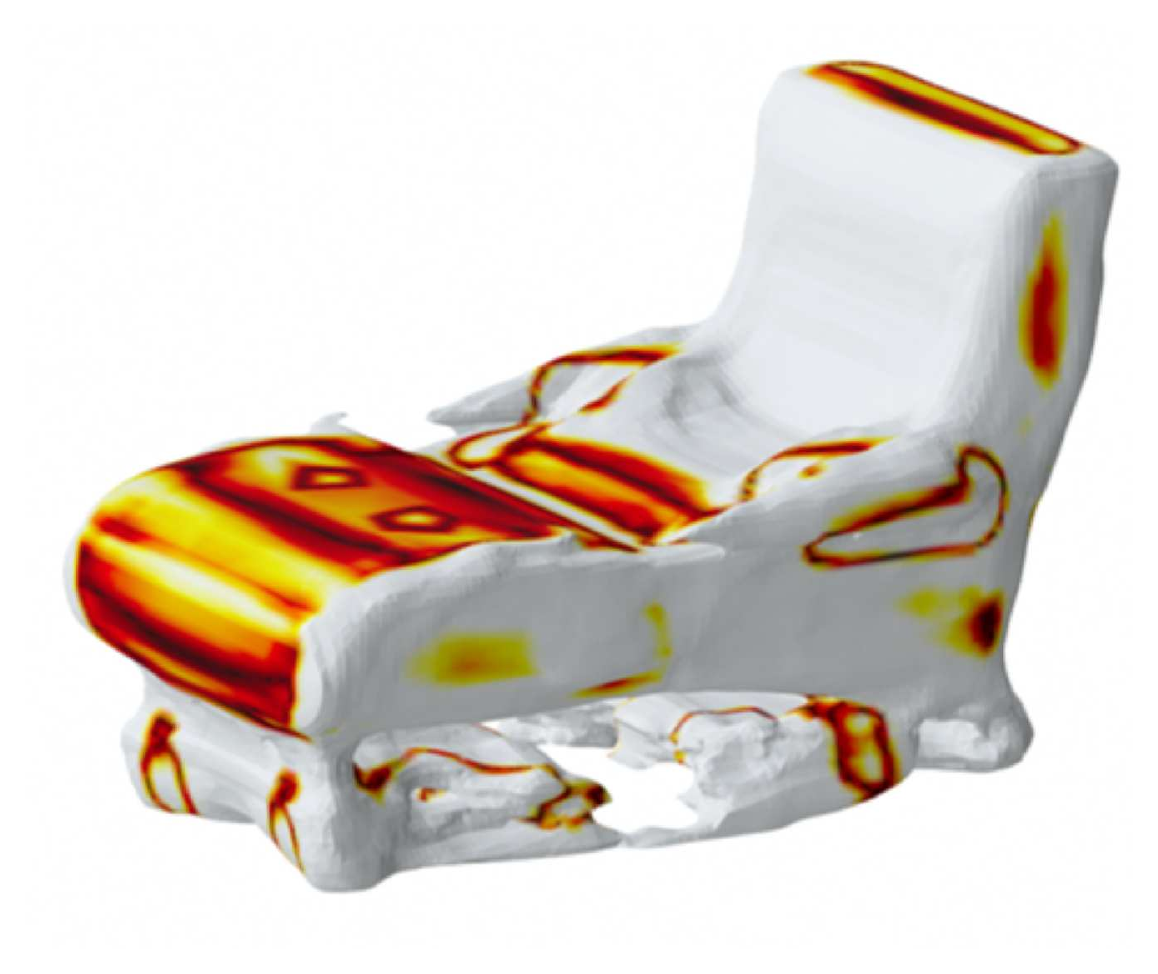} };
				\node[] (c) at (15/8*2,6.4) {\includegraphics[width=0.09\textwidth]{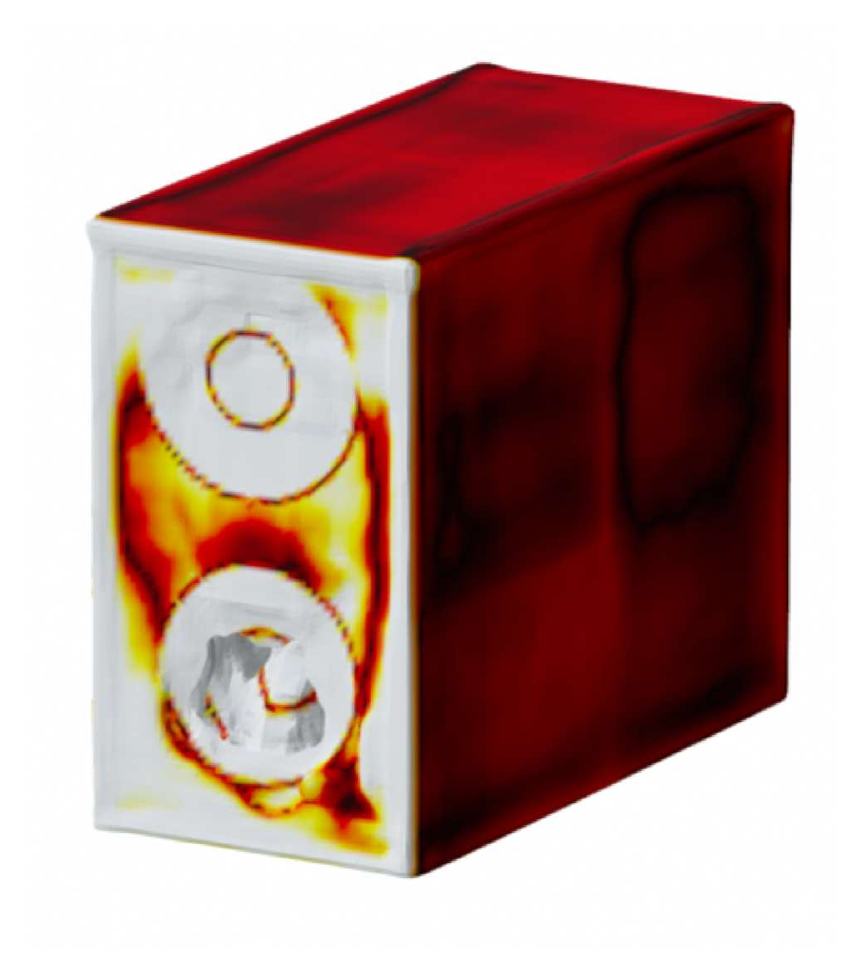} };
				\node[] (c) at (15/8*2,0) {\small (c) ONet \cite{OCCNET} };
				
				\node[] (d) at (15/8*3,0.6) {\includegraphics[width=0.09\textwidth]{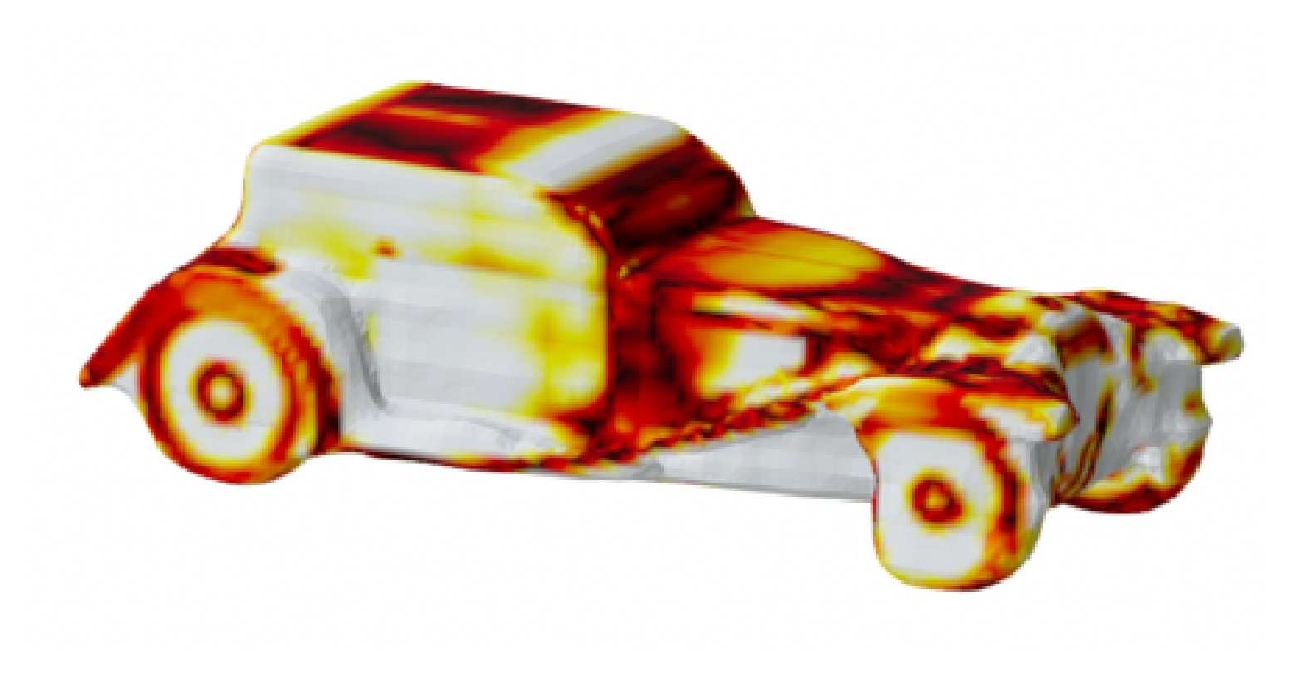}};
				\node[] (d) at (15/8*3,1.5) {\includegraphics[width=0.09\textwidth]{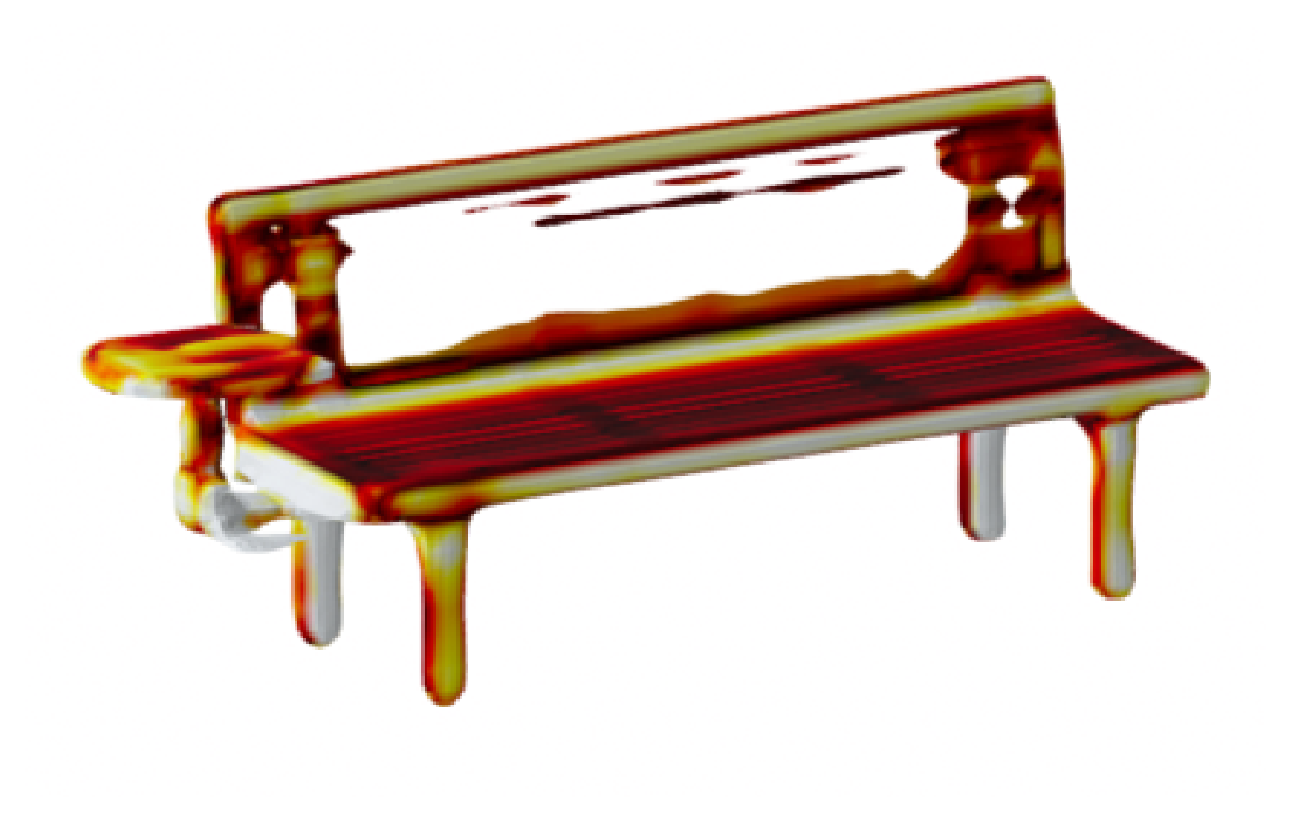}};
				\node[] (d) at (15/8*3,3) {\includegraphics[width=0.09\textwidth]{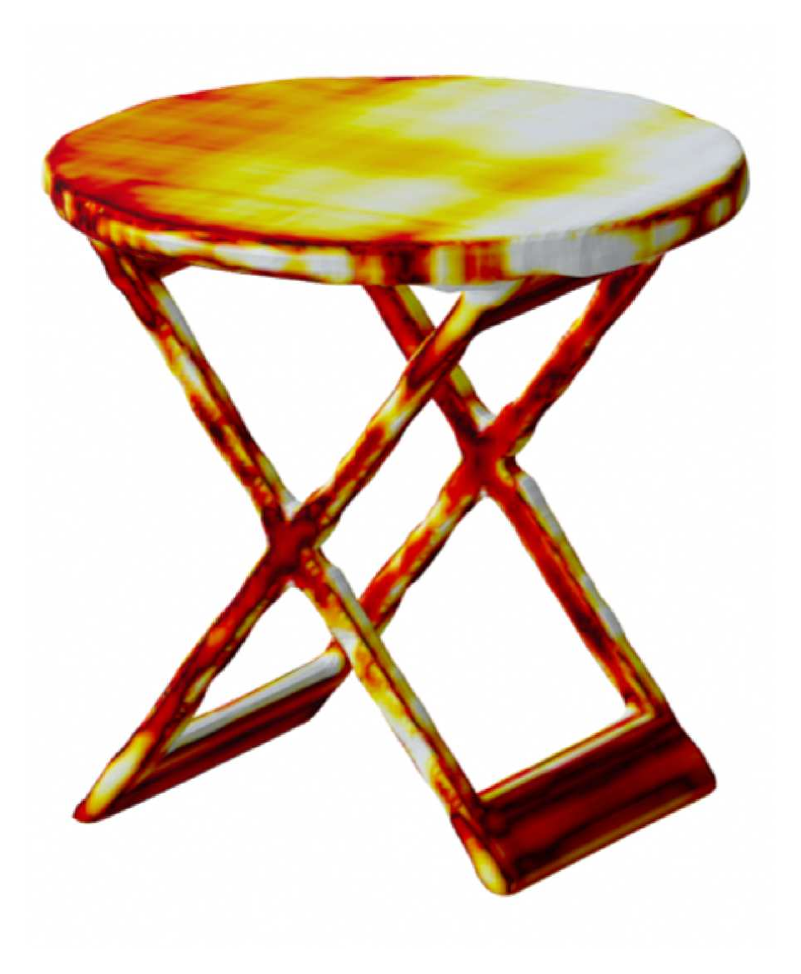} };
				\node[] (d) at (15/8*3,4.7) {\includegraphics[width=0.09\textwidth]{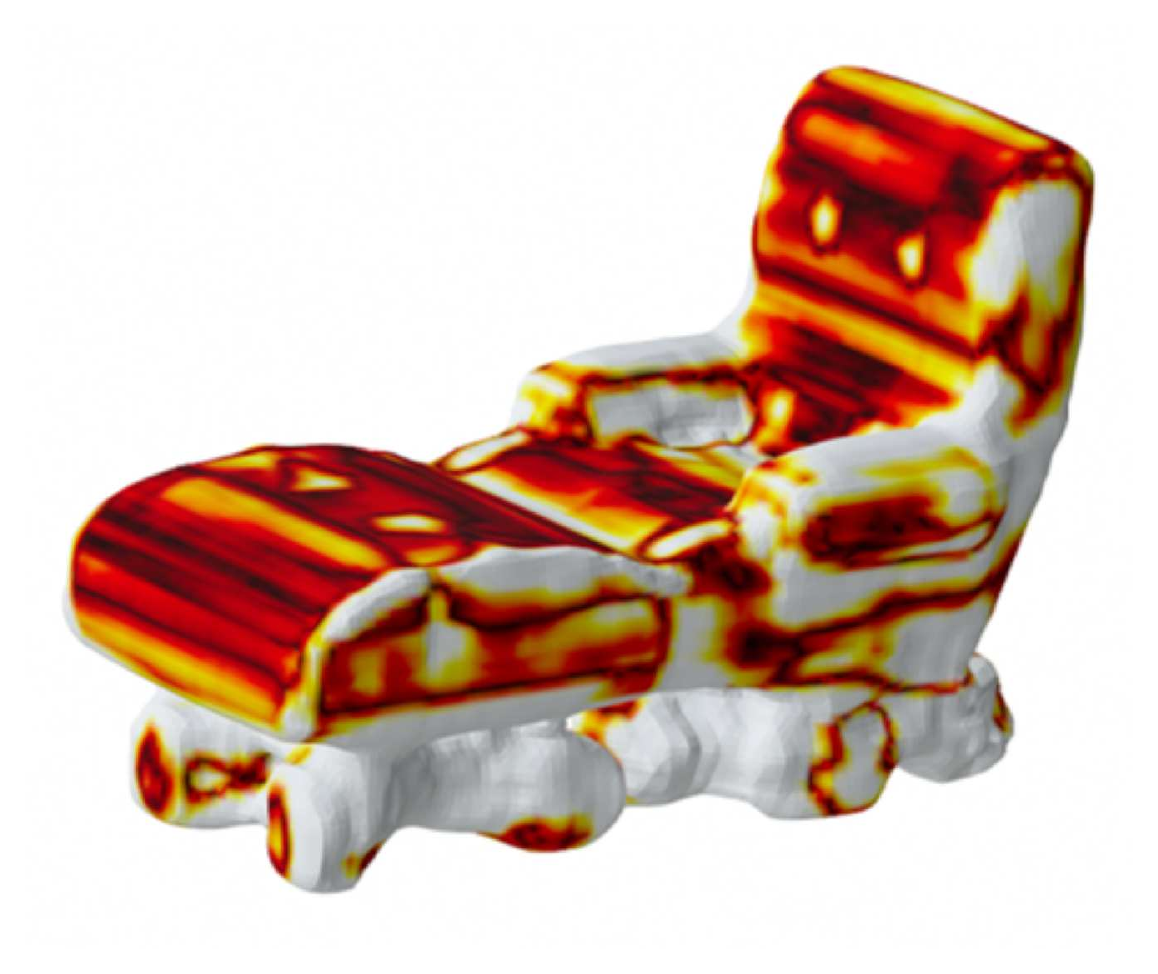} };
				\node[] (d) at (15/8*3,6.4) {\includegraphics[width=0.09\textwidth]{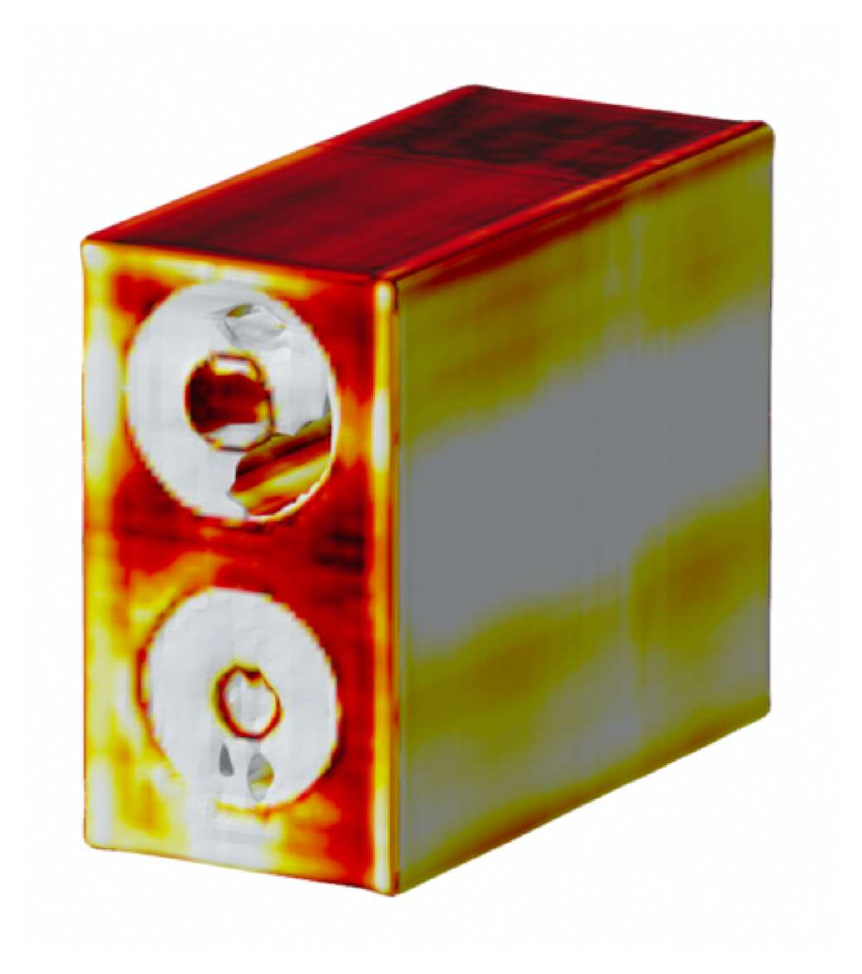} };
				\node[] (d) at (15/8*3,0) {\small (d) CONet \cite{CONVOCCNET} };
				
				\node[] (e) at (15/8*4,0.6) {\includegraphics[width=0.09\textwidth]{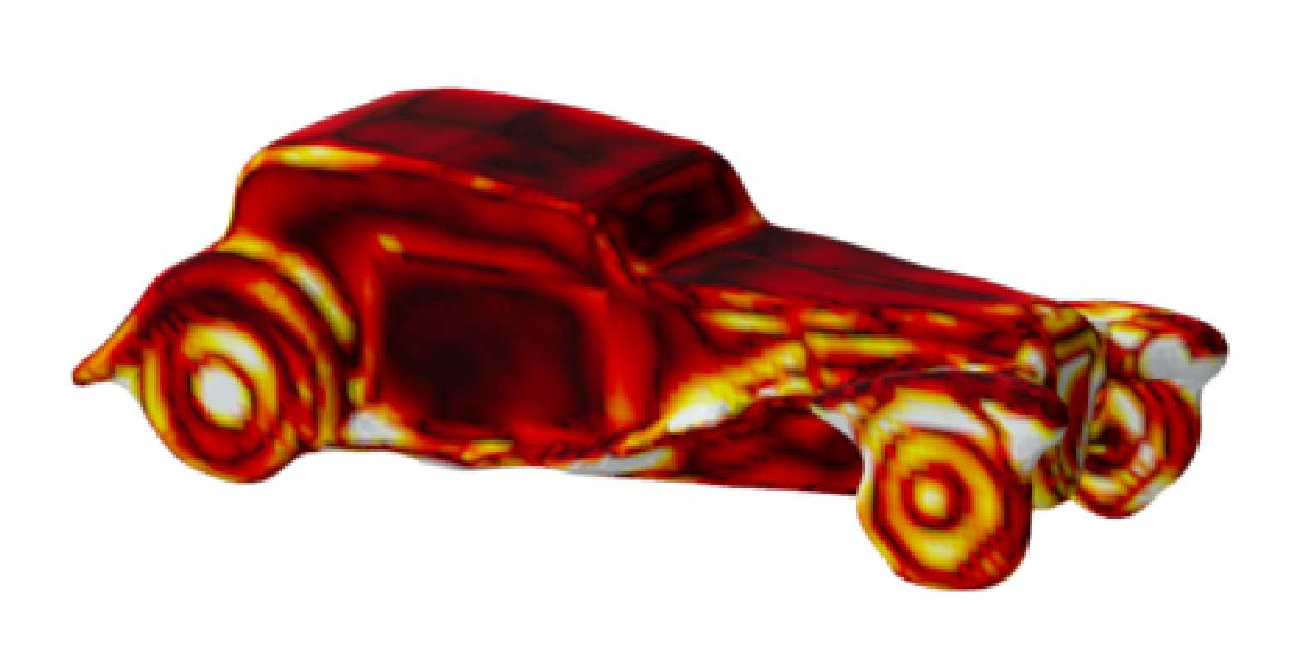}};
				\node[] (e) at (15/8*4,1.5) {\includegraphics[width=0.09\textwidth]{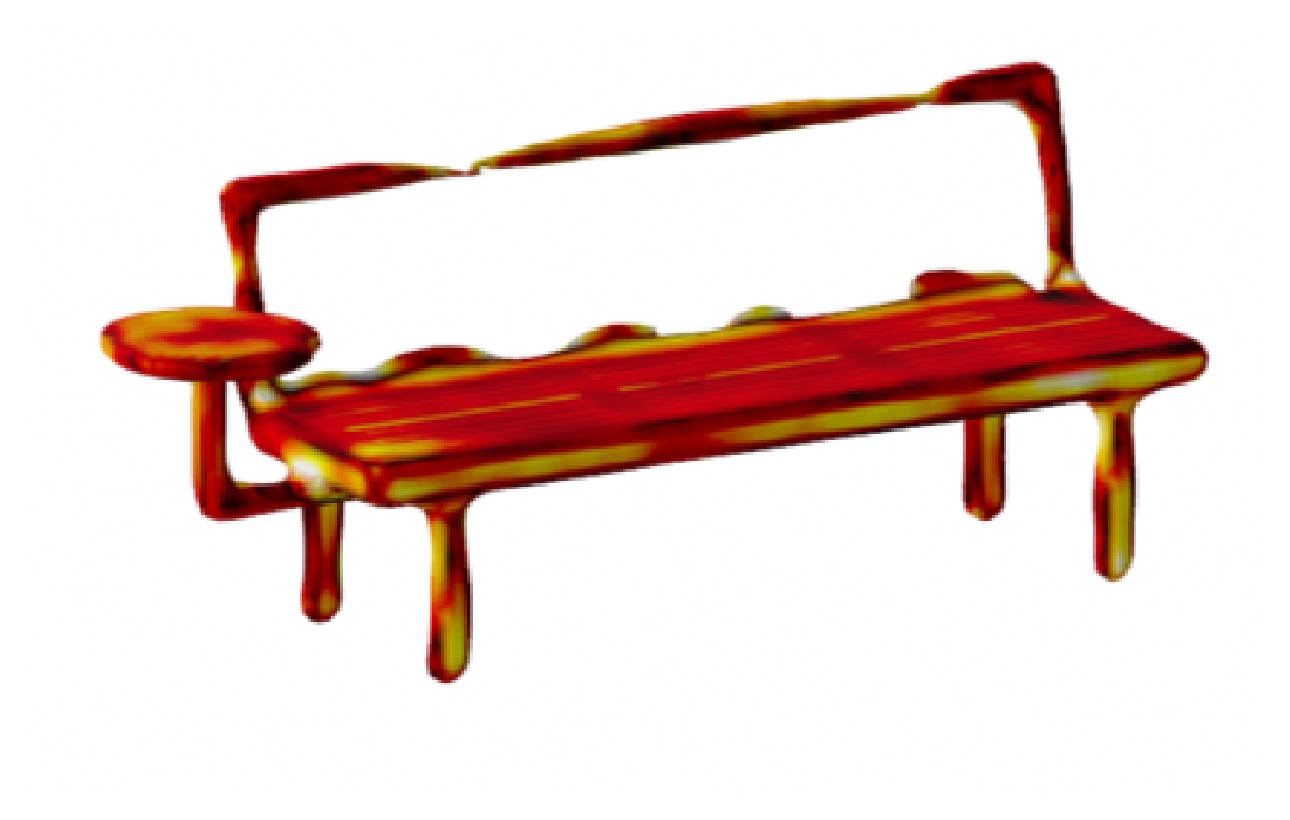}};
				\node[] (e) at (15/8*4,3) {\includegraphics[width=0.09\textwidth]{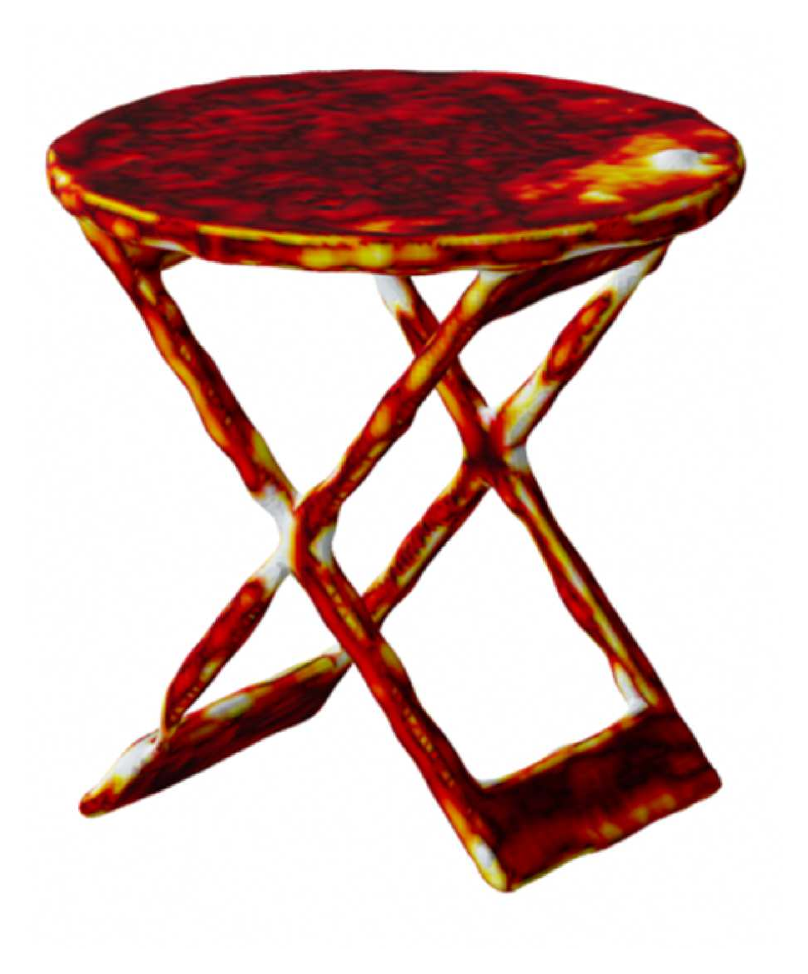}};
				\node[] (e) at (15/8*4,4.7) {\includegraphics[width=0.09\textwidth]{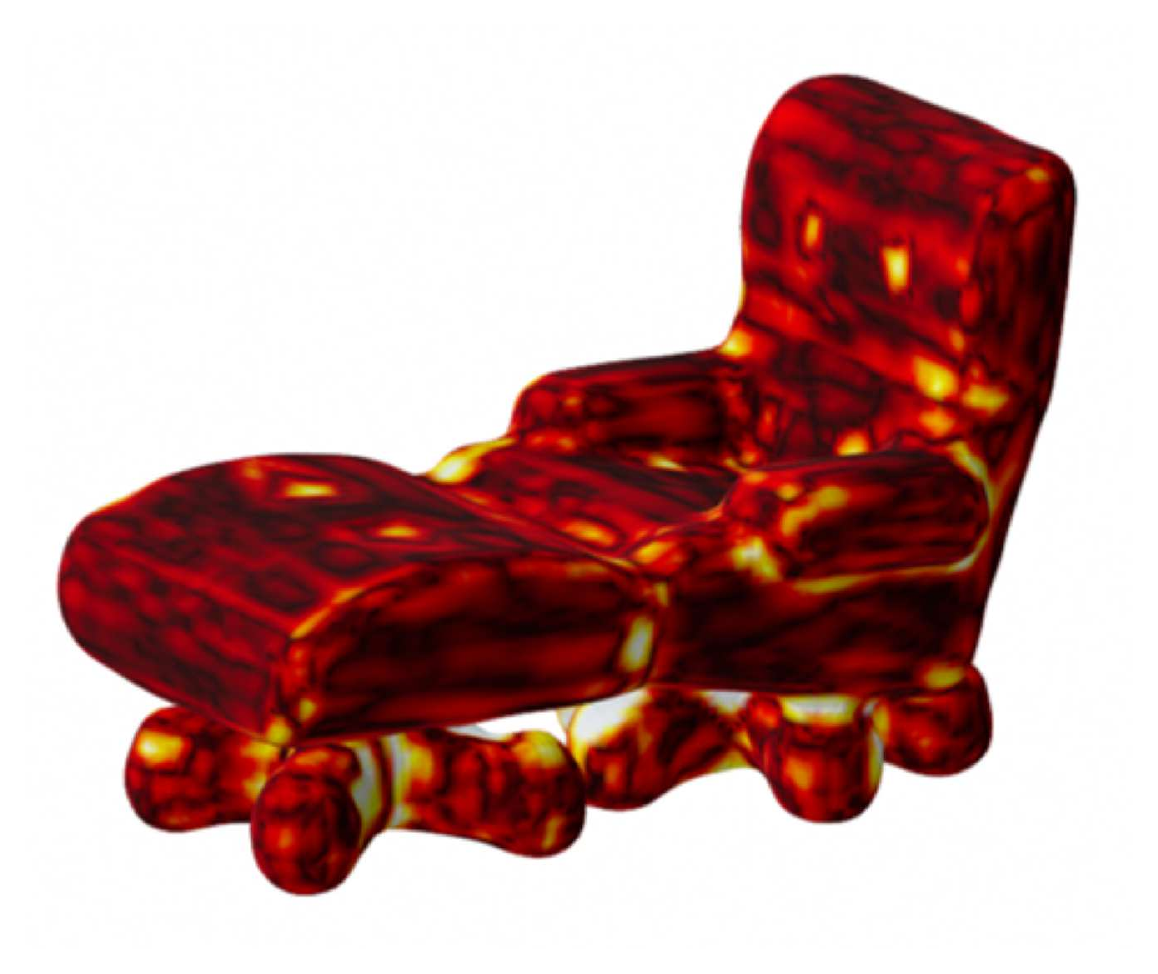}};
				\node[] (e) at (15/8*4,6.4) {\includegraphics[width=0.09\textwidth]{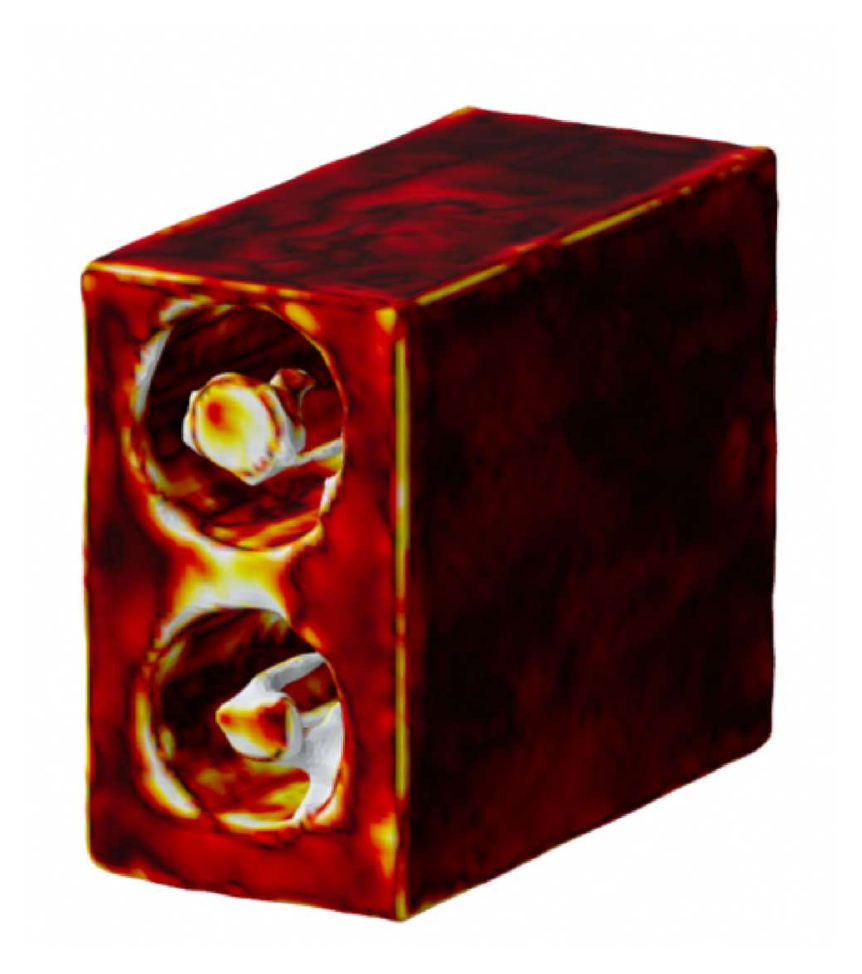}};
				\node[] (e) at (15/8*4,0) {\small (e) SAP \cite{SAP} };
				
				\node[] (f) at (15/8*5,0.6) {\includegraphics[width=0.09\textwidth]{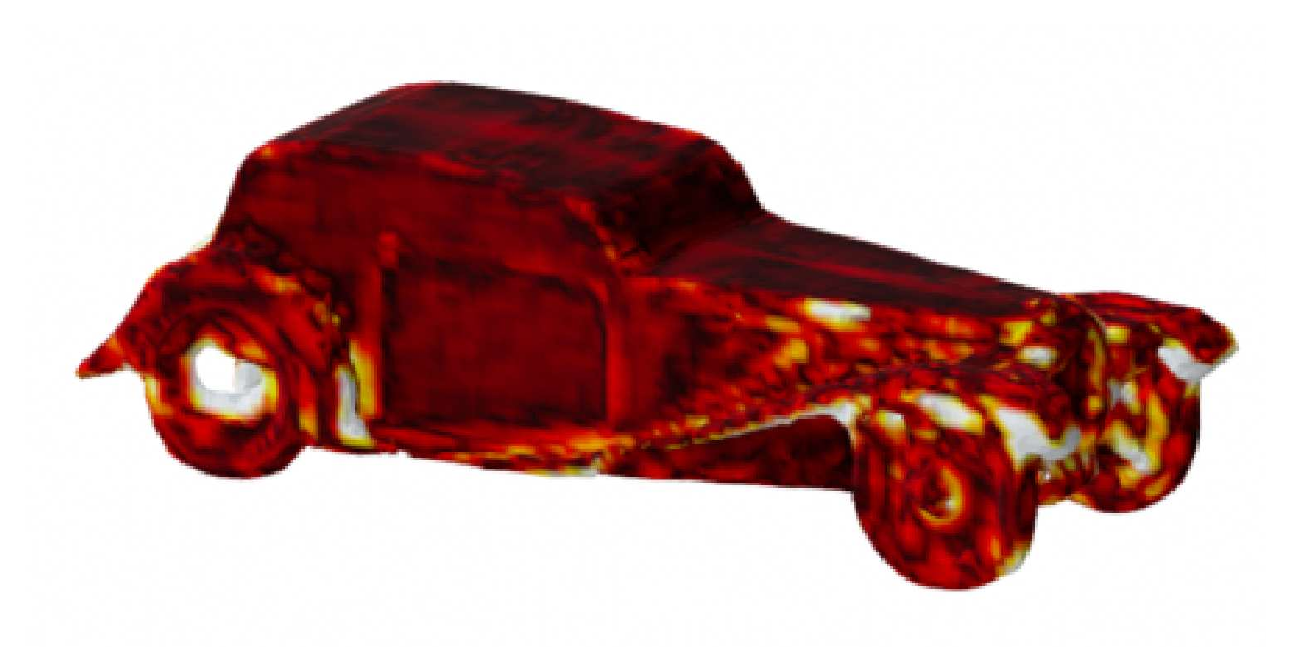}};
				\node[] (f) at (15/8*5,1.5) {\includegraphics[width=0.09\textwidth]{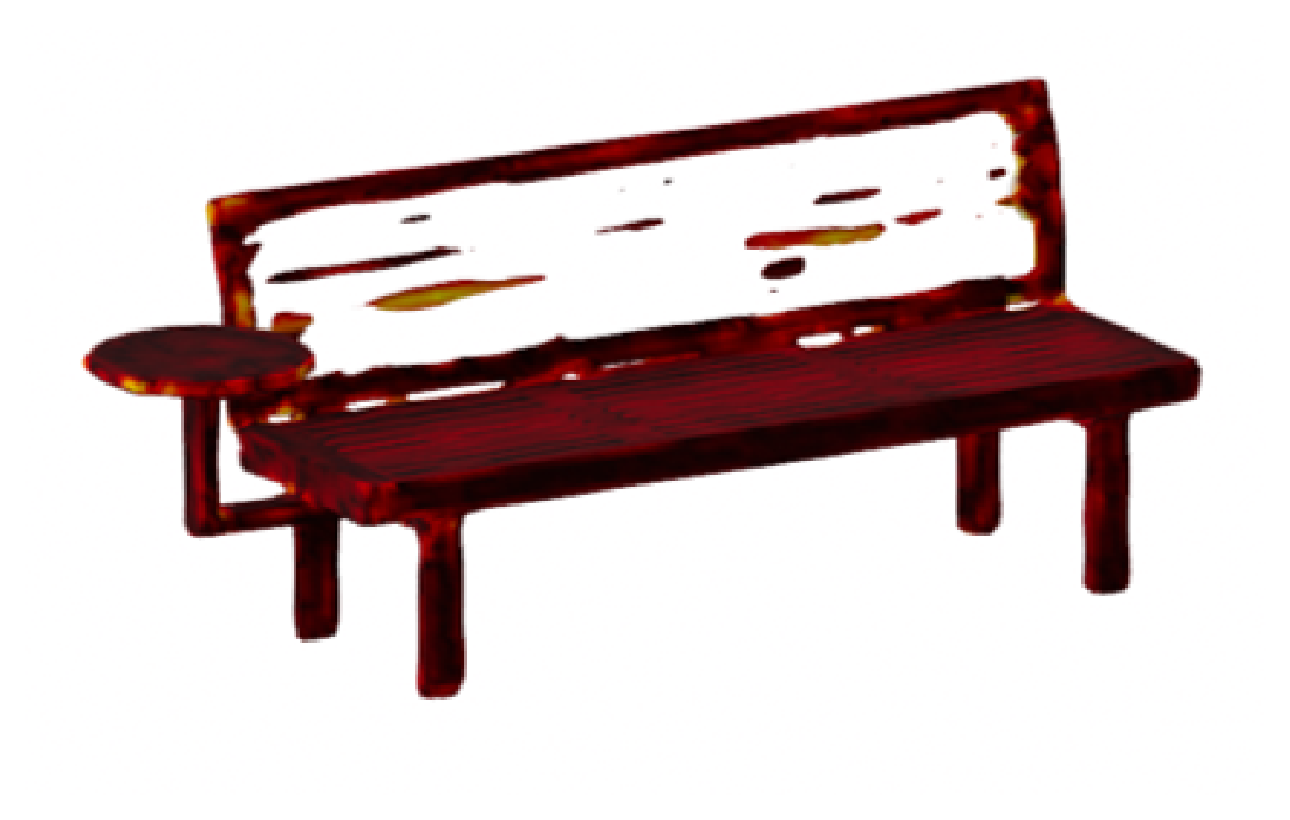}};
				\node[] (f) at (15/8*5,3) {\includegraphics[width=0.09\textwidth]{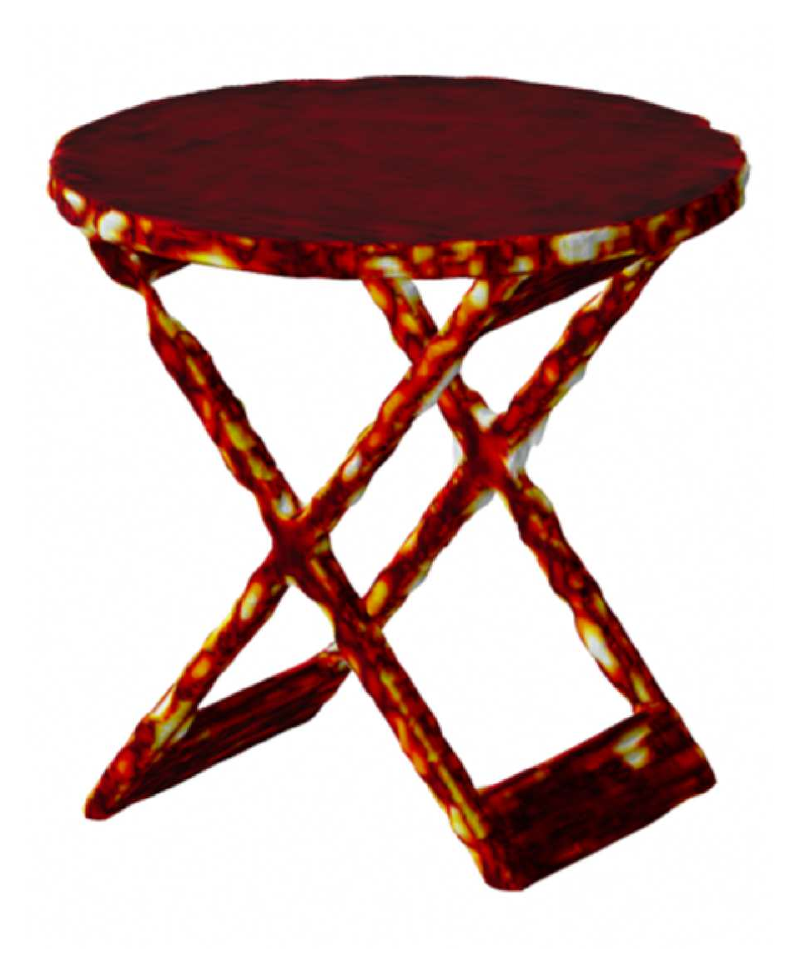}};
				\node[] (f) at (15/8*5,4.7) {\includegraphics[width=0.09\textwidth]{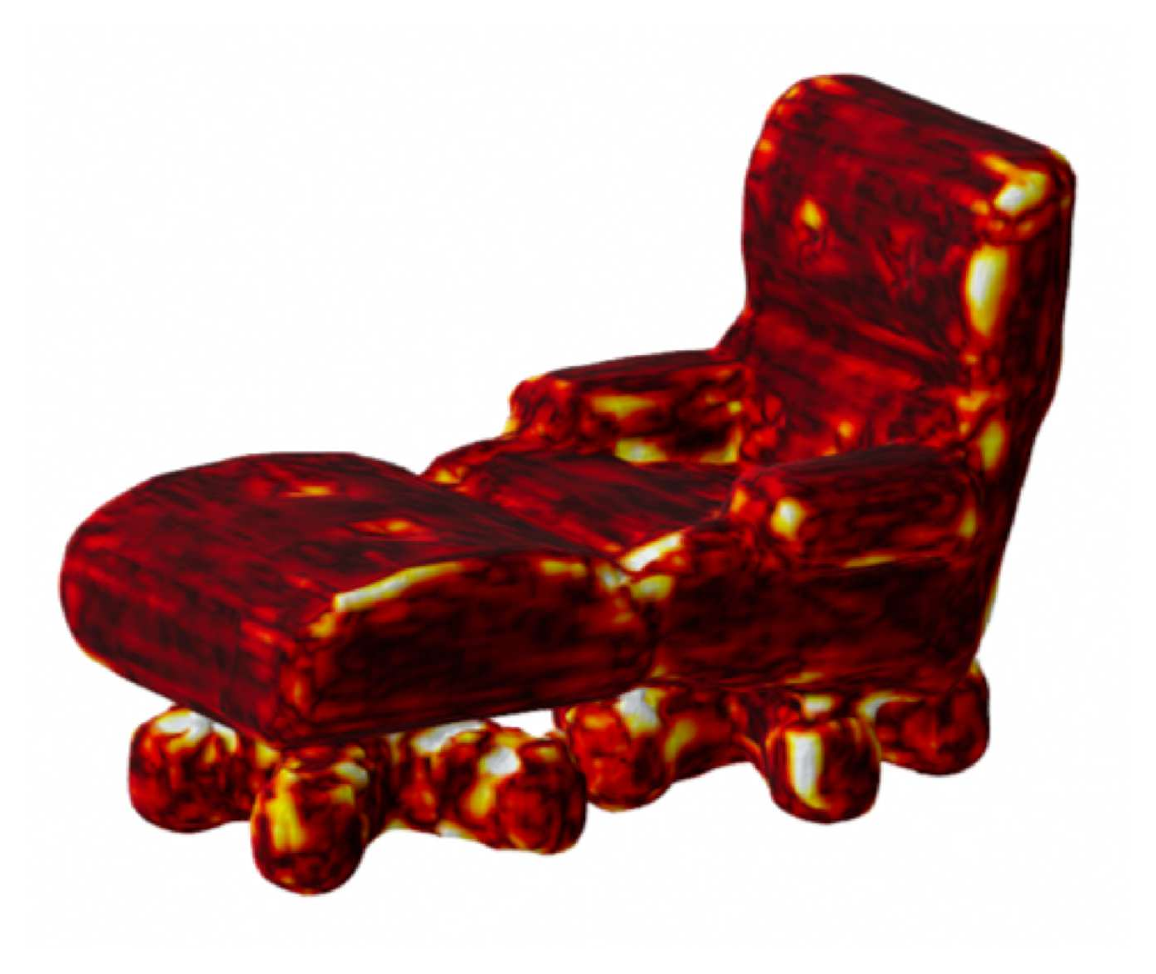}};
				\node[] (f) at (15/8*5,6.4) {\includegraphics[width=0.09\textwidth]{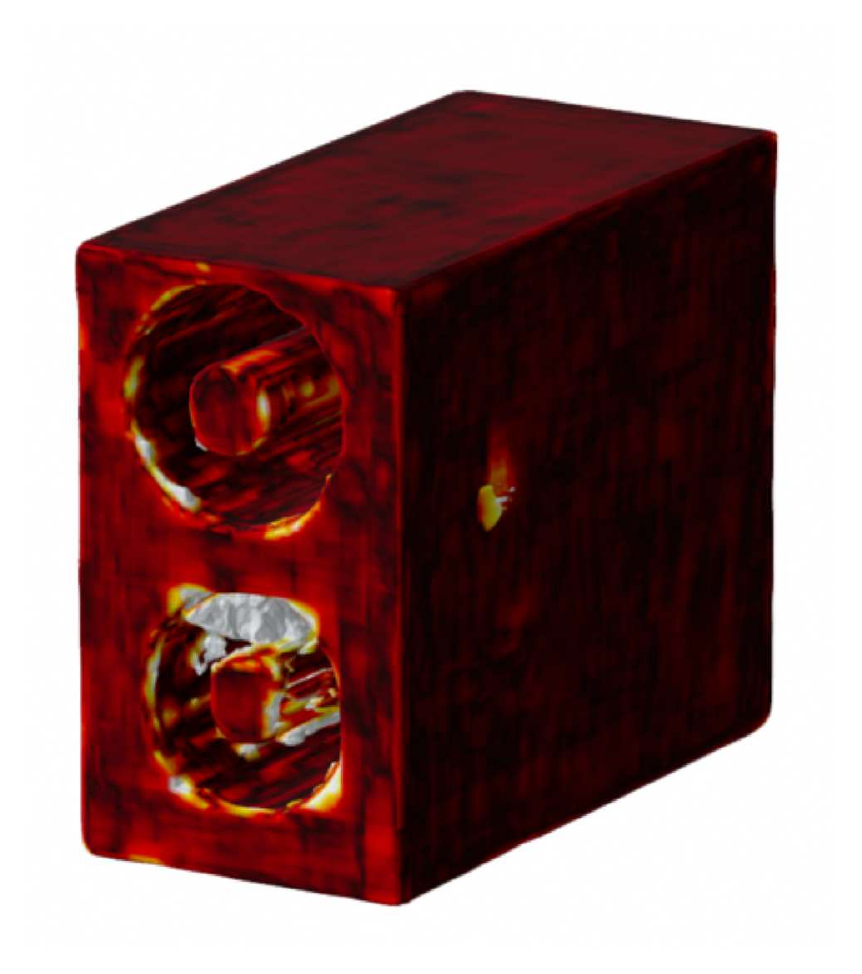}};
				\node[] (f) at (15/8*5,0) {\small (f) POCO \cite{POCO} };
				
				\node[] (g) at (15/8*6,0.6) {\includegraphics[width=0.09\textwidth]{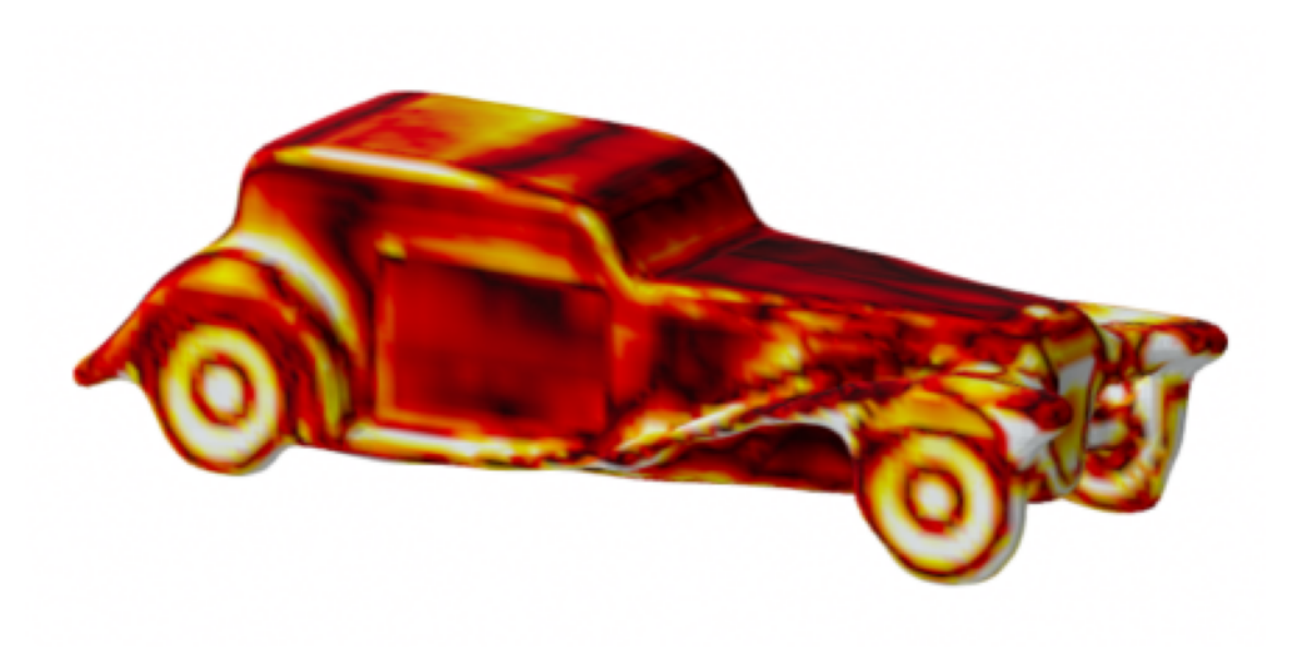}};
				\node[] (g) at (15/8*6,1.5) {\includegraphics[width=0.09\textwidth]{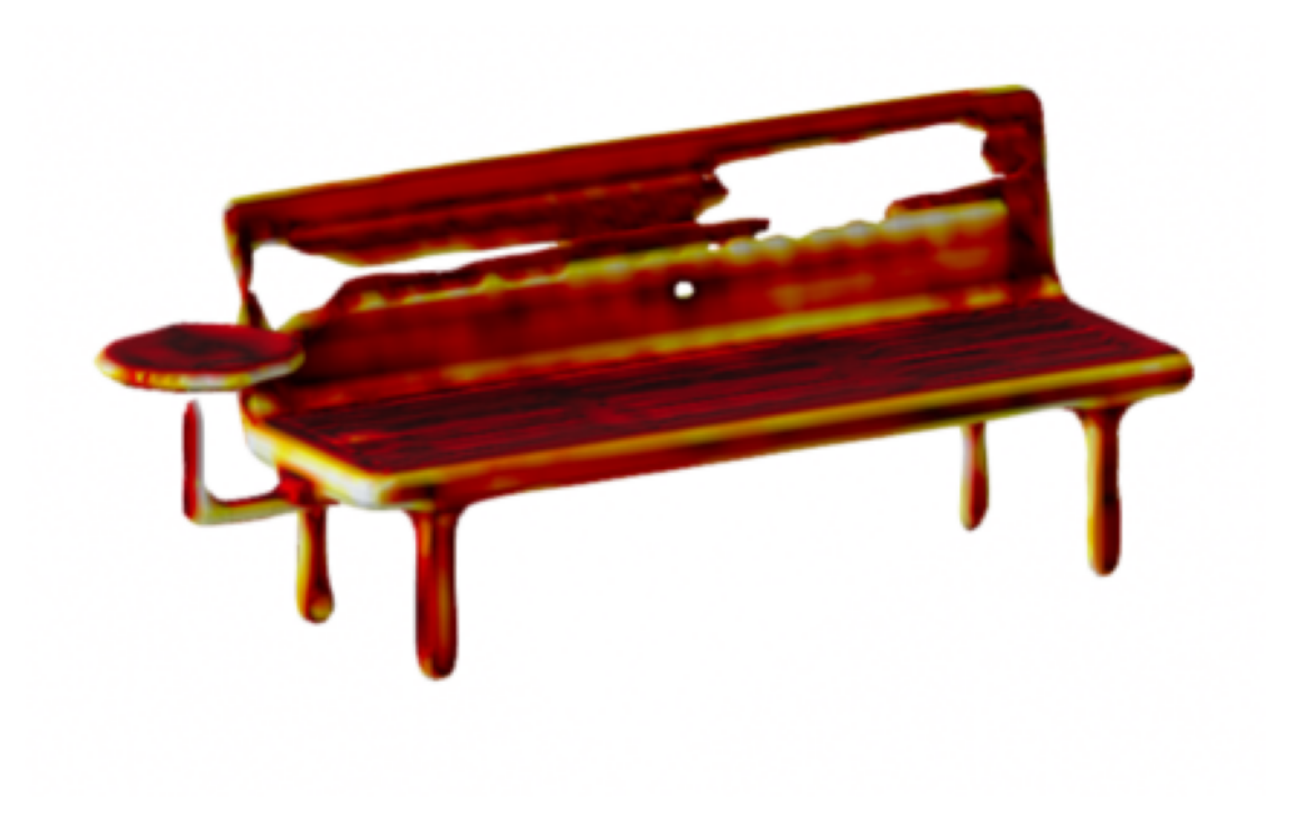}};
				\node[] (g) at (15/8*6,3) {\includegraphics[width=0.09\textwidth]{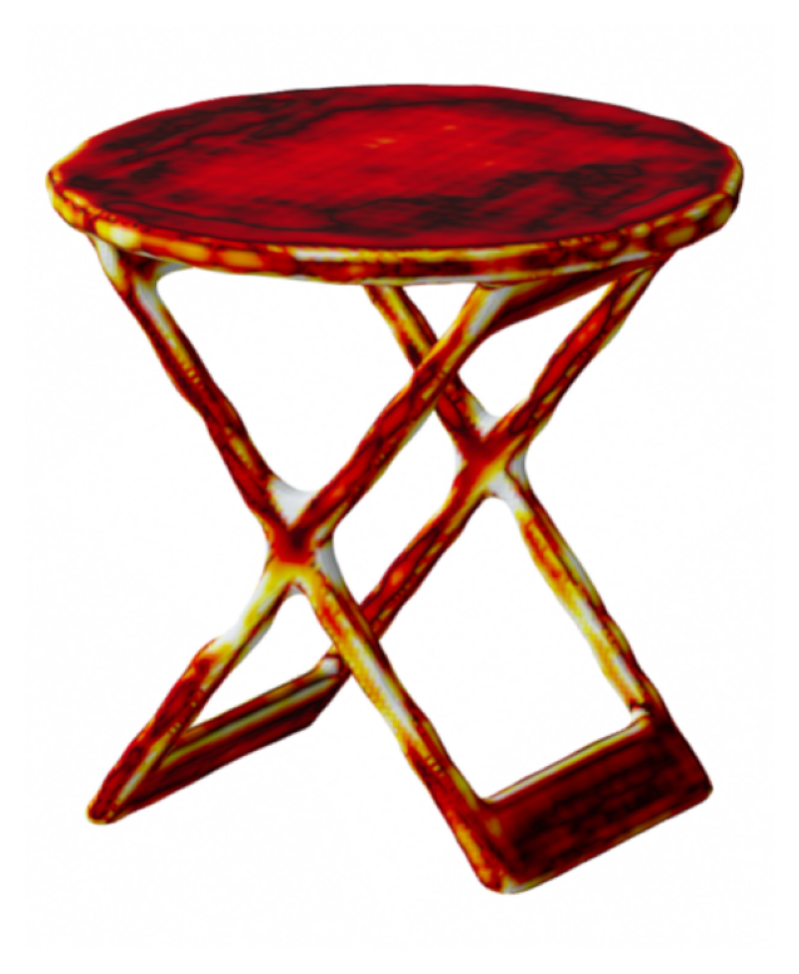}};
				\node[] (g) at (15/8*6,4.7) {\includegraphics[width=0.09\textwidth]{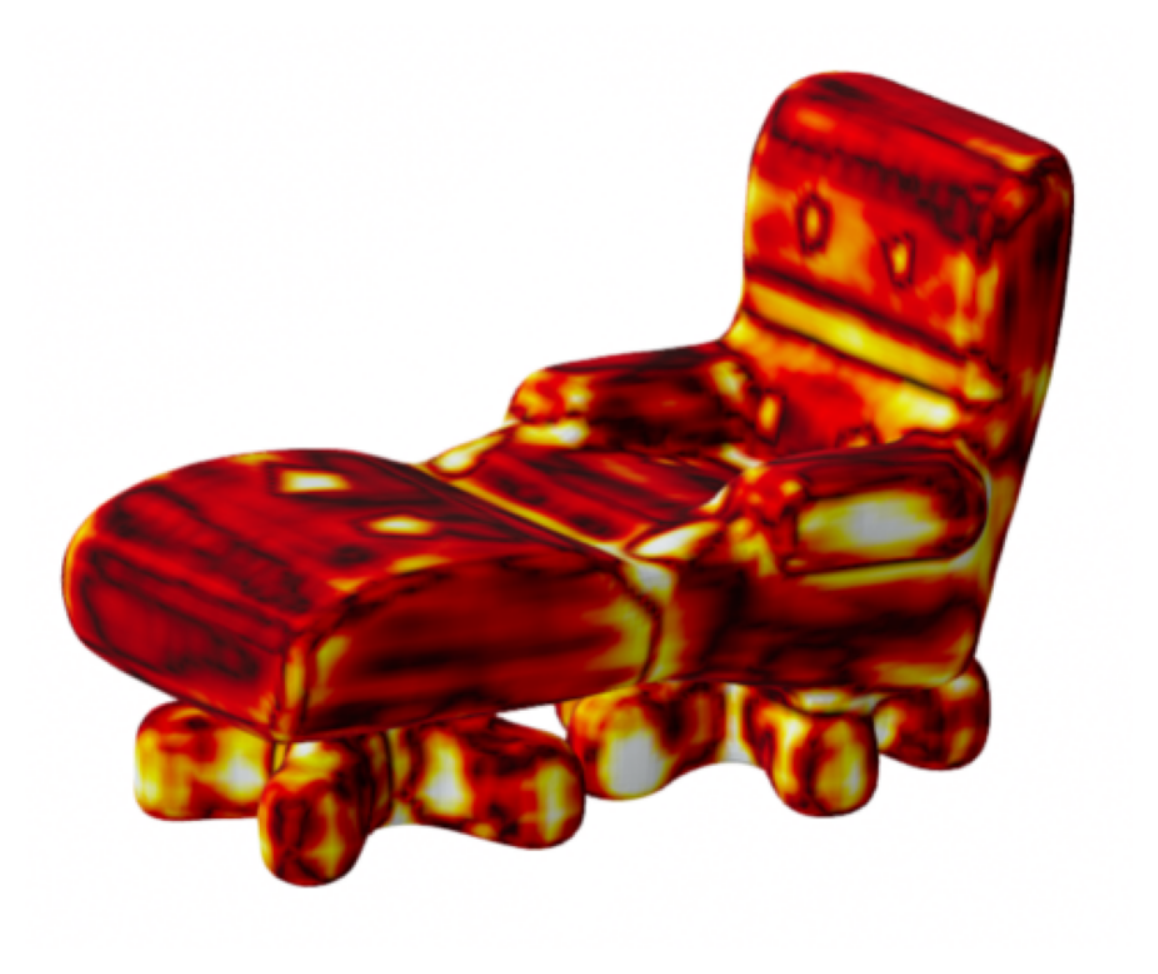}};
				\node[] (g) at (15/8*6,6.4) {\includegraphics[width=0.09\textwidth]{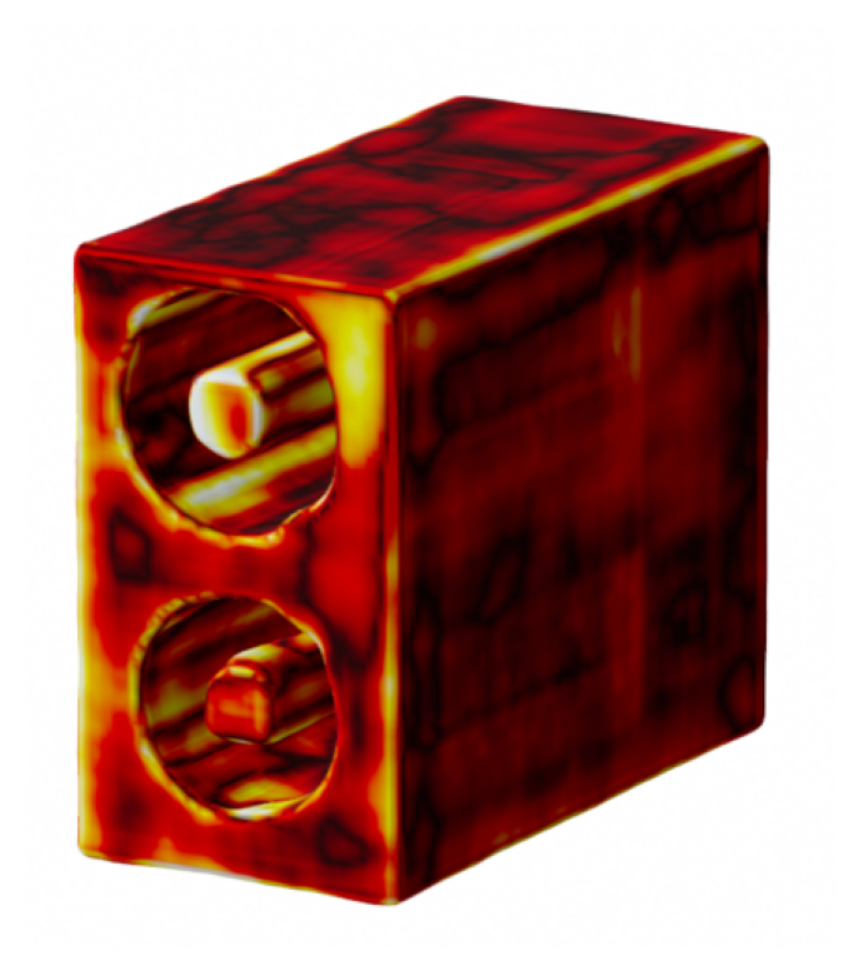}};
				\node[] (g) at (15/8*6,0) {\small (g) {DOG \cite{DOG}} };
				
				\node[] (h) at (15/8*7,0.6) {\includegraphics[width=0.09\textwidth]{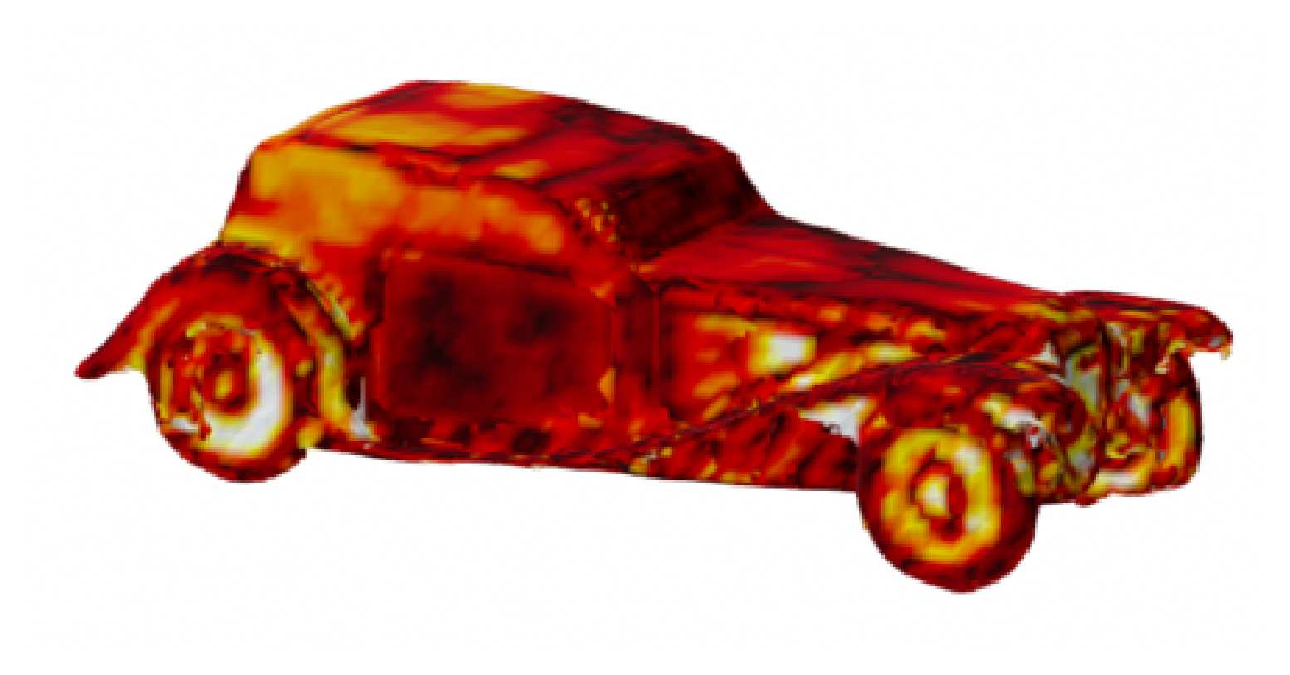}};
				\node[] (h) at (15/8*7,1.5) {\includegraphics[width=0.09\textwidth]{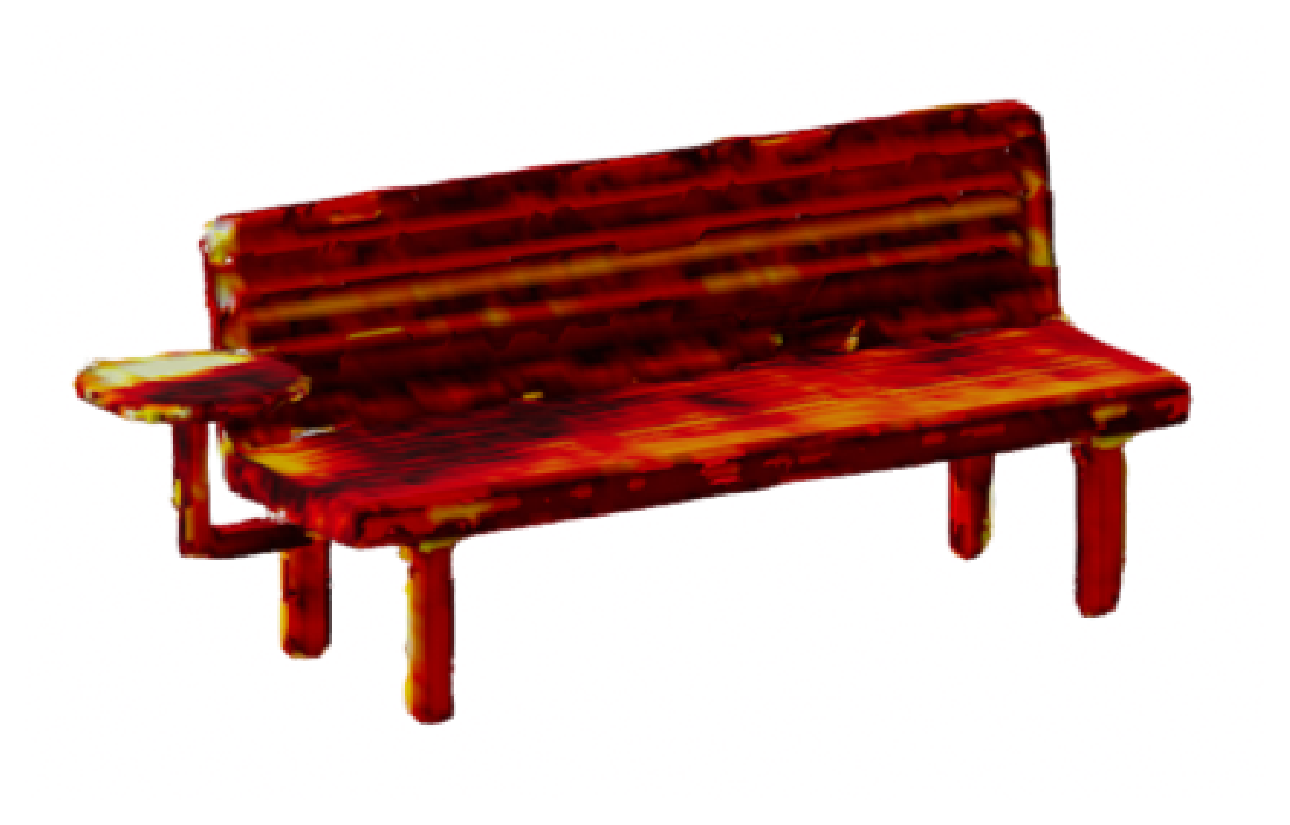}};
				\node[] (h) at (15/8*7,3) {\includegraphics[width=0.09\textwidth]{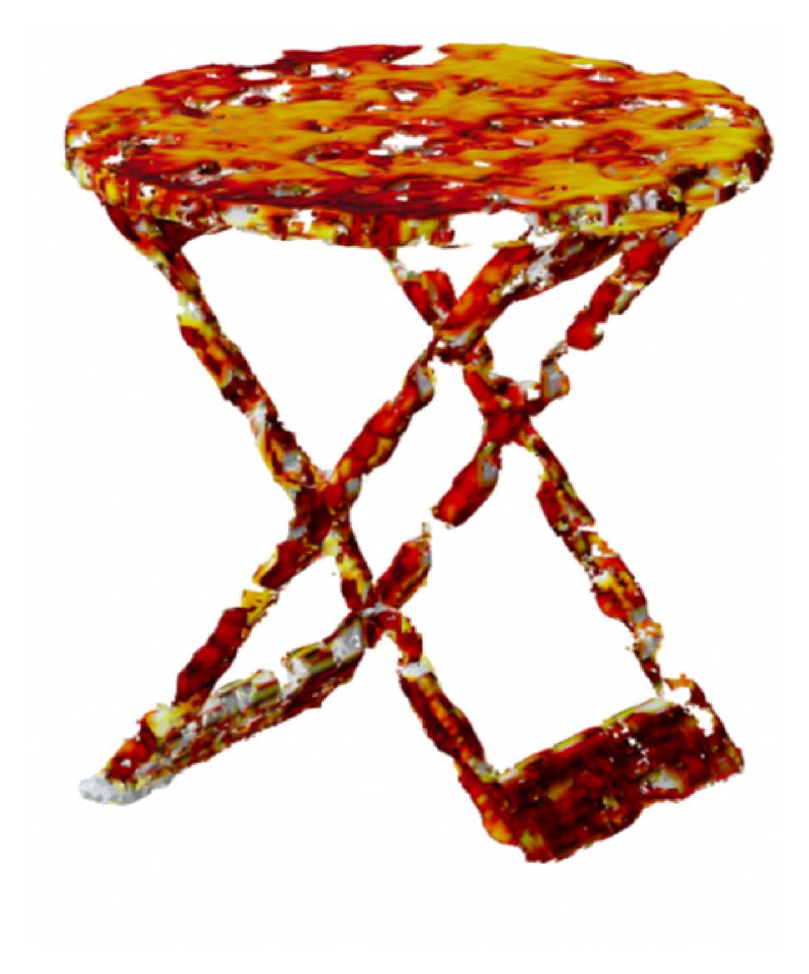}};
				\node[] (h) at (15/8*7,4.7) {\includegraphics[width=0.09\textwidth]{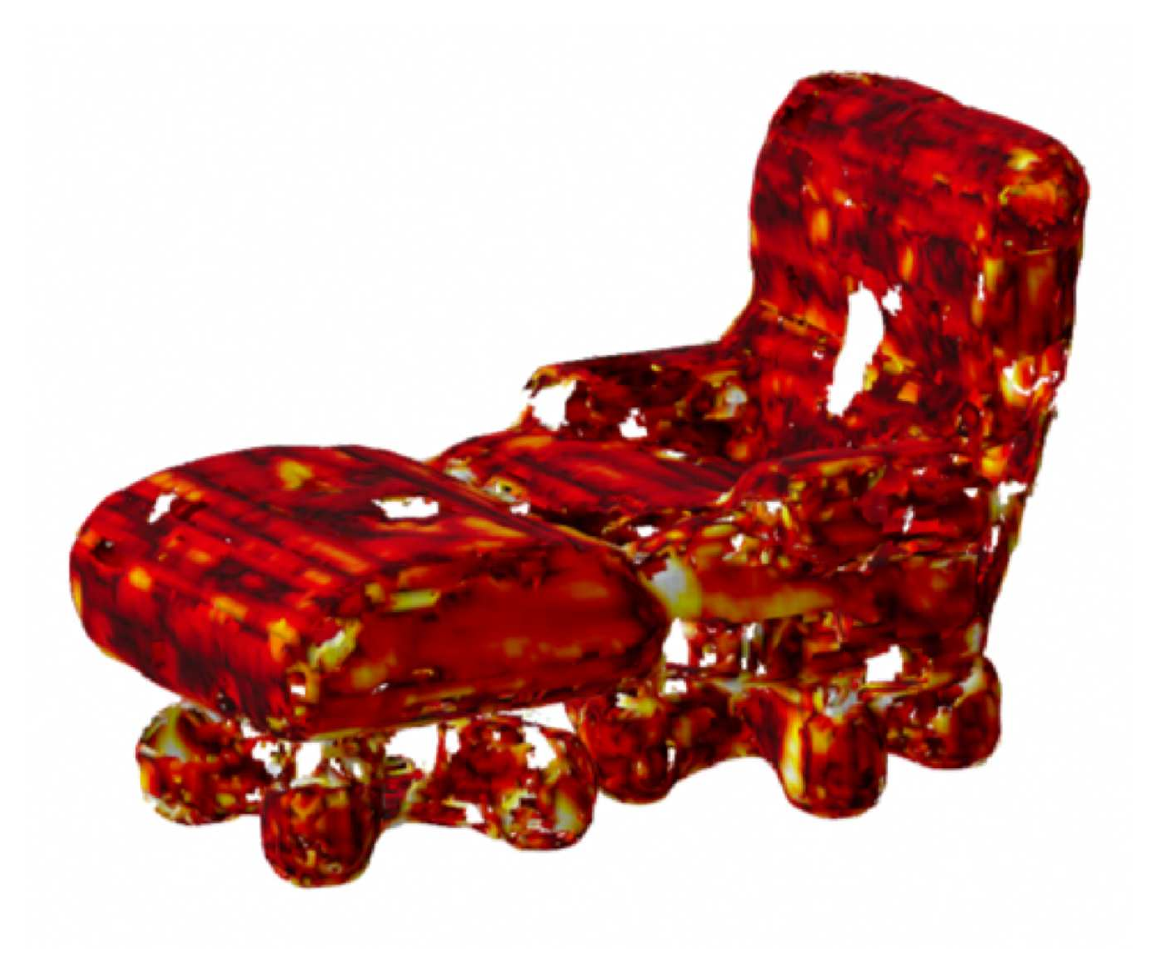}};
				\node[] (h) at (15/8*7,6.4) {\includegraphics[width=0.09\textwidth]{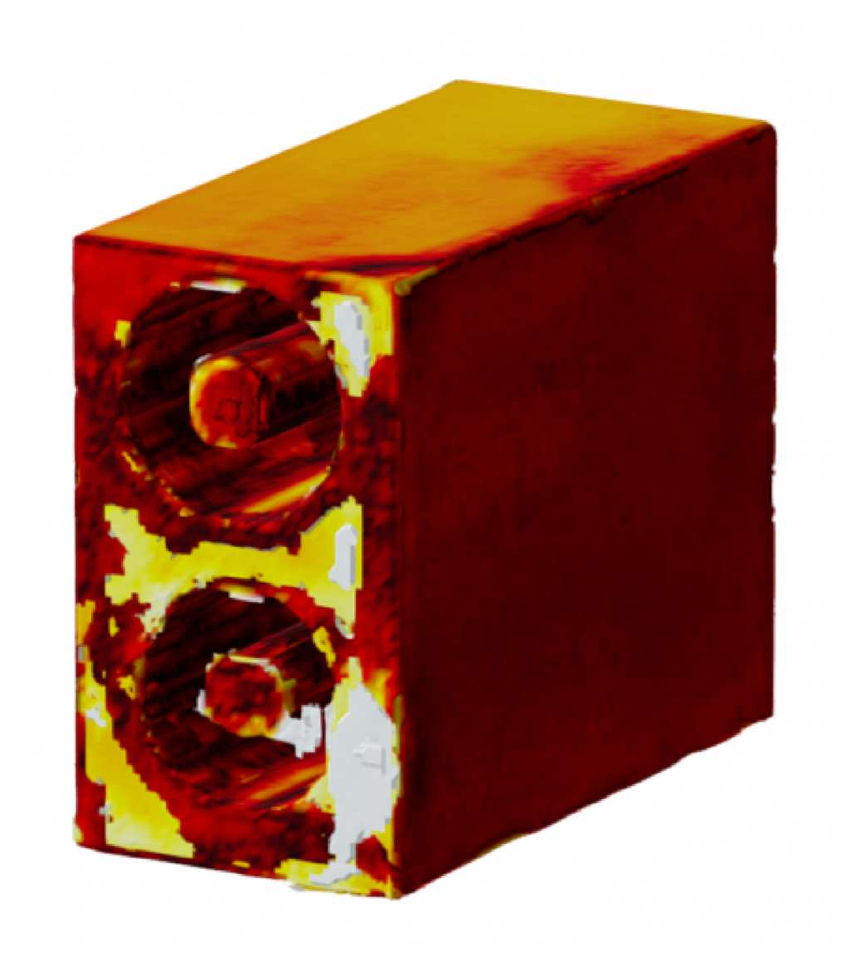}};
				\node[] (h) at (15/8*7,0) {\small (h) GIFS \cite{GIFS} };
				
				\node[] (i) at (15,0.6) {\includegraphics[width=0.09\textwidth]{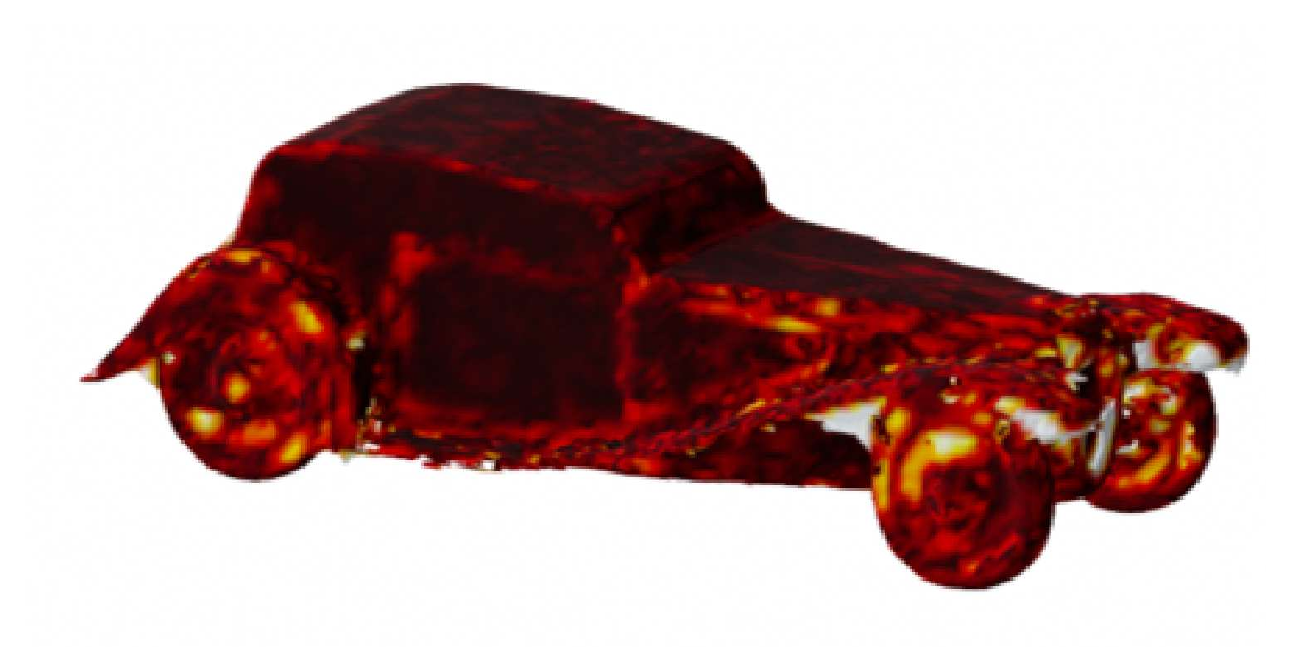}};
				\node[] (i) at (15,1.5) {\includegraphics[width=0.09\textwidth]{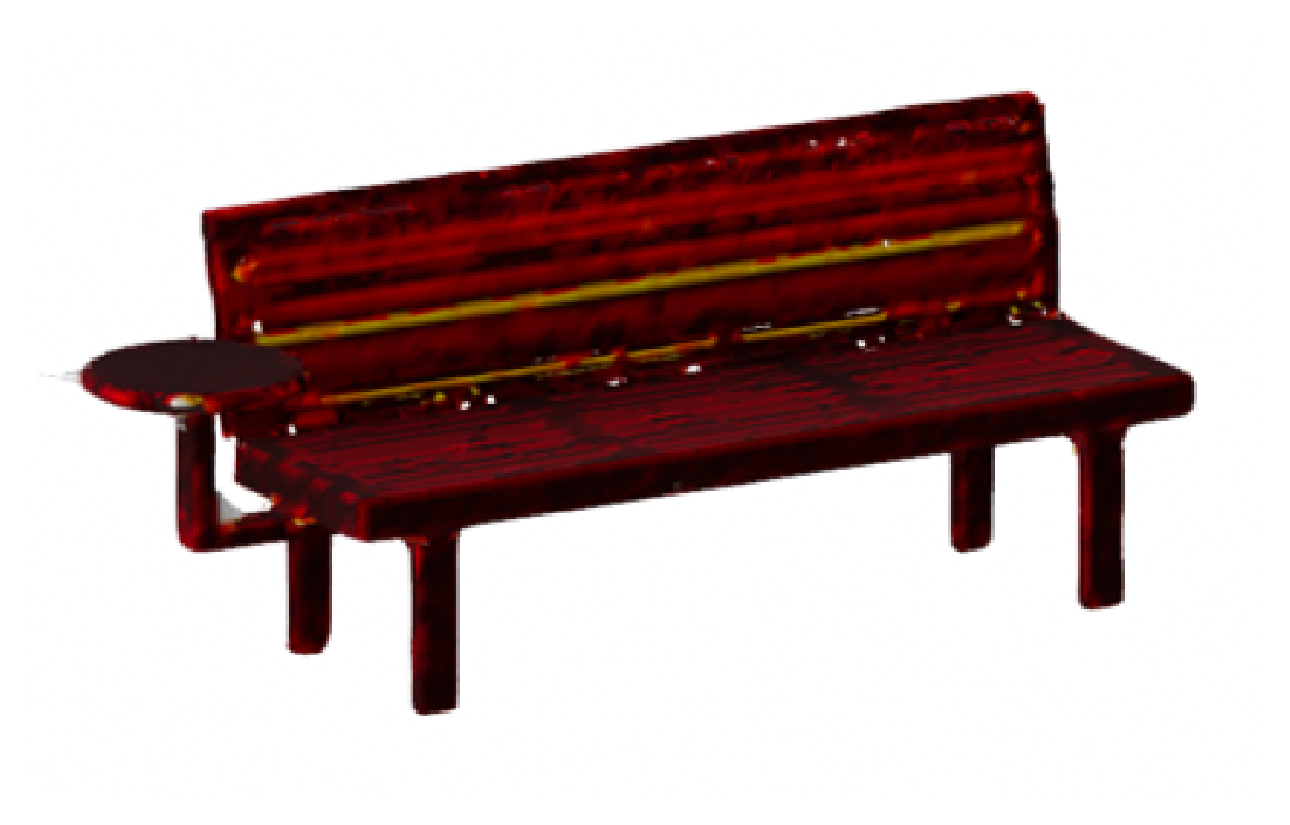}};
				\node[] (i) at (15,3) {\includegraphics[width=0.09\textwidth]{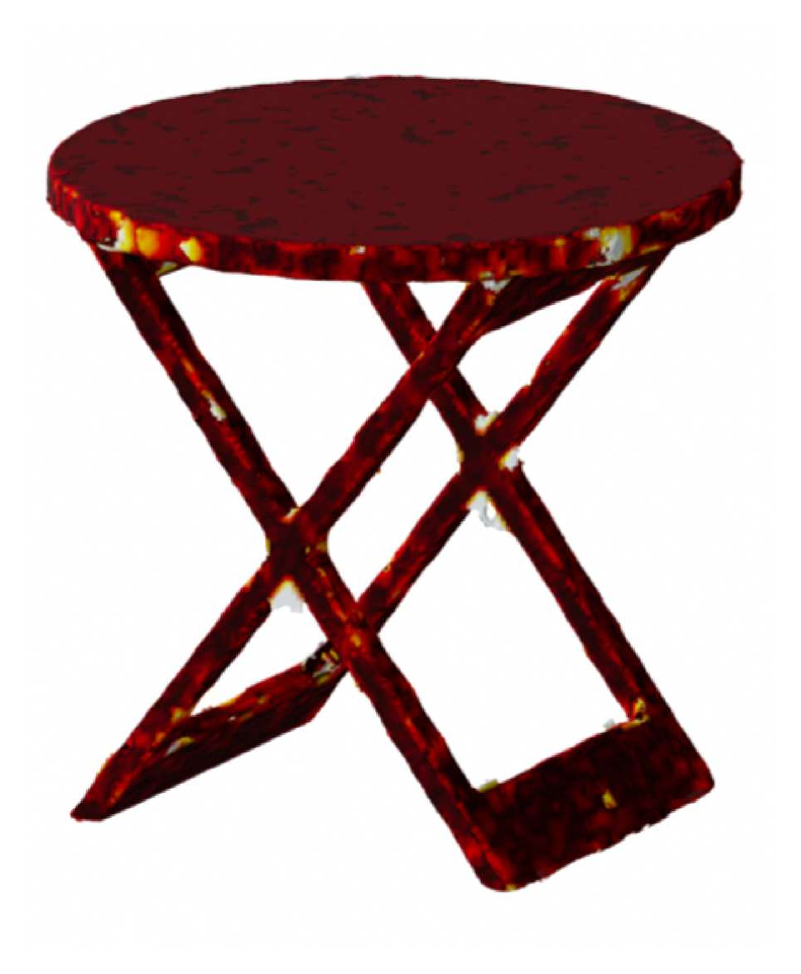}};
				\node[] (i) at (15,4.7) {\includegraphics[width=0.09\textwidth]{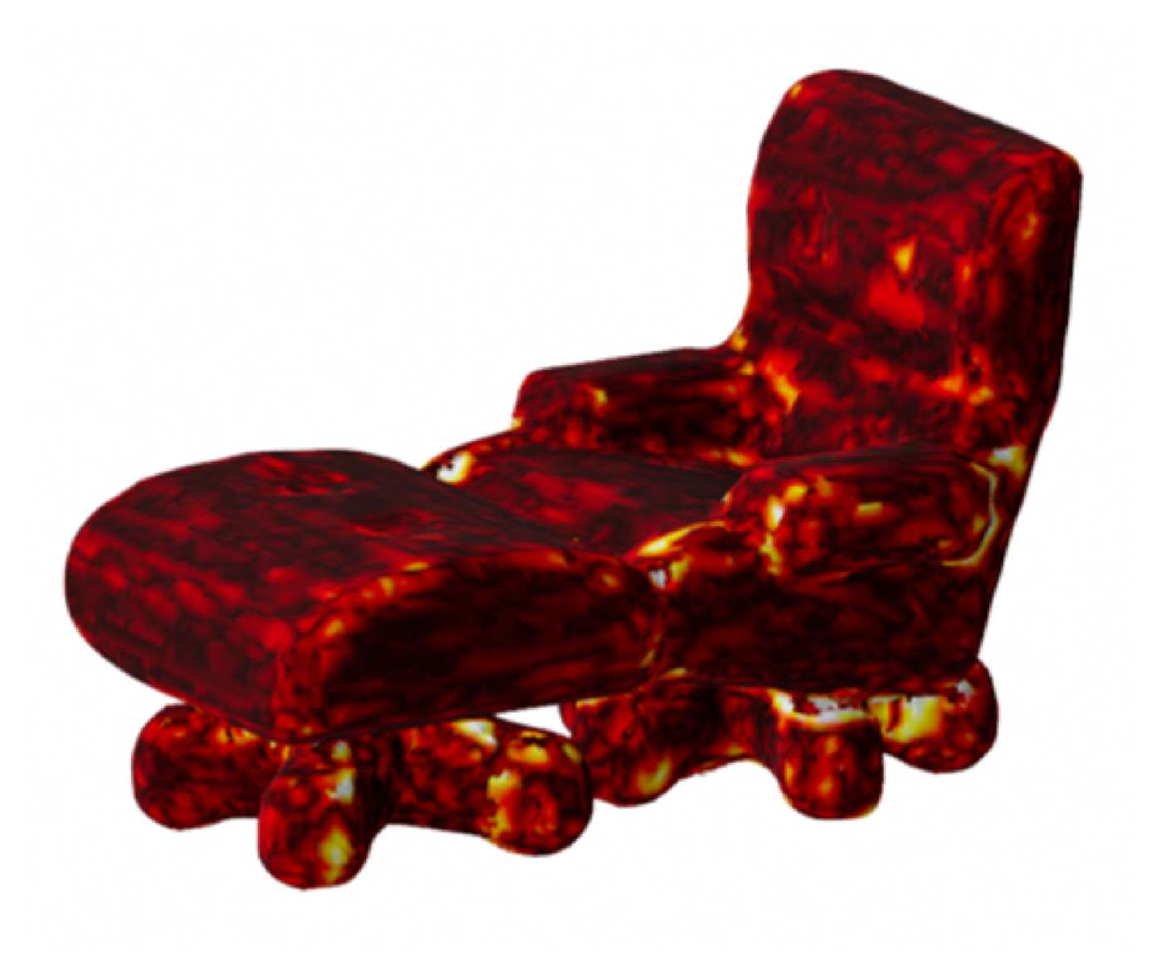}};
				\node[] (i) at (15,6.4) {\includegraphics[width=0.09\textwidth]{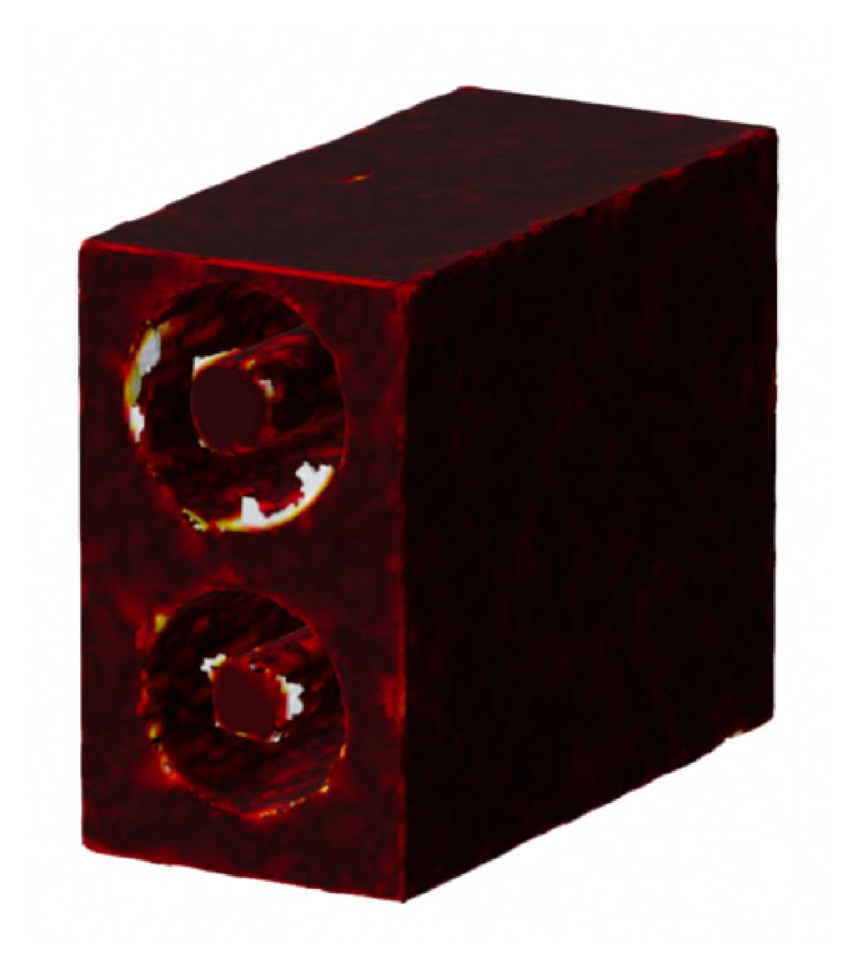}};
				\node[] (i) at (15,0) {\small (i) Ours };
				
				\node[] (h) at (14,7.7) {\includegraphics[width=0.11\textwidth]{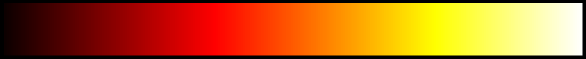}};
				\node[] (h) at (13,7.4) {\small 0};
				\node[] (h) at (15,7.4) {\small 0.008};
				
			\end{tikzpicture}
		}
		\setlength{\abovecaptionskip}{-0.35cm} \vspace{0.4cm}
		\caption{Error Maps of the shapes in the ShapeNet dataset \cite{SHAPENET}. }
		\label{ERRORMAP}
	\end{figure}

	\noindent\textbf{Error Map.} For each reconstructed shape in Fig. \textcolor{red}{6} of the manuscript, we calculated the distance of its vertices to the ground-truth mesh, then used \textit{hot} colormap to assign colors for the vertices. The visual results are shown in Fig. \ref{ERRORMAP}.\\

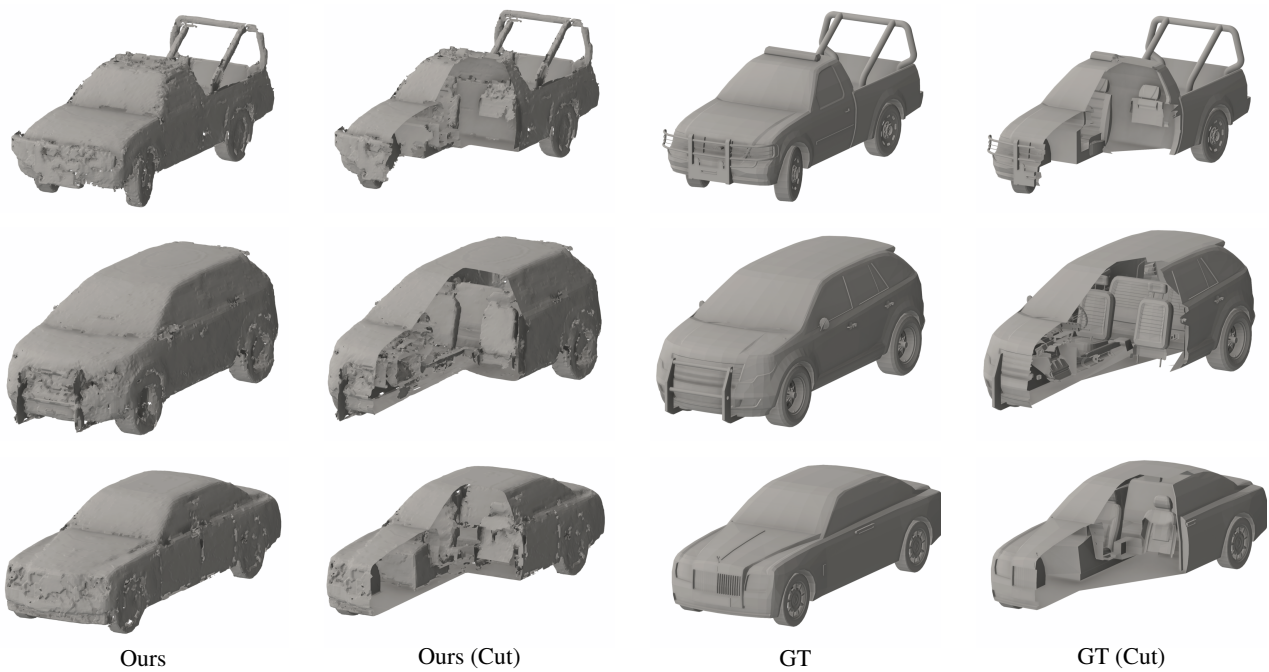
\begin{figure}[h]
		\centering
		{
			\begin{tikzpicture}[]
				\node[] (a) at (-6.5,0.6) {\includegraphics[width=0.23\textwidth]{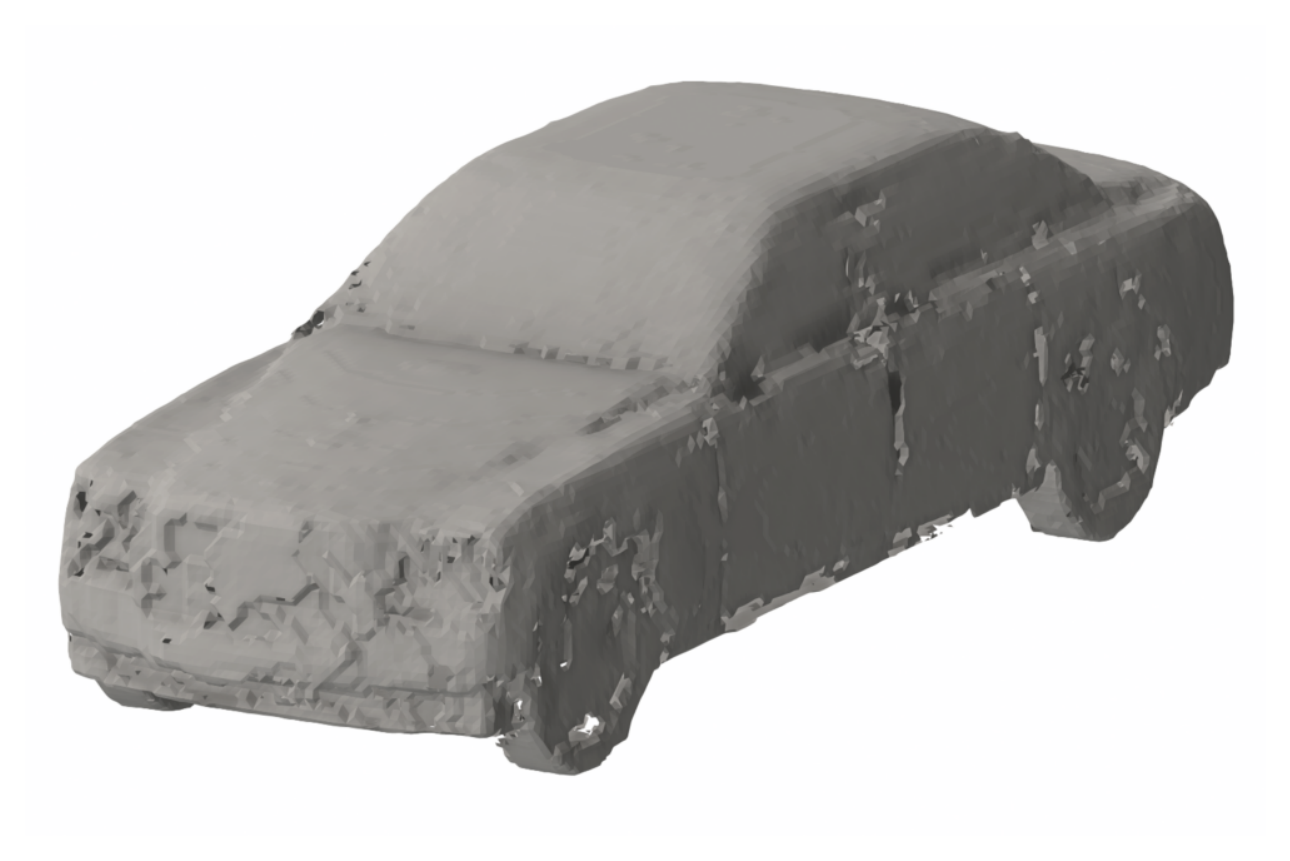} };
				\node[] (a) at (-6.5,3.5) {\includegraphics[width=0.23\textwidth]{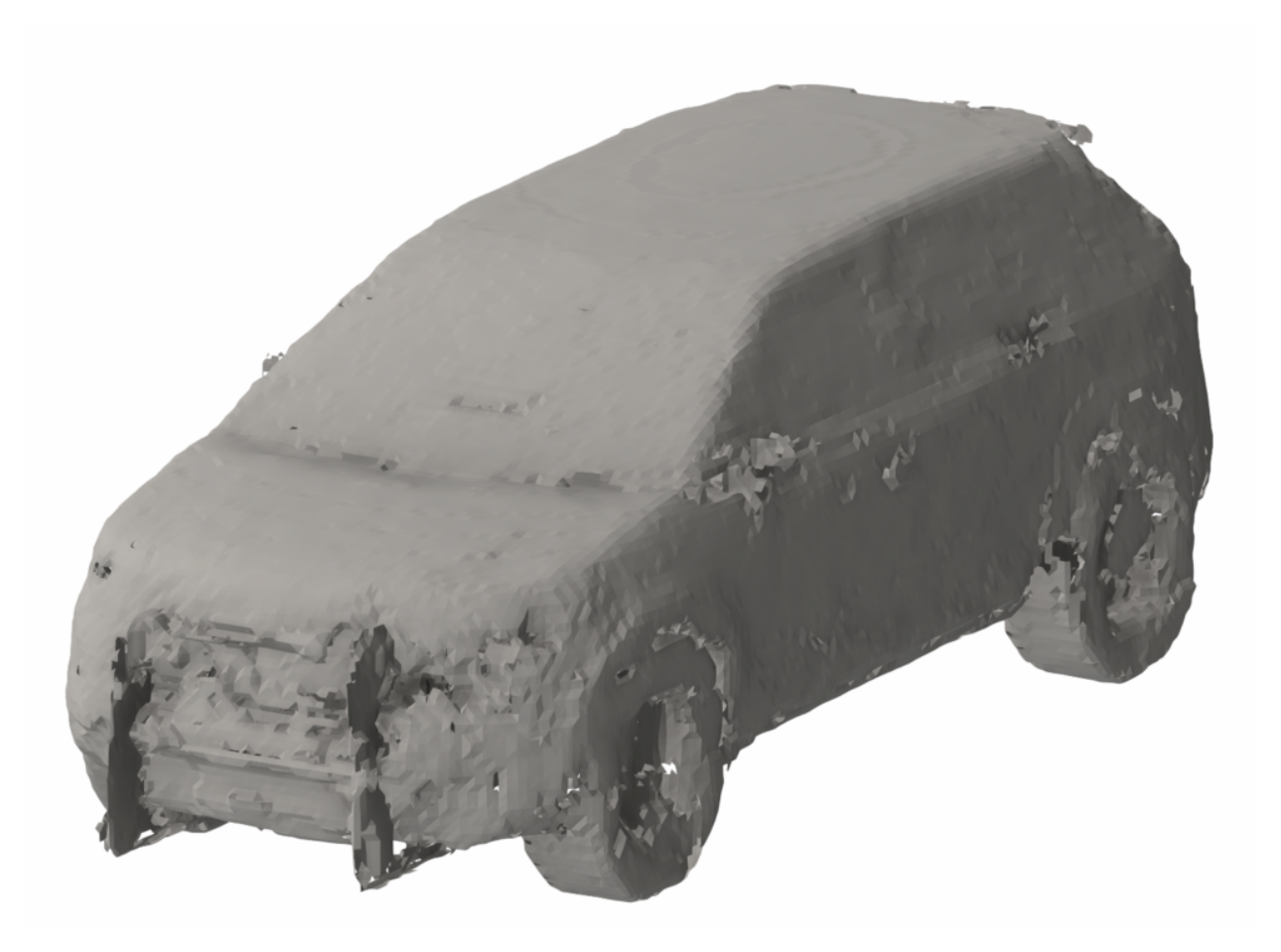} };
				\node[] (a) at (-6.5,6.5) {\includegraphics[width=0.23\textwidth]{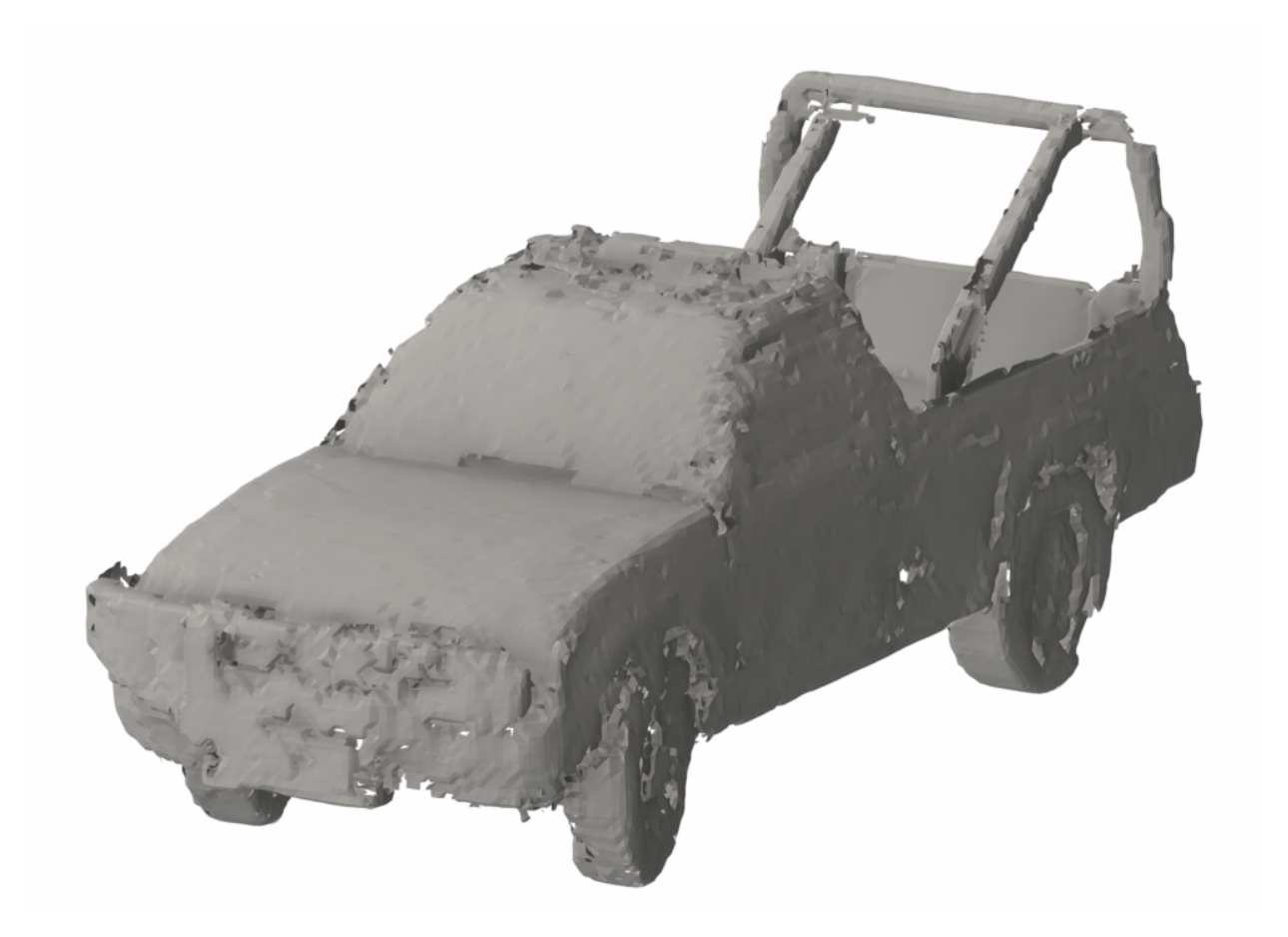} };
				\node[] (a) at (-6.5,-0.8) {\small  Ours};
				
				\node[] (b) at (-6.5+13/3,0.6) {\includegraphics[width=0.23\textwidth]{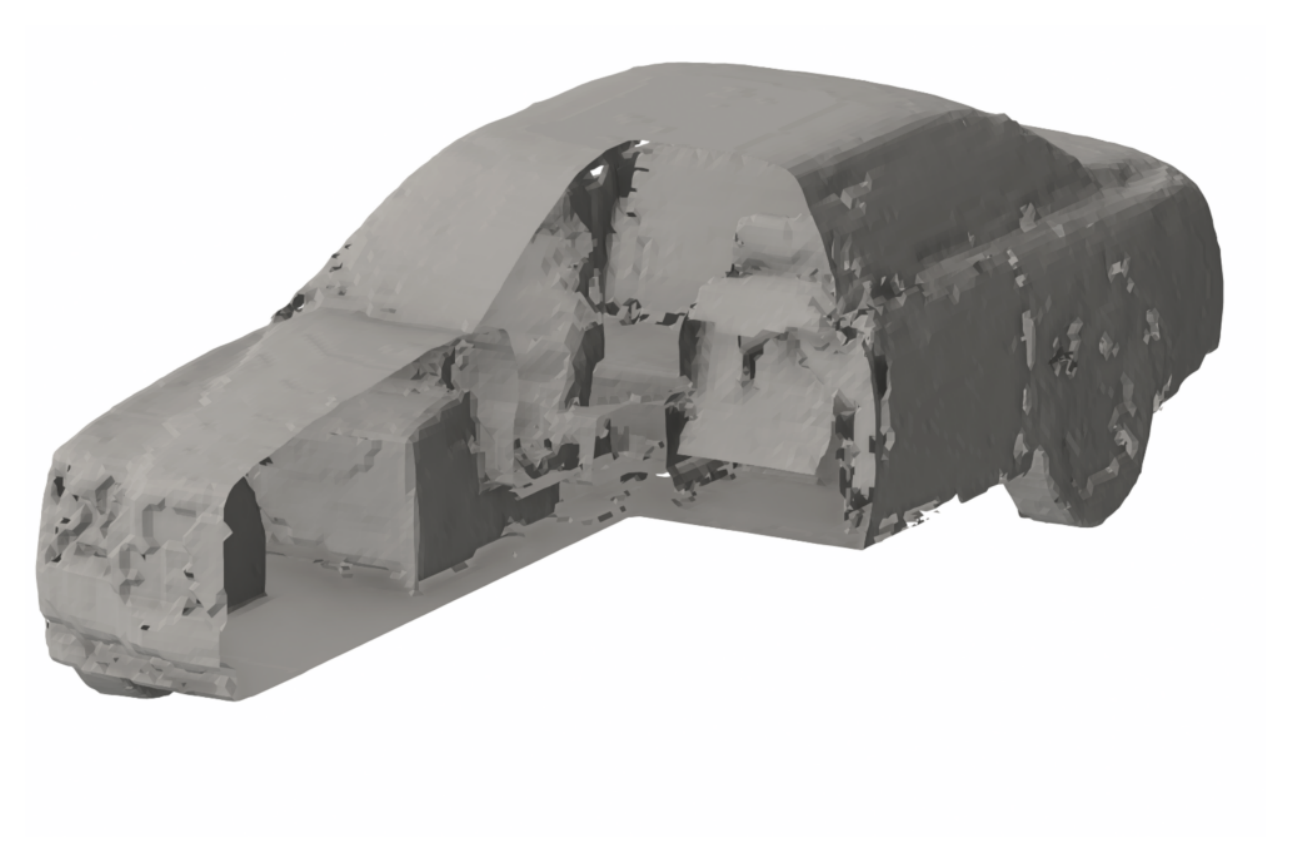}};
				\node[] (b) at (-6.5+13/3,3.5) {\includegraphics[width=0.23\textwidth]{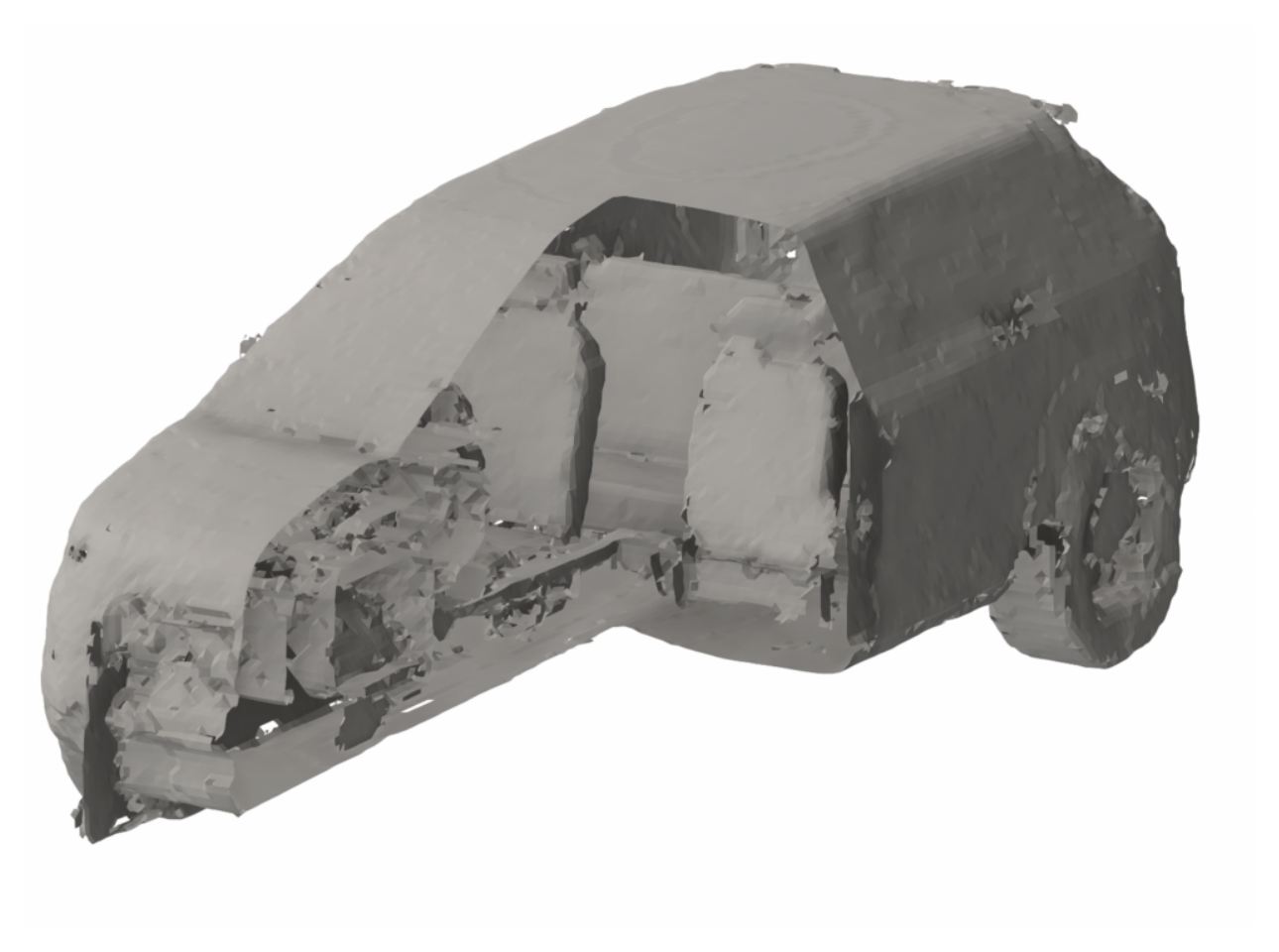}};
				\node[] (b) at (-6.5+13/3,6.5) {\includegraphics[width=0.23\textwidth]{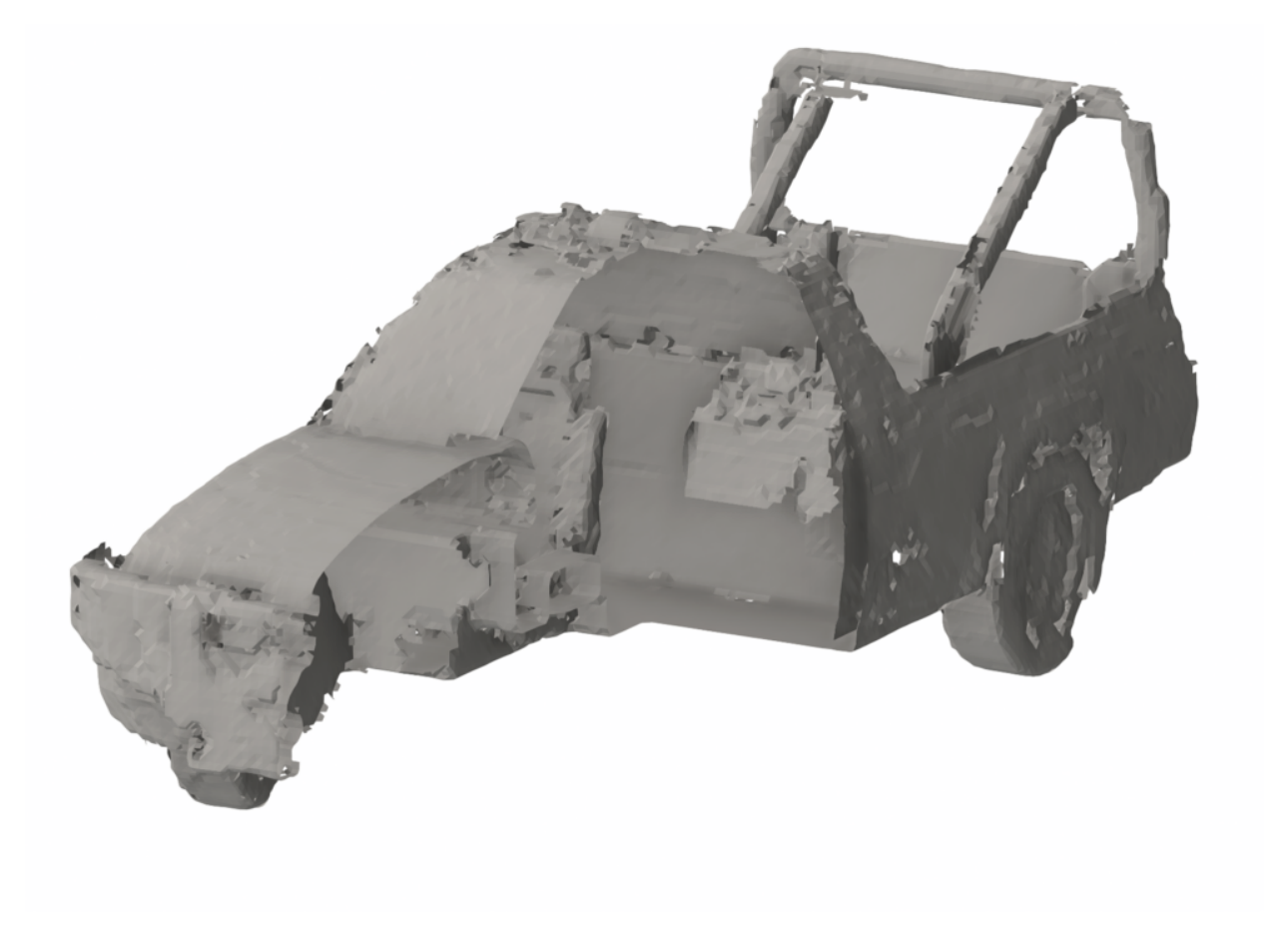}};
				\node[] (b) at (-6.5+13/3,-0.8) {\small  Ours (Cut) };
				
				\node[] (c) at (-6.5+13/3*2,0.6) {\includegraphics[width=0.23\textwidth]{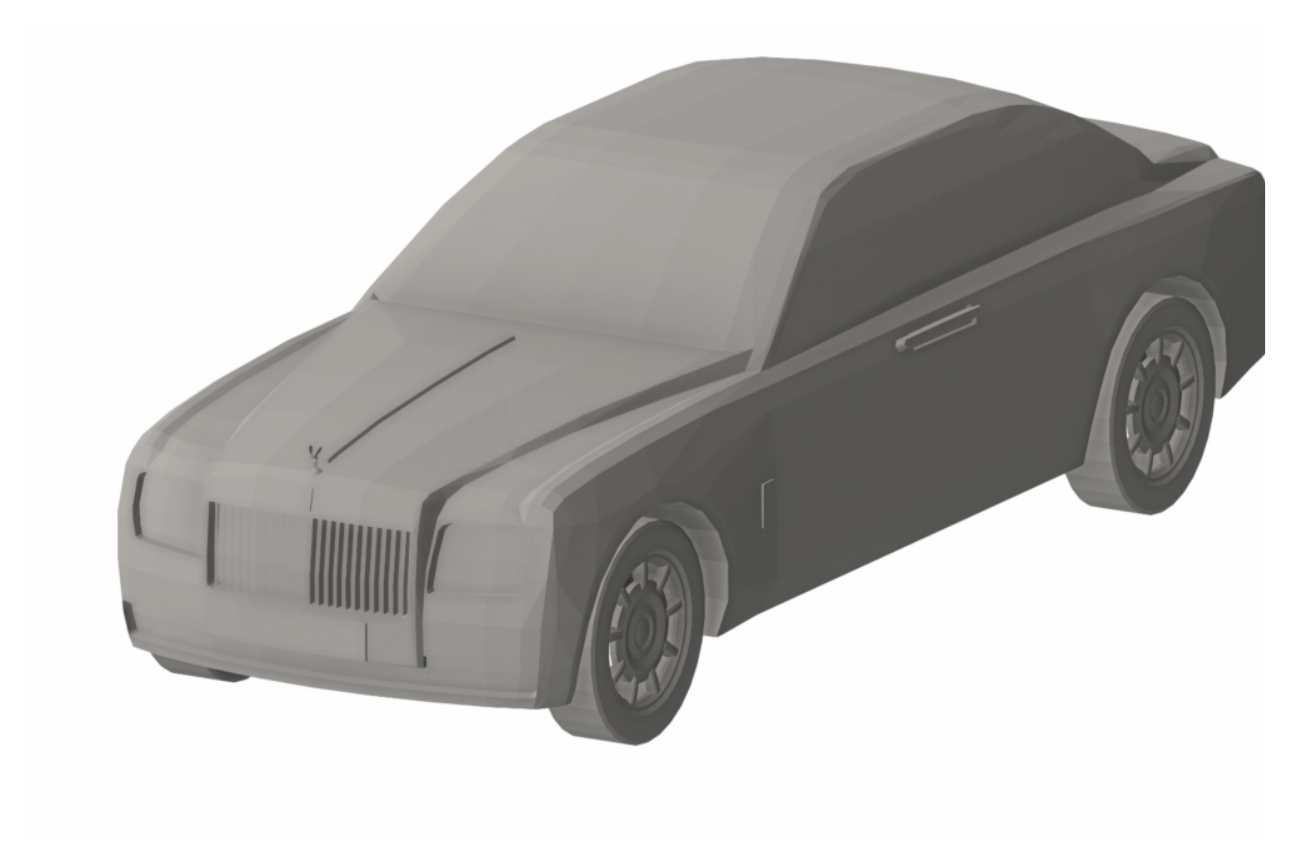}};
				\node[] (c) at (-6.5+13/3*2,3.5) {\includegraphics[width=0.23\textwidth]{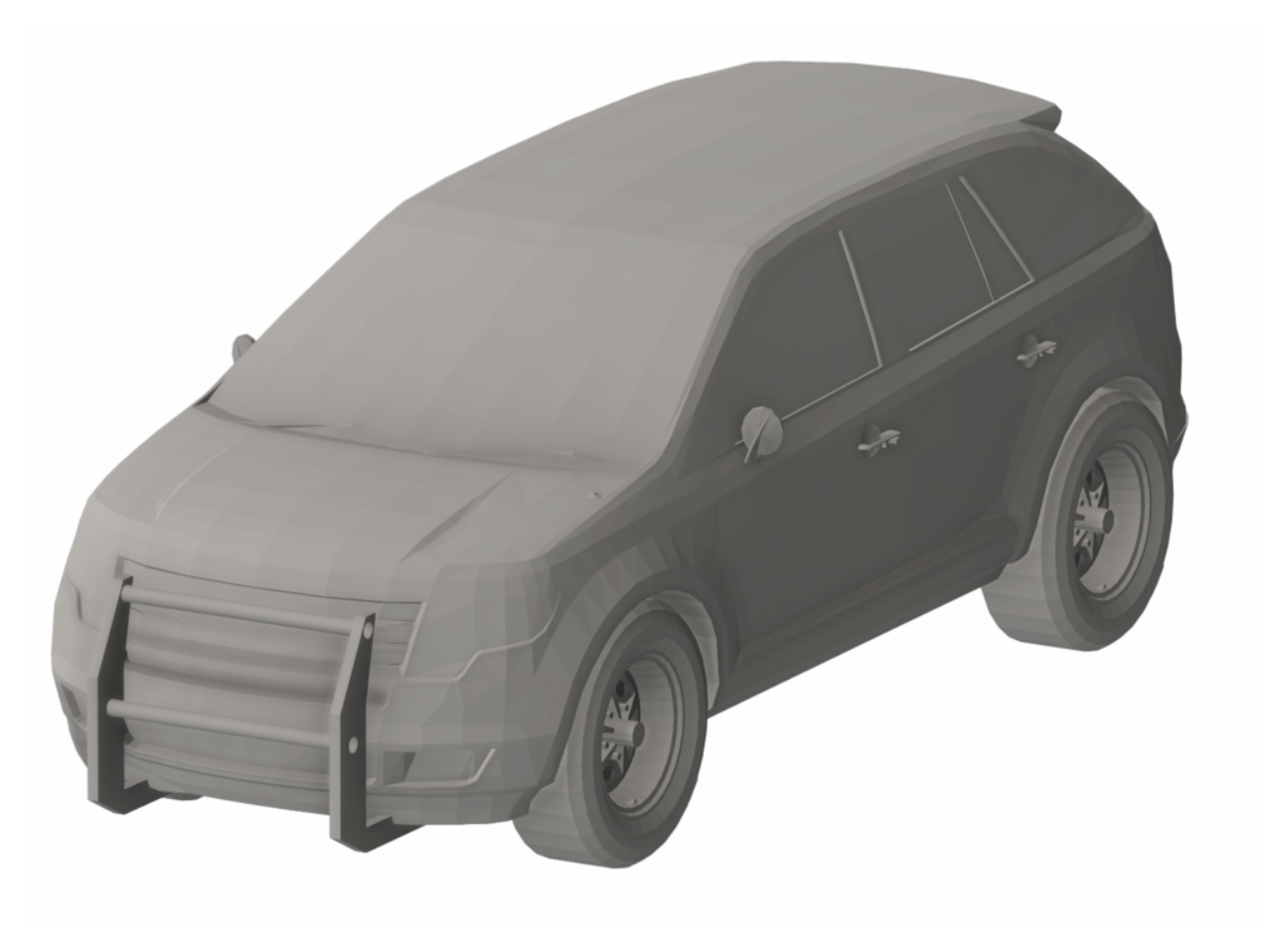}};
				\node[] (c) at (-6.5+13/3*2,6.5) {\includegraphics[width=0.23\textwidth]{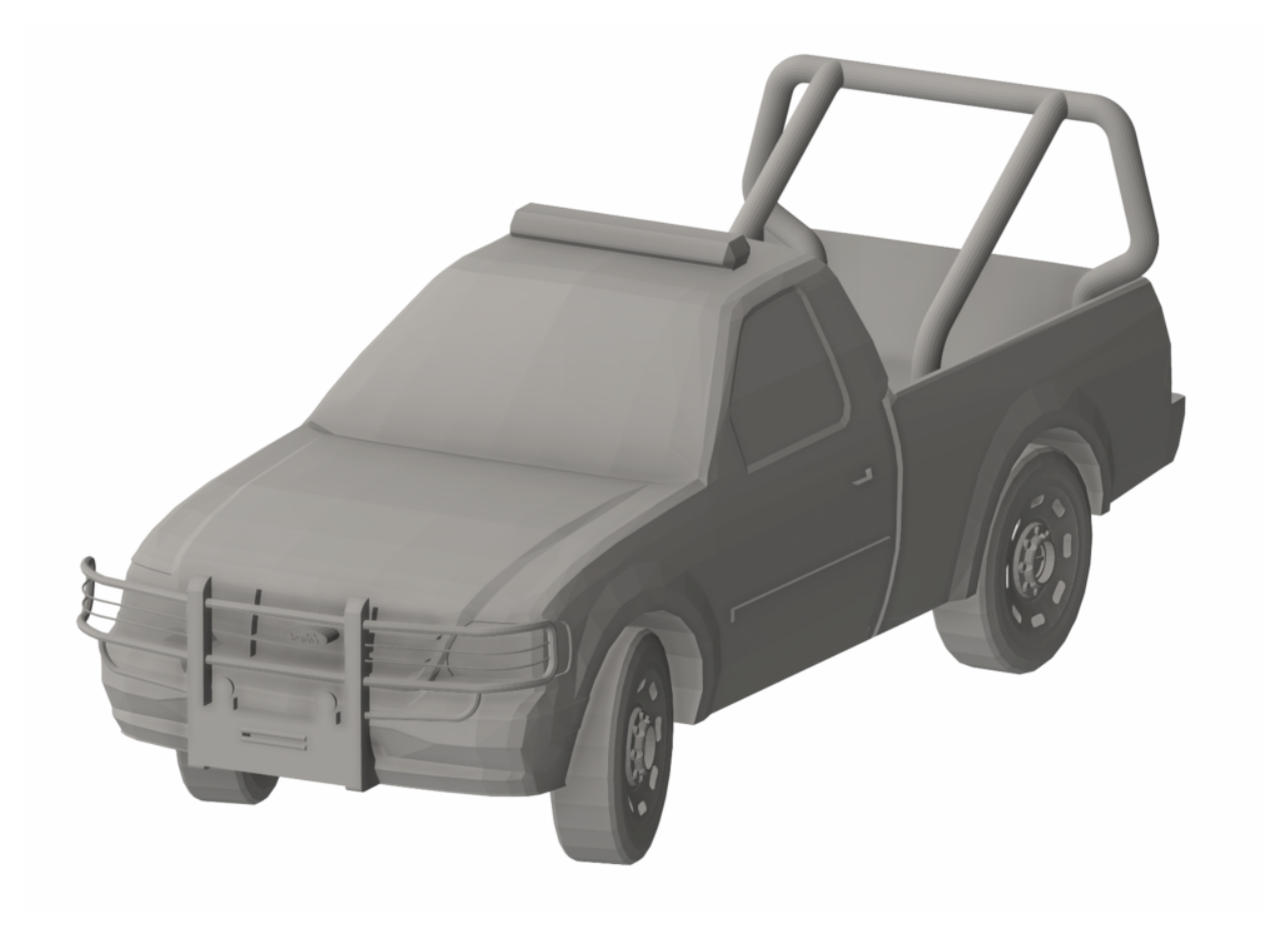} };
				\node[] (c) at (-6.5+13/3*2,-0.8) {\small  GT  };
				
				\node[] (d) at (-6.5+13/3*3,0.6) {\includegraphics[width=0.23\textwidth]{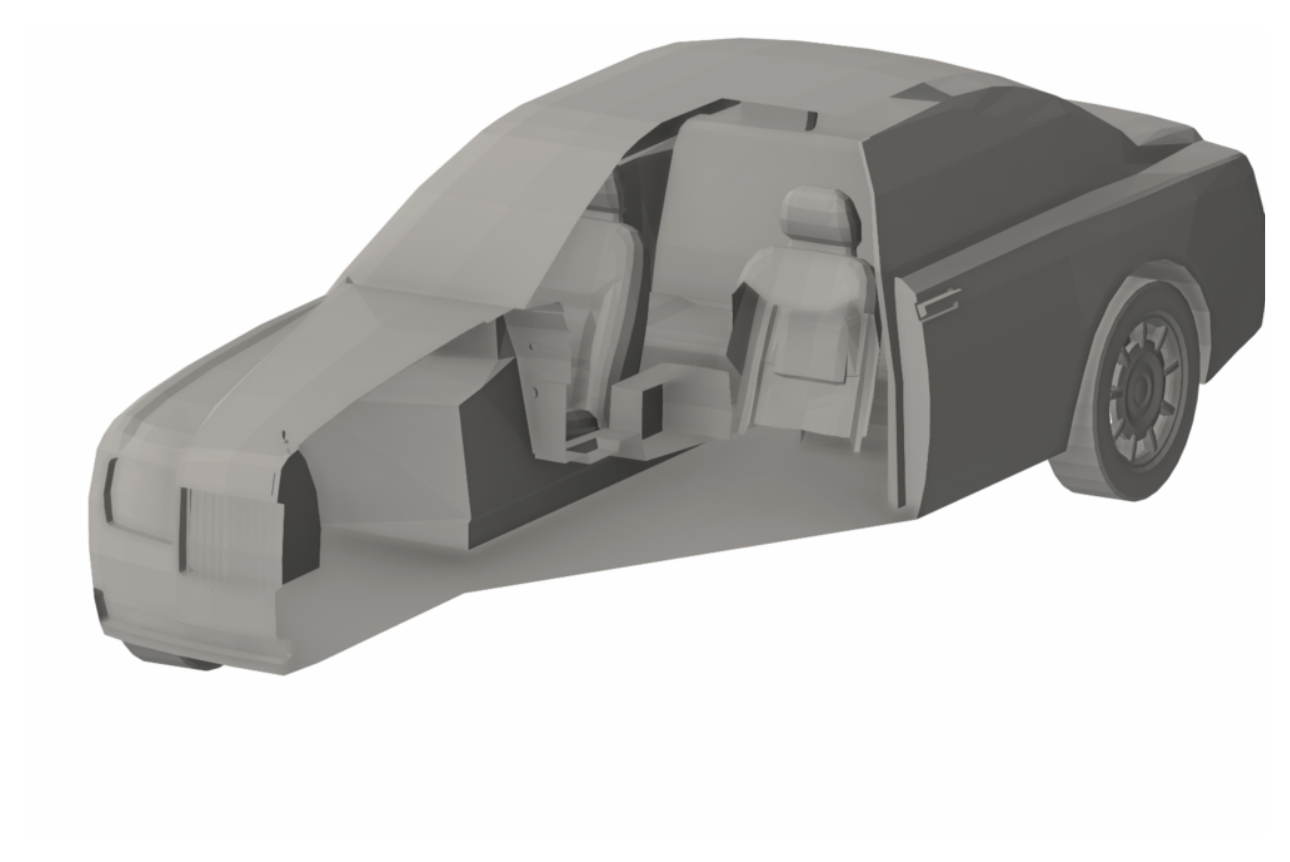}};
				\node[] (d) at (-6.5+13/3*3,3.5) {\includegraphics[width=0.23\textwidth]{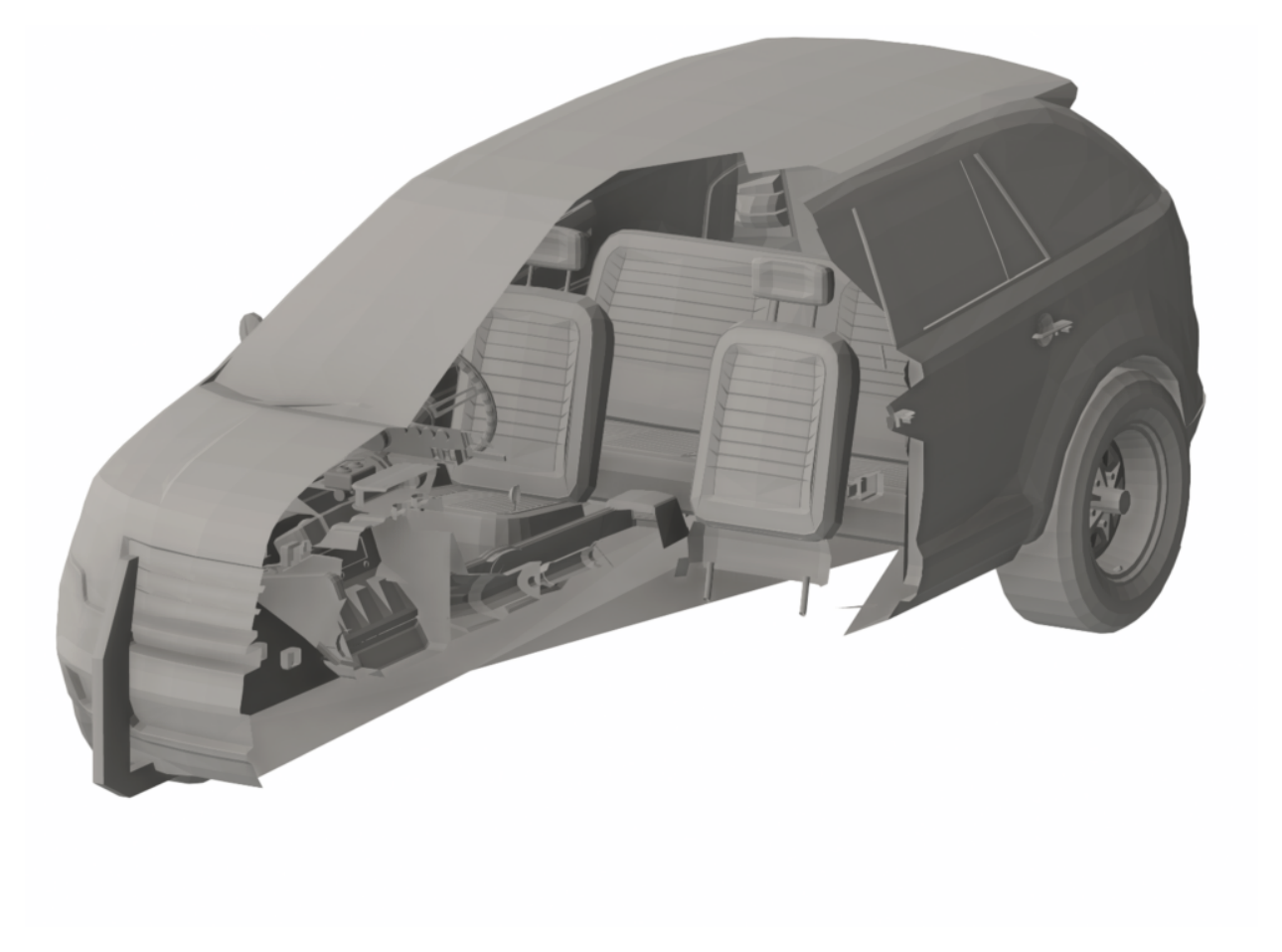}};
				\node[] (d) at (-6.5+13/3*3,6.5) {\includegraphics[width=0.23\textwidth]{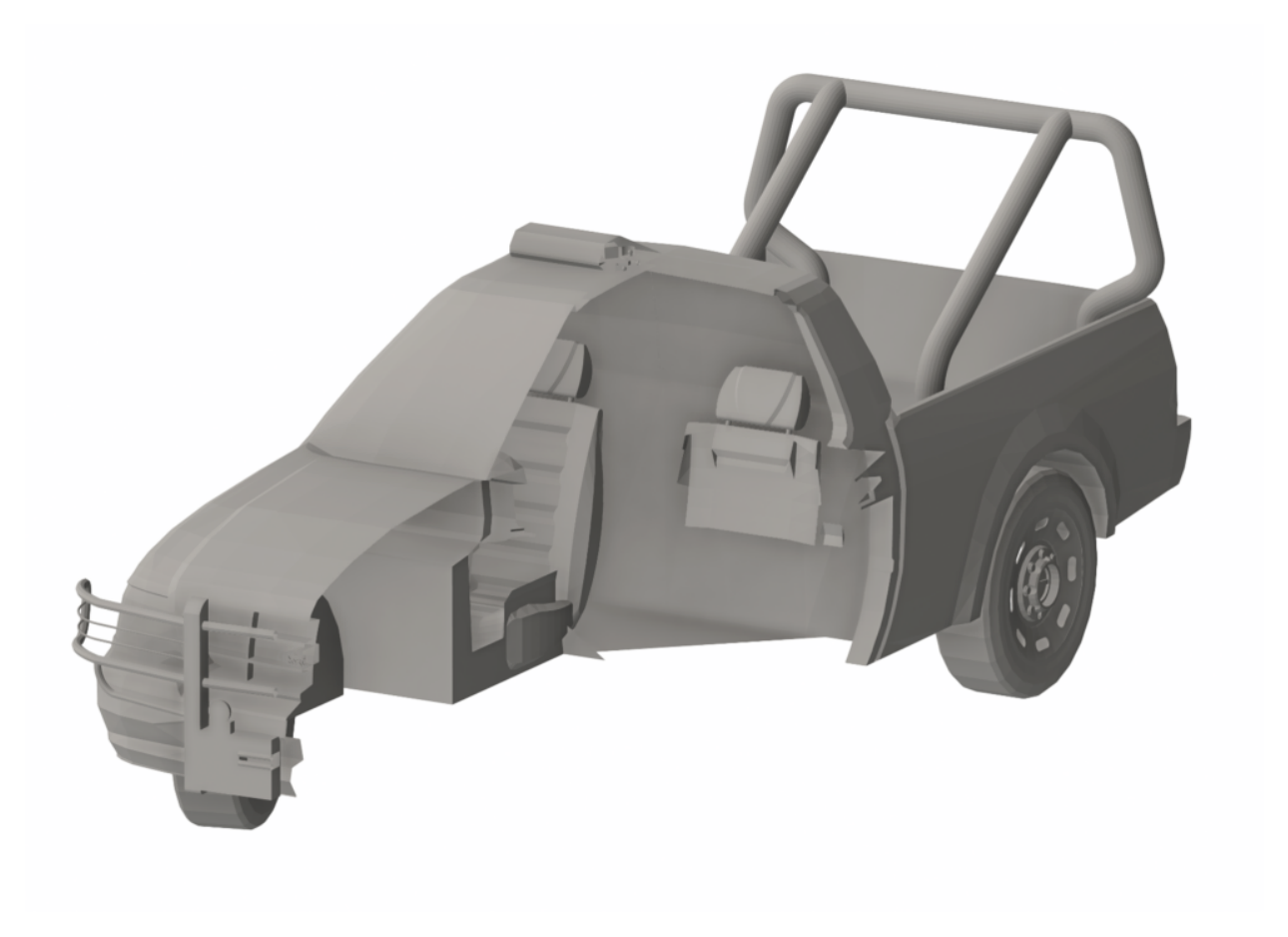} };
				\node[] (d) at (-6.5+13/3*3,-0.8) {\small  GT (Cut) };
				
			\end{tikzpicture}
		}
		\caption{More results of reconstructed shapes with multi-layer structures. }
		\label{MORE:MULTILAYER}
	\end{figure}

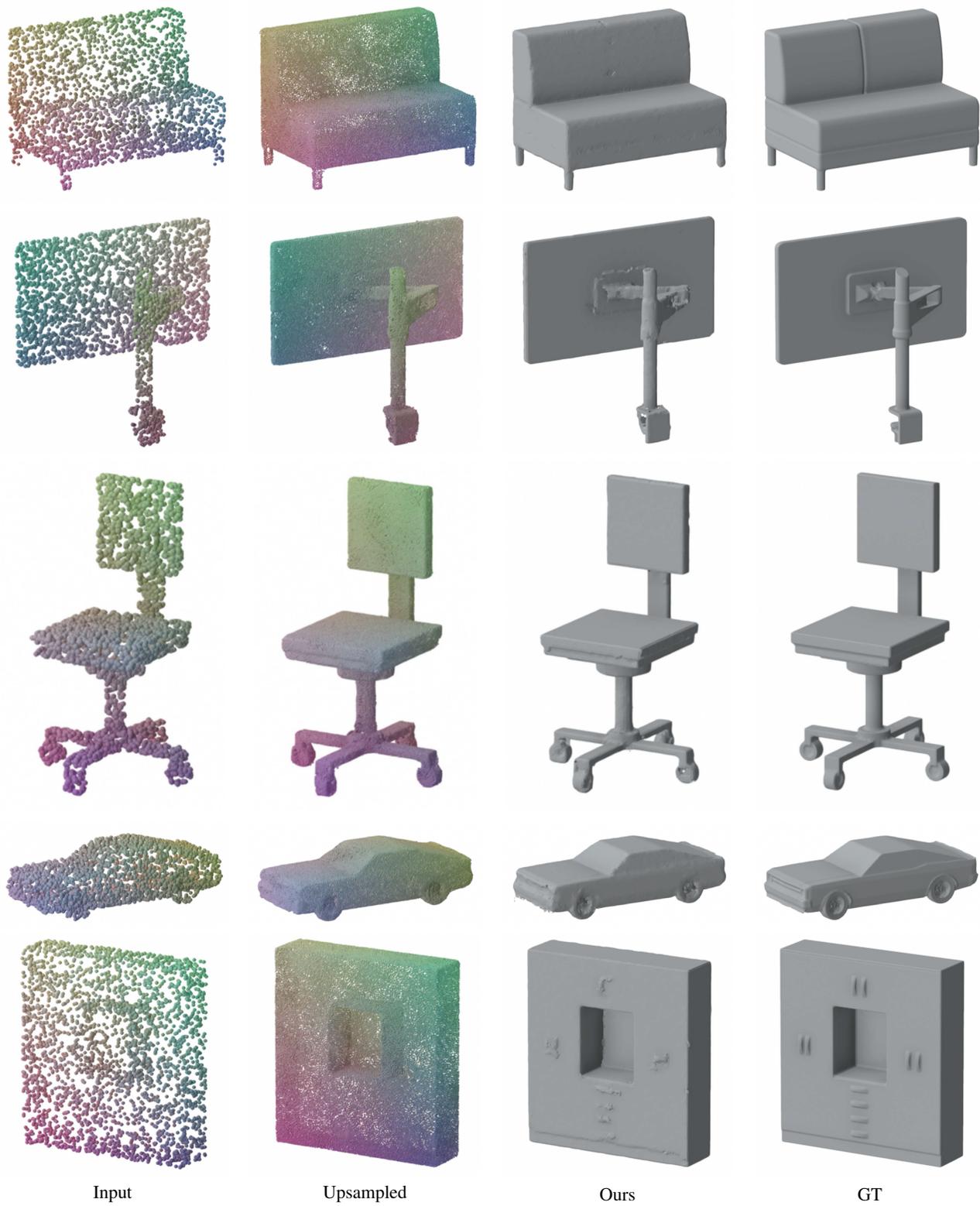
\begin{figure}[h]
		\centering
		{
			\begin{tikzpicture}[]
				\node[] (a) at (-6.5,0.6) {\includegraphics[width=0.24\textwidth]{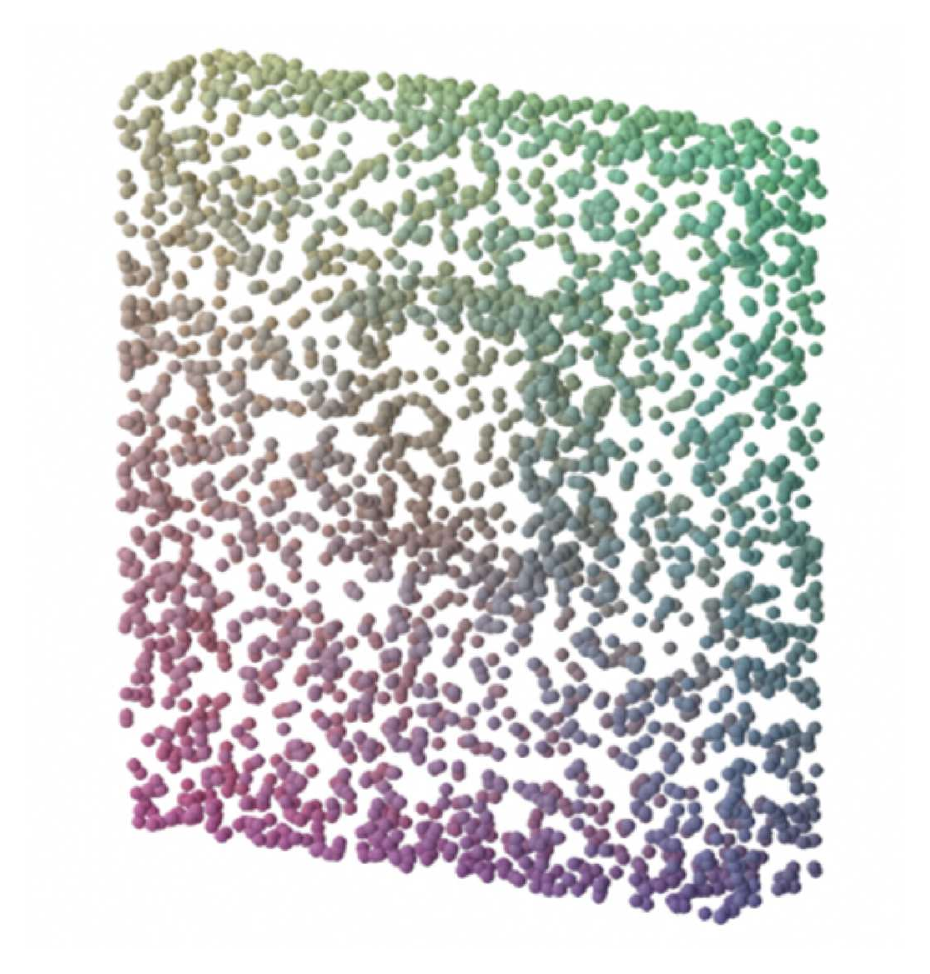} };
				\node[] (a) at (-6.5,3.7) {\includegraphics[width=0.24\textwidth]{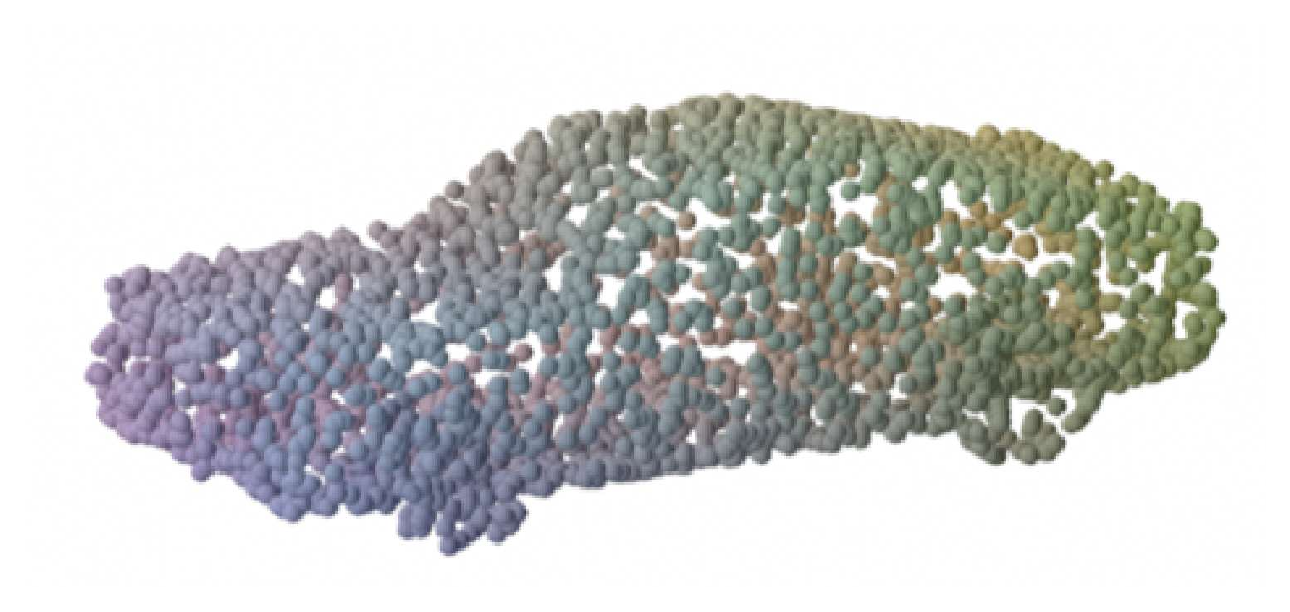} };
				\node[] (a) at (-6.5,7.8) {\includegraphics[width=0.24\textwidth]{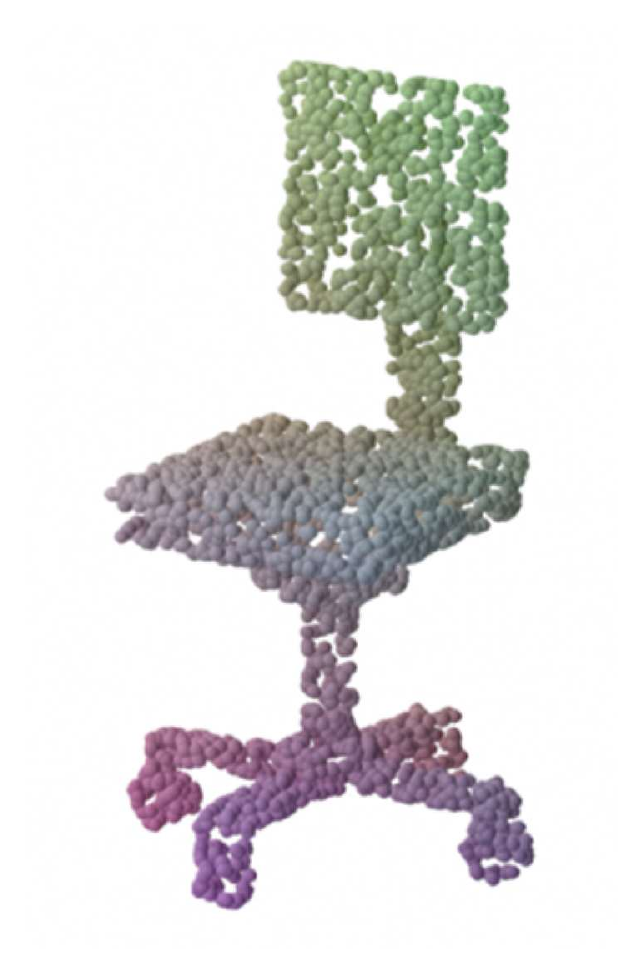} };
				\node[] (a) at (-6.5,13) {\includegraphics[width=0.24\textwidth]{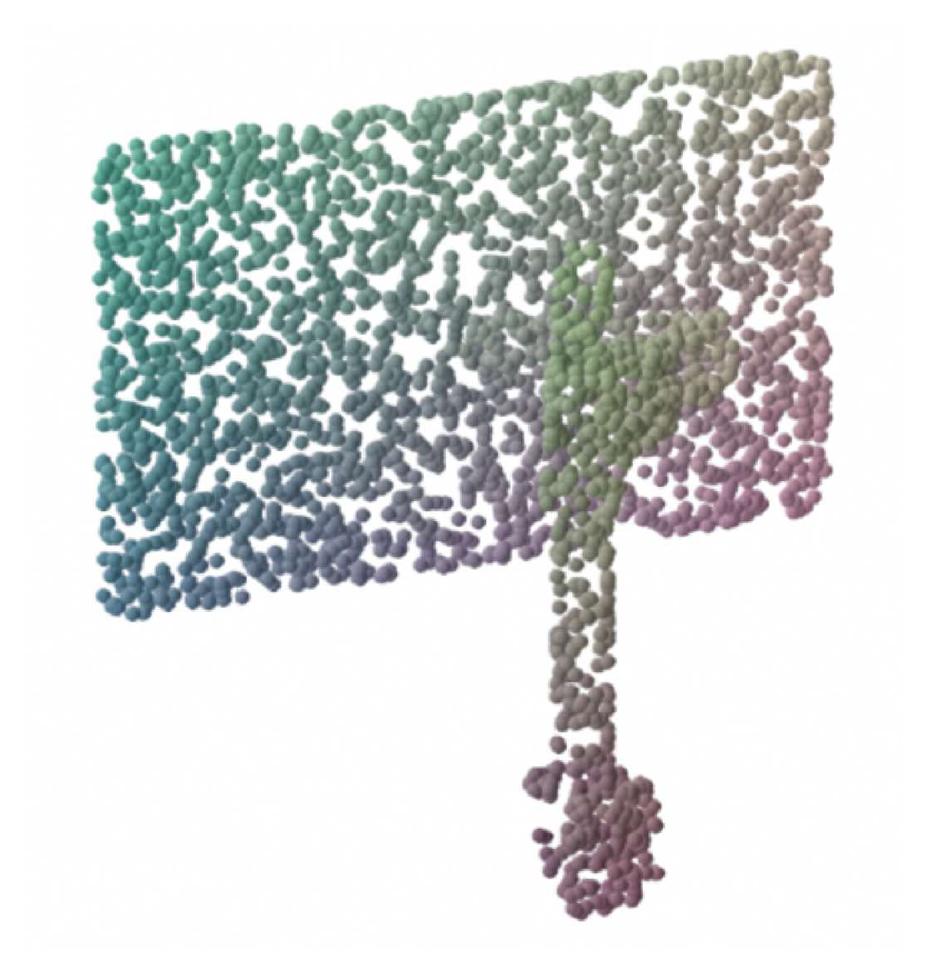} };
				\node[] (a) at (-6.5,17) {\includegraphics[width=0.24\textwidth]{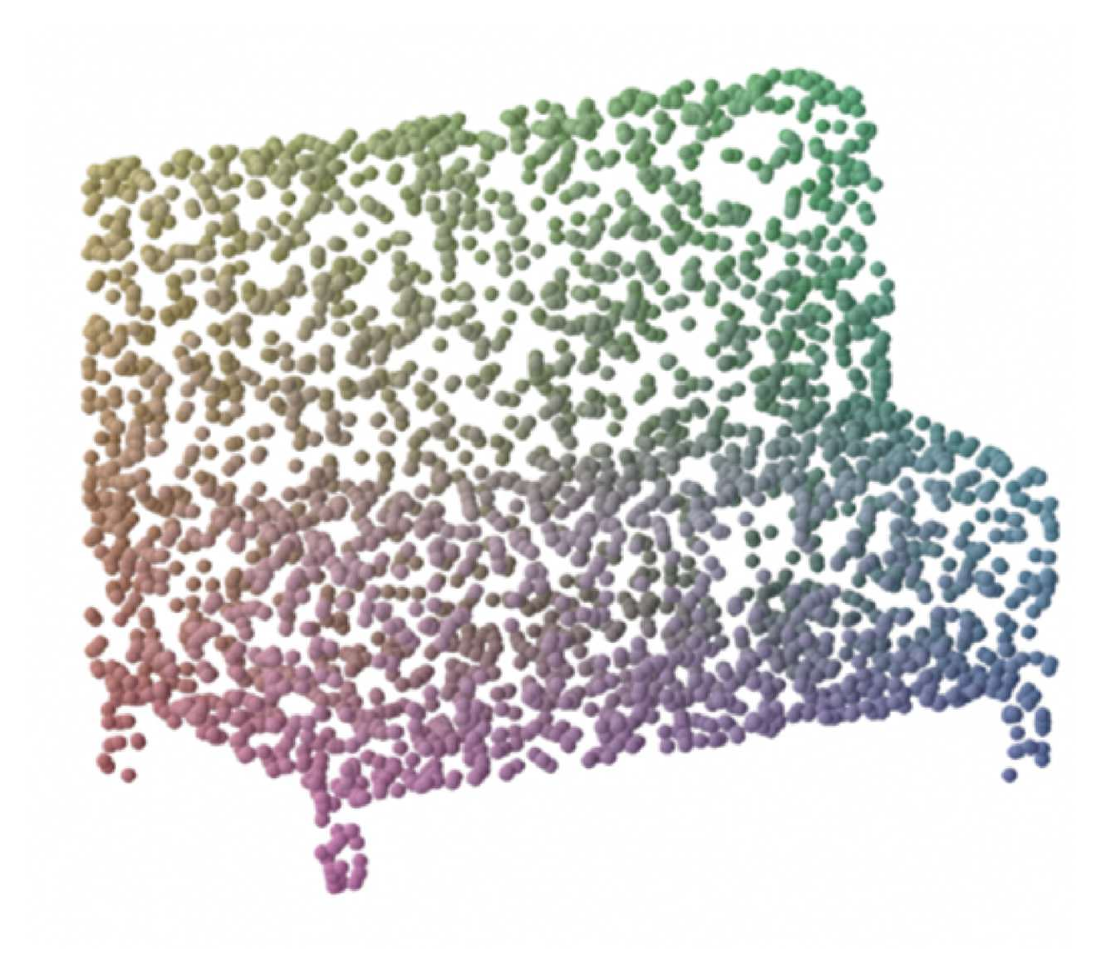} };
				\node[] (a) at (-6.5,-1.8) {\small  Input};
				
				\node[] (b) at (-6.5+13/3,0.6) {\includegraphics[width=0.24\textwidth]{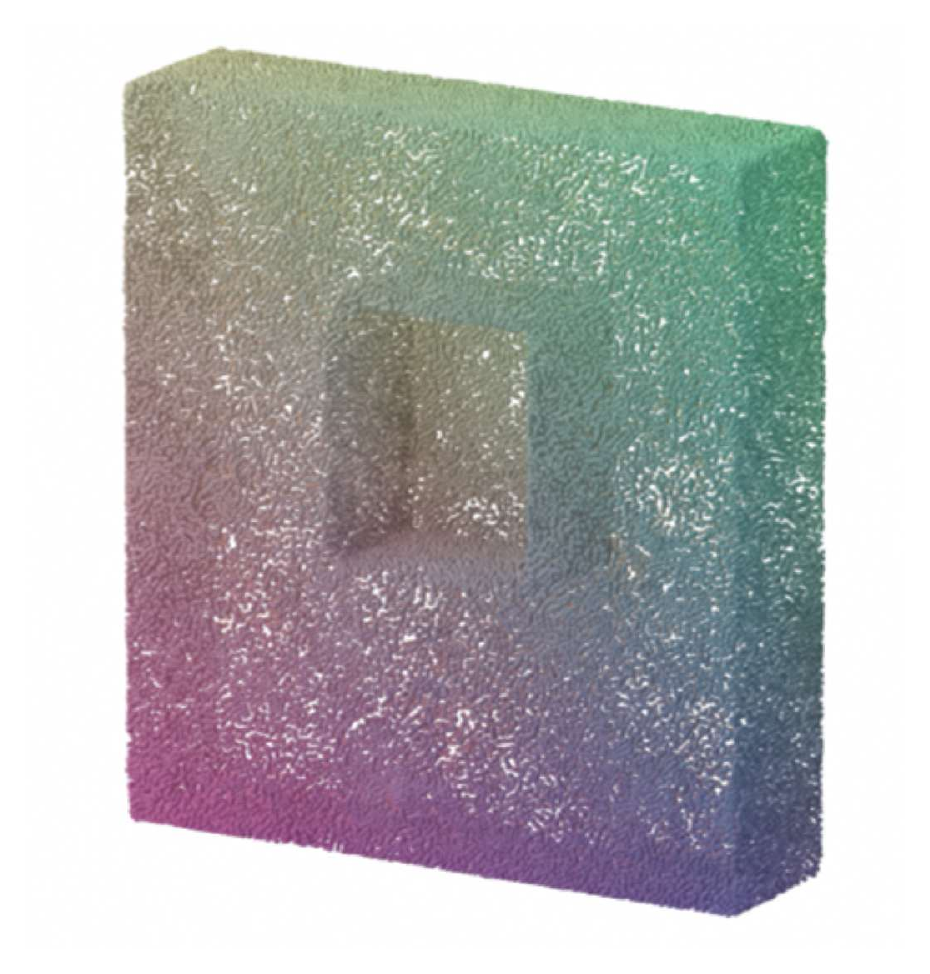}};
				\node[] (b) at (-6.5+13/3,3.7) {\includegraphics[width=0.24\textwidth]{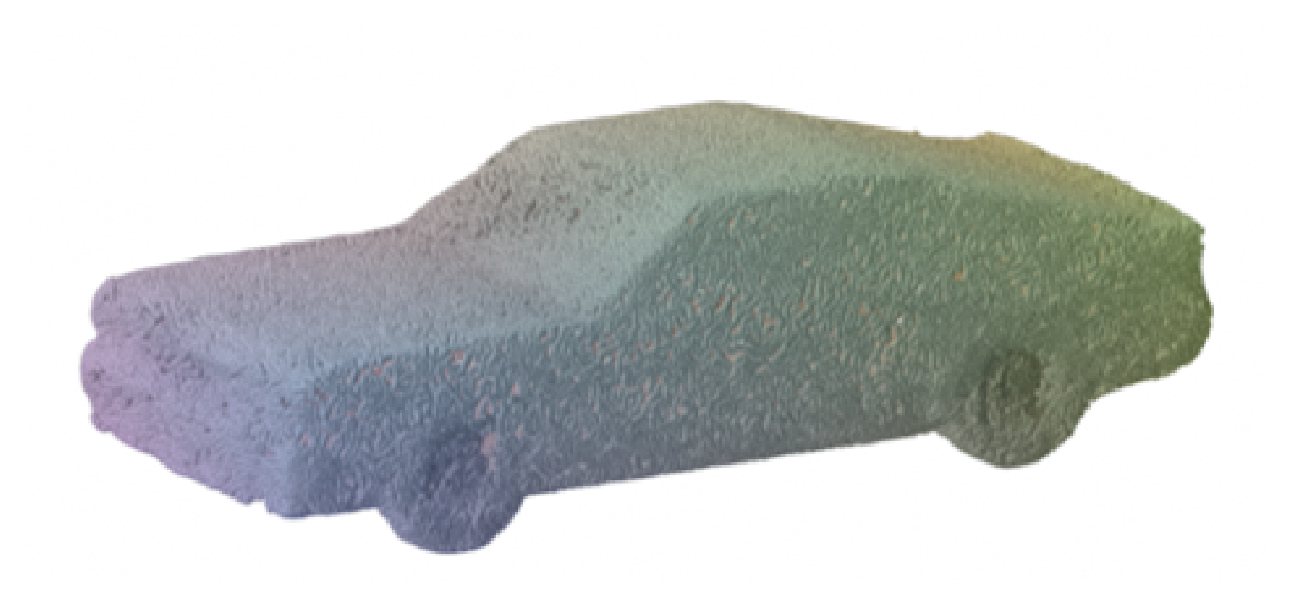}};
				\node[] (b) at (-6.5+13/3,7.8) {\includegraphics[width=0.24\textwidth]{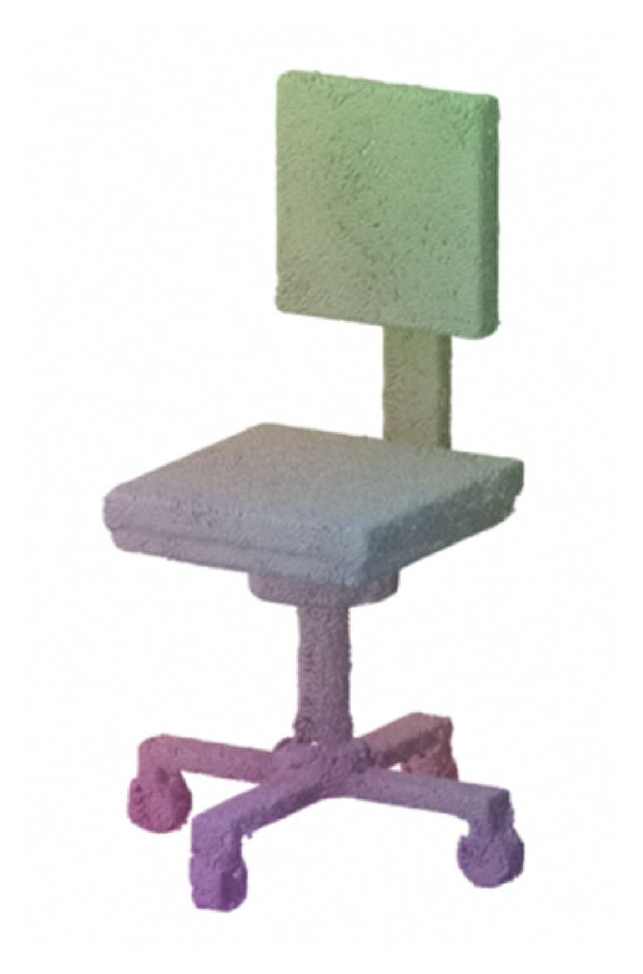}};
				\node[] (b) at (-6.5+13/3,13) {\includegraphics[width=0.24\textwidth]{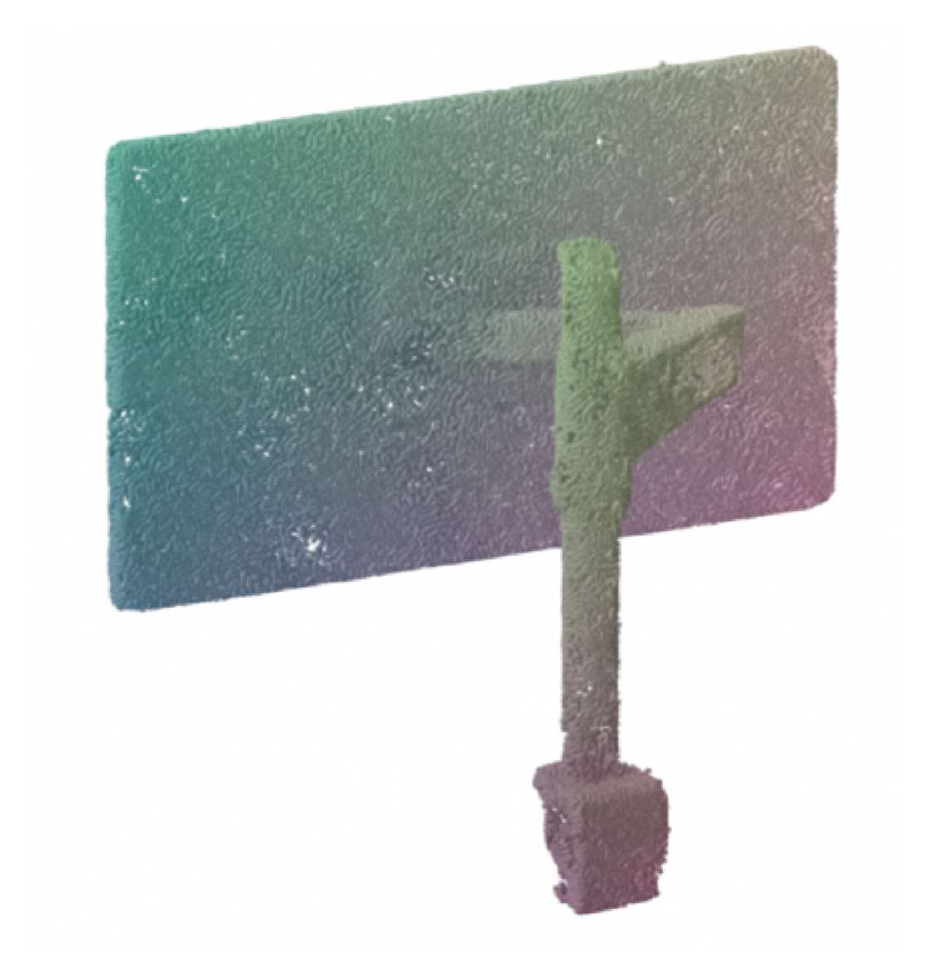}};
				\node[] (b) at (-6.5+13/3,17) {\includegraphics[width=0.24\textwidth]{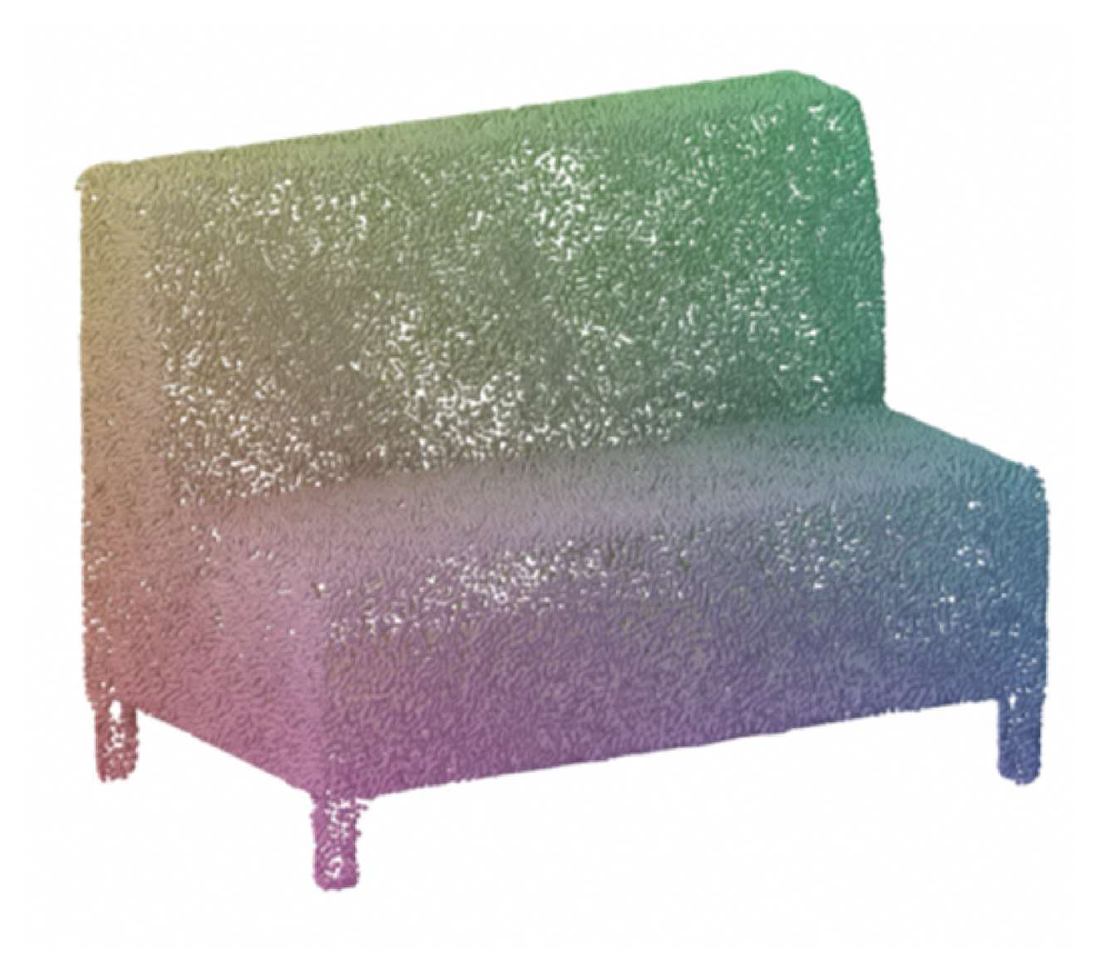}};
				\node[] (b) at (-6.5+13/3,-1.8) {\small  Upsampled };
				
				\node[] (c) at (-6.5+13/3*2,0.6) {\includegraphics[width=0.24\textwidth]{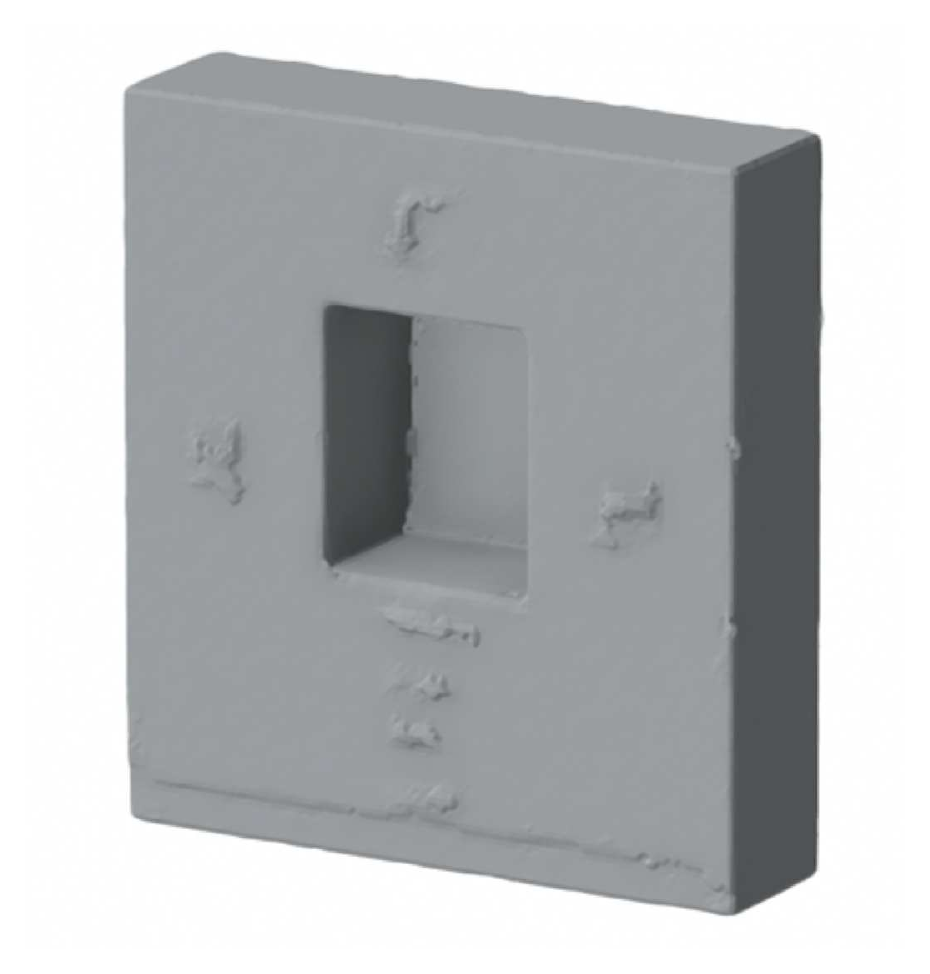}};
				\node[] (c) at (-6.5+13/3*2,3.7) {\includegraphics[width=0.24\textwidth]{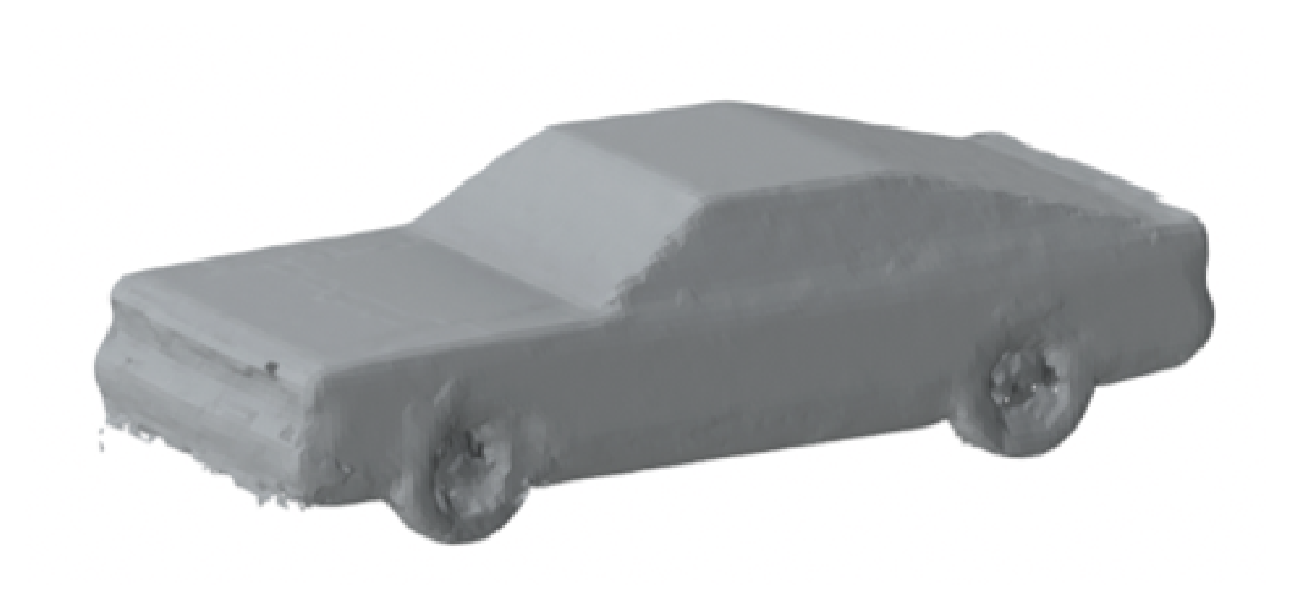}};
				\node[] (c) at (-6.5+13/3*2,7.8) {\includegraphics[width=0.24\textwidth]{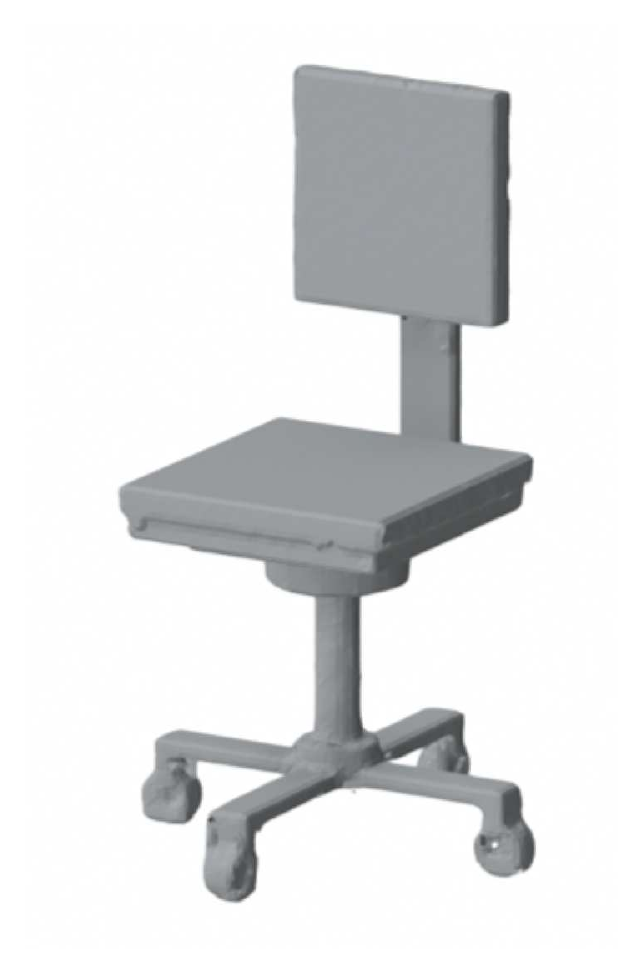} };
				\node[] (c) at (-6.5+13/3*2,13) {\includegraphics[width=0.24\textwidth]{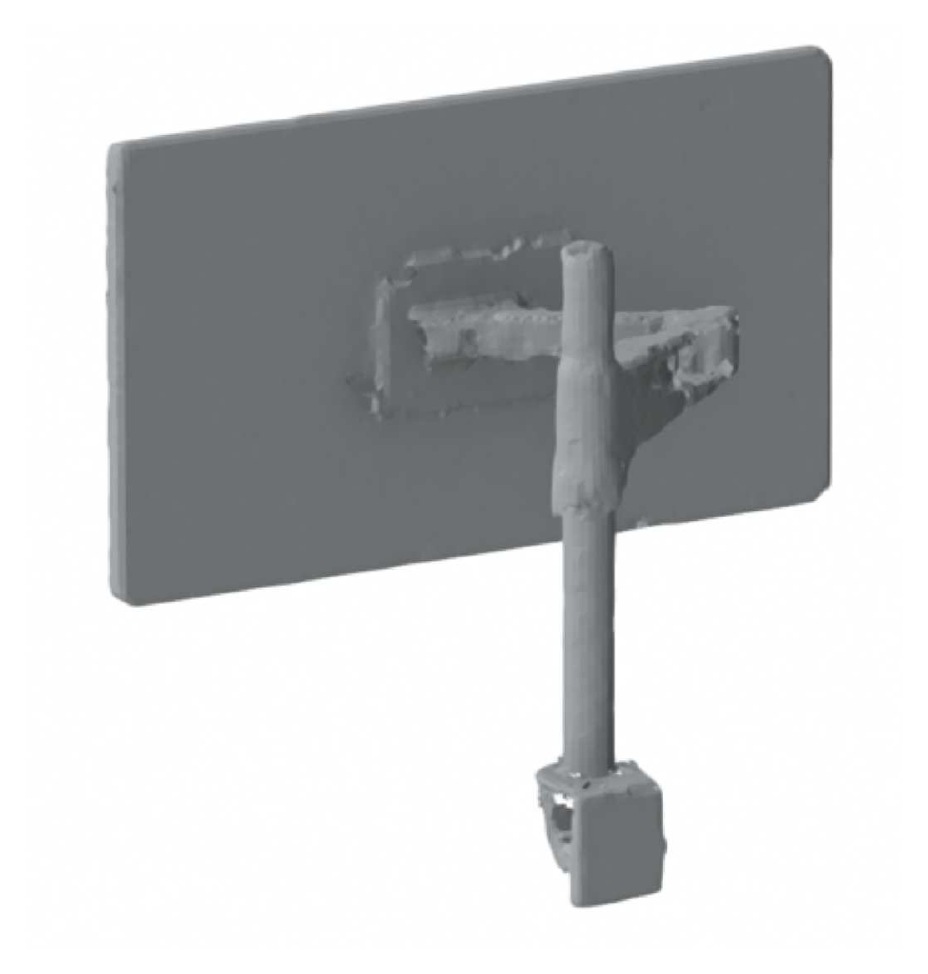} };
				\node[] (c) at (-6.5+13/3*2,17) {\includegraphics[width=0.24\textwidth]{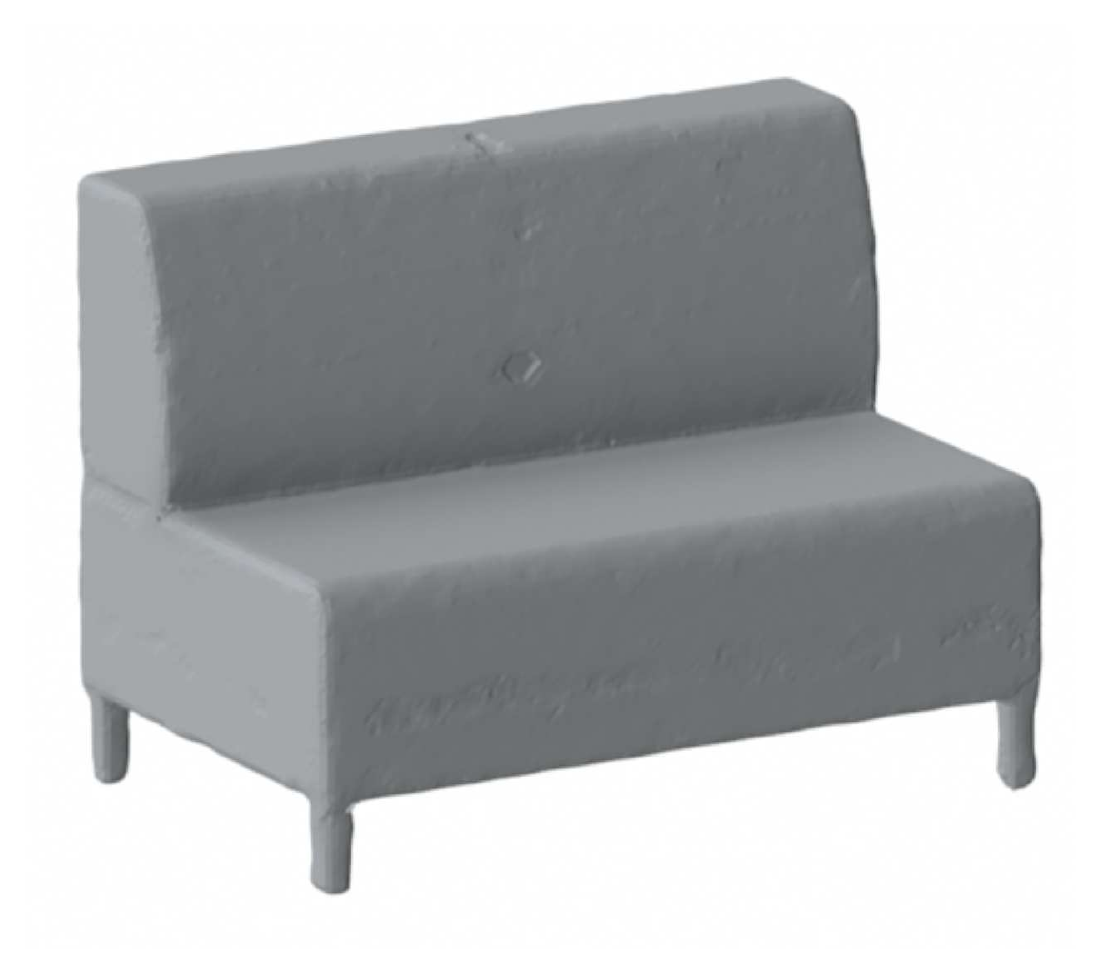} };
				\node[] (c) at (-6.5+13/3*2,-1.8) {\small  Ours  };
				
				\node[] (d) at (-6.5+13/3*3,0.6) {\includegraphics[width=0.24\textwidth]{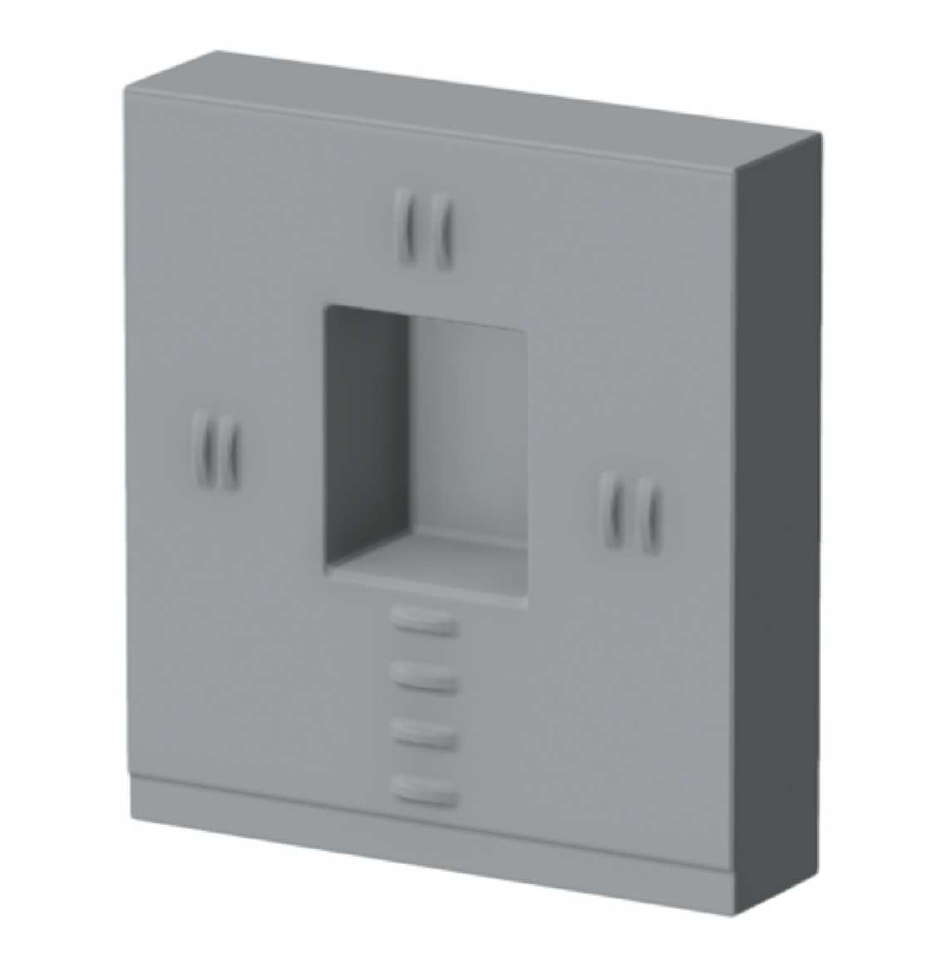}};
				\node[] (d) at (-6.5+13/3*3,3.7) {\includegraphics[width=0.24\textwidth]{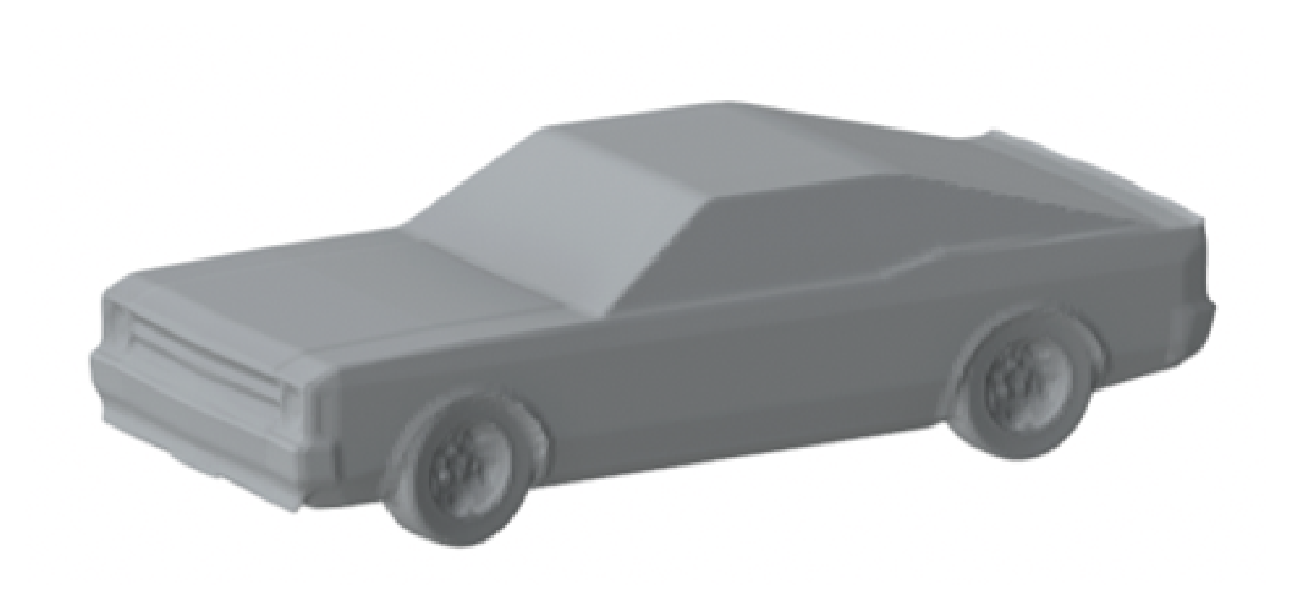}};
				\node[] (d) at (-6.5+13/3*3,7.8) {\includegraphics[width=0.24\textwidth]{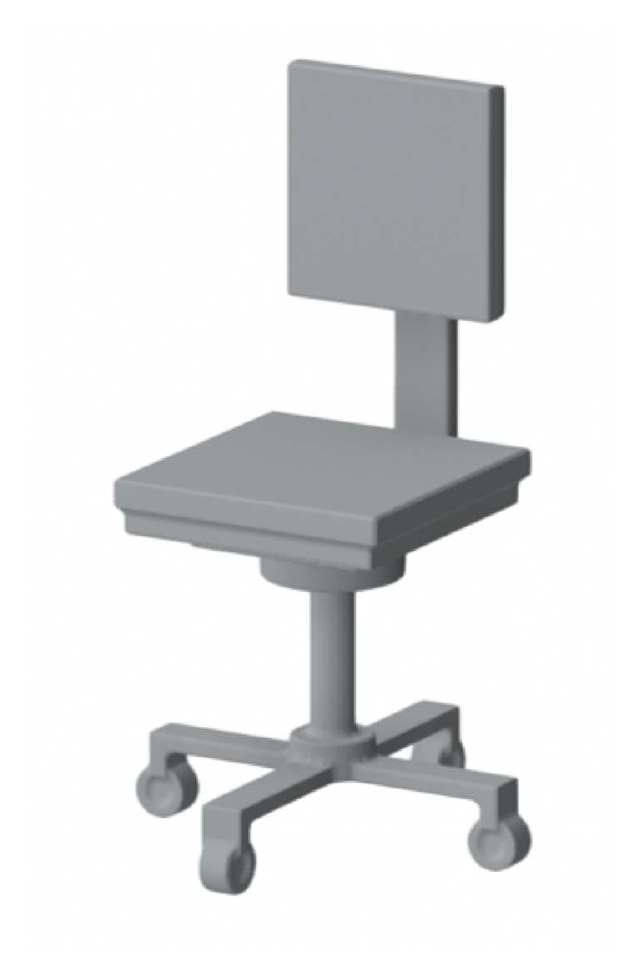} };
				\node[] (d) at (-6.5+13/3*3,13) {\includegraphics[width=0.24\textwidth]{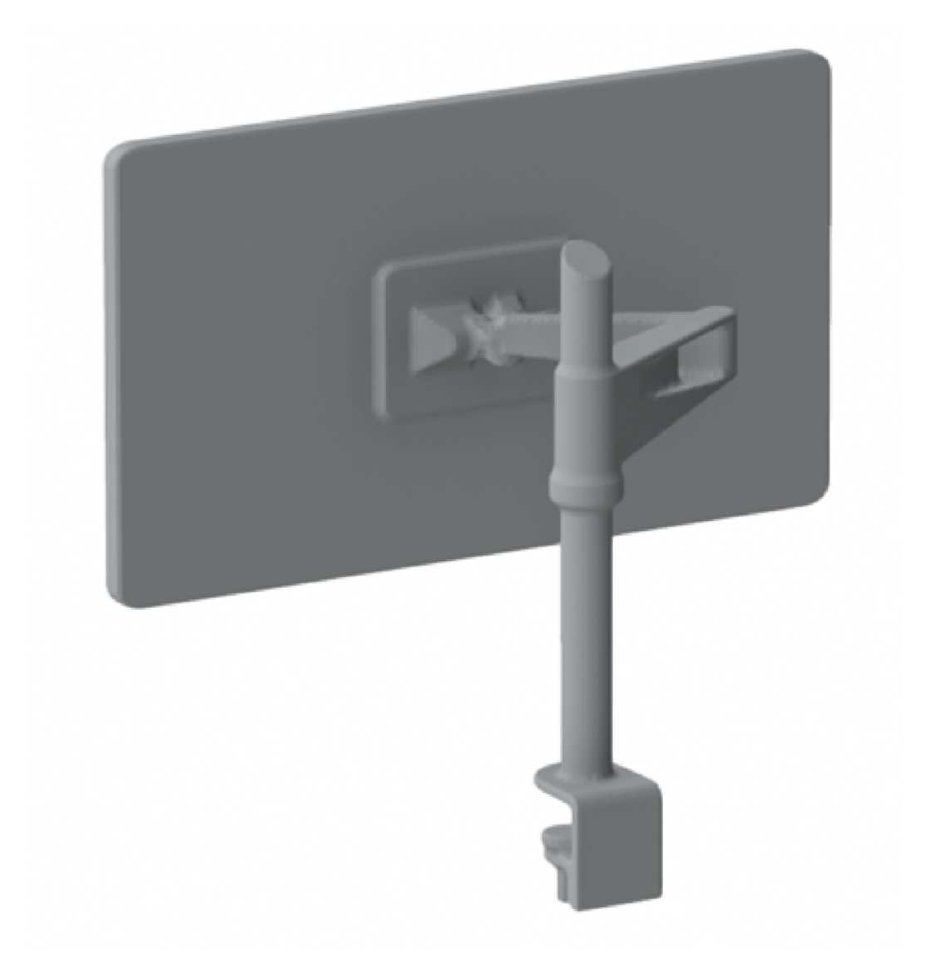} };
				\node[] (d) at (-6.5+13/3*3,17) {\includegraphics[width=0.24\textwidth]{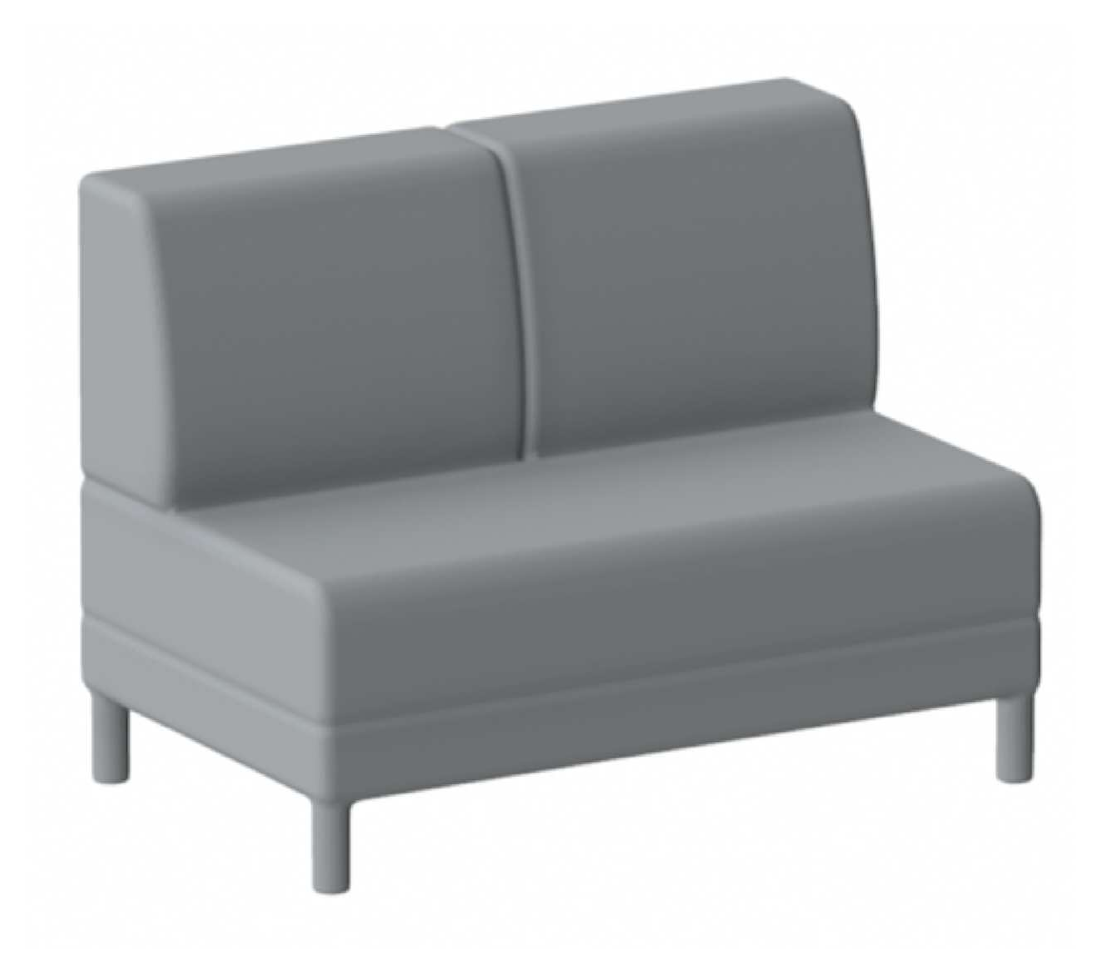} };
				\node[] (d) at (-6.5+13/3*3,-1.8) {\small  GT };
				
			\end{tikzpicture}
		}
		\setlength{\abovecaptionskip}{-0.35cm} \vspace{0.4cm}
		\caption{More results of reconstructed water-tight shapes. }
		\label{MORE:SHAPENET}
	\end{figure}

\begin{figure}[h]
		\centering
		{
			\begin{tikzpicture}[]
				\node[] (a) at (-6.5,0.6) {\includegraphics[width=0.24\textwidth]{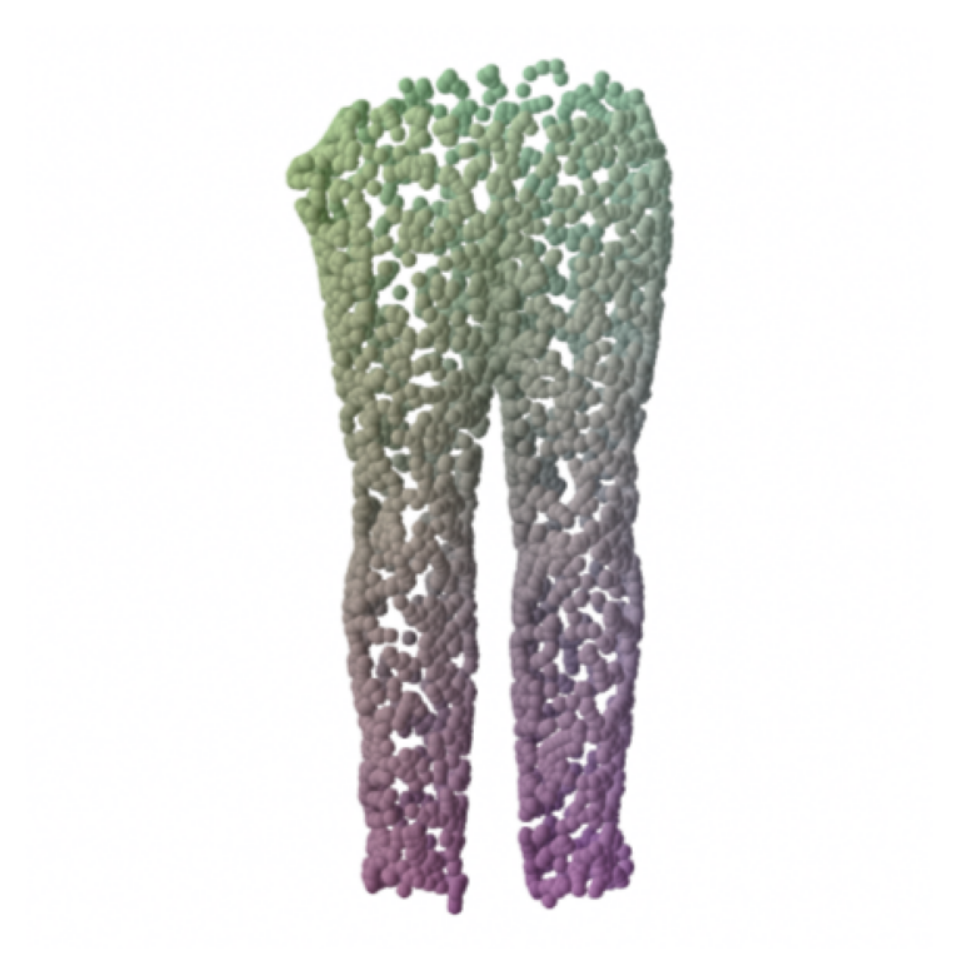} };
				\node[] (a) at (-6.5,3.7) {\includegraphics[width=0.24\textwidth]{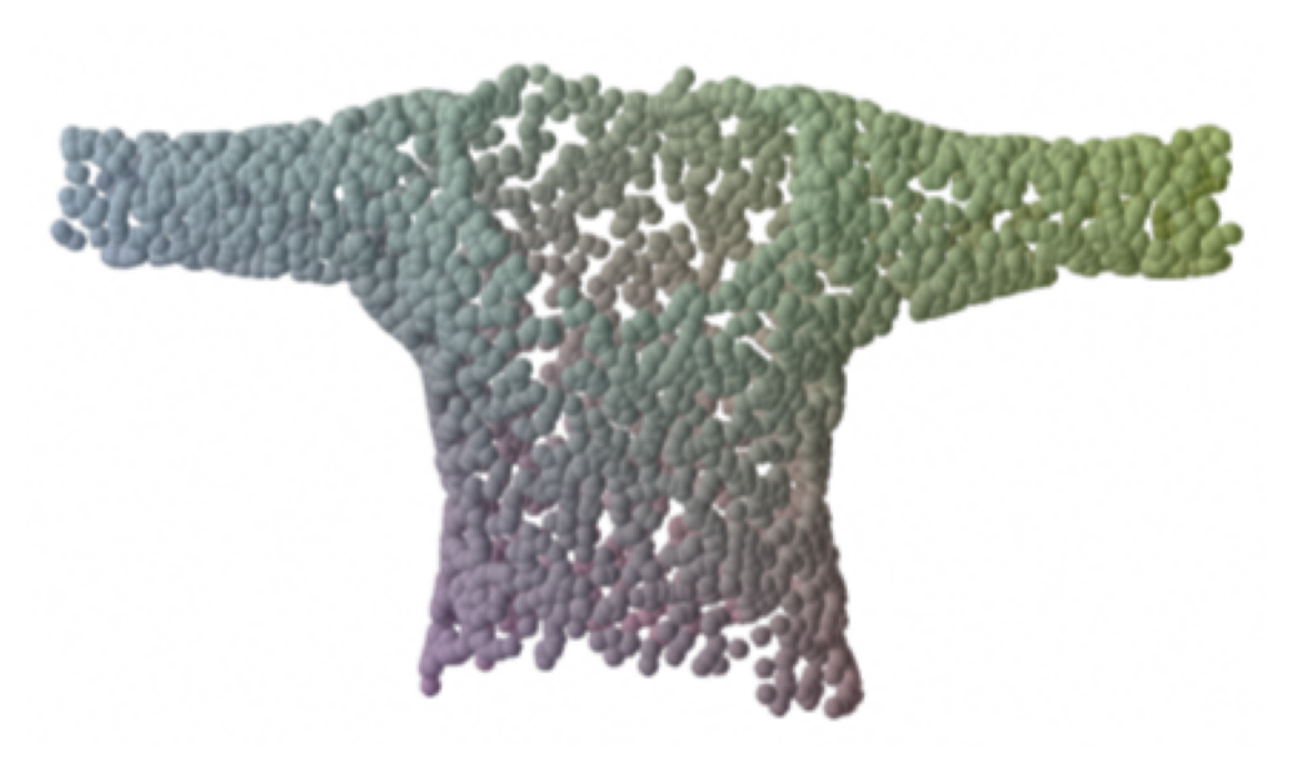} };
				\node[] (a) at (-6.5,-1.6) {\small  Input};
				
				\node[] (b) at (-6.5+13/3,0.6) {\includegraphics[width=0.24\textwidth]{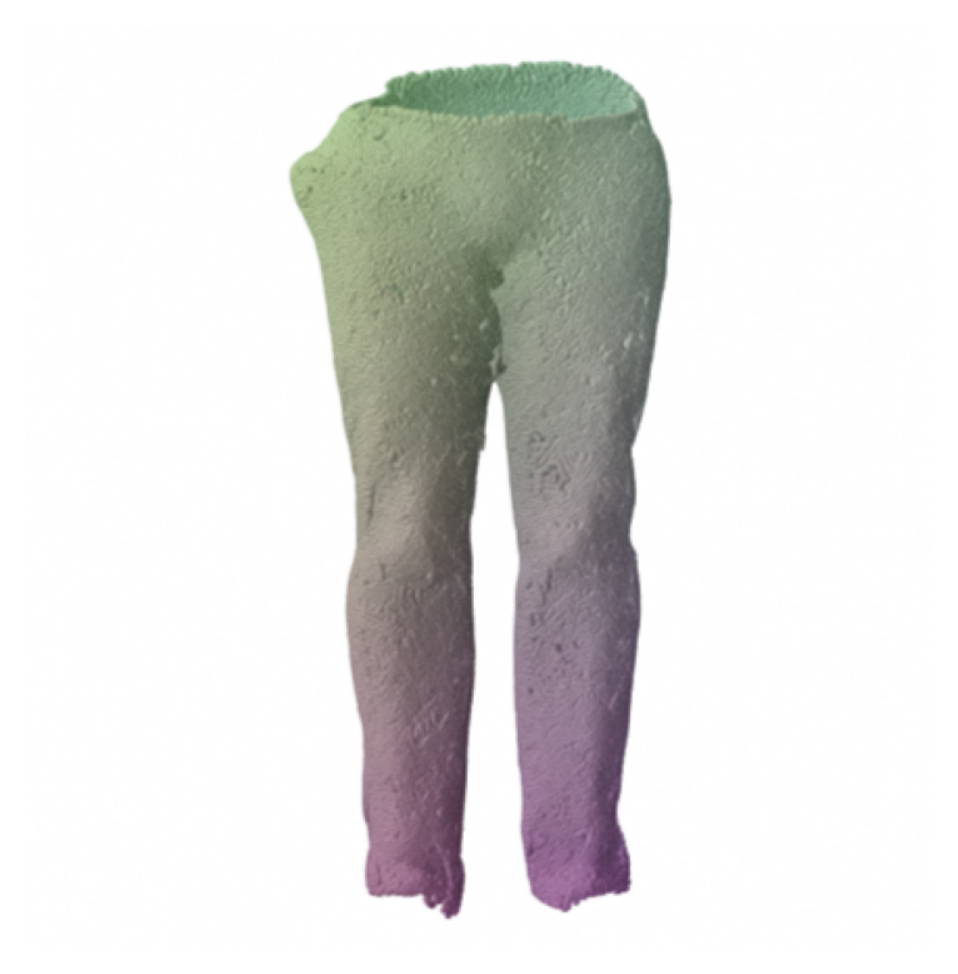}};
				\node[] (b) at (-6.5+13/3,3.7) {\includegraphics[width=0.24\textwidth]{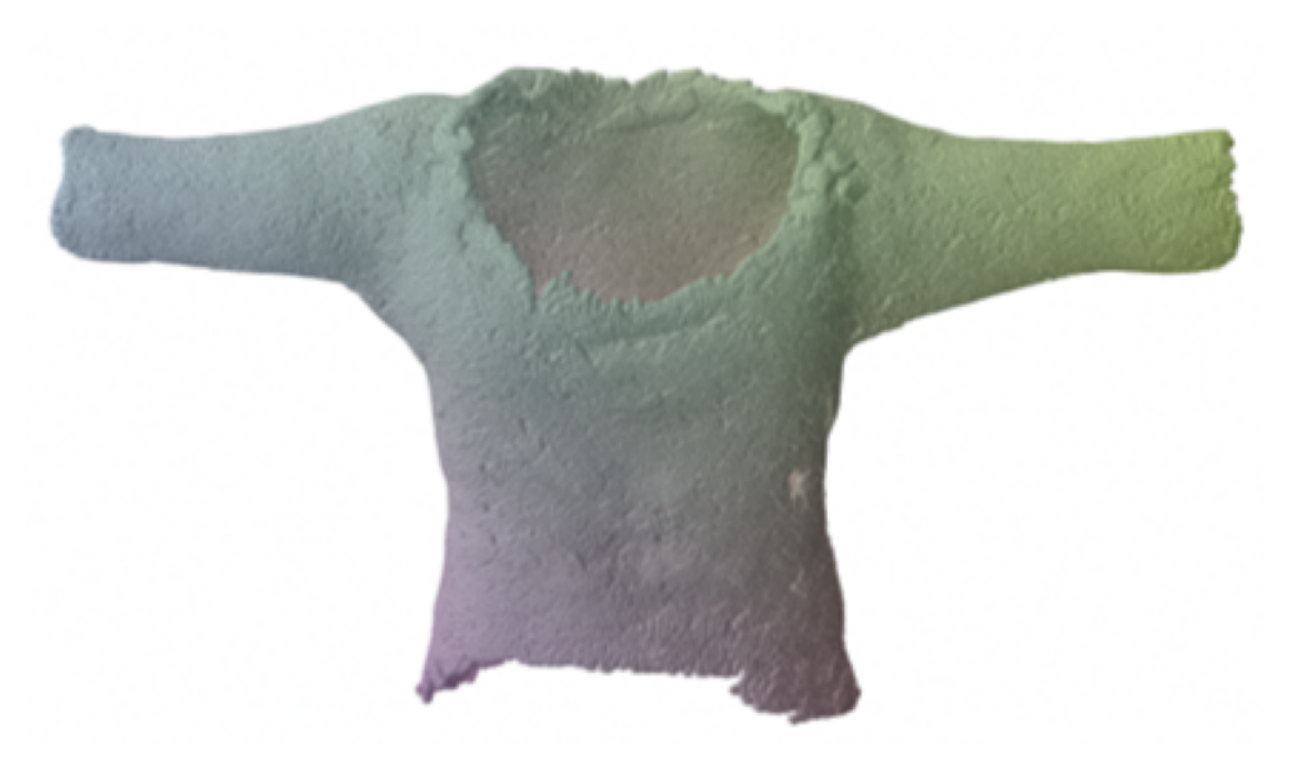}};
				\node[] (b) at (-6.5+13/3,-1.6) {\small  Upsampled };
				
				\node[] (c) at (-6.5+13/3*2,0.6) {\includegraphics[width=0.24\textwidth]{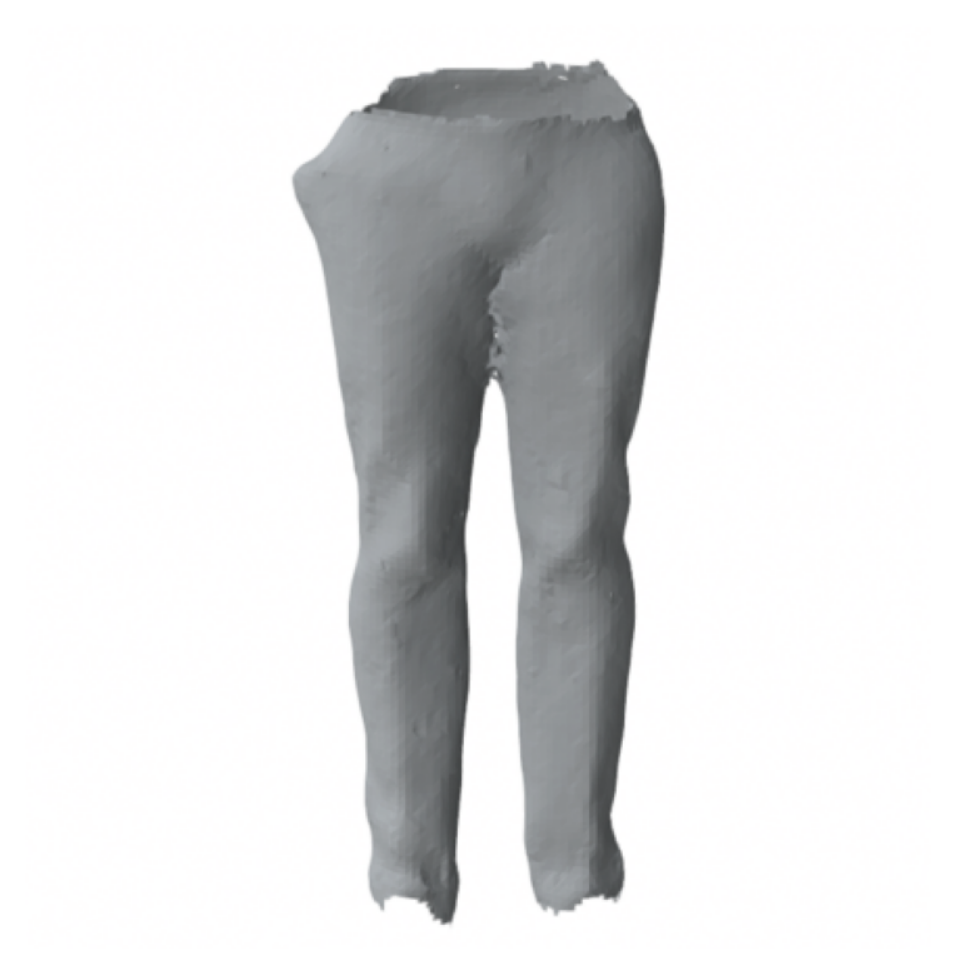}};
				\node[] (c) at (-6.5+13/3*2,3.7) {\includegraphics[width=0.24\textwidth]{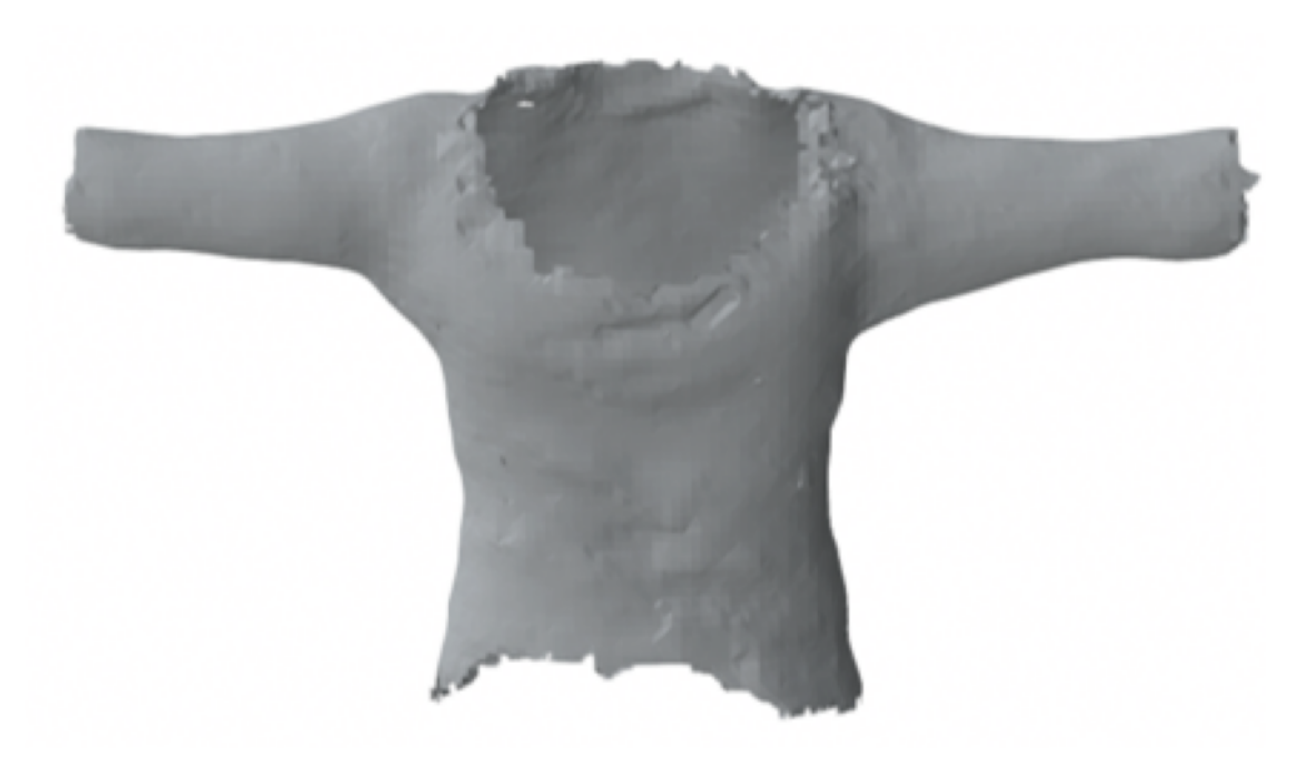}};
				\node[] (c) at (-6.5+13/3*2,-1.6) {\small  Ours  };
				
				\node[] (d) at (-6.5+13/3*3+0.2,0.6) {\includegraphics[width=0.24\textwidth]{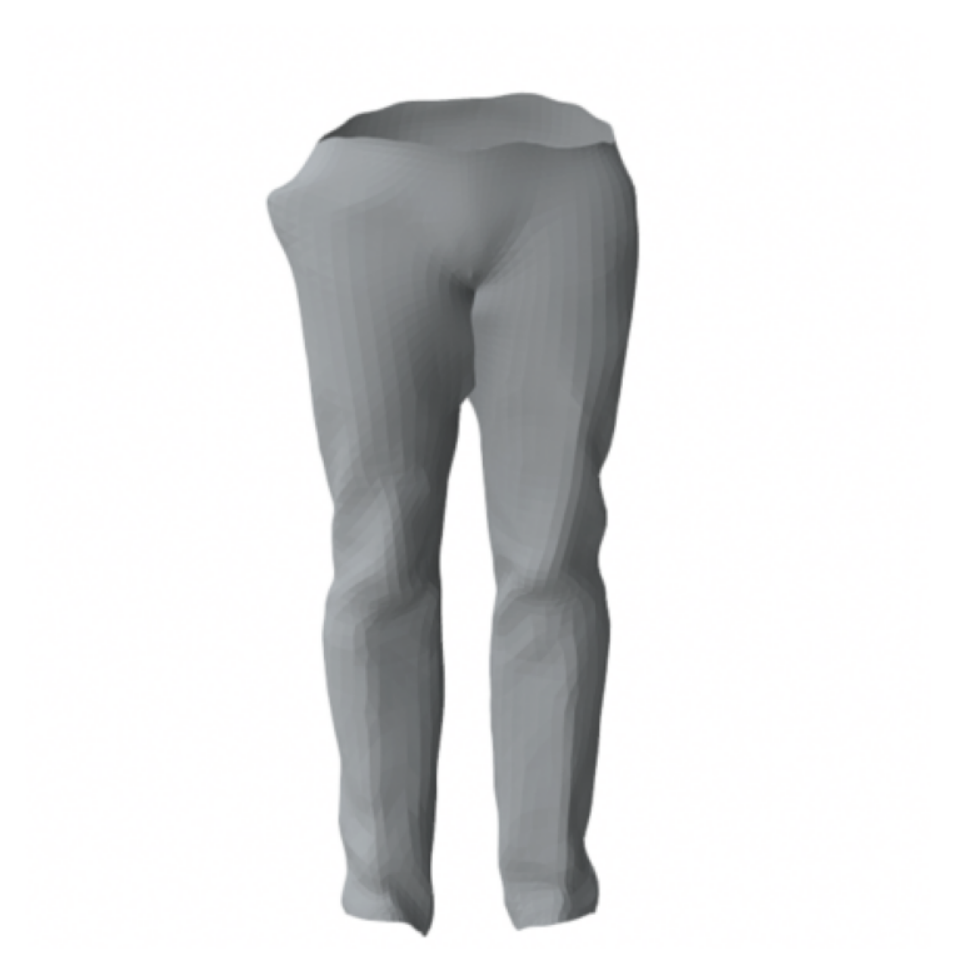}};
				\node[] (d) at (-6.5+13/3*3,3.7) {\includegraphics[width=0.24\textwidth]{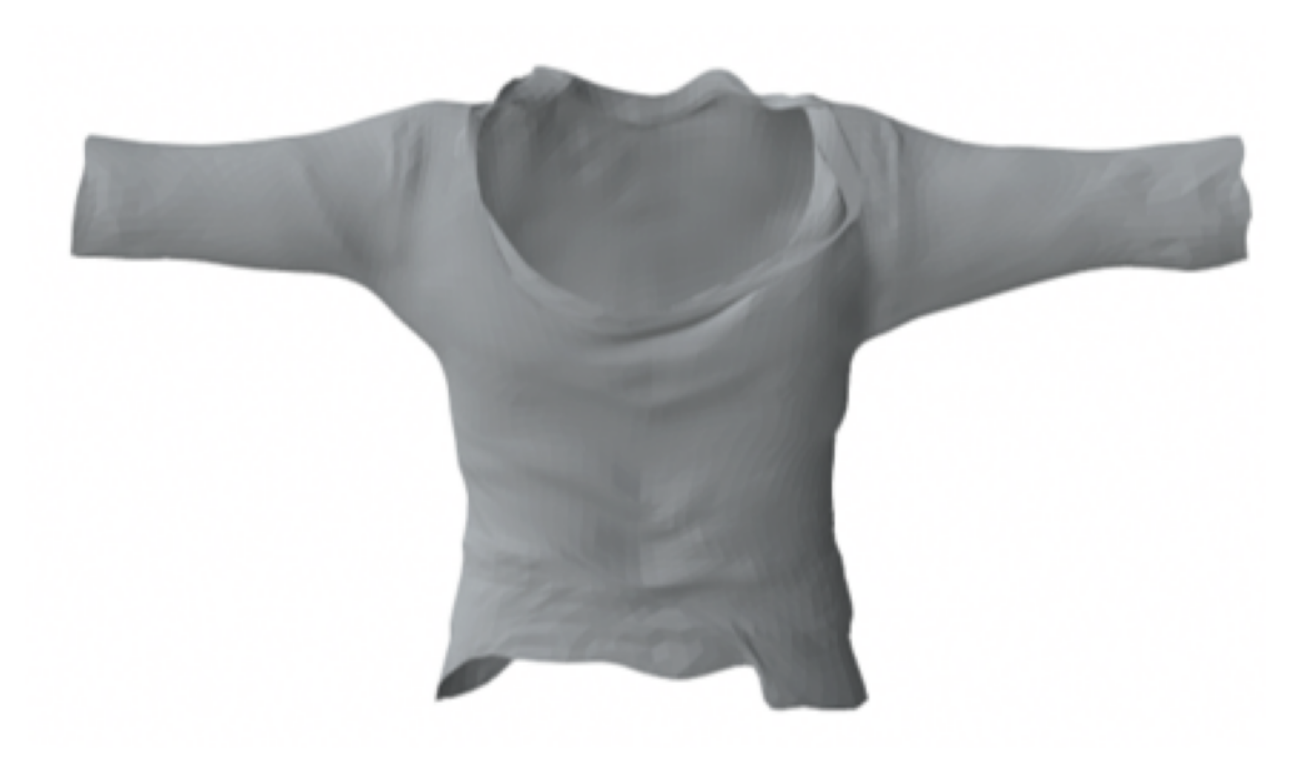}};
				\node[] (d) at (-6.5+13/3*3,-1.6) {\small  GT };
				
			\end{tikzpicture}
		}
		\setlength{\abovecaptionskip}{-0.35cm} 
		\caption{More results of reconstructed open shapes. }
		\label{MORE:OPEN}
	\end{figure}
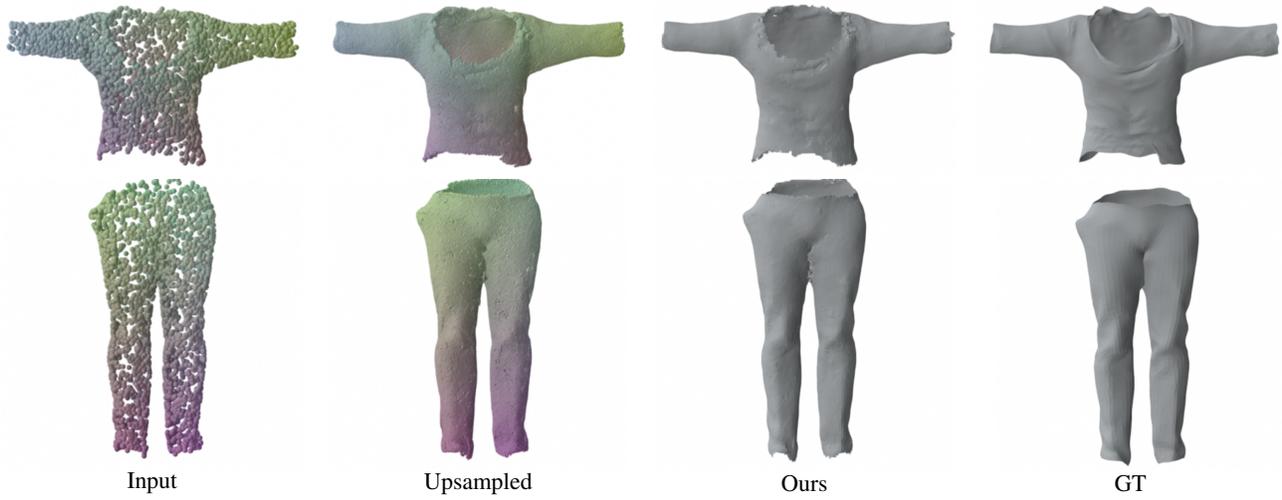

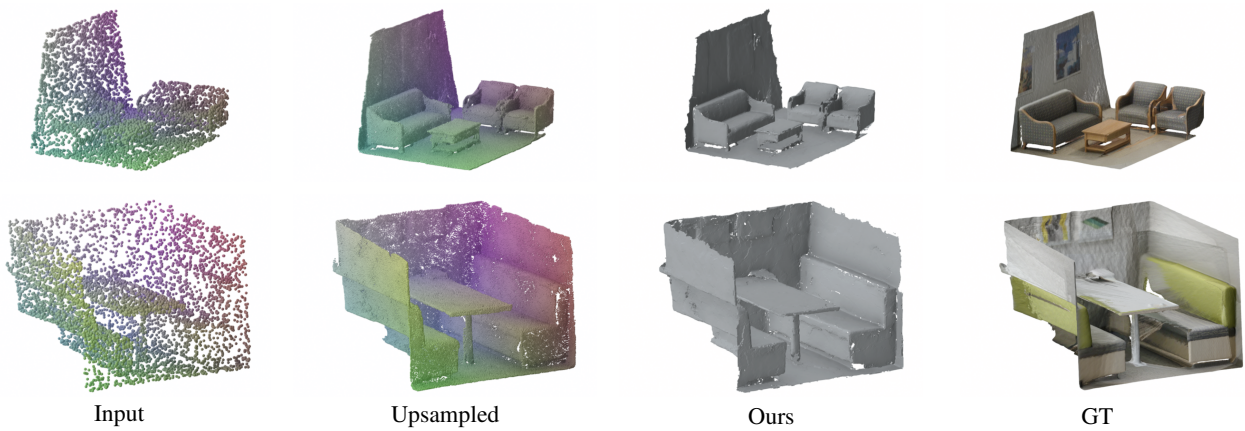
\begin{figure}[h]
		\centering
		{
			\begin{tikzpicture}[]
				\node[] (a) at (-6.5,0.6) {\includegraphics[width=0.24\textwidth]{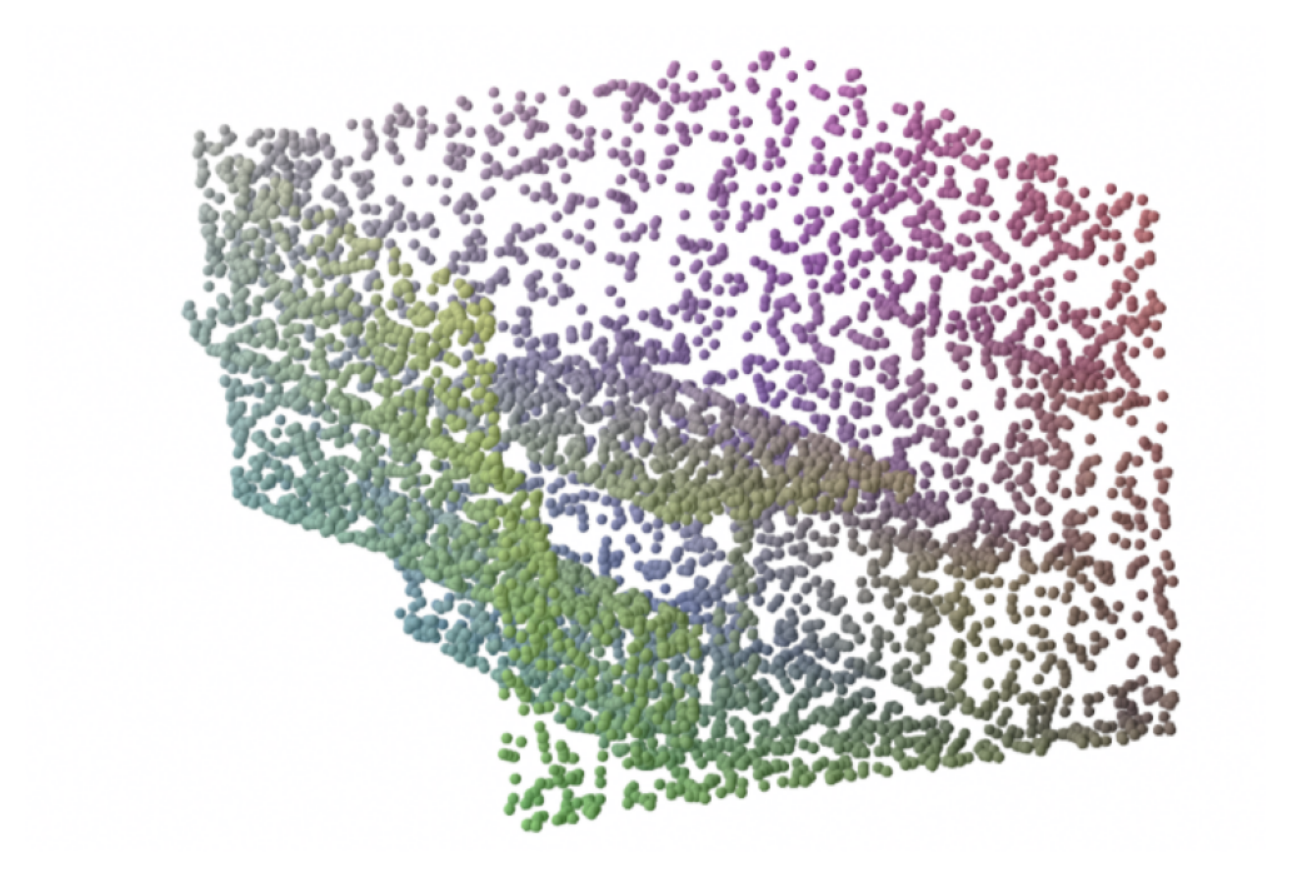} };
				\node[] (a) at (-6.5,3.3) {\includegraphics[width=0.24\textwidth]{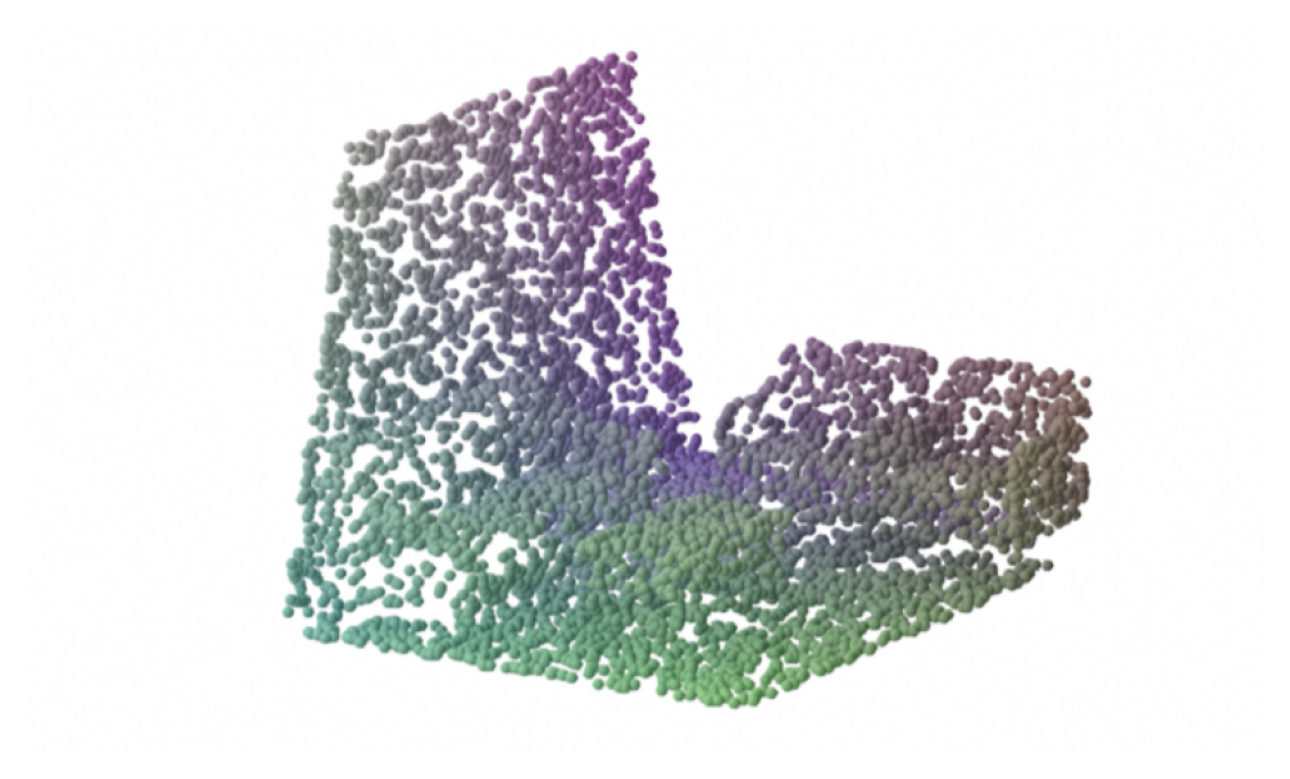} };
				\node[] (a) at (-6.5,-1) {\small  Input};
				
				\node[] (b) at (-6.5+13/3,0.6) {\includegraphics[width=0.24\textwidth]{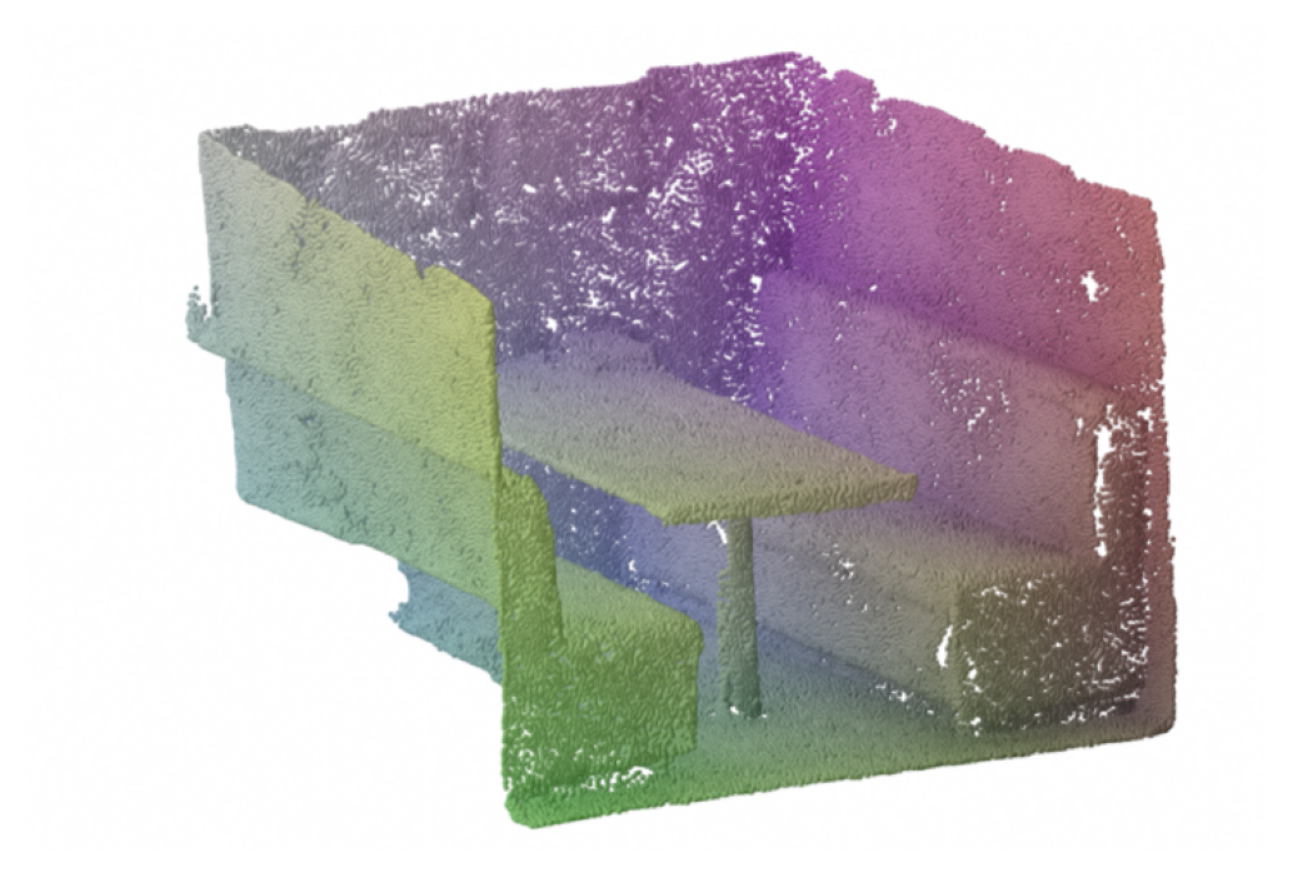}};
				\node[] (b) at (-6.5+13/3,3.3) {\includegraphics[width=0.24\textwidth]{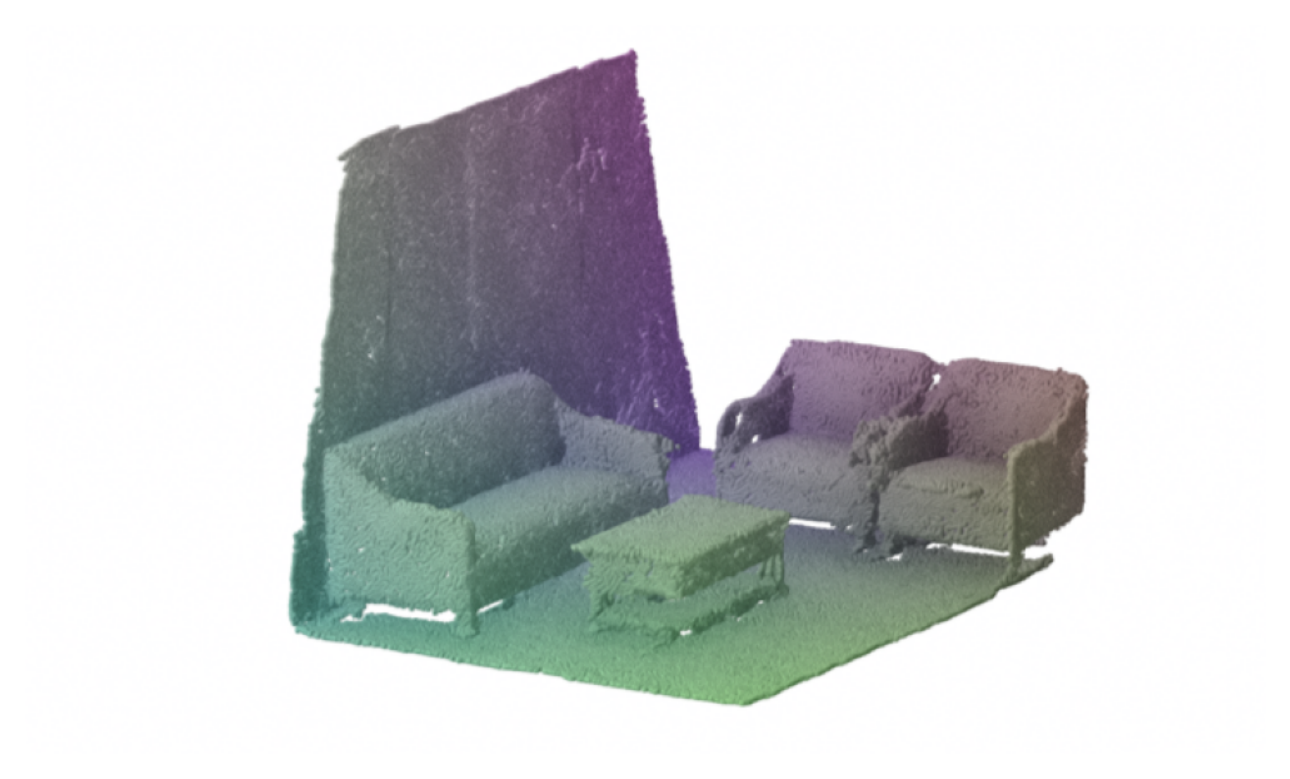}};
				\node[] (b) at (-6.5+13/3,-1) {\small  Upsampled };
				
				\node[] (c) at (-6.5+13/3*2,0.6) {\includegraphics[width=0.24\textwidth]{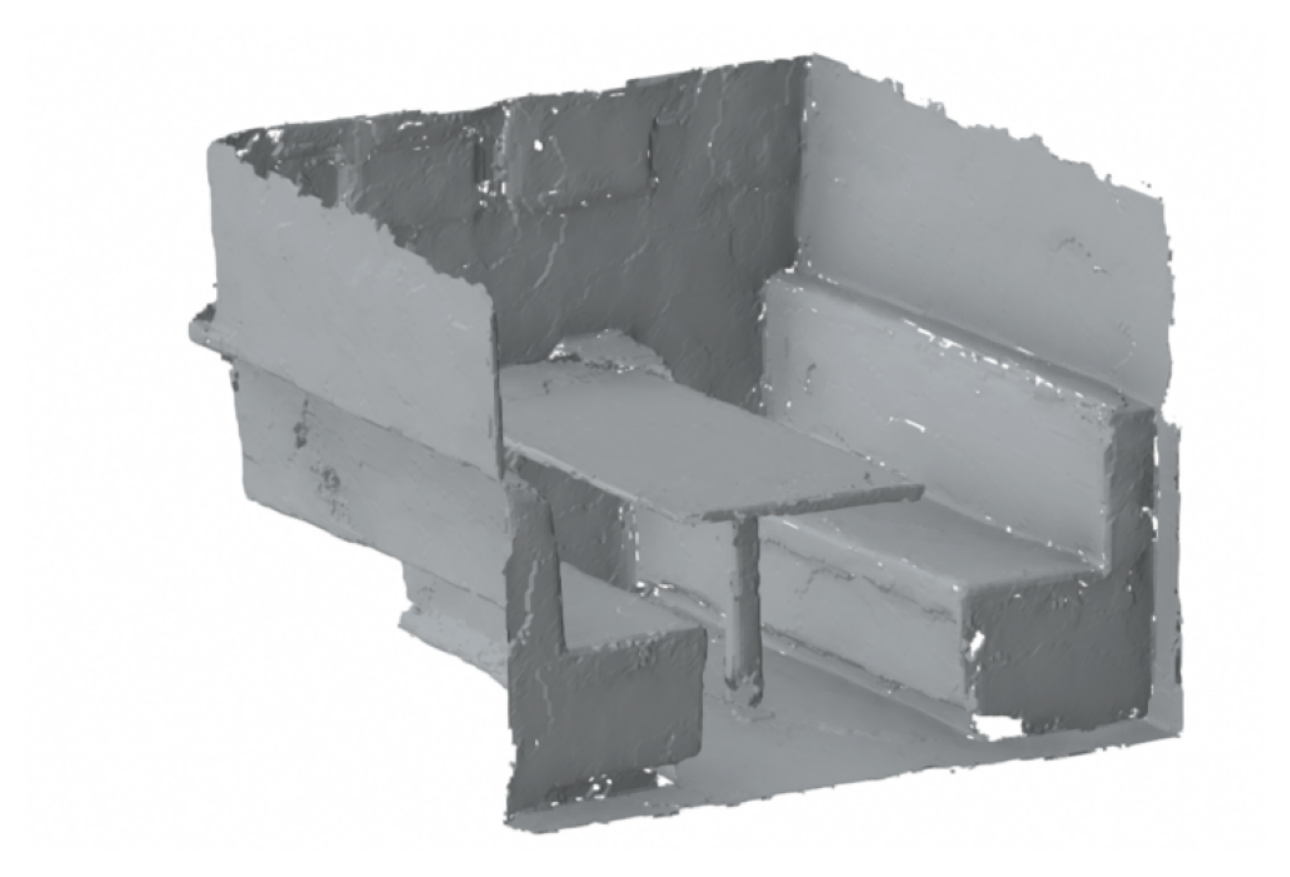}};
				\node[] (c) at (-6.5+13/3*2,3.3) {\includegraphics[width=0.24\textwidth]{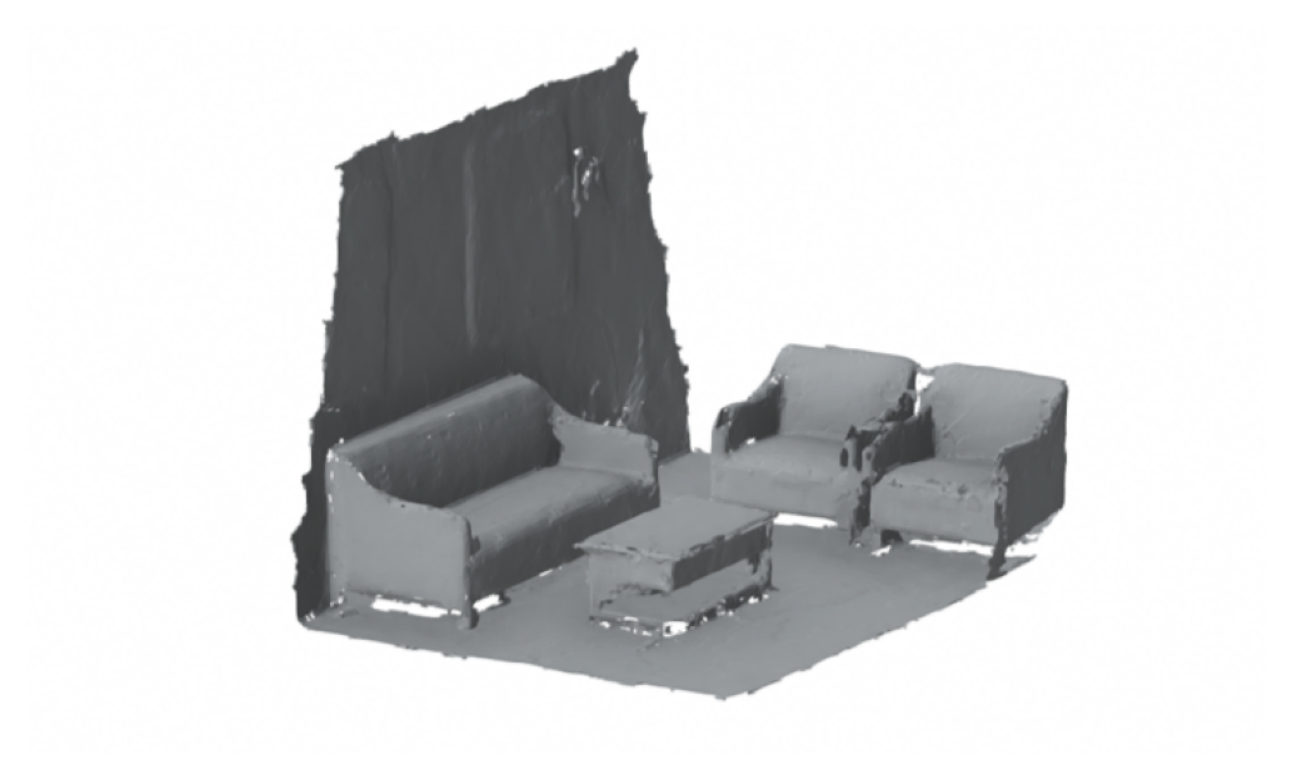}};
				\node[] (c) at (-6.5+13/3*2,-1) {\small  Ours  };
				
				\node[] (d) at (-6.5+13/3*3+0.2,0.6) {\includegraphics[width=0.24\textwidth]{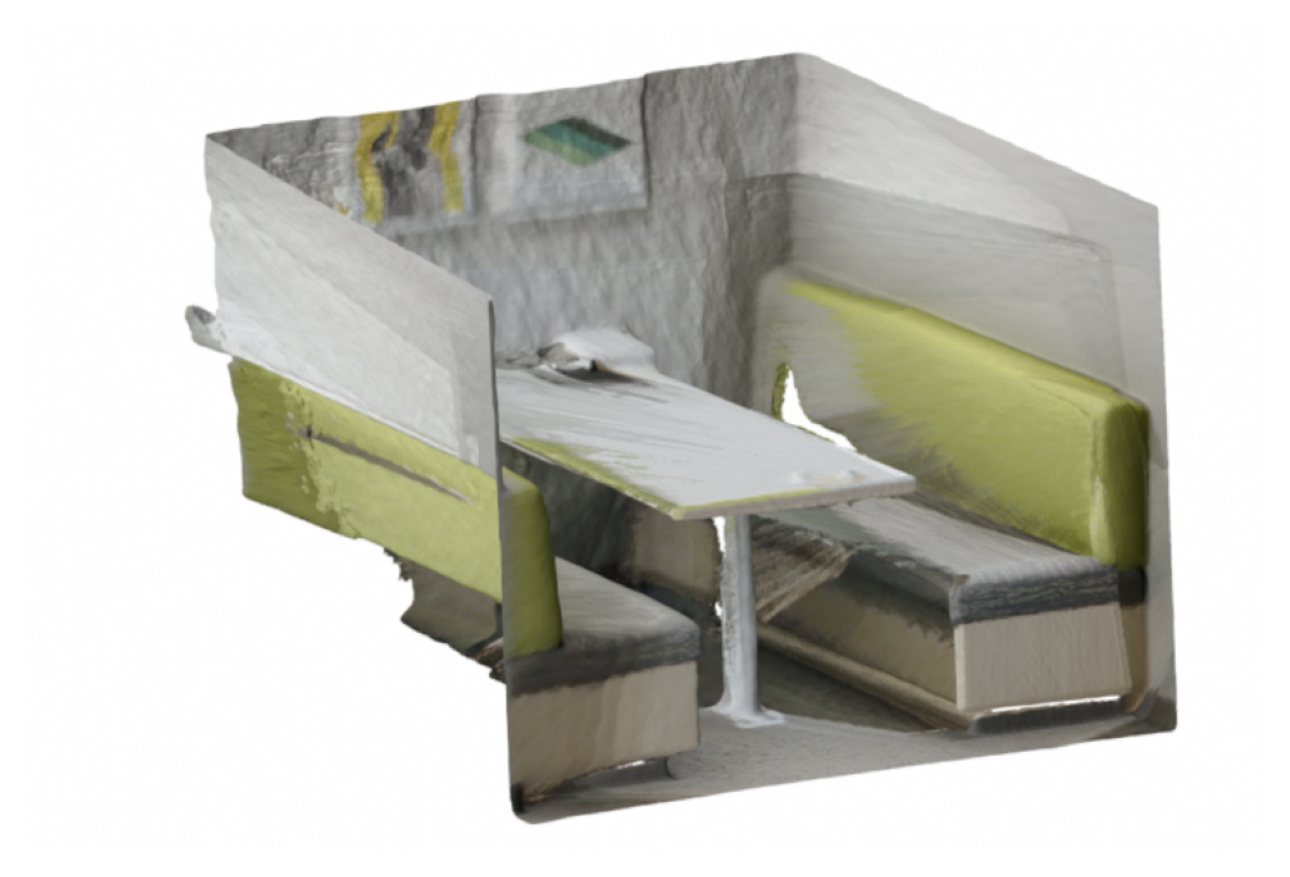}};
				\node[] (d) at (-6.5+13/3*3,3.3) {\includegraphics[width=0.24\textwidth]{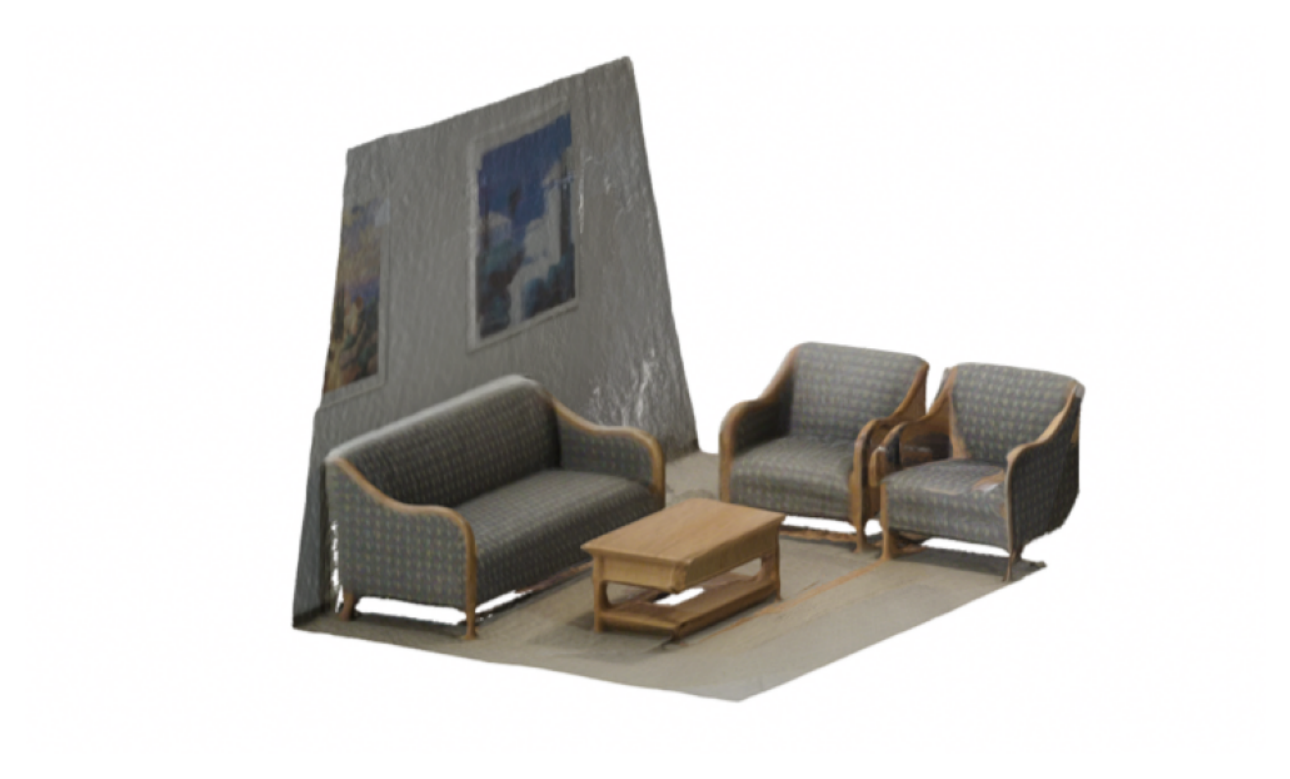}};
				\node[] (d) at (-6.5+13/3*3,-1) {\small  GT };
				
			\end{tikzpicture}
		}
		\setlength{\abovecaptionskip}{-0.35cm} 
		\caption{More results of reconstructed scene-level shapes. }
		\label{MORE:SCENE}
	\end{figure}

	\noindent\textbf{More Visual Results.} We present more visual results in Figs. \ref{MORE:MULTILAYER}, \ref{MORE:SHAPENET}, \ref{MORE:OPEN}, and \ref{MORE:SCENE}.
	
	\subsection{Ablation Study}
	\noindent\textbf{Upsampling Process.} As mentioned in our manuscript, the quadratic polynomial is sufficient to approximate a small surface area. To verify this, we trained a network by replacing Eq. \textcolor{red}{(1)} of the manuscript with the cubic polynomial as a comparison. The results are shown in the following Table \ref{CUBIC}, where it can be seen that the cubic polynomial decreases the upsampling  accuracy because of overfitting. \\

\begin{figure}[h]
		\centering
		\includegraphics[width=0.8\textwidth]{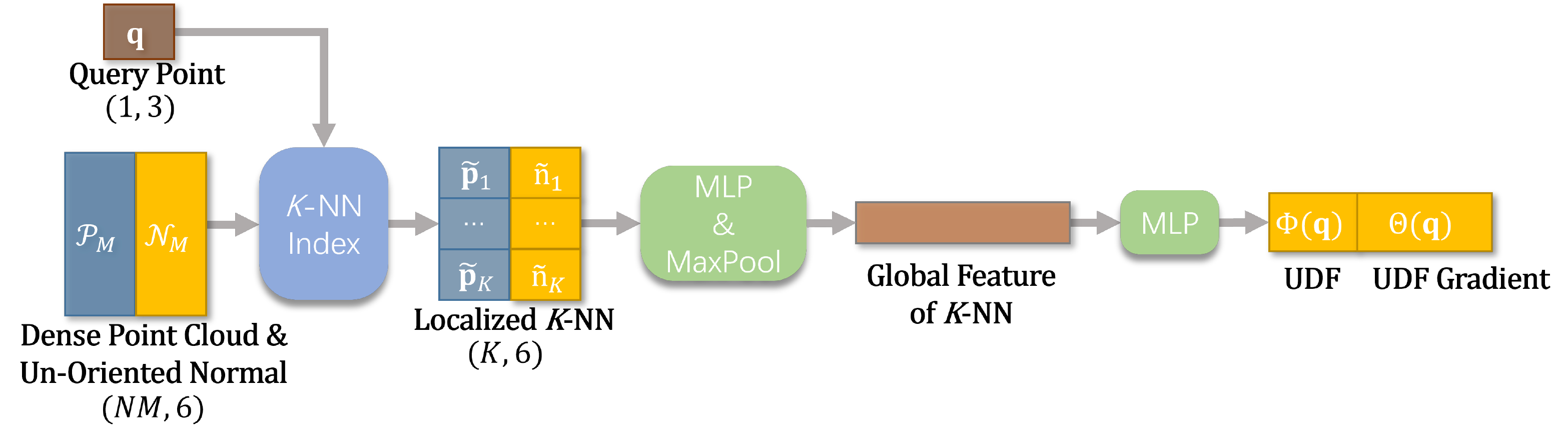}
		\caption{The network architecture of the baseline method ``Regress" in Table \textcolor{red}{6} of the manuscript.}
		\label{BASELINE}
	\end{figure}

	\noindent\textbf{Flowchart of the ``Regress" Method.} In Table \textcolor{red}{6} of the manuscript, we set a baseline named ``Regress" to predict UDFs and gradients by using an MLP.  Fig. \ref{BASELINE} shows the detailed flowchart.
	
	\begin{table}[h] 
		\centering
		\caption{Comparisons of Quadratic and Cubic Polynomials. }\vspace{-0.3cm}
		\begin{tabular}{l|  c c  }
			\toprule[1.2pt]
			Method    &  \makecell[c]{CD ($\downarrow$) \\ ($\times 10^{-2}$)} &  \makecell[c]{P2F ($\downarrow$) \\ ($\times 10^{-3}$)}  \\
			\hline
			Quadratic   & ${0.245}$    & ${0.512}$ \\
			Cubic       & ${0.257}$    & ${0.522}$ \\
			\bottomrule[1.2pt]
		\end{tabular}
		\label{CUBIC}
	\end{table}

	\clearpage
	{\small
		\bibliographystyle{ieee_fullname}
		\bibliography{egbib}
	}